%% file: 9arxiv.tex
\documentclass[review]{elsarticle}

\usepackage{hyperref}
\usepackage{times}
\usepackage{natbib}
\usepackage{amsmath}  
\usepackage{amsfonts}
\usepackage{amssymb} 
\let\checkmark\undefined
\usepackage{mathtools}
\usepackage{subcaption} 
\usepackage{helvet}
\usepackage{courier}
\usepackage{longtable}
\usepackage[dvipsnames]{xcolor}  
\usepackage{rotating}
\usepackage{graphicx}
\usepackage{float}
\usepackage{diagbox}
\usepackage{wrapfig}
\usepackage{dingbat}
 
\usepackage{pifont}
 
\usepackage{multirow}
\usepackage{makecell}
\usepackage[ruled,linesnumbered,vlined]{algorithm2e}
\usepackage{algpseudocode}

\usepackage{titleps}
\renewpagestyle{plain}{%
\sethead{}{}{}
\setfoot[\thepage][][]{}{}{\thepage}
}%
\usepackage{dblfloatfix}
\usepackage{url} 
\usepackage{pgf} 
\usepackage{dingbat}

\usepackage{changes}
\newtheorem{Thm}{Theorem}
\newtheorem{Lma}{Lemma}
\newtheorem{Cly}{Corollary}
\newtheorem{Def}{Definition} 
\newtheorem{Prf}{Proof} 
\newtheorem{PrfLma}{Proof (lemma)} 
\newtheorem{PrfCly}{Proof (corollary)} 
\newtheorem{rmk}{Remark}

\newcommand\syllogism[3][]{%
  \begin{center}
  \def\tmp{#1}%
  \ifx\tmp\empty\else(#1)\quad\fi
  \begin{tabular}{@{}l@{}}#2\\\hline#3\quad$\therefore$
  \end{tabular}
  \end{center}
}

\newcommand\reasoningQA[3][]{%
  \begin{center}
  \def\tmp{#1}%
  \ifx\tmp\empty\else(#1)\quad\fi
  \begin{tabular}{@{}l@{}}
    #2\\\hline#3\quad$?$
  \end{tabular}
  \end{center}
}

\journal{Journal of \LaTeX\ Templates}

\begin{document}

\begin{frontmatter}

\title{
Sphere Neural Networks for Rational Reasoning
}
\author{Tiansi Dong, Mateja Jamnik, Pietro Li\`o} 

\address{Department of Computer Science and Technology\\The University of Cambridge\\
15 JJ Thomson Ave, Cambridge, UK\\
\{td540, mj201,pl219\}@cam.ac.uk}

\input{9arxiv/abstract_no_ref}


\end{frontmatter}
 

\newpage
\tableofcontents 
\newpage
\input{9arxiv/visual_introduction}

\newpage
\input{9arxiv/methodology}

\input{9arxiv/syllogism_1}

\input{9arxiv/syllogism}

\input{9arxiv/proofs}

\input{9arxiv/power_of_rep}

\input{9arxiv/conclusion}

\input{9arxiv/experiments_new}

\section{Data and Code availability}

Datasets and codes will be published for open access after the formal publication of this work.

\section{Acknowledgement}

This work is one of the follow-ups of the Dagstuhl Seminar \href{https://www.dagstuhl.de/seminars/seminar-calendar/seminar-details/21362}{Structure and Learning} \cite{dong_et_al22} held during September 5 -- 10, 2021. The authors thank all participants of the Dagstuhl Seminar, for their generous contribution and active participation.

\bibliography{XBib_NN_s,XBib}
\appendix

\input{9arxiv/syllogism_list}

\input{9arxiv/map_arch}

\end{document}

%% file: 9arxiv/abstract_no_ref.tex
\begin{abstract} 
The success of Large Language Models (LLMs) is witnessed by their planetary popularity and their steadily improved reasoning performance. However, it remains unclear whether LLMs reason. It is an open problem how traditional neural networks can be qualitatively extended to go beyond the statistic paradigm and achieve high-level cognition. 
Here, we present a novel qualitative extension by generalising computational building blocks from vectors to spheres. Sphere boundaries introduce contact and non-contact relations, which are primitive relations for both spatial and non-spatial reasoning. We propose Sphere Neural Networks (SphNNs) for human-like reasoning through model construction and inspection, and develop  $\mathcal{S}$phNN for syllogistic reasoning, a microcosm of
human rationality. $\mathcal{S}$phNN has a Kolmogorov-Arnold Neural (KAN) architecture, where each edge has an activation function to realise transitions between neighbourhood spatial relations. 
$\mathcal{S}$phNN is the first neural model that can determine
the validity of long-chained syllogistic reasoning in one epoch with the worst computational complexity of ${O}(N)$ (where $N$ is the length of the chain). $\mathcal{S}$phNN achieved $100\%$ accuracy in determining the validity of every atomic syllogistic reasoning (Experiment~\ref{exp2}) and $100\%$ accuracy in long-chained syllogistic reasoning, while both ChatGPT (gpt-3.5-turbo and gpt-4o) cannot (Experiment~\ref{exp3}). $\mathcal{S}$phNN can serve as System 2 to evaluate and feedback ChatGPT through prompts, through which ChatGPT improves the accuracy in deciding the satisfiability of atomic syllogistic reasoning (Experiment~\ref{exp4}). Pre-trained vectors can serve as System 1 to hypothesize orientations of sphere centres and shorten the time cost of $\mathcal{S}$phNN in sphere construction (around 90\% were shortened in Experiment ~\ref{exp5}). $\mathcal{S}$phNN can evolve into 
various types of reasoning, such as spatio-temporal reasoning, logical reasoning with negation and disjunction, Bayesian reasoning, and humour understanding (the highest level of cognition). All these suggest a new Herbert A. Simon's scissors with two neural blades: the representation blade using spheres to represent tasks and environments and the reasoning blade solving problems through the transformation of sphere configurations.  
SphNNs will tremendously enhance interdisciplinary collaborations and realise human-bounded rationality. This work suggests that the non-zero radii of spheres are the missing components that prevent traditional deep-learning systems from reaching the realm of rational reasoning and cause LLMs to be trapped in the swamp of indeterminacy and hallucination. 
\end{abstract}

%% file: 9arxiv/visual_introduction.tex
 
\section{Introduction}

\begin{quote}
\centering
    {\em The origin of reason is gradually developed. 
    
    \hfill --- Charles Darwin} 
\end{quote}

It is the rational reasoning that distinguishes humans from animals. This idea can be traced back to Aristotle's time, who invented syllogistic reasoning and founded the science of logic to guarantee the validity of reasoning -- a valid logical conclusion explicitly states what is already implicit in the premises \cite{Simon19}. Thus, the information within the syllogistic premises is sufficient to determine the validity of the syllogistic conclusion. Other training data shall not be needed. In general, people only have insufficient information and limited capabilities, and demonstrate {\em bounded rationality}. They are prone to subgoals or heuristic solutions \citep{Simon19}. Herbert A. Simon raised an inquiry {\em how can the standard theories of global rationality be simplified to render bounded rationality more tractable}?  In the era of deep-learning neural networks, rationality is no longer bounded by learning resources and computing power but by ways of representations, vectorial or symbolic. They are heterogeneous building blocks to construct decision-making {\em artifacts}. Using vectorial representation, deep learning neural networks have succeeded in doing a variety of tasks \citep{LeCunNature15}, including defeating human champions in playing Go \citep{alphaGo2017,alphaGo2020}, and are moving towards simulating high-level cognition, e.g., rational reasoning \citep{Bengio22}. 
Here, we focus on a sub-question of Herbert A. Simon's inquiry, namely, {\em how deterministic syllogistic reasoning can be tractably rendered into neural computing} and moves on to develop novel neural computing upon traditional neural computing, which can evolve to various rational reasoning, including statistic reasoning and heuristic reasoning. 

\paragraph{{\bf Non-deterministic judgements of GPT-4}} The success of deep learning is witnessed by its planetary usage, especially LLMs that demonstrate the fantastic capability of human-like question-answering. Here is a sample communication with GPT-4. 
\begin{quote}
Pietro: {\em Which city is located further east, San Diego or Reno?}

GPT-4: {\em Reno, Nevada, is further east than San Diego, California. While San Diego is located on the coast of the Pacific Ocean, Reno is situated more inland, towards the eastern side of the Sierra Nevada mountain range.}
\end{quote}
\begin{figure}[!t] 
\centering 
\includegraphics[width=1\linewidth]{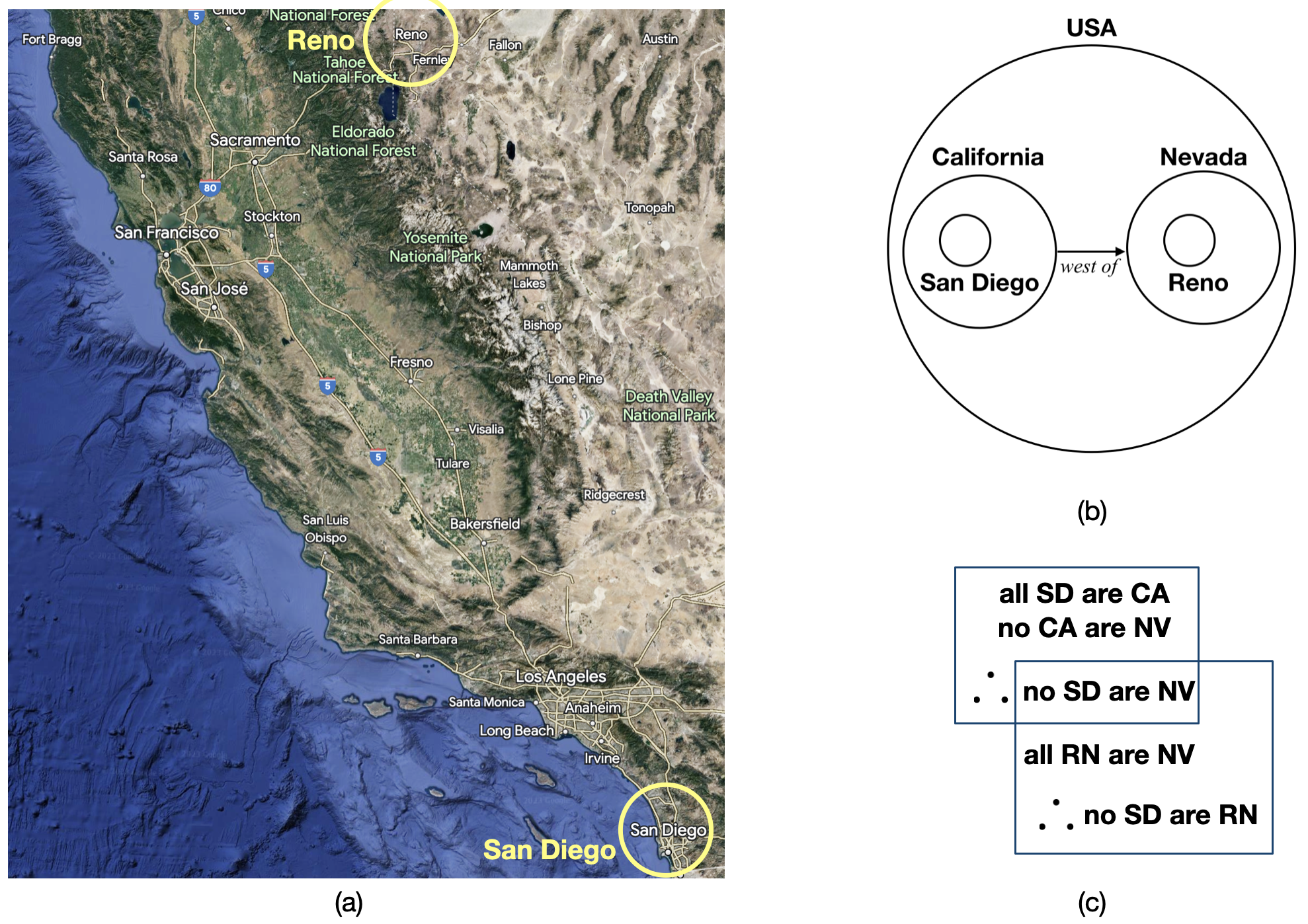} 
\caption{(a) The geographical location of San Diego and Reno; 
(b) the region-based mental spatial representation explains why people mistakenly judge the spatial relation between San Diego and Reno; (c) two-step syllogistic reasoning to judge the relation between San Diego and Reno.}
\label{reno}
\end{figure}
Recent exploration shows that LLMs, e.g., ChatGPT, may pass the Turing test in human-like chatting but have limited capability even for simple reasoning tasks \cite{chatgpt_nature2023}. It remains unclear whether LLMs can reason or only do indeterministic heuristic association \cite{Mitchell_science.adj5957}. 
Let us continue the chat as follows.
\begin{quote}
Pietro: {\em Are you sure?}

GPT-4: {\em I apologize for the confusion. You're right. Although San Diego, California, is geographically south of Reno, Nevada, when considering the longitudinal coordinates, San Diego is actually further east. San Diego sits at a longitude of approximately 117.2 degrees west, while Reno is at approximately 119.8 degrees west. So, contrary to intuitive geography based on north-south positioning, San Diego is actually further east than Reno. Thank you for prompting me to double-check this information.}

Pietro: {\em But your first answer is correct. }

GPT-4: {\em I apologize for any confusion. You're correct. Reno, Nevada, is indeed further east than San Diego, California. San Diego, despite being south of Reno, is more towards the west because it is on the coast. I appreciate your patience, and I'm sorry for any confusion caused earlier. Thank you for helping ensure the accuracy of the information.}
\end{quote}

\paragraph{{\bf Sphere Neural-Network for deterministic syllogistic reasoning}} 

In the example, GPT-4 simulates perfect human-styled question-answering, but is not sure about its answers. Other experiments show that LLMs may demonstrate irrationality by making correct answers with incorrect explanations \cite{creswell2022selectioninference,zelikman2022star}. Many deep-learning systems achieved extremely high accuracy in various tasks \cite{ImageNet_Nips12,mnih2013,alphaGo2020}, including reasoning syllogistic statements \cite{WangJL18,WangJL20}. Yet, their judgements are unreliable, even with 100\% accuracy on the test data \citep{Bengio22}. Traditional deep-learning neural networks must go beyond the statistic learning framework and make qualitative extensions to simulate high-level cognition, e.g., rational reasoning, like humans and animals do \citep{Bengio22}.

Humans' rational reasoning can be metaphorised as scissors \cite{simon90scissor}, whose two blades represent the task environment and the cognition capability, respectively. Being asked {\em which city is located further east, San Diego or Reno?} people construct a nested region structure as the task environment as follows: San Diego is inside California, Reno is inside Nevada, and California is west to Nevada. This representation bounds their rational reasoning and leads to the mistaken judgement that San Diego (SD) is located further west to Reno (RN) \citep{StevensCoupe78,McNamara86}. To judge the relation between San Diego and Reno, people inspect the constructed spatial model and inherit the relation between San Diego and Reno from their parent regions, as illustrated in Figure~\ref{reno}(b). 
If we 
understand {\em being further west to} as a specified relation of {\em being disconnected from}, as illustrated in Figure~\ref{reno}(a), the above reasoning becomes the following rougher reasoning. 
\syllogism{
  All {\em San Diego}  is {\em California}.\\
  No {\em California} is {\em Nevada}. 
}{
  No {\em San Diego} is {\em Nevada}.
} 
\syllogism{
   No {\em San Diego} is {\em Nevada}.\\ All {\em Reno} is {\em Nevada}.
}{
  No {\em San Diego} is {\em Reno}.
}  
This is Aristotelian syllogistic reasoning, the other blade of the scissors. Syllogistic reasoning made logic {\em 
unable to take a single step forward [since Aristotle], and therefore seems to all
appearance to be finished and complete}, as described by Immanuel Kant in {\em the Critique of Pure Reason}. Consider the well-known syllogistic reasoning as follows.  
\syllogism{
 all {\em men} are {\em mortal}.\\
 all {\em Greeks} are {\em men}. 
}{
 all {\em Greeks} are {\em mortal}.
} 
The conclusion that {\em all Greeks are mortal} can be obtained by constructing Euler diagrams.
Reasoning by mental model construction and inspection is best supported by empirical experiments \cite{LairdByrne91,knauf03,GoodwinLaird05,Knauff09,laird2012,Knauff13,RagniMarkus13,enigma2018,Khemlani2021}. 
Mental models are first developed for spatial environments and used as references for domain-general reasoning \citep{Bellmundeaat6766,tversky19}. For example, to infer which city has more inhabitants, San Diego or San Antonio? German students reached $100\%$ accuracy while American students only reached $62\%$ because all German students heard about San Diego but not San Antonio, and assumed San Diego was larger and, thus, had more inhabitants 
\cite[p.43]{Gigerenzer99}. To infer the performances of soccer teams, people will reference the sizes of cities and assume recognised cities are likely to be larger than unrecognised cities, thus having more qualified soccer teams  \citep{Gigerenzer99}. Philosophically, size relations can be formalised by the connection relation~\cite{Laguna22a,DongJPL}. Eminent philosophers and psychologists advocated the fundamental roles of connection relation and regions in cognitive modelling \cite{Laguna22a,whitehead29,carey09,Smith94,tversky19}. In contrast,
traditional neural networks use vectors and similarity relations \cite{LeCunNature15}. 
Here, we computationally reconcile them by extending vectors into spheres and consequently extending traditional neural networks into sphere neural networks (SphNNs). Spheres are diameter-fixed geometric entities and can be used as Euler diagrams to represent set-theoretic relations in the vector space, in this way, SphNNs have the genealogy from both Minsky's diameter-limited perceptron \cite{minsky88} and Rosenblatt's set-diagrammatic network architecture \cite{rosenblatt62}.   The capability of set-theoretic knowledge representation endows SphNNs with the power of {\em model construction and inspection} for {\em deterministic reasoning}. 

\begin{sidewaysfigure}
\centering
\vspace*{0mm}
\hspace*{-25mm}
\includegraphics[width=1.3\textwidth]{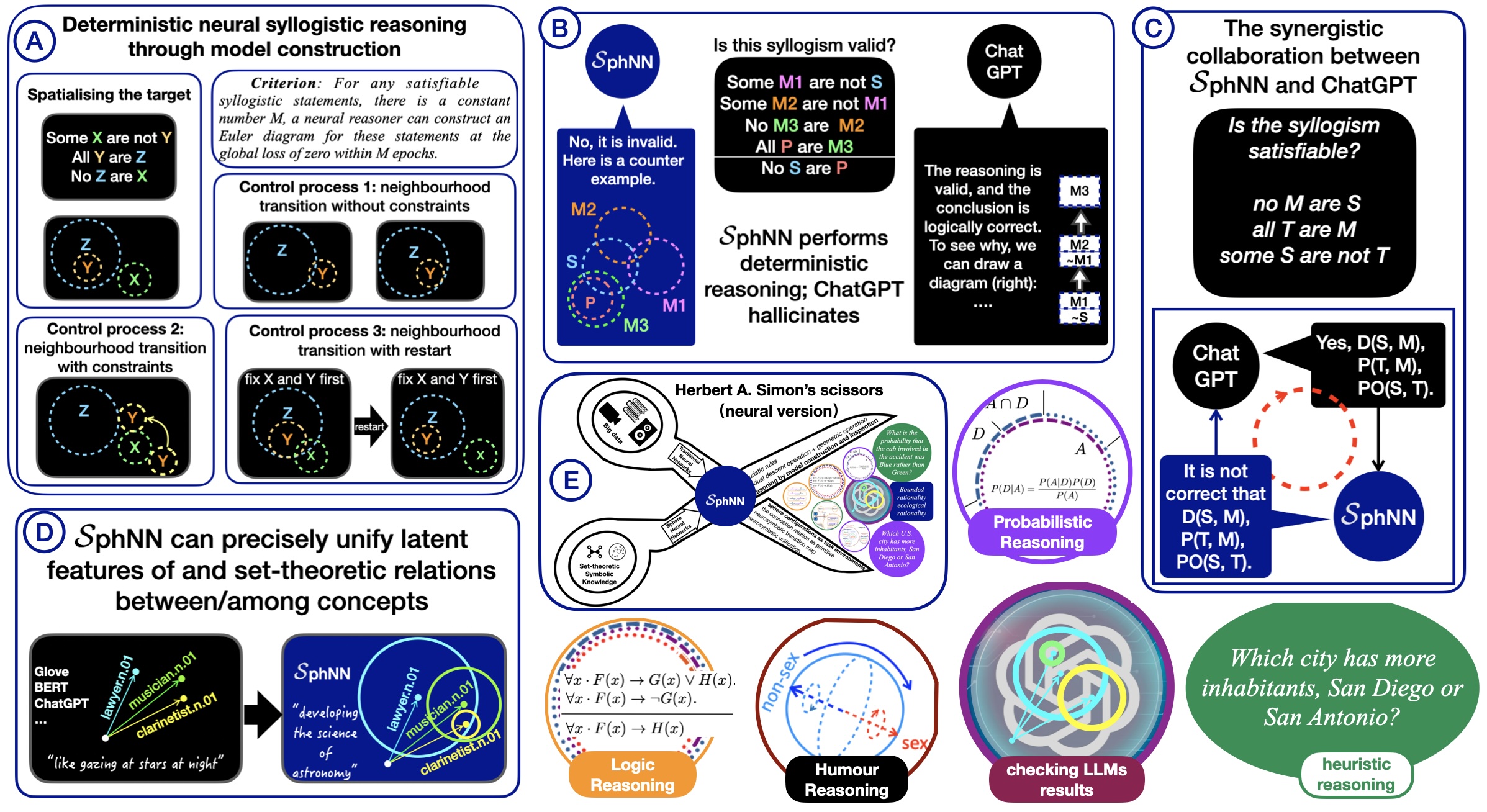}
\caption{(A)  $\mathcal{S}$phNN for deterministic neural syllogistic reasoning equipped with a strict criterion; (B) Comparative experiments about $\mathcal{S}$phNN and ChatGPT; (C) $\mathcal{S}$phNN greatly improves the performance of ChatGPT through prompt engineering; (D) Spheres can precisely unify latent features of and set-theoretic relations between/among concepts; (E) 
$\mathcal{S}$phNN can evolve into a neural version of Herbert A. Simon's scissors for bounded rationality, covering various types of reasoning, e.g., logic reasoning, probabilistic reasoning, reasoning (checking) the outputs of LLMs, and reach humour reasoning (the highest level of human cognition. A demo for humour understanding via sphere representation is available at \url{https://www.ml2r.de/joke/\#title}.) and heuristic reasoning.}
\label{visual_intro}
\end{sidewaysfigure}

We develop a particular $\mathcal{S}$phNN that can validate {\em all Greeks are mortal} from {\em all men are mortal}, and {\em all Greeks are men} (each is a syllogistic statement) without training data\footnote{Traditional supervised deep learning systems cannot reach the determinacy of syllogistic reasoning. We can show that training data automatically generates new out-of-distribution data that makes a well-trained deep-learning system indeterministic.}, as 
follows:
\begin{sidewaysfigure}
\centering
\includegraphics[width=1\textwidth]{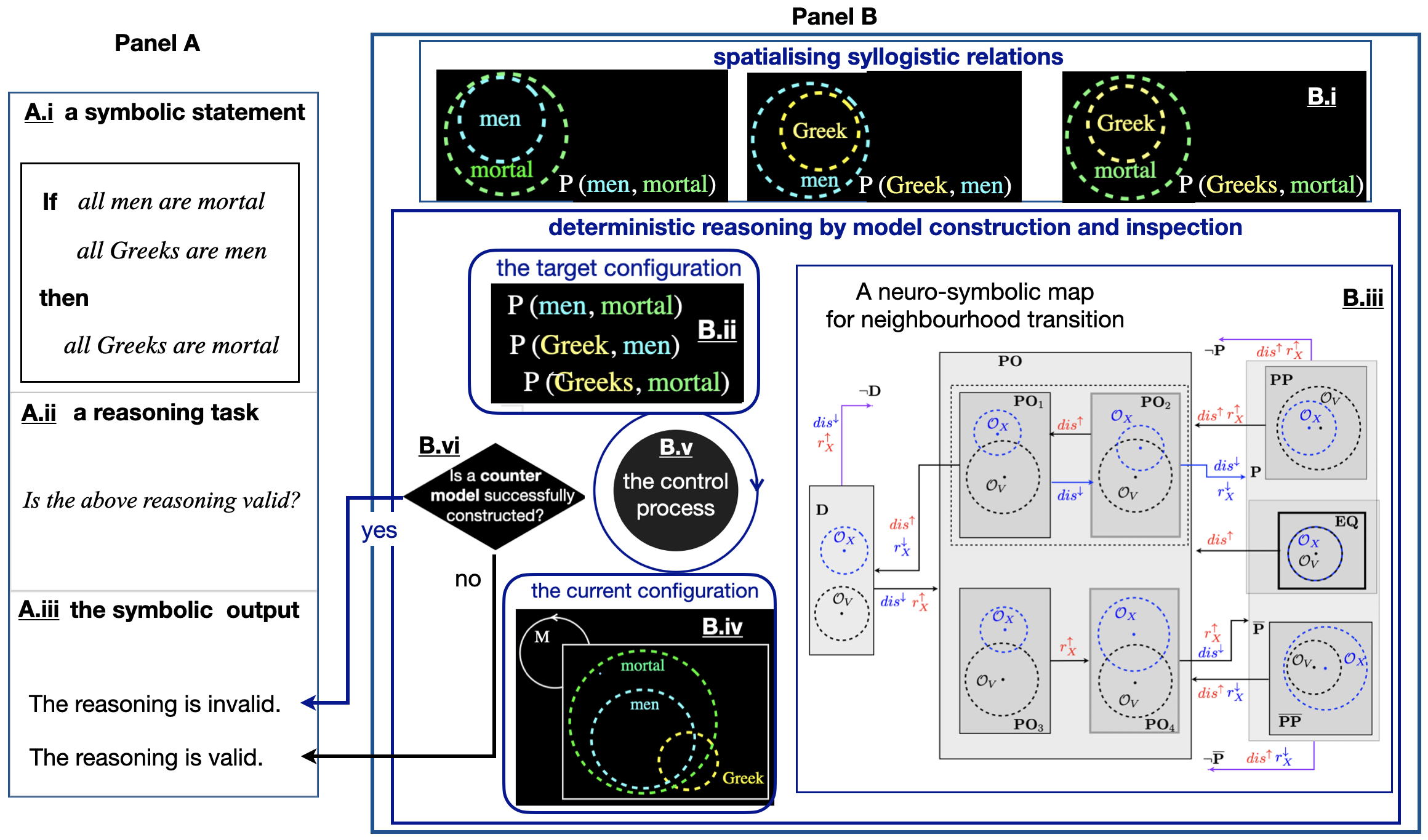}
\caption{Deterministic neural reasoning through model construction. Panel A delivers a symbolic reasoning task (A.i and A.ii) to our $\mathcal{S}$phNN, and receives results from it (A.iii). Panel B outlines the main components of $\mathcal{S}$phNN: (B.i) a module that spatialises symbolic relations;  (B.ii) the target spatial configuration for the reasoning task; (B.iii) a neuro-symbolic map for the transition between neighbourhood spatial relations; (B.iv) a blackboard where a configuration of spheres is constructed; (B.v) a control process that initialises a sphere configuration, and repeatedly transforms the current configuration towards the target, guided by the transition map; (B.vi) after the construction process, a conclusion is made through the inspecting the constructed model.}
\label{valid_enn}
\end{sidewaysfigure} 
$\mathcal{S}$phNN firstly spatialises each statement into a spatial relation between spheres, Figure~\ref{valid_enn} (B.i). To decide the validity, $\mathcal{S}$phNN negates the conclusion and tries to show that the following sphere configuration does not exist: {\em the men sphere is inside the mortal sphere; the Greek sphere is inside the men sphere; the Greek sphere is not inside the mortal sphere}, Figure~\ref{valid_enn} (A.ii, B.ii). Its reasoning process consists of three control processes that synergistically transform sphere configurations, Figure~\ref{visual_intro}(A). It starts with initialising a {\em men} sphere, a {\em Greek} sphere, and a {\em mortal} sphere and then transforms the current sphere configuration to the neighbourhood configuration towards the target. This is guided by a neuro-symbolic  transition map of qualitative spatial relations, Figure~\ref{valid_enn} (B.iii-B.v). 
A neural reasoner is {\em deterministic} for the classic syllogism (syllogistic reasoning with three statements) if it has the property as follows.
   
\begin{quote}
{\em
For any three {\em satisfiable} syllogistic statements, there is \underline{a constant number $M$}, a neural reasoner can construct an Euler diagram for these statements in vector space \underline{at the global loss of zero} \underline{within $M$ epochs}. 
}
\end{quote}

For $\mathcal{S}$phNN we prove that $M$ exists, and $M = 1$.  With this {\em deterministic}  property, $\mathcal{S}$phNN inspects the constructed configuration after the first epoch, Figure~\ref{valid_enn} (B.vi). If it is not a target configuration, $\mathcal{S}$phNN will conclude the three statements are unsatisfiable. This proves the validity of the original reasoning, Figure~\ref{valid_enn} (A.iii). This {\em deterministic} property can be extended to long-chained syllogistic reasoning: $\mathcal{S}$phNN can determine the validity (the satisfiability) of any long-chained syllogistic reasoning in one epoch, with the computational complexity of ${O}(N)$ (where $N$ is the length of the chain), as shown in Figure~\ref{visual_intro}(B). 

\paragraph{{\bf 
$\mathcal{S}$phNN is slow, but determinate}} 

Besides theoretical proofs in Section~\ref{proofs}, we demonstrated in Experiment~\ref{exp2} that in the first epoch, $\mathcal{S}$phNN correctly determines all 24 valid syllogistic reasoning among 256 possible syllogistic deductions and successfully constructed a counter-example for each invalid syllogistic reasoning.  
In Experiment~\ref{exp3}, $\mathcal{S}$phNN is compared with ChatGPT, to determine the validity of syllogistic reasoning. Although slower, $\mathcal{S}$phNN correctly determined all 240 long-chained (ranging from 3 to 12 terms) valid syllogistic reasoning among 1200 candidates without a time limit. In contrast, ChatGPT (gpt-3.5-turbo) is faster, whose response time is almost irrelevant to the length of syllogistic reasoning, achieves maximum $80\%$ accuracy for classic (atomic) syllogistic reasoning (with three terms), and drops to $51.67\%$ for syllogism with 12 terms. The performance of ChatGPT (gpt-4o) is stable, ranging from 89.17\% to 95.83\%. ChatGPT (gpt-3.5-turbo and gpt-4o) might give conclusions inconsistent with its descriptions, as shown in Figure~\ref{visual_intro}(B). {\bf $\mathcal{S}$phNN can identify such inconsistency through constructing a model described by ChatGPT}.  Experiment~\ref{exp4} shows that $\mathcal{S}$phNN can inform ChatGPT of the consistency information through prompt engineering. Although ChatGPT (gpt-3.5-turbo and gpt-4o) might neglect such feedback, $\mathcal{S}$phNN still helped ChatGPT (gpt-3.5-turbo) improve the accuracy from 47.27\% to 56.64\% and ChatGPT (gpt-4o) improve the accuracy from 77.39\% to 80.01\% in deciding the satisfiability of atomic syllogistic reasoning (3-statement syllogism). The communication between ChatGPT and $\mathcal{S}$phNN demonstrates a micro neural world that mirrors the synergistic collaboration between System 1 and System 2 of the mind: the former proposes candidate solutions using fast associative thinking; the latter slowly examines the correctness by following rules (here, constructing models) \citep{Kahneman11}, as shown in Figure~\ref{visual_intro}(C).  

\paragraph{{\bf $\mathcal{S}$phNN has the representational capacity for neuro-symbolic unification}} 

 
Vector embeddings learned from traditional neural networks can serve as {\em content addressable memory} \cite{Smolensky88, Hunter88}, which means that concepts with similar meanings are represented by vectors close to each other. A sphere configuration may extend the representation power of {\em content addressable memory} as follows: all sphere centres represent traditional {\em content addressable memory}, and the boundary relations introduce explicit and precise set-theoretic relations among concepts. 
This way, $\mathcal{S}$phNN can evolve into unified neuro-symbolic models that simulate both System 1 (using sphere centres) and System 2 (using boundary relations). Experiment~\ref{exp5} shows the existence of this unification by successfully {\em extending} pre-trained vectors of GLOVE, BERT, and ChatGPT into spheres whose boundary relations precisely represent set-theoretic relations between syllogistic statements, as shown in Figure~\ref{visual_intro}(D).  This unified representation will allow SphNNs to directly update vector embeddings of LLMs, instead of using external prompt communication that LLMs might ignore (as observed in Experiment~\ref{exp4}). This provides a new way to work around open problems in prompt engineering \cite{prompt_survey23}. 

\paragraph{{\bf $\mathcal{S}$phNN evolves into Herbert A. Simon's neural scissors}}

We systematically show how $\mathcal{S}$phNN can evolve into various types of representation and reasoning that serve as two neural blades of Herbert A. Simon's scissors, as shown in Figure~\ref{visual_intro}(E). The context blade represents task environments constructed by spheres. The evolution starts with spatiotemporal structure, moves onto event structure, and then arrives at neurosymbolic unification that can host latent feature vectors into sphere centres and lands at Descartes's product of sphere configurations for heterogeneous knowledge. 
The rational reasoning blade represents computational procedures that select parts of sphere configuration, transform them into the target, and decide the stop criterion. Along with the evolution, various computational procedures can be developed to simulate logic reasoning, Bayesian reasoning,  heuristic reasoning, reasoning with LLM results and humour (the highest level of cognition \cite{joke2020}).  

%% file: 9arxiv/methodology.tex
\section{The methodology: 
bounded rational reasoning
through the construction of sphere configurations}

\begin{quote} 
{\em 
The representation of space and
of things in space will necessarily be a central topic in a science of design.
    
    \hfill --- Herbert A. Simon  \cite{Simon19}
     } 
\end{quote}

\subsection{Bounded rationality}

In his seminal Ph.D thesis {\em Administrative Behaviour: A Study of Decision-Making Processes in Administrative Organisation}, Herbert A. Simon coined the term {\em bounded rationality} with two main tenets as follows: (1) humans only have bounded rationality, bounded by the available information given by problems, by different individual capacities, and by the situation where the problem is being solved, e.g. under time pressure, without auxiliary computing devices, and (2) consequently, people are prone to switch to subgoals and seek satisfactory, instead of optimal solutions \cite{Simon19}. Considering the intractable number of chess positions, Herbert A. Simon used the game of chess as a nice example to illustrate the necessity of developing computationally efficient methods. 
A solution is ``good enough'' in his terms, if it has a search procedure and a stopping criterion and tells how information is integrated to make a decision. This raises two open questions\footnote{\url{https://plato.stanford.edu/entries/bounded-rationality/}} as follows. 
\begin{itemize}
    \item How do human beings actually make decisions ``in the wild''?
    \item How can the standard theories of global rationality be simplified to render them more tractable?
\end{itemize}
In recent years, deep learning neural networks have successfully solved problems in various fields and won Go against world champions \cite{alphaGo2017, alphaGo2020}. This is primarily due to high-performance GPUs and significant memories and storage, which allow neural networks to learn heuristic patterns from almost all data on this planet. This puts Herbert A. Simon's {\em bounded rationality} into a new situation, where rationality is no longer bounded by learning resources and computing power. Though
 LLMs even demonstrate human-like rationality in question-answering, they remain unexplainable and have limited capabilities in doing simple logical reasoning \cite{Mitchell_science.adj5957,chatgpt_nature2023}. Following Herbert A. Simon's slogan ``problem solving as change in representation'' \cite{Simon19}, rationality in the era of neural computing is bounded by representations used for decision-making, namely, vectorial or symbolic. Heuristic rationalities obtained by vectorial neural computing do not have the deterministic rationality of symbolic logical reasoning. We introduce a sub-inquiry of Herbert A. Simon's second inquiry, as follows.    
\begin{itemize}
\item How can the determinacy of standard theories of global rationality be tractably rendered into the process of a heuristic inference and form the end pole of the continuum of various reasoning processes, when information is steadily accumulating and becomes sufficient; thus, uncertain inference problems become certain?
\end{itemize}
Observing the success of deep learning neural computing, here we consider two basic questions as follows.
\begin{itemize}
    \item How can deterministic syllogistic reasoning be tractably rendered into neural computing?  
   \item How can the above neural computing serve as a common place to evolve and integrate three major decision-making methods, namely, neural (heuristic), probabilistic, and logical? 
\end{itemize}
To answer the two questions, we need to systematically develop semantics for syllogistic reasoning and statistical inference in the vector space and show that they are consistent with the current vector semantics of traditional deep-learning neural networks. We will provide our solution in Section~\ref{sphere_power}.  

\subsection{Why do we focus on syllogistic reasoning?}

\begin{quote}
    {\em The relevance of the whole battery of Aristotelian syllogisms to psychology is, we are tempted to quip, equally mysterious.
\begin{flushright} --- Mercier and Sperber  \cite{enigma2018}
        \end{flushright}
        }
\end{quote}

Scientists prefer to taking microcosms to disclose complex phenomena and ground fundamental theories \cite{Khemlani2021}. 
Due to its simplicity, synthetic reasoning is used as the microcosm by psychologists to explore the rationality of the mind \cite{Khemlani2021}. In over one hundred years of research, various psychological theories have been proposed; some account for it as a heuristic phenomenon, and some account for it as deliberative reasoning processes with formal rules or diagrams. The mental model theory is the most promising but still can not account for all the aspects. Modern theories are developed as unified theories by integrating several accounts, e.g., mReasoner \cite{Khemlani2021}. Despite these, developing neural models for syllogism was extremely hard and considered in the psychological community to be utopian ten years ago. In recent years, LLMs, e.g. ChatGPT, demonstrate human-like performance in question-answering, including syllogistic reasoning questions. Although trained by almost all the datasets on this planet, LLMs, unfortunately, have not reached the determinacy of syllogistic reasoning. On the other hand, syllogistic reasoning is easily solved in mathematical logic \cite{VukmirovicBCS19,BentkampBTV21}. Therefore, we focus on syllogistic reasoning and stand at the logic perspective to explore novel neural computing that can achieve deterministic syllogistic reasoning and show how this novel neural computing can develop various kinds of rational reasoning.  

\subsection{Spheres as the  building blocks for knowledge representation and neural computing}
\label{sphere}
 
The research about mental models can be traced back to Tolman's experiments with rats in 1948 \cite{Tolman48}. Early psychological research found that mental spatial objects are grouped into ``regions'', and their hierarchical structures are represented as nested regions where relations among siblings can be explicitly represented \citep{Davis81,McDermott81,StevensCoupe78,McNamara86}. Advances in neuroscience 
indicated that mental spatial models 
represent cognitive spaces for domain-general tasks, representing properties and concepts as convex regions constrained by geometric features \citep{Bellmundeaat6766}. 

Though the tradition of geometry uses the imperceptible abstract concept {\em point} as the primitive to develop other concepts, e.g., lines, triangles, circles, cones, and spheres, spatial knowledge can also be developed by utilising perceptible concrete {\em regions} \cite{Laguna22a,whitehead29,Clarke85,Cohn93,DongJPL}, governed by the connection relation with three features\footnote{The main literature of qualitative spatial representation assumes that two regions should be connected if they satisfy feature (i) and feature (ii), e.g., \cite{Randell92a,Clarke85,CohnRenz07}, which is unfortunately insufficient and will introduce issues in the theory and applications \cite{DongJPL,GaryACM15}. A counter-example will be the relation ``region A is less than one meter away from region B'' -- they satisfy both features but may be disconnected. Feature (iii) is the characteristic feature of the connectedness relation and can be formalised \cite{DongJPL}.}: (i) any region connects with itself (reflexive), (ii) if region A connects with region B, then region B connects with region A (symmetric), and (iii) if two regions are connected, any third region can be moved to a place where it connects with the first two regions \cite{DongJPL}. The three features can be axiomatised to develop various qualitative relations (topology, distance, and orientation) and abstract concepts, such as {\em point} \cite{DongJPL}. 

Events are four-dimensional entities and can be understood in the same way as knowledge of extended objects \cite{Quine1985}. They have parts (in space and time) and taxonomies. Time is more natural and straightforward to be represented by intervals than by points \cite{Dowty79,Allen83,Benthem83}. Temporal relationships among events develop causal relations \cite{ZacksTversky01}. In this way, the connection relation and regions can be the building blocks for representing events, times, and causalities. 

Vectors can be understood as spheres evolved by shrinking the radii to zero. So, traditional neural networks can be understood as being evolved from more primitive {\em neural networks} that use {\em spheres} as the computational building block. Such primitive {\em neural networks} represent and reason with spatial and non-spatial knowledge, simulating the way humans do before they learn abstract concepts, such as {\em points} and {\em vectors}. We name such primitive neural networks {\em Sphere Neural Networks} (SphNNs).

\subsection{Unified representation for heuristic and deliberative reasoning}
\label{unified_reasoning}

\begin{quote}
    {\em The now dominant view of reasoning (``dual process'' or ``fast and slow thinking''), however appealing, is but a makeshift construction amid the ruins of old ideas. 
    
    \hfill --- Mercier and Sperber \cite{enigma2018}
    }
\end{quote}

How people reason is a central topic in the research of human rationality. Over one hundred years, eminent psychologists chose Aristotelian syllogistic reasoning as a microcosm to investigate human rationality. 
Among 12 competing theories for syllogistic reasoning, none provides an adequate account \cite{laird2012}. Still, three sorts are promising: heuristic theories, deliberative theories with rules, and deliberative theories with set-theoretic diagrams or models \cite{laird2012}. Heuristic reasoning is {\em fast and shallow} and often occurs before {\em slow and deep} deliberative reasoning that validates or refutes heuristic results. Falsifying heuristic results may generate new heuristics. More general accounts shall operate heuristic and deliberative reasoning in parallel and interact with each other \cite{EVANS2003,Kahneman11,Khemlani2021}.

\begin{figure}[!t] 
\centering \includegraphics[width=1\linewidth]{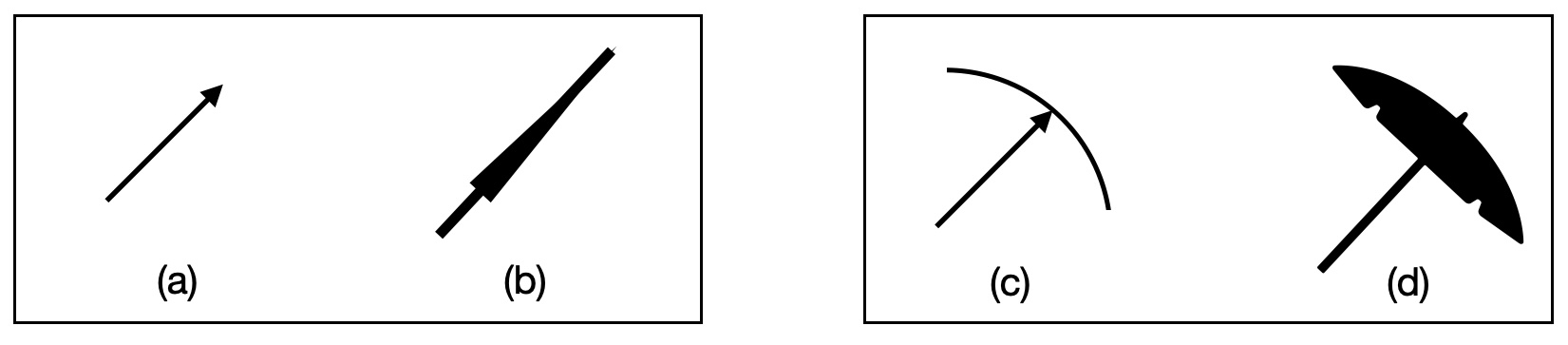} 
\caption{(a) a vector; (b) a closed umbrella; (c) an arc with its centre vector; (d) an open umbrella.}
\label{umbrella}
\end{figure} 

Heuristic is usually simulated by neural networks or the Bayesian rule.  
We will illustrate in Section~\ref{sphere_power} that the Bayesian rule and disjunctive syllogistic reasoning (a kind of deliberative reasoning) can be introduced into neural computing through evolving a vector embedding into an arc embedding -- somehow, like opening an umbrella -- a vector like a closed umbrella, an arc like an open umbrella (see Figure~\ref{umbrella}).  
In this way, heuristic and deliberative reasoning can be carried out by a unified representation, which will realise the interaction and paralleling of different styles of reasoning. The main challenge is to develop a family of neural operations on spheres to realise deliberative reasoning and to prove that these operations indeed reach the determinacy of deliberative reasoning. Finally, we need to explain the relationship between our novel neural proof and the classic symbolic proof, as did in diagrammatic reasoning \cite{Jamnik99}. 

\subsection{Deliberative reasoning through model construction and inspection}
\label{reasoning_via_model_construction}

Deliberative reasoning by model construction and inspection is best supported by empirical experiments \cite{LairdByrne91,knauf03,GoodwinLaird05,Knauff09,Knauff13} -- {\em 
What we have is a procedure to represent and integrate in our mind the content of premises by means of models comparable to schematic pictures of the situation. We then read the conclusions of these models} \cite{enigma2018}. 
In the standard model theory, the reasoning is carried out as a process of {\em model construction, model inspection, and model variation}  \cite{LairdByrne91}. In the {\em model variation} phase, people try to construct alternative models to refute the conclusion. 
If a counter-model is found (the premises are true, and the conclusion is false), the conclusion will be proven to be {\em invalid}. The preferred mental model theory argues that people construct a preferred and simplified model in mind, in a deterministic manner, while ignoring other possibilities \cite{RagniMarkus13,Knauff13} -- The construction of the first model shall not be a stochastic process that {\em produces one model this time and another the next time} \cite[p.563-564]{RagniMarkus13}, the next model will be revised following the principle of minimal changes from the current one \cite{Harman1986,Gardenfors1988,GAeRDENFORS90,knauffetc13}, and generated by a local transformation of the current model, whose similarity is measured by a neighbourhood graph \cite{Freksa91,GoodayCohn94}. 

\begin{figure}[!t] 
\centering \includegraphics[width=1\linewidth]{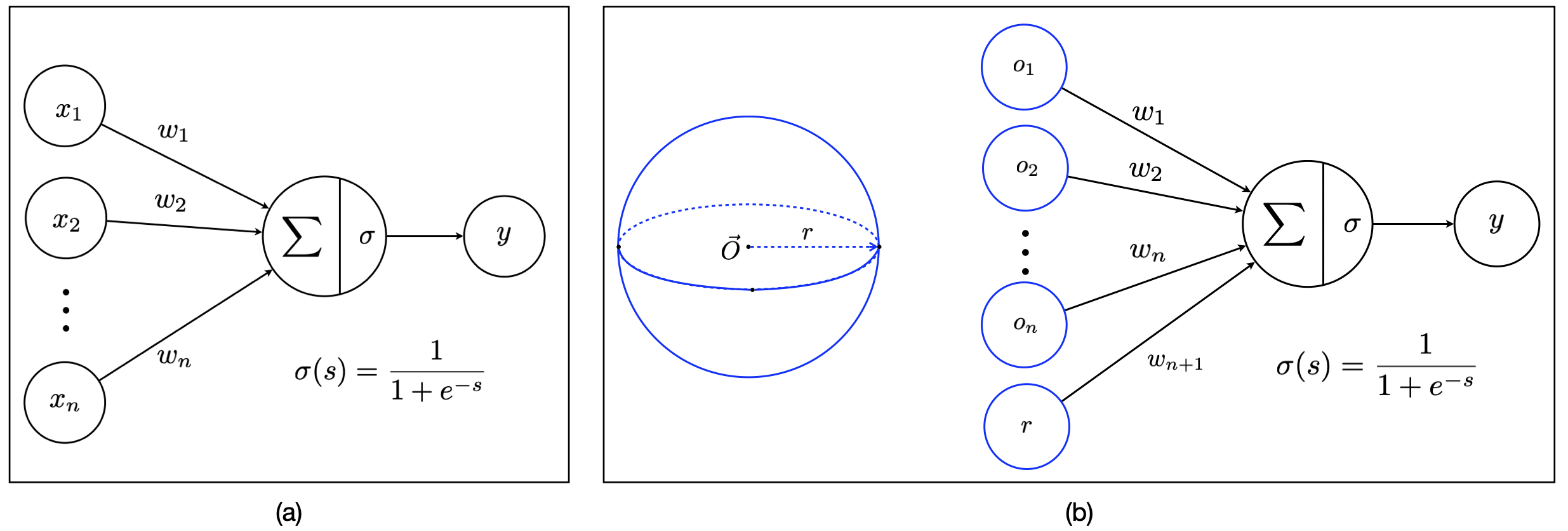} 
\caption{(a) the input of a traditional perceptron is a vector $\vec{x}=[x_1\dots x_n]$; (b) the input of a diameter-limited perceptron is restricted inside a sphere with the centre $\vec{O}$ and the radius $r$.}
\label{fsphere1}
\end{figure} 

\subsection{{\em Sphere Neural Networks} simulate mental model construction}

Mental models represent assertions as a set of observed or imaged possibilities that mirror what they represent \cite{laird83}. These representations are iconic and discrete \cite{Khemlani2021}. Like {\em tips of the iceberg}, they are supported by the part under the water, which are neural routines \cite{Bengio22}. We represent mental models as discrete configurations of spheres and seek a family of habitual neural routines that manipulate these spheres. 

Neurons can be distinguished through their output encodings \cite{Ballard86}: one type of neuron outputs {\em frequency} signals, for example, oculomotor neurons that control eye movements. Their outputs are one-dimensional,  characterized as {\em integration} devices, and simulated by {\em perceptron} \cite{minsky88,Anderson95}. A specific {\em perceptron} is the {\em diameter-limited perceptron} whose input signals $x_i$ are restricted within a certain fixed diameter \cite[p.12]{minsky88}: Each $x_1,\dots,x_n$ is located within a sphere, as shown in Figure~\ref{fsphere1}(a, b). 
We define the input of a Sphere Neural Network as the input domain of a {\em diameter-limited perceptron} $[o_1, \dots, o_n, r]$, where $[o_1, \dots, o_n]$ is the central vector $\vec{O}$ and $r \ge 0$ is the radius, that is, $\|x_i-\vec{O}\|\le r$, where $1\le i \le n$. If $r=0$, it degrades into a {\em random perceptron} \cite[p.12]{minsky88}.
\begin{figure}[!t] 
\centering \includegraphics[width=1\linewidth]{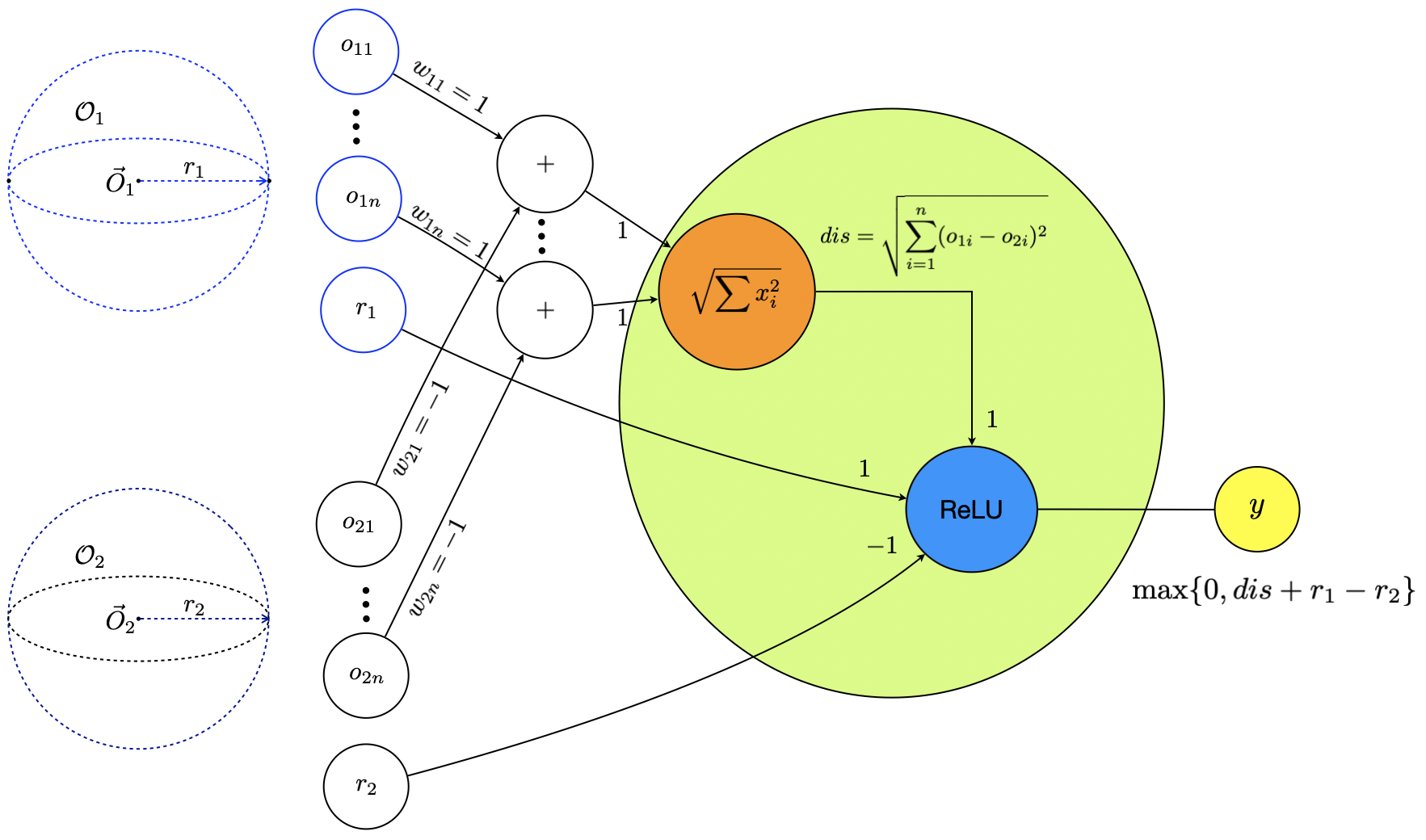} \caption{The inputs of the neural network are two spheres, $o_{11}\dots o_{1n}, r_1$ and $o_{21}\dots o_{2n}, r_2$, respectively, each is represented by its centre and its radius. The network computes the distance between their centres $dis = \sqrt{\sum_{i=1}^n (o_{1i}-o_{2i})^2}$. The output of the network is the value of $\max\{0, dis + r_1 - r_2\}$, which equals 0 when $\mathcal{O}_1$ is inside $\mathcal{O}_2$, and greater than 0, if not.}
\label{fsphere2}
\end{figure} 
The second type of neuron outputs {\em spatial} or {\em place}  encoding, e.g., neurons in the visual cortex \cite{Ballard86}. The output shows the degree of the matching between the input stimuli and the receptive field and is simulated by non-monotonic activation functions. We may represent the input stimuli and the receptive field as two spheres $\mathcal{O}_1$ and $\mathcal{O}_2$, with the non-monotonic activation function $f(\mathcal{O}_1,\mathcal{O}_2)\triangleq \max\{0, \|\vec{O}_1-\vec{O}_2\| + r_1 - r_2\}$, as illustrated in Figure~\ref{fsphere2}. When $f(\mathcal{O}_1,\mathcal{O}_2)=0$, $\mathcal{O}_1$ is inside $\mathcal{O}_2$; the larger the value of $f(\mathcal{O}_1,\mathcal{O}_2)$ is, the further away is $\mathcal{O}_1$ from $\mathcal{O}_2$. This way, Sphere Neural Networks can explicitly signal spatial relations between two spheres, e.g., {\em inside, partial overlapping}. 

\begin{figure}[!t] 
\centering \includegraphics[width=1\linewidth]{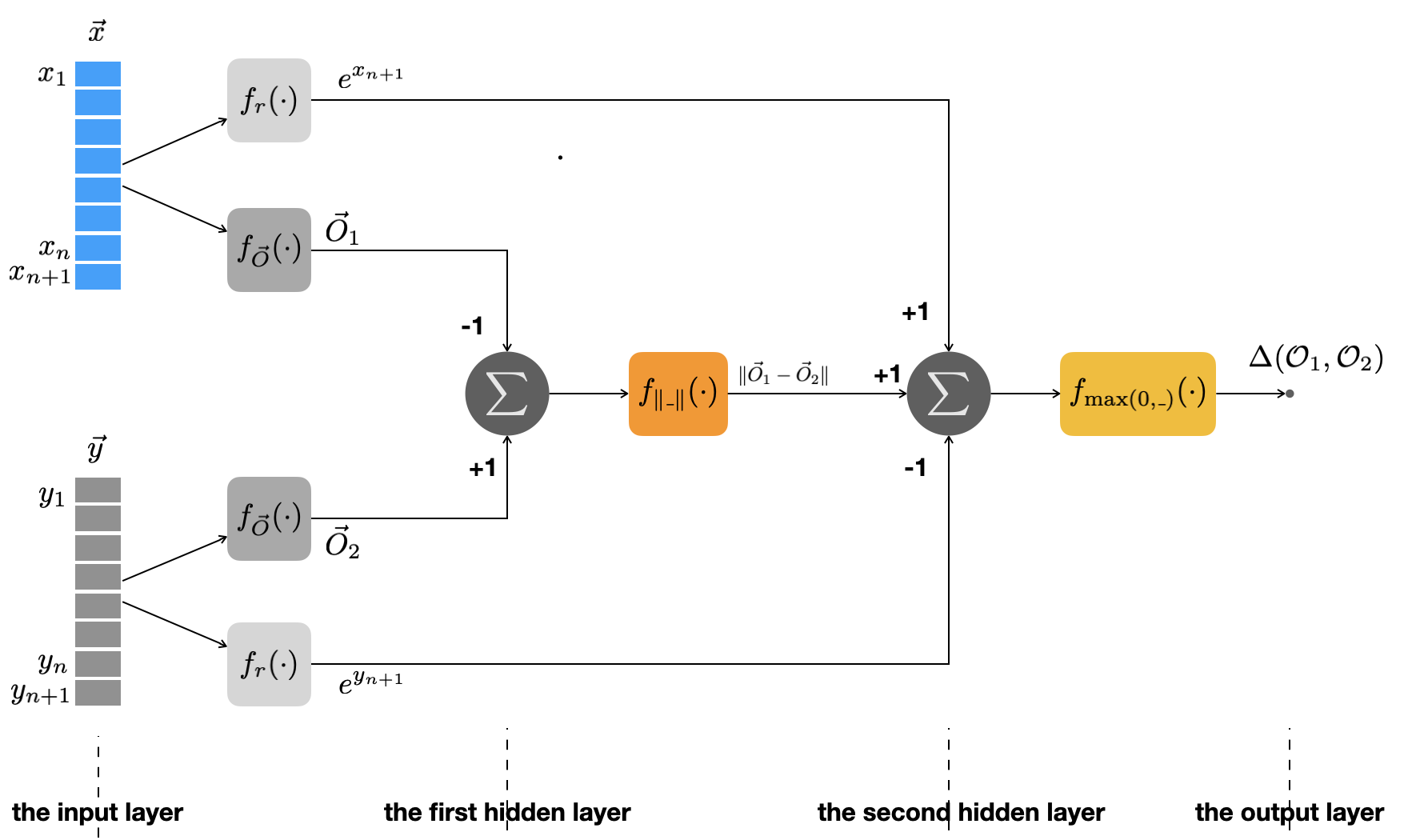} \caption{The Kolmogorov-Arnold neural architecture of $\Delta(\mathcal{O}_1,\mathcal{O}_2)\triangleq \max(0, \|\vec{O}_1-\vec{O}_2\| + r_1 - r_2)$. Given two $n+1$-dimensional vectors $\vec{x}$ and $\vec{y}$ representing $n$-dimensional spheres, the $f_r(\cdot)$ selects the $n+1^{th}$ element $x_{n+1}$ and $y_{n+1}$, and returns the radius $e^{x_{n+1}}$ and $e^{y_{n+1}}$ of $\mathcal{O}_1$ and $\mathcal{O}_2$, respectively; $f_{\vec{O}}(\cdot)$ selects the first $n$ elements as the centre of a sphere; $f_{\|\_\|}(\cdot)$ computes the Euclidean norm of a vector; the output of the first hidden layer is $\vec{O}_1-\vec{O}_2$; the output of the second hidden layer is $\|\vec{O}_1-\vec{O}_2\| + r_1 - r_2$; the final output is this network is zero, if $\mathcal{O}_1$ is inside $\mathcal{O}_2$, otherwise the output is greater than zero.}
\label{kan_fsphere}
\end{figure}

The non-monotonic activation function $\Delta(\mathcal{O}_1,\mathcal{O}_2)\triangleq \max(0, \|\vec{O}_1-\vec{O}_2\| + r_1 - r_2)$ is a deviation of the non-monotonic activations of traditional neural networks and better described as a kind of Kolmogorov–Arnold Networks (KANs) \cite{liu2024kan}, whose learnable activation functions are on edges, as shown in Figure~\ref{kan_fsphere}.

\subsection{What is $\mathcal{S}$phNN about, and not about?}
\label{spatial2nonspatial}
 
\begin{quote}
    {\em Reason is not a superpower implausibly grafted onto an animal mind; it is, rather, a well integrated component of the extraordinarily developed mind that characterizes the human animal.
    
       \hfill --- Mercier and Sperber \cite{enigma2018}
   
    }
\end{quote}


$\mathcal{S}$phNN is the first neural model to determine the validity ({\em valid} or {\em invalid}) and the satisfiability ({\em satisfiable} or {\em unsatisfiable}) of syllogistic reasoning. It only constructs configurations of spheres whose radii are greater than zero and does not construct abstract concepts, such as {\em empty sphere} \cite{Renz02} or {\em point} \cite{Smith96}. Its control processes suffice to determine the satisfiability of long-chained syllogistic reasoning in the psychological literature \cite{laird2012,laird2015}, which is slightly different from the usual description in the logic textbooks \cite{laird2012}. A syllogistic reasoning being {\em valid} means that the {\em conclusion is true in every case in which all its premises are true} \cite[p.1]{jeffrey81}:  
that $r_1(X,\, Y)\, r_2(Y,\, Z)\,\therefore r_3(X,\, Z)$ is {\em valid} means that $r_3(X,\, Z)$ is true in every case in which $r_1(X,\, Y)$ and $r_2(Y,\, Z)$ are true. 
$\mathcal{S}$phNN can evolve into a variety of 
rational reasoning, 
and can seamlessly integrate latent feature vectors learned by  traditional neural networks. 



%% file: 9arxiv/syllogism_1.tex
\section{Spatialising syllogistic statements in the vector space}
\label{spatialisation}

\begin{quote} 
{\em 
Spatial thinking is the foundation of abstract thought.   

    \hfill --- Barbara Tversky  \cite{tversky19}
     } 
\end{quote}

\begin{figure}
\centering  
  \centering 
\includegraphics[width=0.8\textwidth]{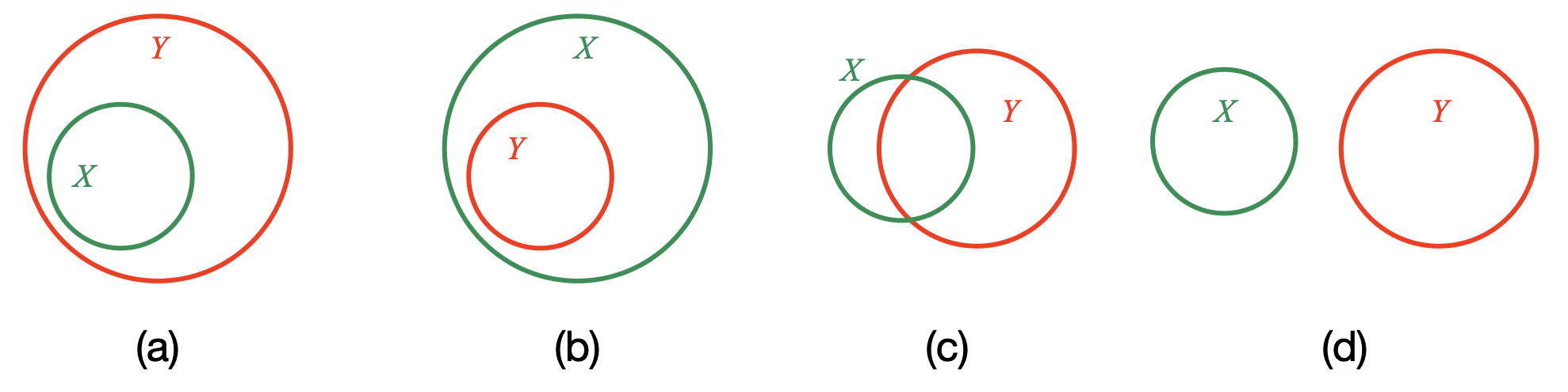}  
\caption{Basic diagrammatic representation for syllogistic statements. (1) that {\em all X are Y} is represented by (a) $X\subset Y$; (2) that {\em some X are Y} is represented by (a) $X\subset Y$ or (b) $Y\subset X$ or (c) $X\cap Y\neq\emptyset$; (3) that {\em no X are Y} is represented by (d) $X\cap Y=\emptyset$; (4) that {\em some X are not Y} is represented by (b) $Y\subset X$ or (c) $X\cap Y\neq\emptyset$ or (d) $X\cap Y=\emptyset$. 
}
\label{euler_diagram}
\end{figure}

\begin{figure}
\centering 
\includegraphics[width=1.\textwidth]{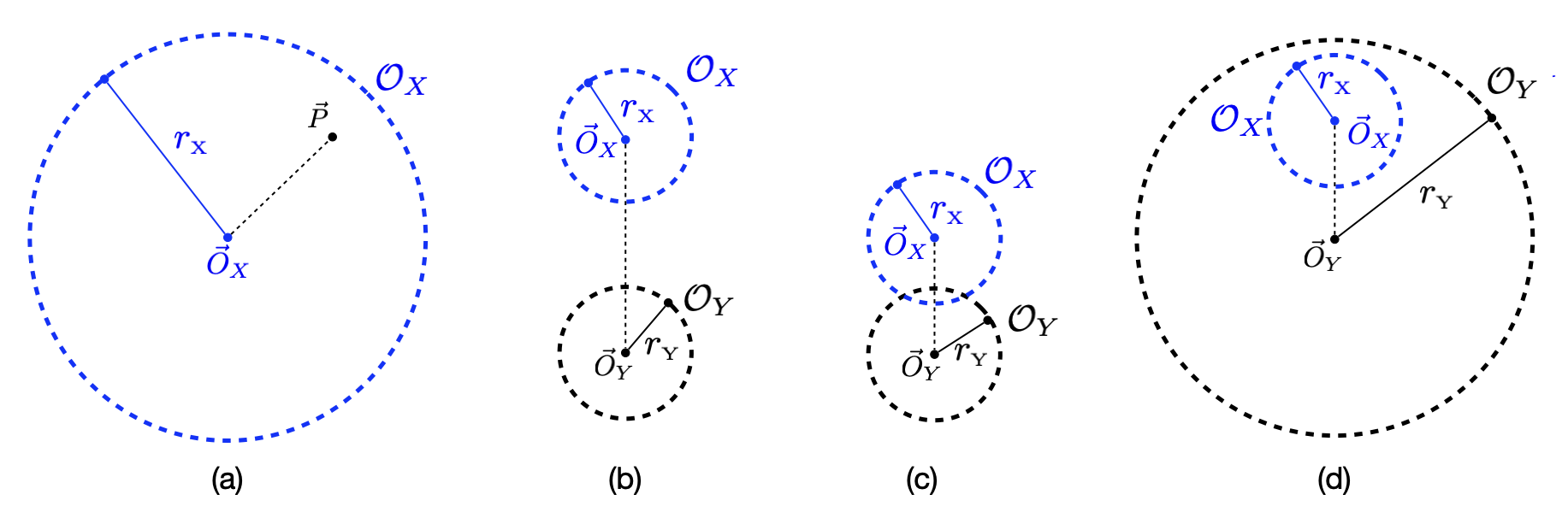}
\caption{(a) Sphere $\mathcal{O}_X$ is an open sphere; (b) Sphere $\mathcal{O}_X$ disconnects from Sphere $\mathcal{O}_Y$;  (c) Sphere $\mathcal{O}_X$ partially overlaps with Sphere $\mathcal{O}_Y$; (d) Sphere $\mathcal{O}_X$ is part of Sphere $\mathcal{O}_Y$.}
\label{sphdef}
\end{figure}
\subsection{Spatialising syllogistic statements}
\label{spatialising1}

Classic syllogistic reasoning \cite{jeffrey81} is a form of deductive reasoning with only two premises, three terms, and four possible relations (or ``moods'' used in the psychological literature): (1) {\em universal affirmative}: all $X$ are $Y$, $all(X,\, Y)$; (2) {\em particular affirmative}: some $X$ are $Y$, $some(X,\, Y)$; (3) {\em universal negative}: no $X$ are $Y$, $no(X,\, Y)$; (4) {\em particular negative}: some $X$ are not $Y$, $some\_not(X,\, Y)$.
Moods can be reduced to four basic set relations in the forms of Euler diagrams \cite{hammer98}: (a) $X$ is part of $Y$ ($X\subset Y$), (b) $X$ contains $Y$ ($Y\subset X$), (c) $X$ partially overlaps with $Y$ ($X\cap Y\neq \emptyset$), and (d) $X$ is disjoint from $Y$ ($X\cap Y=\emptyset$), as shown in Figure~\ref{euler_diagram}. If we fix moods and allow terms to exchange places in the premises, there will be four kinds of syllogistic reasoning, and enumerating moods will total 256 different syllogistic reasoning types.

\begin{eqnarray*}
 r_1(X, Y)\quad r_2(Y, Z)\quad\therefore r_3(X, Z)\\ 
 r_1(Y, X)\quad r_2(Y, Z)\quad\therefore r_3(X, Z)\\
 r_1(X, Y)\quad r_2(Z, Y)\quad\therefore r_3(X, Z)\\
 r_1(Y, X)\quad r_2(Z, Y)\quad\therefore r_3(X, Z)
\end{eqnarray*}\noindent where  $r_i\in\{all, some, no, some\_not \}$ ($i=1,2,3$).  
We map each syllogistic relation to spatial relations, define them in vector space, and use neural operations to update them. Concretely,
we ground a syllogistic term $X$ to an $n$-dimensional sphere $\mathcal{O}_X$, with the central vector $\vec{O}_X=[x_1, \dots, x_n]$, and the radius $r_X =  e^{x_{n+1}}$. Sphere $\mathcal{O}_X$ is defined as open, that is, a point $\vec{P}$ is inside a sphere $\mathcal{O}_X$, if and only if $\|\vec{P}-\vec{O}_X\| < r_X$, as shown in Figure~\ref{sphdef}(a). Sphere $\mathcal{O}_X$ disconnects from sphere $\mathcal{O}_Y$, $\mathbf{D}(\mathcal{O}_X, \mathcal{O}_Y)$, if and only if the distance between their central vectors is greater than or equal to the sum of their radii\footnote{By defining spheres as open, the externally connected relation $\mathbf{EC}$ ($\|\vec{O}_X - \vec{O}_Y\| - (r_X + r_Y) = 0$) in \cite{Randell92a} is included into the $\mathbf{D}$ relation. This way, we do not explicitly introduce the $\mathbf{EC}$ relation. This reduces the number of neighbourhood transitions and makes the proofs of theorems clean.}, as shown in Figure~\ref{sphdef}(b).

\begin{eqnarray*}
  \mathbf{D}(\mathcal{O}_X, \mathcal{O}_Y)\mbox{ holds} 
    , \mbox{ if and only if }  \|\vec{O}_X - \vec{O}_Y\| - (r_X + r_Y) \ge 0.    
\end{eqnarray*}
To reach the disconnectedness relation, we shall either increase the distance $dis_{X,Y}=\|\vec{O}_X - \vec{O}_Y\|$ or decrease the radius, $r_X$, or $r_Y$, or both. 
We define $\mathcal{I}^\mathbf{D}(\mathcal{O}_X, \mathcal{O}_Y)=\max\{0, -dis_{X,Y} + r_X + r_Y\}$, such that $\mathcal{O}_X$ disconnects from $\mathcal{O}_Y$, if and only if $\mathcal{I}^\mathbf{D}(\mathcal{O}_X, \mathcal{O}_Y)= 0$, otherwise, $\mathcal{I}^\mathbf{D}(\mathcal{O}_X, \mathcal{O}_Y)>0$. 
Generally, we design an inspection function $\mathcal{I}^\mathbf{R}(\mathcal{O}_X, \mathcal{O}_Y)$ to inspect whether the relation $\mathbf{R}$ is held between $\mathcal{O}_X$ and $\mathcal{O}_Y$.  It returns zero, if and only if the relation $\mathbf{R}(\mathcal{O}_X, \mathcal{O}_Y)$ is satisfied; otherwise, it returns a positive real number. So, a target configuration is reached when the sum of all inspection functions equals zero. 


$\mathcal{O}_X$ is part of $\mathcal{O}_Y$, $\mathbf{P}(\mathcal{O}_X, \mathcal{O}_Y)$, if and only if the distance between their centres plus $r_X$ is less than or equals to $r_Y$, as illustrated in Figure~\ref{sphdef}(d). $\mathcal{O}_Y$ containing $\mathcal{O}_X$, $\mathbf{\overline{P}}(\mathcal{O}_Y,\mathcal{O}_X)$, is equivalent to $\mathcal{O}_X$ being part of $\mathcal{O}_Y$.
\begin{eqnarray*}
&&\mathbf{P}(\mathcal{O}_X,\mathcal{O}_Y)\mbox{ holds}, \mbox{ if and only if }     \|\vec{O}_X - \vec{O}_Y\| + r_X \le  r_Y
\\
&&\mathbf{\overline{P}}(\mathcal{O}_Y,\, \mathcal{O}_X)\equiv\mathbf{P}(\mathcal{O}_X,\, \mathcal{O}_Y)
\end{eqnarray*}
To reach the relation $\mathbf{P}(\mathcal{O}_X, \mathcal{O}_Y)$, we shall either decrease the distance $dis_{X,Y}$ or decrease $r_X$, or increase $r_Y$. The inspection function $\mathcal{I}^\mathbf{P}(\mathcal{O}_X,\mathcal{O}_Y)\triangleq \max\{0, dis_{X,Y} + r_X - r_Y\}$, and $\mathcal{I}^{\overline{\mathbf{P}}}(\mathcal{O}_X,\mathcal{O}_Y)\triangleq\max\{0, dis_{X,Y}+ r_Y- r_X\}$. 
We call $\mathbf{P}$ and $\overline{\mathbf{P}}$ being {\em transposition}, written as $\overline{\mathbf{P}}$
and $\overline{\overline{\mathbf{P}}}=\mathbf{P}$. In general, the {\em transposition} of $\mathbf{R}(\mathcal{O}_X, \mathcal{O}_Y)$ is to switch the order of the parameters and name the new relation as $\overline{\mathbf{R}}(\mathcal{O}_Y, \mathcal{O}_X)$, e.g., $\overline{\mathbf{D}}(\mathcal{O}_Y, \mathcal{O}_X)=\mathbf{D}(\mathcal{O}_X, \mathcal{O}_Y)$, $\overline{\neg\mathbf{D}}(\mathcal{O}_Y, \mathcal{O}_X)=\neg\mathbf{D}(\mathcal{O}_X, \mathcal{O}_Y)$. $\overline{\overline{\mathbf{R}}}(\mathcal{O}_X, \mathcal{O}_Y)$ is to switch two times the order of $\mathcal{O}_X$ and $\mathcal{O}_Y$, that will be the same as before switching, so, $\overline{\overline{\mathbf{R}}}(\mathcal{O}_X, \mathcal{O}_Y)=\mathbf{R}(\mathcal{O}_X, \mathcal{O}_Y)$. A syllogistic statement can be spatialised to exactly one qualitative spatial relation if the order of two spheres is fixed. This one-to-one mapping prevents the complexity of the model construction process from exploding exponentially. Thus:
 \begin{itemize}
	\item {\em All $X$ are $Y$}, $all(X,\, Y)$, is spatialised to $\mathbf{P}(\mathcal{O}_X,\, \mathcal{O}_Y)$.
\item  {\em Some $X$ are $Y$}, $some(X ,Y)$, is spatialised to $\neg\mathbf{D}(\mathcal{O}_X,\, \mathcal{O}_Y)$.
	\item {\em No $X$ are $Y$}, $no(X,\, Y)$, is spatialised to $\mathbf{D}(\mathcal{O}_X,\, \mathcal{O}_Y)$.
	\item {\em Some $X$ are not $Y$}, $some\_not(X,\, Y)$, is spatialised to $\neg\mathbf{P}(\mathcal{O}_X,\, \mathcal{O}_Y)$.
\end{itemize}
 
We introduce the spatialisation function $\psi$ that 
maps $\{all, some, no, some\_not\}$ to $\{\mathbf{P},\mathbf{D},\neg\mathbf{P},\neg\mathbf{D}\}$, namely, $\psi(all)=\mathbf{P}, \psi(some) = \neg\mathbf{D}, \psi(no)=\mathbf{D}, \psi(some\_not)=\neg\mathbf{P}$. 

\begin{rmk}
Two spheres are coincided $\mathbf{EQ}(\mathcal{O}_X,\mathcal{O}_Y)$, if and only if $\vec{O}_X=\vec{O}_Y$ and $r_X = r_Y$.  
Sphere $\mathcal{O}_X$ is a proper part of Sphere $\mathcal{O}_Y$ $\mathbf{PP}(\mathcal{O}_X,\mathcal{O}_Y)$, if and only if $\mathcal{O}_X$ is part of $\mathcal{O}_Y$ and they are not coincided, that is, $\mathbf{PP}(\mathcal{O}_X,\mathcal{O}_Y)\triangleq\mathbf{P}(\mathcal{O}_X,\mathcal{O}_Y)\land \neg\mathbf{EQ}(\mathcal{O}_X,\mathcal{O}_Y)$. Its transposition relation is written as $\overline{\mathbf{PP}}(\mathcal{O}_X,\mathcal{O}_Y)\triangleq\mathbf{PP}(\mathcal{O}_Y, \mathcal{O}_X)$.
That sphere $\mathcal{O}_X$ partially overlaps with Sphere $\mathcal{O}_Y$ $\mathbf{PO}(\mathcal{O}_X, \mathcal{O}_Y)$, if and only if the distance between their centres is (1) greater than the absolute difference between their radii, and (2) less than the sum of their radii, that is, $\mathbf{PO}(\mathcal{O}_X,\mathcal{O}_Y)\triangleq |r_X - r_Y| <\|\vec{O}_X - \vec{O}_Y\| < r_X + r_Y. 
$
If {\em some $X$ are $Y$}, $\mathcal{O}_X$ can either be proper part of $\mathbf{PP}$, or partially overlap with $\mathbf{PO}$, or equal to $\mathbf{EQ}$, or be {\em transposition} proper part $\overline{\mathbf{PP}}$ of $\mathcal{O}_Y$. We can prove  $\mathbf{PP}(\mathcal{O}_X, \mathcal{O}_Y)\lor\mathbf{PO}(\mathcal{O}_X, \mathcal{O}_Y)\lor\mathbf{EQ}(\mathcal{O}_X, \mathcal{O}_Y)\lor\overline{\mathbf{PP}}(\mathcal{O}_X, \mathcal{O}_Y)$ is equivalent to   $\neg\mathbf{D}(\mathcal{O}_X, \mathcal{O}_Y)$. If {\em some $X$ are not $Y$},  $\mathcal{O}_X$ can either partially overlap with ($\mathbf{PO}$), or be {\em transposition} proper part of ($\overline{\mathbf{PP}}$), or disconnect from ($\mathbf{D}$)  $\mathcal{O}_Y$. We can prove  $\mathbf{PO}(\mathcal{O}_X, \mathcal{O}_Y)\lor\overline{\mathbf{PP}}(\mathcal{O}_X, \mathcal{O}_Y)\lor\mathbf{D}(\mathcal{O}_X, \mathcal{O}_Y)$ is equivalent to  $\neg\mathbf{P}(\mathcal{O}_X, \mathcal{O}_Y)$.


\end{rmk}



\subsection{Syllogistic reasoning through model construction in the vector space}
\label{spatialising2}

With the above spatialisation, we transform the task of syllogistic reasoning into the task of model construction in the vector space as follows. 
A syllogistic reasoning 
\syllogism{
  $r_1(X,Y)$.\\
  $r_2(Y,Z)$. 
}{
  $r_3(X,Z)$.
}is {\em satisfiable}, where $r_i\in\{all, no, some, some\_not\}$, if and only if there are three spheres $\mathcal{O}_X$, $\mathcal{O}_Y$, and $\mathcal{O}_Z$ that satisfy spatial relations as follows: $\psi(r_1)(\mathcal{O}_X,\,\mathcal{O}_Y)$, $\psi(r_2)(\mathcal{O}_Y,\,\mathcal{O}_Z)$, and $\psi(r_3)(\mathcal{O}_X,\,\mathcal{O}_Z)$. This syllogistic reasoning is {\em valid}, if and only if, there are no spheres $\mathcal{O}_X$, $\mathcal{O}_Y$, and $\mathcal{O}_Z$ that satisfy all spatial relations as follows: $\psi(r_1)(\mathcal{O}_X,\,\mathcal{O}_Y)$, $\psi(r_2)(\mathcal{O}_Y,\,\mathcal{O}_Z)$, and $\neg\psi(r_3)(\mathcal{O}_X,\,\mathcal{O}_Z)$. $\mathcal{S}$phNN determines the original reasoning {\em valid}, if after $M$ epochs, it cannot construct a configuration of $\mathcal{O}_{X_1}$, $\mathcal{O}_{X_2}$, and $\mathcal{O}_{X_3}$ satisfying all relations $\psi(r_1)(\mathcal{O}_{X_1},\, \mathcal{O}_{X_2})$, $\psi(r_2)(\mathcal{O}_{X_2},\, \mathcal{O}_{X_3})$, and $\neg\psi(r_3)(\mathcal{O}_{X_1},\, \mathcal{O}_{X_3})$.

%% file: 9arxiv/syllogism.tex
\section{$\mathcal{S}$phNN: A hierarchical GNN}
\label{transition_map}

\begin{quote} 
{\em 
We conceive GPS as moving through
a large maze. The nodes of the maze represent situations, described afferently; the paths joining one node to another are the actions, described as
motor sequences, that will transform the one situation into the other. At
any given moment GPS is always faced with a single question: ``What
action shall I try next?''
    
    \hfill --- Herbert A. Simon  \cite{Simon19}
     } 
\end{quote}

\begin{sidewaysfigure} 
\centering
\includegraphics[width=1\textwidth]{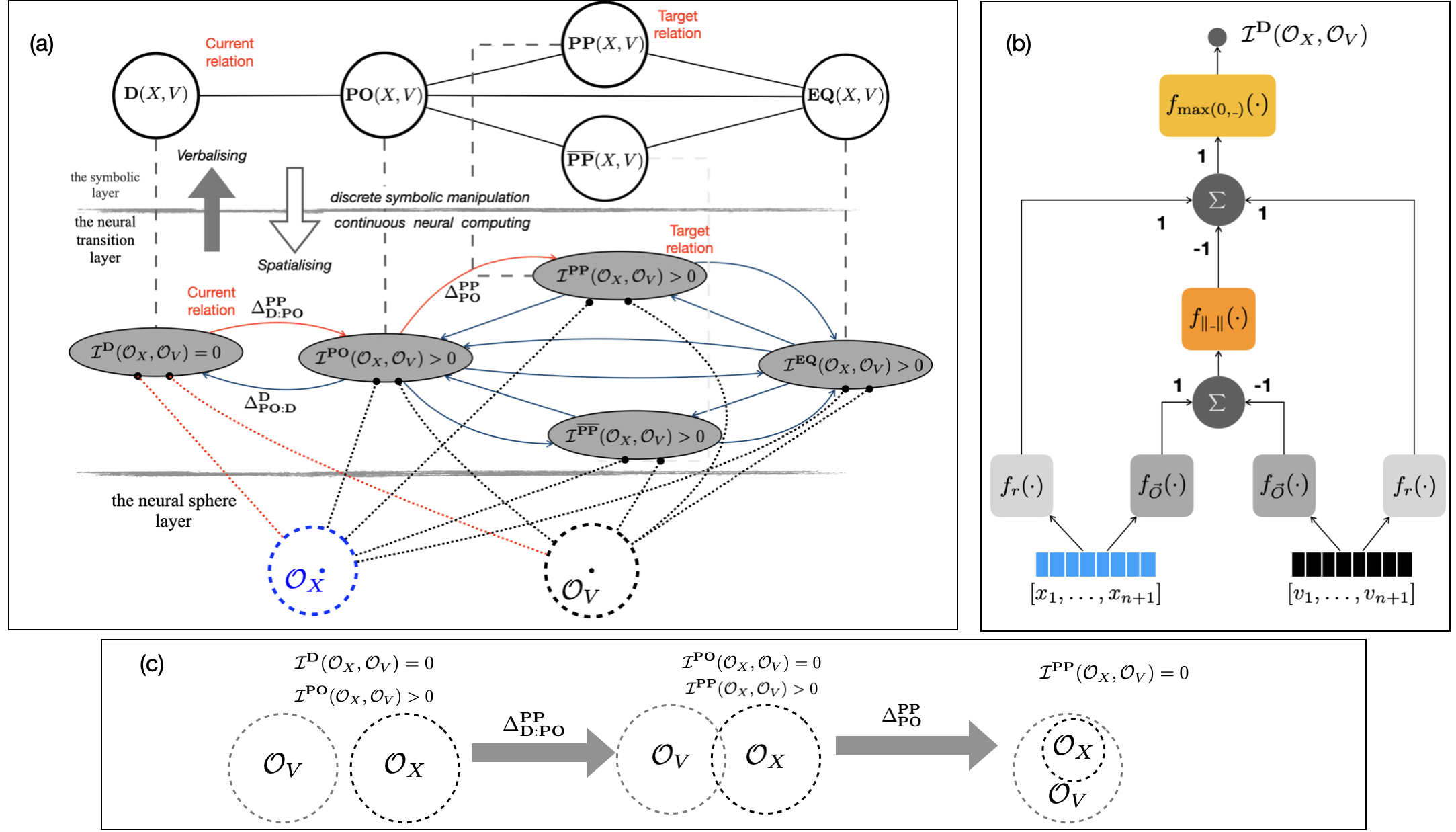}
\caption{(a) An architecture $\mathcal{S}$phNN has three layers. The bottom neural sphere layer is the current sphere configuration in the vector space. The top symbolic layer symbolic graph about neighbourhood spatial relations, e.g., $\mathbf{D}$, $\mathbf{PO}$, $\mathbf{PP}$, $\overline{\mathbf{PP}}$ and $\mathbf{EQ}$. The middle spatial transition layer is a GNN that realises the spatial transformation of the current sphere configuration in the bottom layer to the target configuration defined in the top layer. The GNN has a Kolmogorov-Arnold-styled architecture, in the sense that its directed edge has a gradual descent function that realises neighbourhood transitions by updating a sphere's location and size in the bottom layer. For example, two {\color{red}{red arrows}} are the two-step neighbourhood transitions from $\mathcal{O}_X$ disconnecting from $\mathcal{O}_Y$ to $\mathcal{O}_X$ being part of $\mathcal{O}_Y$. $\Delta_{\mathbf{D}:\mathbf{PO}}^{\mathbf{PP}}(\mathcal{O}_X, \mathcal{O}_V)$ means ``with the current relation $\mathbf{D}$ and the target $\mathbf{PP}$, first transform to the $\mathbf{PO}$''; (b) The Kolmogorov-Arnold neural architecture of $\mathcal{I}^\mathbf{D}(\mathcal{O}_X,\mathcal{O}_V)$, where $f_r([x_1,\dots,x_{n+1}])=e^{x_{n+1}}$, $f_{\vec{O}}([x_1,\dots,x_{n+1}])=[x_1,\dots,x_{n}]$, $f_{\|\_\|}(\vec{x})= \|\vec{x}\|$, $f_{\max(0, \_)}(s)=\max(0, s)$; (c) updating the location and the size of $\mathcal{O}_{X}$ to transform the spatial relation from disconnecting from to being part of $\mathcal{O}_{V}$.}
\label{fig:hsgnn}
\end{sidewaysfigure}

$\mathcal{S}$phNN is a hierarchical {\em neuro-symbolic} {\em Kolmogorov-Arnold}~\cite{liu2024kan} {\em Geometric}~\cite{GGNN23} {\em Graph} Neural Network that explicitly constructs sphere configurations in the vector space. $\mathcal{S}$phNN has three layers -- the top symbolic layer, the spatial transition layer, and the geometric sphere layer, as illustrated in Figure~\ref{fig:hsgnn}(a). The top symbolic layer is a symbolic graph of neighbourhood relations \cite{Randell92a,DongJPL}. These relations are jointly exhaustive and pairwise disjoint -- at any time, there is one and only one relation being true. The bottom geometric sphere layer is the current sphere configuration, each sphere is represented by an $n+1$ dimensional vector whose first $n$ elements represent the centre and whose last element represents the radius; between the top and the bottom layers is the spatial transition layer that transforms the current sphere configuration to the target configuration through neighbourhood transitions. The spatial transition layer is the graph neighbourhood network whose nodes are spatial relations and whose edges are spatial neighbourhood relations. Every node in the spatial transition layer corresponds to a node in the symbolic layer and has a function that determines whether the symbolic spatial relation is held between the corresponding spheres in the bottom layer. The values of this function are non-minus, it returns zero if and only if this spatial relation is held between the two spheres. For example, $\mathcal{I}^\mathbf{D}(\mathcal{O}_X,\mathcal{O}_V)\triangleq\max\{0, (r_X+r_V) - \|\vec{O}_X-\vec{O}_V\|\}$ is the function that inspects locations and sizes of $\mathcal{O}_X$ and $\mathcal{O}_V$ and decides whether $\mathcal{O}_X$ disconnects from $\mathcal{O}_V$. We call $\mathcal{I}^\mathbf{D}$ an inspection function and define $\mathcal{I}^\mathbf{D}(\mathcal{O}_{X}, \mathcal{O}_{V})$ and implement it in a Kolmogorov-Arnold-styled architecture, in the sense that each edge is associated with a gradual descent function that can transform the current spatial relation into the neighbourhood relation, as shown in Figure~\ref{fig:hsgnn}(b). We introduce $\Delta_{\mathbf{T}_1:\mathbf{T}_2}^{\mathbf{T}}(\mathcal{O}_X, \mathcal{O}_V)$ as the gradual descent function that realises the neighbourhood transition from the current relation $\mathbf{T}_1 (\mathcal{O}_X, \mathcal{O}_V)$ to its neighbourhood relation $\mathbf{T}_2 (\mathcal{O}_X, \mathcal{O}_V)$ with the target $\mathbf{T} (\mathcal{O}_X, \mathcal{O}_V)$. If the context is clear, $\Delta_{\mathbf{T}_1:\mathbf{T}_2}^{\mathbf{T}}(\mathcal{O}_X, \mathcal{O}_V)$ can be written as $\Delta_{\mathbf{T}_1}^{\mathbf{T}}(\mathcal{O}_X, \mathcal{O}_V)$, for example, $\Delta_{\mathbf{T}_1:\mathbf{T}_2}^{\mathbf{T}_2}(\mathcal{O}_X, \mathcal{O}_V)=\Delta_{\mathbf{T}_1}^{\mathbf{T}_2}(\mathcal{O}_X, \mathcal{O}_V)$. Each $\Delta_{\mathbf{T}_1:\mathbf{T}_2}^{\mathbf{T}}(\mathcal{O}_X, \mathcal{O}_V)$ satisfies three conditions as follows:
\begin{enumerate}
    \item non-negative, $\Delta_{\mathbf{T}_1:\mathbf{T}_2}^{\mathbf{T}}(\mathcal{O}_X, \mathcal{O}_V)\ge 0$; 
    \item strict monotonic, when $\Delta_{\mathbf{T}_1:\mathbf{T}_2}^{\mathbf{T}}(\mathcal{O}_X, \mathcal{O}_V)> 0$;
    \item  if the target relation is reached, $\Delta_{\mathbf{T}_1:\mathbf{T}_2}^{\mathbf{T}}(\mathcal{O}_X, \mathcal{O}_V)= 0$.
\end{enumerate}
The three features make $\Delta_{\mathbf{T}_1:\mathbf{T}_2}^{\mathbf{T}}(\mathcal{O}_X, \mathcal{O}_V)$ work like a slide where $\mathcal{O}_X$ slips at the height status of $\mathbf{T}_1(\mathcal{O}_X, \mathcal{O}_V)$, where $\mathcal{I}^{\mathbf{T}_1}(\mathcal{O}_{X}, \mathcal{O}_{V})=0$ and $\mathcal{I}^{\mathbf{T}_2}(\mathcal{O}_{X}, \mathcal{O}_{V})>0$, downward to the status of $\mathbf{T}_2(\mathcal{O}_X, \mathcal{O}_V)$, where $\mathcal{I}^{\mathbf{T}_2}(\mathcal{O}_{X}, \mathcal{O}_{V})=0$,  towards the final status (the target value) of $\mathbf{T}(\mathcal{O}_X, \mathcal{O}_V)$, where $\mathcal{I}^{\mathbf{T}}(\mathcal{O}_{X}, \mathcal{O}_{V})=0$. Each transition ($\Delta$ function) starting from $\mathbf{T}_1$ targeting $\mathbf{T}$ with the next neighbourhood transition $\mathbf{T}_2$ can be realised by a linear combination of the radii and the distance between the centre points and implemented in Kolmogorov-Arnold-styled neural architecture. The neighbourhood relations structure a graph structure that allows
$\mathcal{S}$phNN to construct the target sphere configuration through neighbourhood transitions. 
For example, suppose the current graph is $\mathcal{G}=(\mathcal{V}, \mathcal{E})$, where  $\mathcal{V}=\{\mathcal{O}_{X}, \mathcal{O}_{V}\}$ and $\mathcal{E}=\{\mathbf{D}(\mathcal{O}_{X}, \mathcal{O}_{V})\}$, and the target graph is $\mathcal{G}'=(\mathcal{V}, \mathcal{E}')$, where $\mathcal{E}'=\{\mathbf{PP}(\mathcal{O}_{X}, \mathcal{O}_{V})\}$. $\mathcal{S}$phNN firstly uses $\Delta_{\mathbf{D}:\mathbf{PO}}^\mathbf{PP}(\mathcal{O}_{X}, \mathcal{O}_{V})$ to transform into the neighbourhood relation $\mathbf{PO}$, then uses $\Delta_{\mathbf{PO}}^\mathbf{PP}(\mathcal{O}_{X}, \mathcal{O}_{V})$ to transform the $\mathbf{PO}$ relation into its neighbourhood relation $\mathbf{PP}$, as illustrated in Figure~\ref{fig:hsgnn}(c).

\subsection{Geometric operations on spheres}

We introduce a set of geometric operations on a sphere $\mathcal{O}_X$ to update its relation referenced to a fixed Sphere $\mathcal{O}_V$: $\delta(\mathcal{O}_X | \mathcal{O}_V)=\{dis_{X,V}^\downarrow, r_X^\downarrow, dis_{X,V}^\uparrow, r_X^\uparrow\}$, where $dis_{X,V}$ is the distance between their centres $dis_{X,V}=\|\vec{O}_X -\vec{O}_V\|$, $\downarrow$ represents to decrease a value, $\uparrow$ represents to increase a value. 
The target relation $\mathbf{T}$ determines possible operations, either to preserve the already reached target relation or to transform it into a neighbourhood relation towards the target. For example, to preserve $\mathcal{O}_X$ being inside $\mathcal{O}_V$, $\mathbf{P}(\mathcal{O}_X, \mathcal{O}_V)$, the possible operations on $\mathcal{O}_X$ are either to decrease the distance between their centres, or to decrease the radius of $\mathcal{O}_X$, written as   
$\delta^\mathbf{P}(\mathcal{O}_X| \mathcal{O}_V)=\{dis_{X,V}^\downarrow, r_X^\downarrow\}$; to transform $\mathcal{O}_X$ from being disjoint with $\mathcal{O}_V$ to partially overlapping with $\mathcal{O}_V$,  the possible operations on $\mathcal{O}_X$ are either to decrease the distance, or to increase $r_X$, so, $\delta_{\mathbf{D}:\mathbf{PO}}(\mathcal{O}_X| \mathcal{O}_V)=\{dis_{X,V}^\downarrow, r_X^\uparrow\}$. If this neighbourhood transition is targeted at $\mathcal{O}_X$ being inside $\mathcal{O}_V$, the current operation of increasing $r_X$ will violate the possible operations of the target relation and may introduce unnecessary back-and-forth updates of $\mathcal{O}_X$, so, $r_X^{\uparrow}$ will not be selected. The set of possible operations on $\mathcal{O}_X$ to transform $\mathbf{D}(\mathcal{O}_X, \mathcal{O}_V)$ to $\mathbf{PO}(\mathcal{O}_X, \mathcal{O}_V)$ with the target $\mathbf{P}(\mathcal{O}_X, \mathcal{O}_V)$, $\delta^\mathbf{P}_{\mathbf{D}:\mathbf{PO}}(\mathcal{O}_X| \mathcal{O}_V)$, are the intersections of $\delta^{\mathbf{P}}(\mathcal{O}_X| \mathcal{O}_V)$ and $\delta_{\mathbf{D}:\mathbf{PO}}(\mathcal{O}_X| \mathcal{O}_V)$, written as 
$\delta^\mathbf{P}_{\mathbf{D}:\mathbf{PO}}(\mathcal{O}_X| \mathcal{O}_V)\triangleq \delta^{\mathbf{P}}(\mathcal{O}_X| \mathcal{O}_V)\cap \delta_{\mathbf{D}:\mathbf{PO}}(\mathcal{O}_X| \mathcal{O}_V)=\{dis_{X,V}^\downarrow, r_X^\uparrow\}\cap\{dis_{X,V}^\downarrow, r_X^\downarrow\}$ $=\{dis_{X,V}^\downarrow\}$. Possible operations are implemented by gradual descent functions as follows: $x^\downarrow$ is implemented by $+x$, written as $\zeta(x^\downarrow) = +x$; $x^\uparrow$ is implemented by $-x$, written as $\zeta(x^\downarrow) = -x$, where $x\in\{dis_{X,V}, r_X, r_V\}$. This transforms a set of operations into a gradual descent function.  $\Delta_{\mathbf{T}_1:\mathbf{T}_2}^{\mathbf{T}}(\mathcal{O}_X, \mathcal{O}_V)$ is implemented by $\max\{0, C + \sum \zeta(op)\}$, where $op\in \delta_{\mathbf{T}_1:\mathbf{T}_2}^{\mathbf{T}}(\mathcal{O}_X| \mathcal{O}_V)$ and $C$ is a constant such that $\mathbf{T}_2$ is reached, exactly when $C + \sum \zeta(op) = 0$.

\begin{figure}[t]
\centering
\includegraphics[width=1\textwidth]{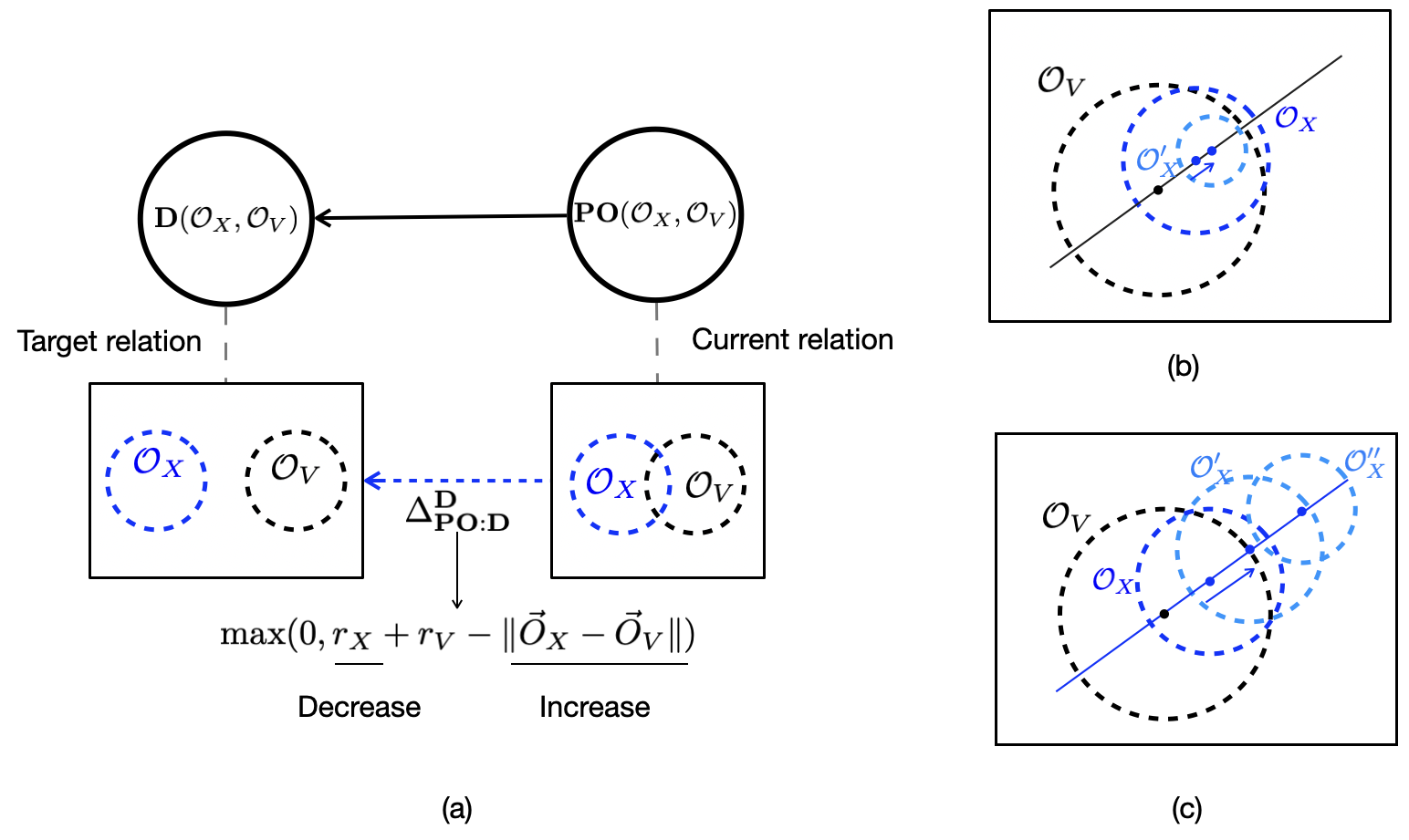}
\caption{(a) $\Delta^\mathbf{D}_{\mathbf{PO}:\mathbf{D}}(\mathcal{O}_X, \mathcal{O}_V)$ implements a neighbourhood transition from $\mathbf{PO}(\mathcal{O}_X, \mathcal{O}_V)$ to $\mathbf{D}(\mathcal{O}_X, \mathcal{O}_V)$. (b) If the centre of $\mathcal{O}_X$ is inside $\mathcal{O}_V$, reducing $r_X$ too fast will cause $\mathcal{O}_X$ being inside $\mathcal{O}_V$; (c) Solution: when the centre of $\mathcal{O}_X$ is inside $\mathcal{O}_V$, $\mathcal{O}_X$ will be moved away from $\mathcal{O}_V$ till its centre is at the boundary of $\mathcal{O}_V$.}
\label{sub_rel1}
\end{figure}

\subsection{Atomic neighbourhood transition}

\label{neighbour_trans}
As a neighbourhood transition, $\Delta_{\mathbf{T}_1:\mathbf{T}_2}^{\mathbf{T}}(\mathcal{O}_X, \mathcal{O}_V)$ needs to guarantee that on the way from  $\mathbf{T}_1(\mathcal{O}_X, \mathcal{O}_V)$ to its neighbour $\mathbf{T}_2(\mathcal{O}_X, \mathcal{O}_V)$, there will not appear a third relation $\mathbf{T}_3(\mathcal{O}_X, \mathcal{O}_V)$, where $\mathbf{T}_3\notin\{\mathbf{T}_1,\mathbf{T}_2\}$. That is, neighbourhood transitions should be {\em atomic}. However,
as gradual descent functions update independently the centre and the radius of $\mathcal{O}_X$, a neighbourhood transition, under some situations, may not be {\em atomic}. For example,
the transition from the partial overlapping relation $\mathbf{PO}(\mathcal{O}_X, \mathcal{O}_V)$ to the disconnectedness relation $\mathbf{D}(\mathcal{O}_X, \mathcal{O}_V)$ is realised by the gradual descent function $\Delta_{\mathbf{PO}:\mathbf{D}}^\mathbf{D}(\mathcal{O}_X, \mathcal{O}_V)=\Delta_\mathbf{PO}^\mathbf{D}(\mathcal{O}_X, \mathcal{O}_V)=\max\{0, r_X + r_V- dis_{X,V}\}$, as shown in Figure~\ref{sub_rel1}(a). To reach the target, $\Delta_\mathbf{PO}^\mathbf{D}(\mathcal{O}_X, \mathcal{O}_V)$ will either reduce $r_X$ or increase $dis_{X,V}$ or both (Sphere $\mathcal{O}_V$ is fixed). When the centre of $\mathcal{O}_X$ is inside $\mathcal{O}_V$, reducing $r_X$ too fast may cause $\mathcal{O}_X$ being inside $\mathcal{O}_V$, as shown in Figure~\ref{sub_rel1}(b). To avoid this situation, we partition $\mathbf{PO}$ into two sub-relations: $\mathbf{PO}_1$ and $\mathbf{PO}_2$: 
$\mathbf{PO}_1(\mathcal{O}_X, \mathcal{O}_V)$ is the sub-relation of $\mathbf{PO}$ when the centre of $\mathcal{O}_X$ is outside $\mathcal{O}_V$; $\mathbf{PO}_2(\mathcal{O}_X, \mathcal{O}_V)$ is the sub-relation of $\mathbf{PO}$ when the centre of $\mathcal{O}_X$ is inside or at the border of $\mathcal{O}_V$. We define $\Delta_{\mathbf{PO}_2:\mathbf{PO}_1}^{\mathbf{D}}(\mathcal{O}_X, \mathcal{O}_V)$ as moving $\mathcal{O}_X$ away from $\mathcal{O}_V$ till $\mathbf{PO}_1(\mathcal{O}_X, \mathcal{O}_V)$, while fixing $r_X$.   $\Delta_{\mathbf{PO}_1}^{\mathbf{D}}(\mathcal{O}_X, \mathcal{O}_V)$ is {\em atomic} even if its centre and radius are optimised independently, as shown in Figure~\ref{sub_rel1}(c). In this way, $\Delta_\mathbf{PO}^\mathbf{D}(\mathcal{O}_X, \mathcal{O}_V)$ is replaced by either  $\Delta_{\mathbf{PO}_1}^{\mathbf{D}}(\mathcal{O}_X, \mathcal{O}_V)$ or $\Delta_{\mathbf{PO}_2:\mathbf{PO}_1}^{\mathbf{D}}(\mathcal{O}_X, \mathcal{O}_V)$ followed with $\Delta_{\mathbf{PO}_1}^{\mathbf{D}}(\mathcal{O}_X, \mathcal{O}_V)$. Each case is {\em atomic}.

\begin{figure}[!t]
\centering
\includegraphics[width=1\textwidth]{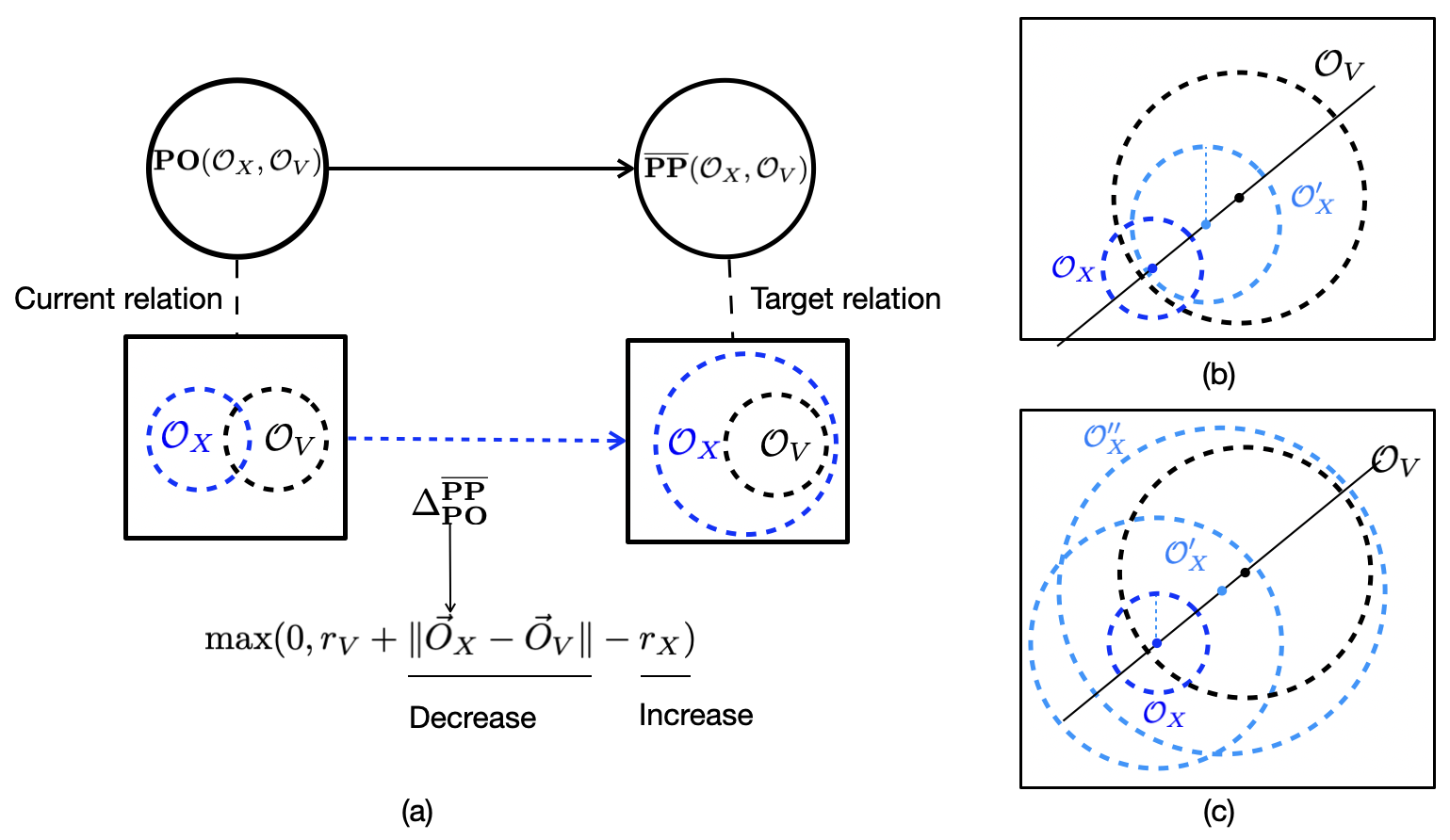} \caption{(a) $\Delta^{\overline{\mathbf{PP}}}_{\mathbf{PO}:{\overline{\mathbf{PP}}}}(\mathcal{O}_X, \mathcal{O}_V)$ implements a neighbourhood transition $\mathbf{PO}(\mathcal{O}_X, \mathcal{O}_V)$ to $\overline{\mathbf{PP}}(\mathcal{O}_X, \mathcal{O}_V)$. (b) If $r_X<r_V$, enlarging  $r_X$ too slow will cause $\mathcal{O}_X$ being inside $\mathcal{O}_V$; (c) Solution: when $r_X<r_V$, firstly enlarge $r_X$ to $r_V$ while fixing the centre of $\mathcal{O}_X$.}
\label{sub_rel2}
\end{figure}
Another case is the transition from the partial overlapping relation to the containing relation, $\Delta^{\overline{\mathbf{PP}}}_{\mathbf{PO}:{\overline{\mathbf{PP}}}}(\mathcal{O}_X, \mathcal{O}_V)=\Delta^{\overline{\mathbf{PP}}}_{\mathbf{PO}}(\mathcal{O}_X, \mathcal{O}_V)$, whose geometric operations are enlarging $r_X$ and decreasing $dis_{X,V}$,  as shown in Figure~\ref{sub_rel2}(a). If $r_X<r_{V}$ and $r_X$ is enlarged too slow,  $\mathcal{O}_{X}$ will be part of $\mathcal{O}_{V}$, instead of containing $\mathcal{O}_{V}$, as shown in Figure~\ref{sub_rel2}(b). To avoid this situation, we split the $\mathbf{PO}$ relation into $\mathbf{PO}_3$ and $\mathbf{PO}_4$: $\mathbf{PO}_3(\mathcal{O}_{X},\mathcal{O}_{V})$ is the sub-relation of $\mathbf{PO}(\mathcal{O}_{X},\mathcal{O}_{V})$ with the condition that $r_{X}< r_{V}$;  $\mathbf{PO}_4(\mathcal{O}_{X},\mathcal{O}_{V})$ is the sub-relation of $\mathbf{PO}(\mathcal{O}_{X},\mathcal{O}_{V})$ with the condition that  $r_{X}\ge r_{V}$. If 
$\mathbf{PO}_3(\mathcal{O}_{X},\mathcal{O}_{V})$ holds, $r_{X}$ will be enlarged to reach the same length as $r_V$, while fixing the centre of $\mathcal{O}_X$, resulting in  $\mathbf{PO}_4(\mathcal{O}_{X},\mathcal{O}_{V})$. After that, $\Delta_{\mathbf{PO}_4}^{\overline{\mathbf{PP}}}$ will transform $\mathbf{PO}_4(\mathcal{O}_{X}, \mathcal{O}_{V})$ into ${\overline{\mathbf{PP}}}(\mathcal{O}_{X}, \mathcal{O}_{V})$,  as illustrated in Figure~\ref{sub_rel2}(c). In this way, $\Delta^{\overline{\mathbf{PP}}}_{\mathbf{PO}:{\overline{\mathbf{PP}}}}(\mathcal{O}_X, \mathcal{O}_V)$ is replaced by either  $\Delta^{\overline{\mathbf{PP}}}_{\mathbf{PO}_4}(\mathcal{O}_X, \mathcal{O}_V)$ or $\Delta^{\overline{\mathbf{PP}}}_{\mathbf{PO}_3:\mathbf{PO}_4}(\mathcal{O}_X, \mathcal{O}_V)$
followed with $\Delta^{\overline{\mathbf{PP}}}_{\mathbf{PO}_4}(\mathcal{O}_X, \mathcal{O}_V)$. Each case is {\em atomic}.

\section{Transition functions between two spheres}
\label{transfun}

\begin{quote} 
{\em 
All mathematical derivation can be viewed simply as change in representation, making evident what was previously true but obscure.
    
    \hfill --- Herbert A. Simon  \cite{Simon19}
     } 
\end{quote}

In this section, we list all transition functions for  $\mathcal{O}_{X}$ and $\mathcal{O}_{V}$, where $\mathcal{O}_{V}$ is fixed. 

\subsection{Targeting at $\mathbf{D}(\mathcal{O}_{X}, \mathcal{O}_{V})$}
\label{D}

Let $\mathcal{O}_{X}$ should disconnect from $\mathcal{O}_{V}$, $\mathbf{D}(\mathcal{O}_{X}, \mathcal{O}_{V})$. Whether this target relation is satisfied can be measured geometrically by the truth value of $(r_X+r_V) - dis_{X,V} \le 0$, where $dis_{X,V}=\|\vec{O}_X -\vec{O}_V\|$.  To make the formula true, $\mathcal{S}$phNN can either gradually descent $r_
X$ or gradually ascent $dis_{X,V}$,  therefore, the possible operations are $dis_{X,V}^\uparrow$ and $r_X^\downarrow$, $\delta^\mathbf{D}(\mathcal{O}_X| \mathcal{O}_V)=\{dis_{X,V}^\uparrow, r_X^\downarrow\}$. The inspection function is defined as $\mathcal{I}^\mathbf{D}(\mathcal{O}_{X}, \mathcal{O}_{V})\triangleq\max\{0, (r_X+r_V) - dis_{X,V}\}$.

\begin{figure}
\centering
\includegraphics[width=1\textwidth]{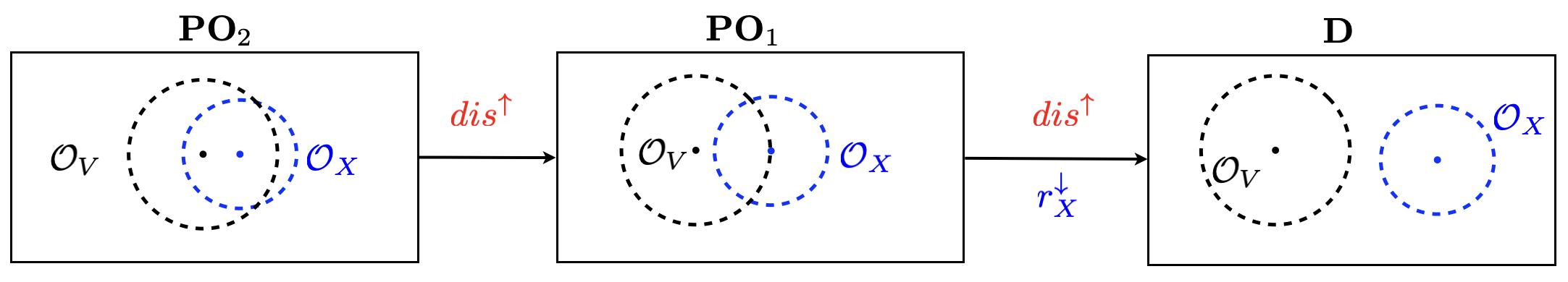}
\caption{If the centre of $\mathcal{O}_X$ is inside $\mathcal{O}_V$, $\mathcal{S}$phNN will move $\mathcal{O}_X$ away from $\mathcal{O}_V$ while fixing $r_X$, till the centre of $\mathcal{O}_X$ is located at the boundary of $\mathcal{O}_V$. Then, $\mathcal{S}$phNN will continue to move $\mathcal{O}_X$ away from $\mathcal{O}_V$ while independently decreasing $r_X$, till reaching the target relation $\mathbf{D}(\mathcal{O}_X, \mathcal{O}_V)$.}
\label{fig:enn_d}
\end{figure}

If $\mathcal{O}_{X}$ and $\mathcal{O}_{V}$ are currently partially overlapped and the centre of $\mathcal{O}_{X}$ is inside $\mathcal{O}_{V}$, it may happen that uncoordinated optimising the centre and $r_X$ will not lead $\mathcal{O}_{X}$ to disconnect from $\mathcal{O}_{V}$. Following the analysis in Section~\ref{neighbour_trans}, we partition the $\mathbf{PO}$ relation into $\mathbf{PO}_1$ and $\mathbf{PO}_2$ and list the related formulas as follows  and illustrated in Figure~\ref{fig:enn_d}.  
\begin{eqnarray*}
&&\mathbf{PO}_1(\mathcal{O}_{X},\mathcal{O}_{V})\triangleq\mathbf{PO}(\mathcal{O}_{X},\mathcal{O}_{V})\land dis_{X,V} >r_V\\ 
&&\mathbf{PO}_2(\mathcal{O}_{X},\mathcal{O}_{V})\triangleq\mathbf{PO}(\mathcal{O}_{X},\mathcal{O}_{V})\land dis_{X,V}\le r_V\\ 
&&\mathcal{I}^{\mathbf{PO}}(\mathcal{O}_{X}, \mathcal{O}_{V})\triangleq\max\{0,  |r_X - r_V| - dis_{X,V} + \epsilon\}+\max\{0,  dis_{X,V}-r_V -r_X +\epsilon\}\\
&&\mathcal{I}^{\mathbf{PO}_1}(\mathcal{O}_{X}, \mathcal{O}_{V})\triangleq\mathcal{I}^{\mathbf{PO}}(\mathcal{O}_{X}, \mathcal{O}_{V})+\max\{0, r_V - dis_{X,V}+\epsilon\}\\
&&\mathcal{I}^{\mathbf{PO}_2}(\mathcal{O}_{X}, \mathcal{O}_{V})\triangleq\mathcal{I}^{\mathbf{PO}}(\mathcal{O}_{X}, \mathcal{O}_{V})+\max\{0,  dis_{X,V}-r_V \}\\
&&\Delta_{\mathbf{PO}_2:\mathbf{PO}_1}^{\mathbf{D}}(\mathcal{O}_{X}, \mathcal{O}_{V})\triangleq   \max\{0, r_V - dis_{X, V}\}\\
&&\Delta_{\mathbf{PO}_1} ^\mathbf{D}(\mathcal{O}_X, \mathcal{O}_V)\triangleq  \max\{0, r_X + r_V  - dis_{X, V}\}
\end{eqnarray*}

\subsection{Targeting at $\overline{\mathbf{P}}(\mathcal{O}_{X}, \mathcal{O}_{V})$}
\label{Pbar}

Let $\mathcal{O}_{X}$ should contain $\mathcal{O}_{V}$, $\overline{\mathbf{P}}(\mathcal{O}_{X}, \mathcal{O}_{V})$. 
Whether this target relation is satisfied can be measured geometrically by the truth value of $ dis_{X,V} + r_V -r_X \le 0$. The inspection function $\mathcal{I}^{\overline{\mathbf{P}}}(\mathcal{O}_{X}, \mathcal{O}_{V})\triangleq\max\{0, dis_{X,V} + r_V -r_X\}$. When $\mathcal{O}_{X}$ contains $\mathcal{O}_{V}$, the value is 0; otherwise,  the value is greater than 0. To reduce the value, $\mathcal{S}$phNN shall increase $r_X$ or reduce $dis_{X,V}$,  therefore, the allowed operations are $r_X^\uparrow$ and $dis_{X,V}^\downarrow$.  

\begin{figure}[!t] 
\centering
\includegraphics[width=0.9\textwidth]{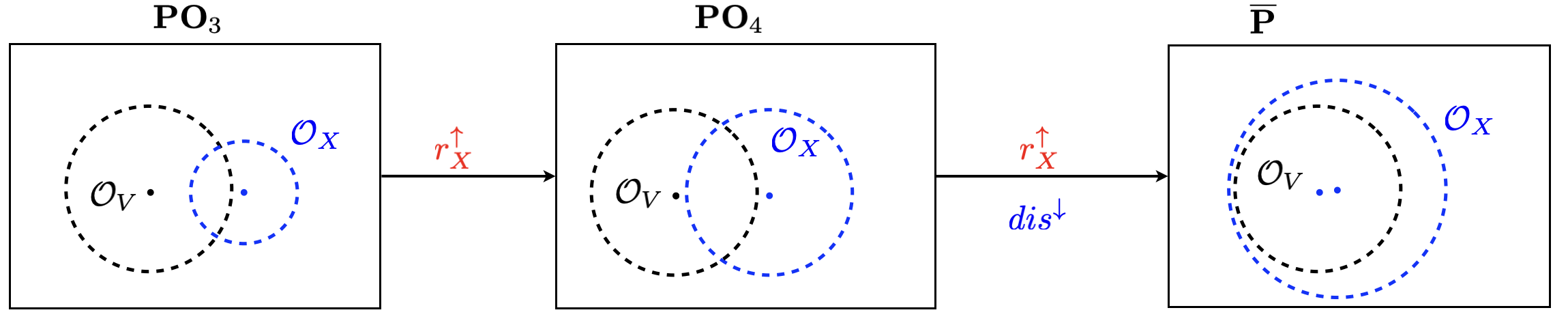}
\caption{If $r_X$ is shorter than $r_V$, $\mathcal{S}$phNN will enlarge $r_X$ till $r_X=r_V$. Then, $\mathcal{S}$phNN will continue to move $\mathcal{O}_X$ towards $\mathcal{O}_V$ while independently increasing $r_X$, till reaching the target relation $\overline{\mathbf{P}}(\mathcal{O}_X, \mathcal{O}_V)$.}
\label{fig:enn_pbar}
\end{figure}
If $\mathcal{O}_{X}$ and $\mathcal{O}_{V}$ are partially overlapped, and $r_X$ is shorter than $r_V$, it may happen that uncoordinated optimising the centre and the radius of $\mathcal{O}_{X}$ will not lead $\mathcal{O}_{X}$ to cover $\mathcal{O}_{V}$. Following the analysis in Section~\ref{neighbour_trans}, we partition the $\mathbf{PO}$ relation into $\mathbf{PO}_3$ and $\mathbf{PO}_4$ and list the related formulas as follows and illustrated in Figure~\ref{fig:enn_pbar}. 

\begin{eqnarray*}
&&\mathbf{PO}_3(\mathcal{O}_{X},\mathcal{O}_{V})\triangleq\mathbf{PO}(\mathcal{O}_{X},\mathcal{O}_{V})\land r_V < r_X\\ 
&&\mathcal{I}^{\mathbf{PO}_3}(\mathcal{O}_{X}, \mathcal{O}_{V})\triangleq\mathcal{I}^{\mathbf{PO}}(\mathcal{O}_{X}, \mathcal{O}_{V})+\max\{0,  r_V -r_X +\epsilon\}\\
&&\mathbf{PO}_4(\mathcal{O}_{X},\mathcal{O}_{V})\triangleq\mathbf{PO}(\mathcal{O}_{X},\mathcal{O}_{V})\land r_V \ge r_X\\
&&\mathcal{I}^{\mathbf{PO}_4}(\mathcal{O}_{X}, \mathcal{O}_{V})\triangleq\mathcal{I}^{\mathbf{PO}}(\mathcal{O}_{X}, \mathcal{O}_{V})+\max\{0,  r_X -r_V\}\\
&&\Delta_{\mathbf{PO}_3:\mathbf{PO}_4}^{\overline{\mathbf{P}}}(\mathcal{O}_{X},\mathcal{O}_{V})\triangleq\max\{0, r_V - r_X\}\\
&&\Delta_{\mathbf{PO}_4}^{\overline{\mathbf{P}}}(\mathcal{O}_X, \mathcal{O}_V)\triangleq\max\{0, dis_{X,V} + r_V - r_X\}
\end{eqnarray*}
 
\subsection{Targeting at $\mathbf{PO}(\mathcal{O}_{X},\mathcal{O}_{V})$}

\begin{figure}[!t]
\centering
\includegraphics[width=1\textwidth]{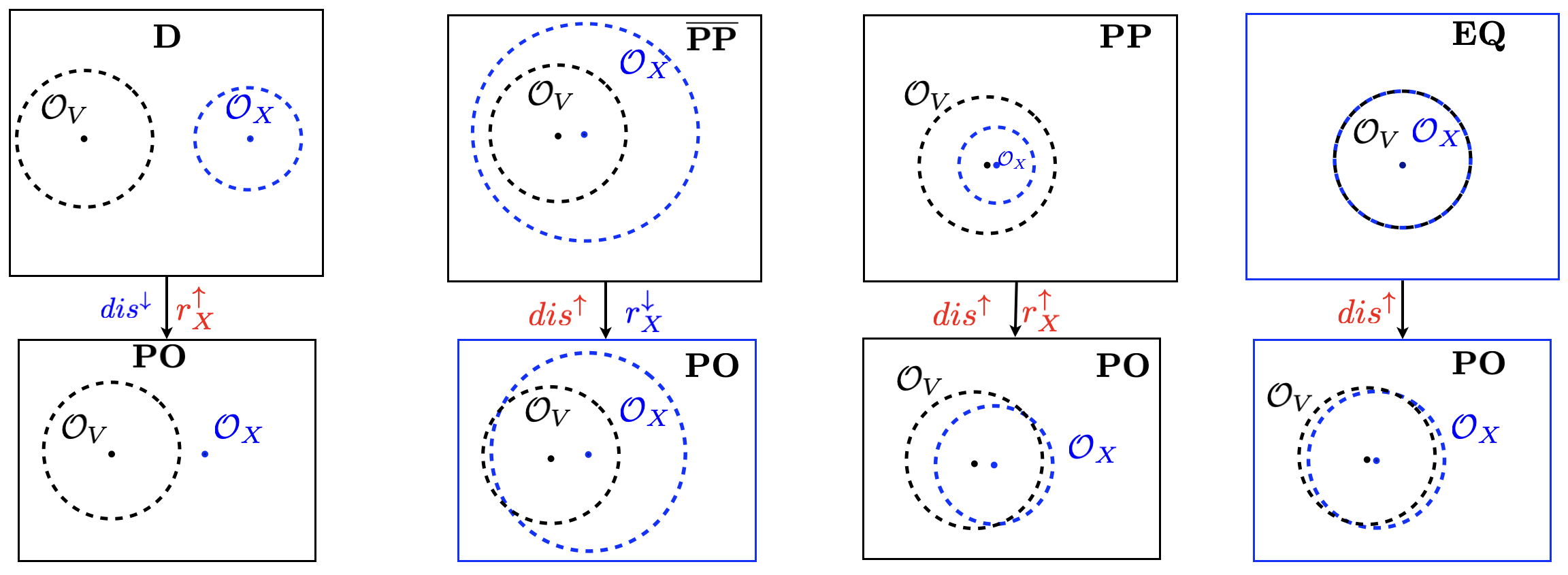}
\caption{The target relation $\mathbf{PO}(\mathcal{O}_X, \mathcal{O}_V)$ can be reached from $\mathbf{D}(\mathcal{O}_X, \mathcal{O}_V)$,  $\mathbf{PP}(\mathcal{O}_X, \mathcal{O}_V)$, and   
$\overline{\mathbf{PP}}(\mathcal{O}_X, \mathcal{O}_V)$ by independently optimising $r_X$ and $dis$. If the current relation is 
$\mathbf{EQ}(\mathcal{O}_X, \mathcal{O}_V)$, $\mathcal{S}$phNN will increase $dis$ by slightly changing the centre of $\mathcal{O}_X$ to reach the target relation.}
\label{fig:enn_po}
\end{figure}

Two spheres being partially overlapped means that the distance between their centres is (1) shorter than the sum of their radii, and (2) longer than the difference between their radii. That is $|r_X - r_V| <dis_{X,V} < r_X + r_V$.
The partial overlapping 
 relation $\mathbf{PO}$ is not a final target relation but rather an intermediate target to reach other target relations. It can be reached from four other relations.  If $\mathcal{O}_X$ currently disconnects from $\mathcal{O}_V$, 
$\mathcal{S}$phNN can perform both $dis_{X,V}^\downarrow$ and $r_X^\uparrow$ operations; if  $\mathcal{O}_X$ currently is a proper part of $\mathcal{O}_V$, $\mathcal{S}$phNN can perform $dis_{X,V}^\uparrow$ and $r_X^\uparrow$ operations; if  $\mathcal{O}_V$ is a proper part of $\mathcal{O}_X$, $\mathcal{S}$phNN can perform $dis_{X,V}^\uparrow$ and $r_X^\downarrow$ operations; if $\mathcal{O}_X$ coincides with $\mathcal{O}_V$, $\mathcal{S}$phNN only need to randomly shift $\mathcal{O}_X$ away from its current location, as illustrated in Figure~\ref{fig:enn_po}. 
\begin{eqnarray*}
&&\Delta^{\mathbf{PO}}_{\overline{\mathbf{PP}}}(\mathcal{O}_{X},\mathcal{O}_{V})\triangleq\max\{0,  r_X -r_V - dis_{X,V}\}\\
&&\Delta^{\mathbf{PO}}_\mathbf{PP}(\mathcal{O}_X, \mathcal{O}_V)\triangleq\max\{0, r_V -r_X - dis_{X,V}\}\\
&&\Delta^{\mathbf{PO}}_\mathbf{D}(\mathcal{O}_X, \mathcal{O}_V)\triangleq\max\{0, dis_{X,V} - r_V -r_X\}\\
&&\Delta^{\mathbf{PO}}_\mathbf{EQ}(\mathcal{O}_X, \mathcal{O}_V)\triangleq  \vec{O}_X + \vec{\epsilon}
\end{eqnarray*}

\subsection{Target at $\mathbf{P}(\mathcal{O}_{X}, \mathcal{O}_{V})$}
\label{PO2P}
\begin{figure}
\centering
\includegraphics[width=1\textwidth]{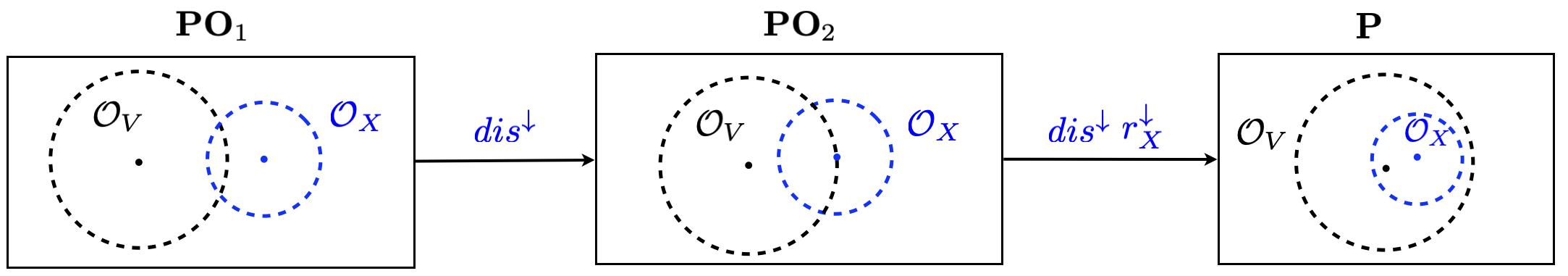}
\caption{If the centre of $\mathcal{O}_X$ is outside $\mathcal{O}_V$, $\mathcal{S}$phNN will move $\mathcal{O}_X$ towards $\mathcal{O}_V$ while fixing $r_X$, till the centre of $\mathcal{O}_X$ is located at the boundary of $\mathcal{O}_V$. Then, $\mathcal{S}$phNN will continue to move $\mathcal{O}_X$ towards $\mathcal{O}_V$ while independently decreasing $r_X$, till reaching the target relation $\mathbf{P}(\mathcal{O}_X, \mathcal{O}_V)$.}
\label{fig:enn_p}
\end{figure}

Let  $\mathcal{O}_{X}$ should be part of $\mathcal{O}_{V}$, $\mathbf{P}(\mathcal{O}_{X}, \mathcal{O}_{V})$. 
Whether this target relation is satisfied can be measured geometrically by the truth value of $ dis_{X,V} + r_X -r_V \le 0$. $\mathcal{I}^{\mathbf{P}}(\mathcal{O}_{X}, \mathcal{O}_{V})\triangleq \max\{0,   dis_{X,V} + r_X -r_V\}$. To observe  $\mathcal{I}^{\mathbf{P}}(\mathcal{O}_{X}, \mathcal{O}_{V}) = 0$, $\mathcal{S}$phNN can perform 
 $dis_{X,V}^\downarrow$ and $r_X^\downarrow$ operations.
In $\mathbf{PO}_2$ status, performing either $dis_{X,V}^\downarrow$ or $r_X^\downarrow$ operation 
will lead to the target status, as illustrated in Figure~\ref{fig:enn_p}. We define $\Delta_{\mathbf{PO}_2}^\mathbf{P}(\mathcal{O}_X, \mathcal{O}_V)\triangleq\max\{0, dis_{X,V} + r_X -r_V\}$. 
In $\mathbf{PO}_1$ status, only doing gradually descent of $r_X$ may lead $\mathcal{O}_{X}$ to disconnect from $\mathcal{O}_{V}$. To prevent this situation, $\mathcal{S}$phNN fixes $r_X$ and only performs the $dis_{X,V}^\downarrow$ operation.
We introduce {\color{red}{$\Delta$}}$_{\mathbf{PO}_1:\mathbf{PO}_2}^{\mathbf{P}}(\mathcal{O}_{X}, \mathcal{O}_{V})\triangleq \max\{0, dis_{X,V}\}$ to transform the relation from $\mathbf{PO}_1$ to $\mathbf{PO}_2$.  

\begin{figure} 
\centering
\includegraphics[width=1\textwidth]{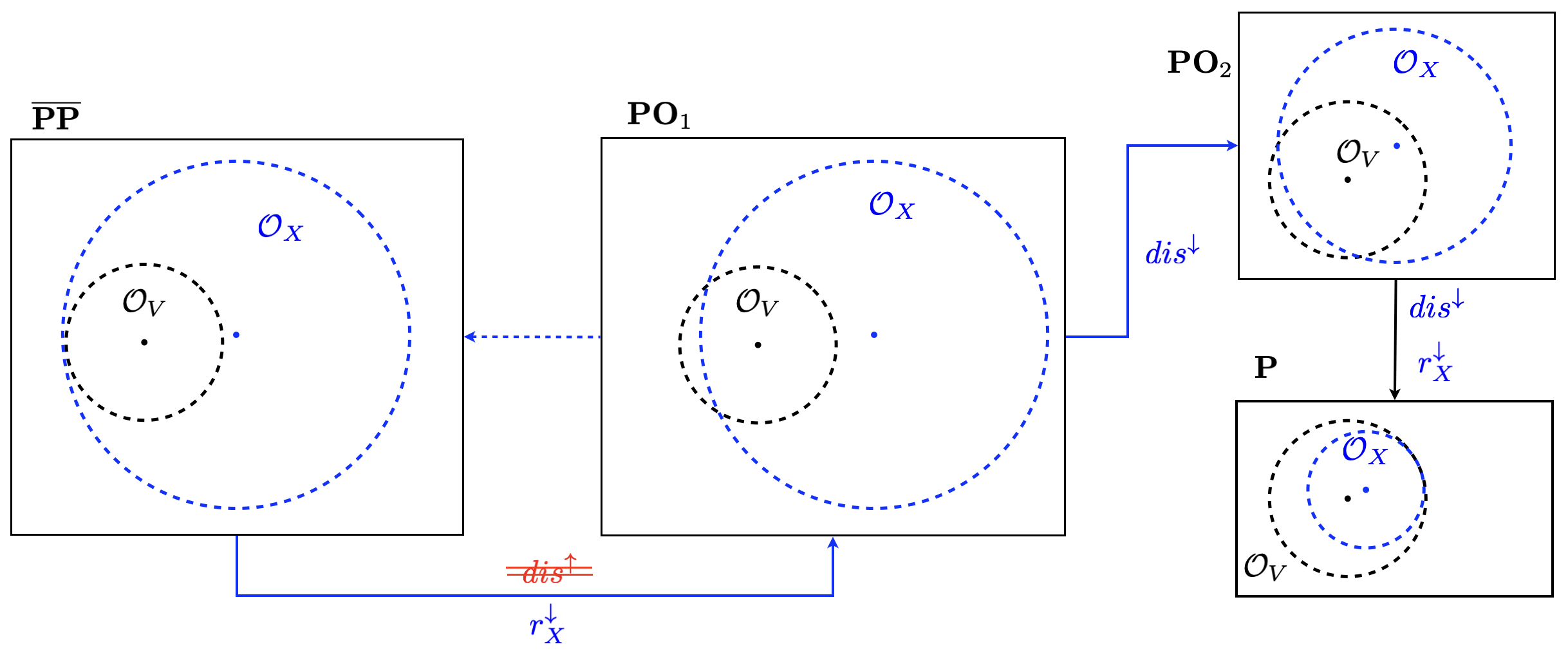}
\caption{If $r_X > 2r_V$, it may happen that reducing the distance between their centres causes $\mathcal{O}_X$ containing $\mathcal{O}_V$, as shown by the blue dotted line. Then, $\Delta_{\overline{\mathbf{PP}}:\mathbf{PO}}^{\mathbf{P}}(\mathcal{O}_X, \mathcal{O}_V)$ will reduce the value of $r_X$. This loop repeats till $\mathbf{PO}_2 (\mathcal{O}_X, \mathcal{O}_V)$ is reached.}
\label{fig:enn_p1}
\end{figure}

The condition for {\color{red}{$\Delta$}}$_{\mathbf{PO}_1:\mathbf{PO}_2}^{\mathbf{P}}(\mathcal{O}_X, \mathcal{O}_V)$ to reach $\mathbf{PO}_2(\mathcal{O}_X, \mathcal{O}_V)$ is that the radius of $\mathcal{O}_X$ should be less than the diameter of $\mathcal{O}_V$, that is, $r_X < 2r_V$. If the condition is not satisfied, repeated operations of {\color{red}{$\Delta$}}$_{\mathbf{PO}_1:\mathbf{PO}_2}^{\mathbf{P}}(\mathcal{O}_X, \mathcal{O}_V)$ will push $\mathcal{O}_X$ to contain $\mathcal{O}_V$, ${\overline{\mathbf{P}}}(\mathcal{O}_X, \mathcal{O}_V)$, and trigger $\Delta_{\overline{\mathbf{PP}}:\mathbf{PO}}^{\mathbf{P}}(\mathcal{O}_X, \mathcal{O}_V)$, whose possible operation is $r_X^\downarrow$. This operation is the intersection of the possible operations from $\overline{\mathbf{PP}}$ to $\mathbf{PO}_1$ and the possible operations of the target relation $\mathbf{P}$, as shown in Figure~\ref{fig:enn_p1}. This works like that  {\color{red}{$\Delta$}}$_{\mathbf{PO}_1:\mathbf{PO}_2}^{\mathbf{P}}(\mathcal{O}_X, \mathcal{O}_V)$ borrows $\Delta_{\overline{\mathbf{PP}}:\mathbf{PO}}^{\mathbf{P}}(\mathcal{O}_X, \mathcal{O}_V)$ to reduce the radius of $\mathcal{O}_X$. {\color{red}{$\Delta$}}$_{\mathbf{PO}_1:\mathbf{PO}_2}^{\mathbf{P}}(\mathcal{O}_X, \mathcal{O}_V)$ is the only $\Delta$ function with a condition. So, we use the {\color{red}{red}} colour to demarcate this feature. To avoid this loop and make all $\Delta$ operations atomic, we can introduce an additional operation: cut $r_X$ to $r_V$, if $r_X>r_V$ and $\mathbf{P}(\mathcal{O}_X, \mathcal{O}_V)$ is targeted. This additional operation might make $\mathcal{O}_X$ disconnect from $\mathcal{O}_V$.  

\subsection{Targeting at negative relations}

If the target is a negative relation $\mathbf{R}\in\{ \neg\mathbf{D}, \neg\mathbf{P}, \neg{\overline{\mathbf{P}}}\}$, there will be only one non-target relation $\neg\mathbf{R}\in\{\mathbf{D}, \mathbf{P}, {\overline{\mathbf{P}}}\}$. If the relation between $\mathcal{O}_X$ and $\mathcal{O}_V$ is $\mathbf{D}(\mathcal{O}_X, \mathcal{O}_V)$, $\mathbf{P}(\mathcal{O}_X, \mathcal{O}_V)$, or $\overline{\mathbf{P}}(\mathcal{O}_X, \mathcal{O}_V)$, following three transition functions will optimise $\mathcal{O}_X$ to reach the target relation with $\mathcal{O}_V$. 
\begin{eqnarray*}
&&\Delta_\mathbf{D}^{\neg\mathbf{D}}(\mathcal{O}_X, \mathcal{O}_V)\triangleq\max\{0, dis_{X,V} - r_V - r_X +\epsilon\}\\
&&\Delta_\mathbf{P}^{ \neg\mathbf{P}}(\mathcal{O}_X, \mathcal{O}_V)\triangleq\max\{0, r_V  - dis_{X,V} - r_X +\epsilon \}\\
&&\Delta_{\overline{\mathbf{P}}}^{\neg\overline{\mathbf{P}}}(\mathcal{O}_X, \mathcal{O}_V)\triangleq\max\{0, r_X  - dis_{X,V} - r_V +\epsilon \}
\end{eqnarray*}

\begin{figure}[t] 
\centering
\includegraphics[width=1\textwidth]{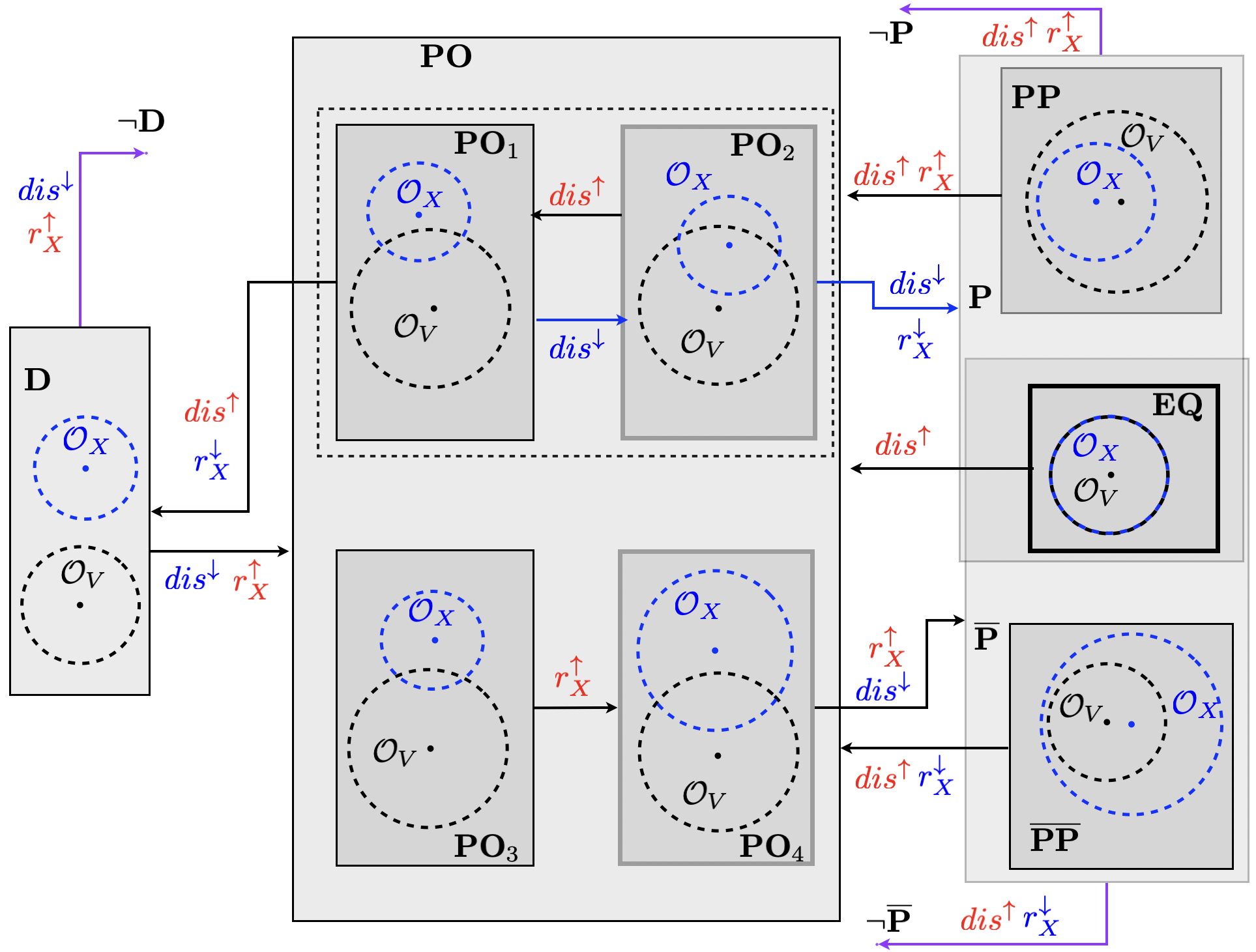} \caption{The full version of the neuro-symbolic transition map of neighbourhood spatial relations. $\mathcal{O}_V$ is fixed, $dis$ is shortened for $dis_{X,V}$. Here, $\mathbf{EQ}$ is set as the initial status, and only takes $\mathbf{PO}$ as its neighbour. Two ways to partition $\mathbf{PO}$: (1) $\mathbf{PO}$ is partitioned into $\mathbf{PO}_1$ and $\mathbf{PO}_2$, and (2) $\mathbf{PO}$ is partitioned into $\mathbf{PO}_3$ and $\mathbf{PO}_4$. Allowed operations between neighbourhood relations are labelled. The target relation determines which of them can be chosen.}
\label{nmap}
\end{figure} 
 
\subsection{Target-oriented spatial partition}
 
For syllogistic reasoning, there are six target spatial relations $\mathcal{T}\triangleq\{\mathbf{D}, \neg\mathbf{D},\mathbf{P},\neg\mathbf{P},$ $\overline{\mathbf{P}},\neg\overline{\mathbf{P}}\}$, and a target relation determines the qualitative partition of the space. For a negative target, e.g.,   $\neg\mathbf{D}, \neg\mathbf{P}$, and $ \neg\overline{\mathbf{P}}$, $\mathcal{S}$phNN only needs to partition the space into two parts: $\mathbf{D}$ and $\neg\mathbf{D}$, or $\mathbf{P}$ and $\neg\mathbf{P}$, or $\overline{\mathbf{P}}$ and $\neg\overline{\mathbf{P}}$. Each case only needs one transition function $\Delta_\mathbf{D}^{\neg\mathbf{D}}(\mathcal{O}_X, \mathcal{O}_V)$, $\Delta_\mathbf{P}^{\neg\mathbf{P}}(\mathcal{O}_X, \mathcal{O}_V)$, and $\Delta_{\overline{\mathbf{P}}}^{\neg\overline{\mathbf{P}}}(\mathcal{O}_X, \mathcal{O}_V)$, respectively. When the target relation is $\mathbf{D}$, $\mathbf{P}$, or $\overline{\mathbf{P}}$, the space will be partitioned hierarchically into two layers, at the top layer are five {\em jointly-exhaustive-and-pairwise-disjoint} relations: $\{\mathbf{D}, \mathbf{EQ}, \mathbf{PO}, \overline{\mathbf{PP}}$, $ \mathbf{PP}\}=\mathcal{T}_5$; at the second layer,  $\mathbf{PO}$ will be partitioned either into $\mathbf{PO}_1$ and  $\mathbf{PO}_2$, or  $\mathbf{PO}_3$ and  $\mathbf{PO}_4$. The transition map is thus hierarchical, as illustrated in Figure~\ref{nmap}.
The whole can be organised into a neuro-symbolic map for the transition of neighbourhood relations and formalised as a six-tuple  $\mathcal{M}\triangleq (\mathcal{T}, f_{tsp}, \mathcal{I},  \mathcal{S},  f_{tn}, \Delta)$. 
\begin{itemize}
    \item $\mathcal{T}$: the set of six target relations;
    \item  $f_{tsp}$: the function that maps a target relation to a set of {\em jointly-exhaustive-and-pairwise-disjoint} qualitative spatial partitions, where ${tsp}$ stands for {\em t}arget-oriented {\em s}patial {\em p}artitions. For example, $f_{tsp}(\mathbf{D})\triangleq \{\mathbf{D}, \mathbf{EQ},  \mathbf{PO}_1, \mathbf{PO}_2, \overline{\mathbf{PP}}, \mathbf{PP}\}$;
    \item $\mathcal{I}$:  a family of inspection functions. Let $\mathcal{O}_1$ and $\mathcal{O}_2$ be two spheres. We distinguish three kinds of inspection functions.
    \begin{enumerate}
        \item inspecting relations with an explicit target relation. $\mathcal{I}(\mathcal{O}_1, \mathcal{O}_2| \mathbf{T})$ returns the relation $\mathbf{R}\in f_{tsp}(\mathbf{T})$ and $\mathbf{T}\in \mathcal{T}$;
        \item inspecting relations with default target relations. $\mathcal{I}(\mathcal{O}_1, \mathcal{O}_2)$ returns the relation $\mathbf{R}\in\mathcal{T}_5=\{\mathbf{D}, \mathbf{EQ},  \mathbf{PO}, \overline{\mathbf{PP}}, \mathbf{PP}\}$;
        \item inspecting whether a given relation holds.  $\mathcal{I}^{\mathbf{R}}(\mathcal{O}_1, \mathcal{O}_2)$ returns 0 if $\mathbf{R}(\mathcal{O}_1, \mathcal{O}_2)$ holds, otherwise, returns a positive real number.
    \end{enumerate} 
    
    \item $\mathcal{S}$: the set of all relations between two spheres. $\mathcal{S}\triangleq\bigcup f_{tsp}(\mathbf{T})$, $\mathbf{T}\in\mathcal{T}$. $\mathcal{S}$ is closed for transposition relations. That is, for any $\mathbf{R}\in\mathcal{S}$, $\overline{\mathbf{R}}\in\mathcal{S}$; 
    \item $f_{tn}$: the function maps the current relation $\mathbf{R}$ to its neighbourhood relation $\mathbf{R}'$, towards the target $\mathbf{T}$, namely, $\mathbf{R}'=f_{tn}(\mathbf{T}, \mathbf{R})$, $tn$ stands for {\em t}arget-oriented {\em n}eighbourhood; 
     \item $\Delta$: the set of neighbourhood transition functions. Let $\mathbf{T}\in\mathcal{T}$ be the target relation and let $\mathcal{O}_1$ and $\mathcal{O}_2$ be two spheres.  $\mathbf{R}=\mathcal{I}(\mathcal{O}_1,\mathcal{O}_2| \mathbf{T})$, where $\mathbf{R}\in f_{tsp}(\mathbf{T})$. The neighbourhood transition function will be  $\Delta^\mathbf{T}_{\mathbf{R}:f_{tn}(\mathbf{T}, \mathbf{R})}(\mathcal{O}_1,\mathcal{O}_2)$ or $\Delta^\mathbf{T}_{\mathbf{R}}(\mathcal{O}_1,\mathcal{O}_2)$ for short. All neighbourhood transition functions are listed in Table~\ref{trt_tb}.
\end{itemize}

\begin{table}[!t]
\caption{Target oriented neighbourhood transition table. `$\emptyset$' means that the target relation is reached; `-' means that the current relation is not in the domain of spatial partition.}
\label{trt_tb}
\centering
\hspace*{-1.2em}
\scalebox{1}{
\begin{tabular}{l|c|c|c|c|c|c}
\hline 
$(\mathcal{O}_X, \mathcal{O}_V)$&
$\mathbf{D}$  & $\mathbf{P}$ & $\overline{\mathbf{P}}$ &$\neg\mathbf{D}$  & $\neg\mathbf{P}$ & $\neg\overline{\mathbf{P}}$ \\ 
\hline 
$\mathbf{D}$ & $\emptyset$ &  $\Delta^{\mathbf{P}}_{\mathbf{D}:\mathbf{PO}}$  & $\Delta^{\overline{\mathbf{P}}}_{\mathbf{D}:\mathbf{PO}}$  &$\Delta^{\neg\mathbf{D}}_\mathbf{D}$ & $\emptyset$ & $\emptyset$ \\ 
$\mathbf{PO}_1$ & $\Delta_{\mathbf{PO}_1}^\mathbf{D}$ & {\color{red}{$\Delta$}}$_{\mathbf{PO}_1:\mathbf{PO}_2}^{\mathbf{P}}$   & - & $\emptyset$ & $\emptyset$& $\emptyset$\\ 
$\mathbf{PO}_2$ &$\Delta_{\mathbf{PO}_2:\mathbf{PO}_1}^{\mathbf{D}}$ &  $\Delta_{\mathbf{PO}_2}^{\mathbf{P}}$  & -  & $\emptyset$ & $\emptyset$ & $\emptyset$ \\
$\mathbf{PO}_3$ & - & -  & $\Delta_{\mathbf{PO}_3:\mathbf{PO}_4}^{\overline{\mathbf{P}}}$  & $\emptyset$ & $\emptyset$ & $\emptyset$ \\ 
$\mathbf{PO}_4$ &- &  - & $\Delta_{\mathbf{PO}_4}^{\overline{\mathbf{P}}}$ & $\emptyset$ & $\emptyset$ & $\emptyset$ \\
$\mathbf{PP}$ & $\Delta_{\mathbf{PP}:\mathbf{PO}}^{\mathbf{D}}$ &  $\emptyset$   & $\Delta_{\mathbf{PP}:\mathbf{PO}}^{\overline{\mathbf{P}}}$   & $\emptyset$ & - & $\emptyset$ \\
$\mathbf{EQ}$ & $\Delta_{\mathbf{EQ}:\mathbf{PO}}^{\mathbf{D}}$ &  $\emptyset$  &  $\emptyset$  & $\emptyset$ & - & -
\\
$\overline{\mathbf{PP}}$ & $\Delta_{\overline{\mathbf{PP}}:\mathbf{PO}}^{\mathbf{D}}$ &  $\Delta_{\overline{\mathbf{PP}}:\mathbf{PO}}^{\mathbf{P}}$   & $\emptyset$  & $\emptyset$ & $\emptyset$&$-$\\
$\mathbf{P}$ & - &  $\emptyset$   & -  & $\emptyset$ & $\Delta^{\neg\mathbf{P}}_\mathbf{P}$& - \\
$\overline{\mathbf{P}}$ & - &  -   & $\emptyset$  & $\emptyset$ & - & $\Delta^{\neg\overline{\mathbf{P}}}_{\overline{\mathbf{P}}}$\\
$\neg\mathbf{D}$ & - &  -   & -& $\emptyset$ & -& -\\
$\neg\mathbf{P}$ & - &  -   & -  & - & $\emptyset$& -\\
$\neg\overline{\mathbf{P}}$& - &  -   & -  & - & -& $\emptyset$\\
\hline
\end{tabular}
}
\end{table}

\newpage
\section{Sequential control processes}
\label{procedure}

\begin{quote} 
{\em 
When we recall that the process will generally be concerned with finding
a satisfactory design, rather than an optimum design, we see that sequence and the division of labor between generators and tests can affect
not only the efficiency with which resources for designing are used but
also the nature of the final design as well.
    
    \hfill --- Herbert A. Simon  \cite{Simon19}
     } 
\end{quote}

The control process that realises deterministic reasoning is a process to determine (1)~the start and the end of the construction process; (2)~the dynamic and static spheres; and (3)~the current neighbourhood transition. This process allows us to prove the existence of the maximum iteration number $M$ and to identify the value of $M$ if the target configuration exists. Thus, this control process is ``good enough'' according to Herbert A. Simon's criterion \cite{Simon19}. 
We inspire decision-making of the control process from the Cumulative Prospect Theory \cite{Amos92}: Instead of using an absolute magnitude of welfare for decision-making, people prefer relative reference points to measure the change in values; people are more sensitive to losses than gains of the same magnitude. Thus, between two choices with the same gain, people will choose the one that won't incur losses. Decision-making in abstract domains, e.g., economics, has its root in the spatial domain \cite{tversky19}.  $\mathcal{S}$phNN does not measure the loss with respect to the final target but measures whether closer to the neighbourhood (Control process 1).  $\mathcal{S}$phNN will not improve the relations of a sphere with two other spheres, as this may cause the loss of one relation, which equals the gain of the other relation. Instead, $\mathcal{S}$phNN first improves one relation till it is satisfied, then improves both relations. If the already satisfied relation is impaired, $\mathcal{S}$phNN will repair it immediately (Control process 2).

\subsection{Control process 1: neighbourhood transition without constraints}
\label{control_1}

Let $p_1$, \dots, $p_{N-1}$ be $N-1$ premises of a long-chained syllogistic reasoning, where $p_i$ can be either $r_i(X_i,\, X_{i+1})$ or $r_i(X_{i+1},\, X_{i})$, $r_i\in\{all,\, some,\, no,\, some\_not\}$. Without loss of generality, they can be spatialised into $N-1$ spatial statements $\psi_1(\mathcal{O}_1,\mathcal{O}_2)$, \dots, $\psi_{N-1}(\mathcal{O}_{N-1},\, \mathcal{O}_{N})$, where 
\begin{eqnarray*}
  \psi_i= 
 \begin{cases}
    \psi(r_i) &  r_i(X_i,\,X_{i+1})    \\
    \overline{\psi(r_i)}      &  r_i(X_{i+1},\, X_i)
\end{cases}
\end{eqnarray*}
$\psi_i\in\{\mathbf{D}, \mathbf{P}, \overline{\mathbf{P}}, \neg\mathbf{D}, \neg\mathbf{P}, \neg\overline{\mathbf{P}}\}$. 
It is easy to construct $N-1$ spheres satisfying $\psi_1(\mathcal{O}_1,\,\mathcal{O}_2)$, \dots, $\psi_{N-1}(\mathcal{O}_{N-1},\, \mathcal{O}_{N})$ in one epoch as follows: We initialise $N-1$ coincided spheres; then, we fix the first one and move the second one till the relation $\psi_1(\mathcal{O}_{1}, \mathcal{O}_{2})$ is satisfied; then we fix the second sphere and move the third sphere till the relation $\psi_2(\mathcal{O}_{2}, \mathcal{O}_{3})$ is satisfied, \dots. Each step can be designed as a gradual descent process (Corollary~\ref{Cly1}).  
This describes the first control process, namely {\em neighbourhood transformation without constraints}. It determines the neighbourhood relation of the current relation and triggers a $\Delta$ function (in Table~\ref{trt_tb}) that gradually optimises the size and the location of a sphere to reach the neighbourhood relation. After one step of optimisation, the control process will inspect whether the neighbourhood relation is reached. This process repeats till the target is reached, as outlined in Algorithm~\ref{algo:control1}.
\begin{algorithm}[!t]
\DontPrintSemicolon 
\KwIn{Two spheres $\mathcal{O}_{X}$ and $\mathcal{O}_{V}$, the target relation:  $\mathbf{T}(\mathcal{O}_{X},\mathcal{O}_{V})$, $\mathcal{O}_{V}$ fixed.}   
\KwOut{Two spheres $\mathcal{O}_{X}$ and $\mathcal{O}_{V}$, satisfying $\mathbf{T}(\mathcal{O}_{X},\mathcal{O}_{V})$} 
$\mathbf{T}_1(\mathcal{O}_{X},\mathcal{O}_{V})\gets$   $\mathcal{I}(\mathcal{O}_{X}, \mathcal{O}_{V}|\mathbf{T})$; \\
\While{$\mathbf{T}_1(\mathcal{O}_{X},\mathcal{O}_{V}) \neq \mathbf{T}(\mathcal{O}_{X},\mathcal{O}_{V})$}
{
one step $\Delta_{\mathbf{T}_1:\mathbf{T}_2}^{\mathbf{T}}(\mathcal{O}_{X}, \mathcal{O}_{V})$;\algorithmiccomment{Corollary~\ref{mono_delta}}\\
$\mathbf{T}_1(\mathcal{O}_{X},\mathcal{O}_{V})\gets$ $\mathcal{I}(\mathcal{O}_{X}, \mathcal{O}_{V}|\mathbf{T})$;
} 
\Return{$\mathcal{O}_{X}$ and $\mathcal{O}_{V}$ 
}
\caption{The procedure to realise  $\mathbf{T}(\mathcal{O}_{X},  \mathcal{O}_{V})$}
\label{algo:control1} 
\end{algorithm} 
Without constraints, $\mathcal{S}$phNN can realise every neighbourhood transition. This follows that $\mathcal{S}$phNN can correctly construct an Euler diagram (in the form of a sphere configuration) for non-cyclic syllogistic statements. We formally describe this in Theorem~\ref{lmanoncyc}.

\begin{Cly}
\label{mono_delta}
Each $\Delta$ function is {\rm linear} concerning the radius and {\rm monotonic} concerning the distance between the centres.  
\end{Cly}
\begin{Thm}
Let $p_1$, \dots, $p_{N-1}$ be $N-1$ premises of a long-chained syllogistic reasoning system, where $p_i$ can be either $r_i(X_i,X_{i+1})$ or $r_i(X_{i+1}, X_{i})$, $(1\le i\le N-1)$, $r_i\in\{all, some, no, some\_not\}$. $\mathcal{S}$phNN can construct a configuration of $N$ spheres, $\mathcal{O}_{i}$ ($i=1\dots N$), as an Euler diagram of the $N-1$ syllogistic statements, such that $X_i$ maps to $\mathcal{O}_{i}$, and $p_i$ maps to $\psi_i(\mathcal{O}_i, \mathcal{O}_{i+1})$, where $\psi_i = \psi(r_i)$ if $r_i(X_i,X_{i+1})$ or $\psi_i = \overline{\psi(r_i)}$ if $r_i(X_{i+1}, X_i)$,  where $\psi_i\in\{\mathbf{D}, \mathbf{P}, \overline{\mathbf{P}}, \neg\mathbf{D},$ $ \neg\mathbf{P}, \neg\overline{\mathbf{P}}\}$.
\end{Thm}


\subsection{Control process 2:   neighbourhood transition with constraints}
\label{neural_cop}

\begin{quote}
    {\em  What we will put to you, then, is an interactionist approach to reason that contrasts with standard intellectualist approaches. 
   \begin{flushright} --- Mercier and Sperber \cite{enigma2018}
        \end{flushright}
    }
\end{quote}

Let $p_1$, \dots, $p_{N-1} \therefore q$ be a long-chained syllogistic reasoning with $N-1$ premises, where $p_i$ can be either $r_i(X_i,X_{i+1})$ or $r_i(X_{i+1}, X_{i})$, $q$ is fixed to the form $r_n(X_1,X_{N})$, $(1\le i\le N)$, where $r_i\in\{all, some, no, some\_not\}$. Without loss of generality, $p_1$, \dots, $p_{N-1}\, \therefore\, q$ can be spatialised into $N$ spatial statements $\psi_1(\mathcal{O}_1,\mathcal{O}_2)$, \dots, $\psi_{N-1}(\mathcal{O}_{N-1}, \mathcal{O}_{N})$, $\psi_{N}(\mathcal{O}_{N}, \mathcal{O}_{1})$, where $\psi_i = \psi(r_i)$ if $r_i(X_i,X_{j})$ or $\psi_i = \overline{\psi(r_i)}$ if $r_i(X_{j}, X_i)$,  and $\psi_i\in\{\mathbf{D}, \mathbf{P}, \overline{\mathbf{P}}, \neg\mathbf{D},$ $ \neg\mathbf{P}, \neg\overline{\mathbf{P}}\}$. $\mathcal{S}$phNN determines the {\em validity} of the reasoning by trying to construct a counter-example, namely, to construct a sphere configuration satisfying $N$ relations 
$\psi_1(\mathcal{O}_1, \mathcal{O}_2)$ \dots $\psi_{N-1}(\mathcal{O}_{N-1}, \mathcal{O}_{N})$, and $\neg\psi_{N}(\mathcal{O}_{N},\mathcal{O}_{1})$, where $\psi_i\in\{\mathbf{D}, \mathbf{P}, {\overline{\mathbf{P}}},\neg\mathbf{D}, \neg\mathbf{P}, \neg{\overline{\mathbf{P}}}\}$. 

\begin{figure}[!t]
\centering
\includegraphics[width=1\textwidth]{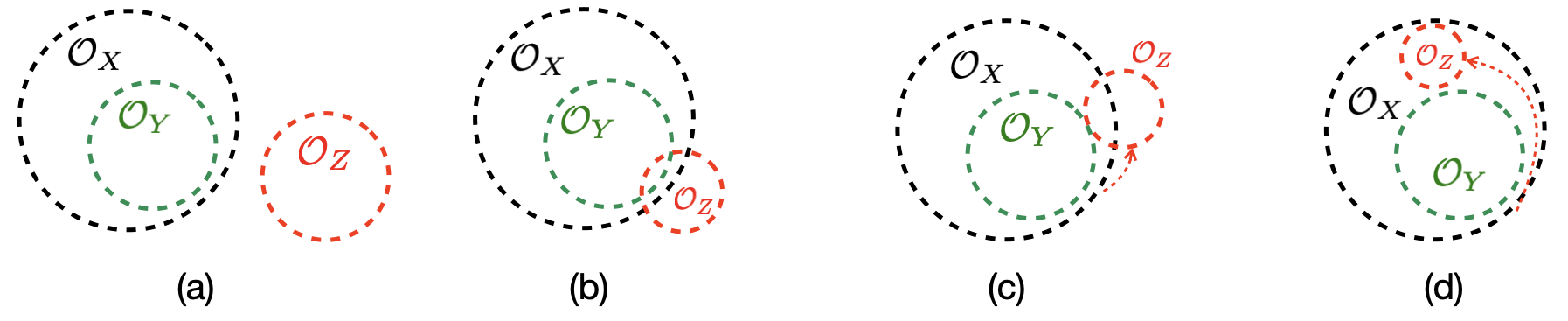} \caption{(a) $\mathcal{O}_Z$ disconnects from  $\mathcal{O}_X$; (b) when $\mathcal{O}_Z$ is approaching to $\mathcal{O}_X$, a non-zero local minimum will be reached; (c) $COP^{\mathbf{P}(\mathcal{O}_Z, \mathcal{O}_X)}_{\mathbf{D}(\mathcal{O}_Z, \mathcal{O}_Y)}(\mathcal{O}_{Z}| \mathcal{O}_{X}; \mathcal{O}_{Y})$ forces $\mathcal{O}_Z$ to rotate around $\mathcal{O}_Y$ to decrease the global loss; (d) $\mathcal{O}_Z$ successfully arrived at a location of a target configuration.}
\label{fig:cop0}
\end{figure} 

\begin{algorithm}[t]
\DontPrintSemicolon 
\KwIn{The target relations:  $\mathbf{T}_{ZX}(\mathcal{O}_{Z},\mathcal{O}_{X})$, $\mathbf{T}_{ZY}(\mathcal{O}_{Z},\mathcal{O}_{Y})$, $\mathbf{T}_{ZX}, \mathbf{T}_{ZY}\in\{\mathbf{P},\overline{\mathbf{P}},\neg\mathbf{P},\neg\overline{\mathbf{P}},\mathbf{D}, \neg\mathbf{D}\}$}   
\KwOut{$gLoss$, $\mathcal{O}_{Z}$} 
\mbox{Optimise $\mathcal{O}_{Z}$ to satisfy  $\mathbf{T}_{ZY}(\mathcal{O}_{Z},\mathcal{O}_{Y})$};\\
\mbox{ $\mathbf{S}_{ZX}(\mathcal{O}_{Z},\mathcal{O}_{X})\gets\mathcal{I}(\mathcal{O}_{Z},\mathcal{O}_{X}|\mathbf{T}_{ZX})$ and $\mathbf{S}_{ZY}(\mathcal{O}_{Z},\mathcal{O}_{Y})\gets\mathcal{I}(\mathcal{O}_{Z},\mathcal{O}_{Y}|\mathbf{T}_{ZY})$;}\\ 
$last\_gLoss\gets +\infty$; \\
$gLoss\gets \mbox{get the loss of }  \Delta^{\mathbf{T}_{ZX}}_{\mathbf{S}_{ZX}}(\mathcal{O}_{Z},\mathcal{O}_{X})$; \\
\While{$gLoss< last\_gLoss$}
{
$last\_gLoss\gets gLoss$;\algorithmiccomment{ln 5-12: Corollary~\ref{mono_delta}, Theorem~\ref{mono_cop}, Lemma~\ref{enn_thm0},  ~\ref{lma_2neg}}\\
one step gradual descent  $\Delta_{\mathbf{S}_{ZX}}^{\mathbf{T}_{ZX}}(\mathcal{O}_{Z}, \mathcal{O}_{X})+\Delta_{\mathbf{S}_{ZY}}^{\mathbf{T}_{ZY}}(\mathcal{O}_{Z},\mathcal{O}_{Y})$;\\
\While{$\Delta_{\mathbf{S}_{ZY}}^{\mathbf{T}_{ZY}}(\mathcal{O}_{Z}, \mathcal{O}_{Y})>0$}
{
Gradual descent  $\Delta_{\mathbf{S}_{ZY}}^{\mathbf{T}_{ZY}}(\mathcal{O}_{Z}, \mathcal{O}_{Y})$;\\
$\mathbf{S}_{ZY}(\mathcal{O}_{Z},\mathcal{O}_{Y})\gets\mathcal{I}(\mathcal{O}_{Z},\mathcal{O}_{Y}|\mathbf{T}_{ZY})$;\\
} 
$\mathbf{S}_{ZX}(\mathcal{O}_{Z},\mathcal{O}_{X})\gets\mathcal{I}(\mathcal{O}_{Z},\mathcal{O}_{X}|\mathbf{T}_{ZX})$;\\ 
$gLoss\gets\mbox{get the loss of } \Delta^{\mathbf{T}_{ZX}}_{\mathbf{S}_{ZX}}(\mathcal{O}_{Z},\mathcal{O}_{X})$;
}  
\Return{$gLoss$,  $\mathcal{O}_{Z}$}
\caption{Update $\mathcal{O}_{Z}$ to optimise its relation with $\mathcal{O}_{X}$ while preserving its relation with $\mathcal{O}_{Y}$.}
\label{algo:cop} 
\end{algorithm}
   
We first use Algorithm~\ref{algo:control1} to construct $N$ spheres satisfying all premises, and observe the relation between $\mathcal{O}_1$ and $\mathcal{O}_N$: If $\neg\psi_N(\mathcal{O}_{N}, \mathcal{O}_{1})$ holds, the current configuration is a target configuration. Otherwise, $\mathcal{S}$phNN tries to update  $\mathcal{O}_N$ to see whether the relation $\neg\psi_N(\mathcal{O}_{N}, \mathcal{O}_{1})$ can be satisfied without breaking the relation $\psi_{N-1}(\mathcal{O}_{N-1}, \mathcal{O}_{N})$.

We design a constraint optimisation process, whose function can be described as follows: Given  $\mathcal{O}_X$, $\mathcal{O}_Y$, and $\mathcal{O}_Z$, with target relations $\mathbf{T}_{XY}(\mathcal{O}_X, \mathcal{O}_Y)$, $\mathbf{T}_{YZ}(\mathcal{O}_Y, \mathcal{O}_Z)$, and $\mathbf{T}_{ZX}(\mathcal{O}_Z, \mathcal{O}_X)$, where $\mathbf{T}_{XY}, \mathbf{T}_{YZ}, \mathbf{T}_{ZX}\in\mathcal{T}=\{\mathbf{D}, \mathbf{P}, \overline{\mathbf{P}}, \neg\mathbf{D}, \neg\mathbf{P}, \neg\overline{\mathbf{P}}\}$. Suppose that  relations $\mathbf{T}_{XY}(\mathcal{O}_X, \mathcal{O}_Y)$ and $\mathbf{T}_{YZ}(\mathcal{O}_Y, \mathcal{O}_Z)$ are satisfied, this control process will fix the two spheres $\mathcal{O}_X$ and $\mathcal{O}_Y$, and optimises $\mathcal{O}_Z$ to satisfy the target relation $\mathbf{T}_{ZX}(\mathcal{O}_Z, \mathcal{O}_X)$, while keeping $\mathbf{T}_{YZ}(\mathcal{O}_Y, \mathcal{O}_Z)$, written as
$COP_{\mathbf{T}_{ZY}}^{\mathbf{T}_{ZX}}(\mathcal{O}_{Z}| \mathcal{O}_{X}; \mathcal{O}_{Y})$, or $COP(\mathcal{O}_{Z}| \mathcal{O}_{X}; \mathcal{O}_{Y})$ for short, where $\mathbf{T}_{ZY}$ is the transposition relation of $\mathbf{T}_{YZ}$. 
$\mathcal{S}$phNN optimises the relation between $\mathcal{O}_Z$ and $\mathcal{O}_X$ by rotating $\mathcal{O}_Z$  around $\mathcal{O}_Y$, as rotating a sphere around another sphere preserves their qualitative spatial relations (Corollary~\ref{topo_preserve}). Concretely,  
let $\mathbf{S}_{ZY}$ and $\mathbf{S}_{ZX}$ be the current inspected relations between $\mathcal{O}_{Z}$ and $\mathcal{O}_{Y}$, and between $\mathcal{O}_{Z}$ and $\mathcal{O}_{X}$, respectively, that is,  $\mathbf{S}_{ZY}=\mathcal{I}(\mathcal{O}_Z, \mathcal{O}_Y| \mathbf{T}_{ZY})$, $\mathbf{S}_{ZX}=\mathcal{I}(\mathcal{O}_Z, \mathcal{O}_X| \mathbf{T}_{ZX})$. $COP_{\mathbf{T}_{ZY}}^{\mathbf{T}_{ZX}}(\mathcal{O}_{Z}| \mathcal{O}_{X}; \mathcal{O}_{Y})$ gradually reduces the value of  $\Delta_{\mathbf{S}_{ZX}}^{\mathbf{T}_{ZX}}(\mathcal{O}_{Z}, \mathcal{O}_{X})+\Delta_{\mathbf{S}_{ZY}}^{\mathbf{T}_{ZY}}(\mathcal{O}_{Z}, \mathcal{O}_{Y})$; After each step, 
if $\mathbf{T}_{ZY}( \mathcal{O}_{Z},\mathcal{O}_{Y})$ is broken, that is, $\Delta_{\mathbf{S}_{ZY}}^{\mathbf{T}_{ZY}}(\mathcal{O}_{Z}, \mathcal{O}_{Y})>0$,  $\mathcal{O}_{Z}$ will be optimised to recover $\mathbf{T}_{ZY}( \mathcal{O}_{Z},\mathcal{O}_{Y})$ till $\Delta_{\mathbf{S}_{ZY}}^{\mathbf{T}_{ZY}}(\mathcal{O}_{Z}, \mathcal{O}_{Y})$ reaches zero (Theorem~\ref{lmanoncyc}). 

The whole process can be understood as such an interactive motion of $\mathcal{O}_Z$ that step-by-step improves the relation with $\mathcal{O}_X$ and immediately repairs the broken relation with $\mathcal{O}_Y$.  
For example, suppose that $\mathcal{O}_Z$ needs to be inside $\mathcal{O}_X$ and disconnect from $\mathcal{O}_Y$, shown in Figure~\ref{fig:cop0}(a). While $\mathcal{O}_Z$ optimises the relation with $\mathcal{O}_X$ by moving towards it, $\mathcal{O}_Z$ may connect with $\mathcal{O}_Y$ and get stuck at a non-zero global loss (due to the broken relation with  $\mathcal{O}_Y$), shown in Figure~\ref{fig:cop0}(b). $COP^{\mathbf{P}}_{\mathbf{D}}(\mathcal{O}_{Z}| \mathcal{O}_{X}; \mathcal{O}_{Y})$ will repair the broken relation first. If the central points of the spheres are not collinear,  $COP^{\mathbf{P}}_{\mathbf{D}}(\mathcal{O}_{Z}| \mathcal{O}_{X}; \mathcal{O}_{Y})$ will force $\mathcal{O}_{Z}$ rotating around $\mathcal{O}_{Y}$, and improve the relation with $\mathcal{O}_{X}$, shown in Figure~\ref{fig:cop0}(c, d). This procedure is outlined in Algorithm~\ref{algo:cop}. We prove that $COP_{\mathbf{T}_{ZY}}^{\mathbf{T}_{ZX}}(\mathcal{O}_{Z}| \mathcal{O}_{X}; \mathcal{O}_{Y})$ is gradual descent and will find a model if the target relation $\mathbf{T}_{ZY}$ and the constraint relation $\mathbf{T}_{ZX}$ are consistent, under the restriction that $\mathcal{O}_X$ and $\mathcal{O}_Y$ are fixed  (Theorem~\ref{mono_cop} and Theorem~\ref{thm_control2}).
\begin{Cly}
\label{topo_preserve}
For any spheres $\mathcal{O}_X$ and $\mathcal{O}_V$, rotating $\mathcal{O}_X$ around the centre of $\mathcal{O}_V$ preserves the qualitative spatial relation between them. 
\end{Cly} 
\begin{Thm} 
Let $\mathcal{O}_X$ and $\mathcal{O}_Y$ be two fixed non-concentric spheres; $\mathcal{O}_Z$ be a movable sphere; $\mathbf{T}_{ZY}$ and $\mathbf{T}_{ZX}$ be the target relations of $\mathcal{O}_Z$ to $\mathcal{O}_Y$ and $\mathcal{O}_X$, respectively, $\mathbf{T}_{ZY}, \mathbf{T}_{ZX}\in\mathcal{T}=\{\mathbf{D}, \mathbf{P}, \overline{\mathbf{P}}, \neg\mathbf{D}, \neg\mathbf{P}, \neg\overline{\mathbf{P}}\}$.  
$COP^{\mathbf{T}_{ZX}}_{\mathbf{T}_{ZY}}(\mathcal{O}_{Z}| \mathcal{O}_{X}; \mathcal{O}_{Y})$  
is monotonic. 
\end{Thm}

\begin{Thm} 
Let $\mathbf{R}_1$, $\mathbf{T}_2$, and $\mathbf{T}_3$ be {\em satisfiable}, where $\mathbf{R}_1\in\{\mathbf{D}, \mathbf{P},\mathbf{PO}, \overline{\mathbf{P}}\}$, $\mathbf{T}_2, \mathbf{T}_3\in\mathcal{T}=\{\mathbf{D}, \mathbf{P}, \overline{\mathbf{P}}, \neg\mathbf{D}, \neg\mathbf{P}, \neg\overline{\mathbf{P}}\}$. Let $\mathcal{O}_X$ and $\mathcal{O}_Y$ be fixed and satisfying $\mathbf{R}_1(\mathcal{O}_X, \mathcal{O}_Y)$.
$\mathcal{S}$phNN can construct $\mathcal{O}_Z$ such that  $\mathbf{T}_2(\mathcal{O}_Y, \mathcal{O}_Z)$, and $\mathbf{T}_3(\mathcal{O}_Z, \mathcal{O}_X)$.
\end{Thm}

\begin{figure}
\centering
\includegraphics[width=0.9\textwidth]{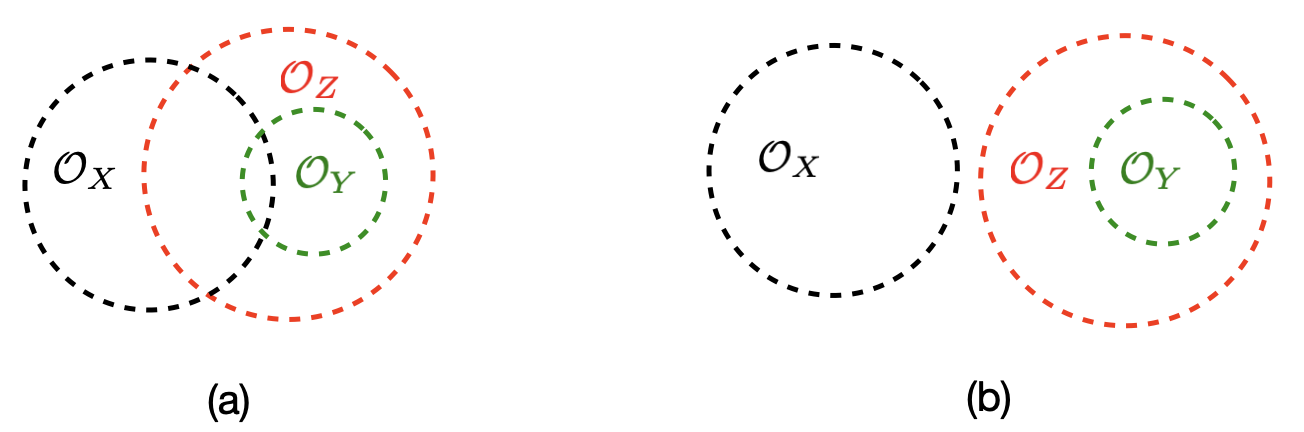} \caption{(a) $\mathcal{O}_X$ and $\mathcal{O}_Y$ are incorrectly fixed to the partial overlapping relation; (b) after restart the process by fixing $\mathcal{O}_Z$ at the beginning, the relation between $\mathcal{O}_X$ and $\mathcal{O}_Y$ will be correct.}
\label{fig:loop0}
\end{figure}

\begin{algorithm}[h]
\DontPrintSemicolon 
\KwIn{Three target relations: $\mathbf{T}_{12}(\mathcal{O}_{1},\mathcal{O}_{2})$, $\mathbf{T}_{23}(\mathcal{O}_{2},\mathcal{O}_{3})$, $\mathbf{T}_{31}(\mathcal{O}_{3},\mathcal{O}_{1})$, where $\mathbf{T}_{ij}\in\{\mathbf{P},\overline{\mathbf{P}},\neg\mathbf{P},\neg\overline{\mathbf{P}},\mathbf{D}, \neg\mathbf{D}\}$, $1\le i, j \le 3$.}  
\KwOut{$\mathtt{SAT}$ or $\mathtt{UNSAT}$} 
Initialise $\mathcal{O}_{1}$, $\mathcal{O}_{2}$, and $\mathcal{O}_{3}$ as being coincided;\\ 
\lIf{all three relations are satisfied} {\Return{$\mathtt{SAT}$}}
break coincide relations with a small random fluctuation;\\ 
fix $\mathcal{O}_{1}$, update $\mathcal{O}_{2}$ to satisfy $\mathbf{T}_{12}(\mathcal{O}_{1},\mathcal{O}_{2})=\overline{\mathbf{T}_{21}}(\mathcal{O}_{2},\mathcal{O}_{1})$; \algorithmiccomment{Algorithm~\ref{algo:control1}}\\ 
fix $\mathcal{O}_{1}$, update $\mathcal{O}_{3}$ to satisfy $\mathbf{T}_{31}(\mathcal{O}_{3},\mathcal{O}_{1})$; \algorithmiccomment{Algorithm~\ref{algo:control1}}\\ 
do $COP^{\mathbf{T}_{32}}_{\mathbf{T}_{31}}(\mathcal{O}_{3}| \mathcal{O}_{2}, \mathcal{O}_{1})$; \algorithmiccomment{Algorithm~\ref{algo:cop}}\\ 
\If{not all three relations are satisfied} {
fix $\mathcal{O}_{2}$, update $\mathcal{O}_{3}$ to satisfy $\overline{\mathbf{T}_{23}}(\mathcal{O}_{3},\mathcal{O}_{2})$; \algorithmiccomment{Algorithm~\ref{algo:control1}}\\ 
fix $\mathcal{O}_{2}$, update $\mathcal{O}_{1}$ to satisfy $\mathbf{T}_{12}(\mathcal{O}_{1},\mathcal{O}_{2})$; \algorithmiccomment{Algorithm~\ref{algo:control1}}\\ 
do $COP^{\mathbf{T}_{13}}_{\mathbf{T}_{12}}(\mathcal{O}_{1}| \mathcal{O}_{3}, \mathcal{O}_{2})$;\algorithmiccomment{Algorithm~\ref{algo:cop}}} 
\lIf{all three relations are satisfied}{
\Return{$\mathtt{SAT}$} \algorithmiccomment{Theorem~\ref{thmcyc}}}
\lElse{\Return{$\mathtt{UNSAT}$}}
\caption{$\mathcal{S}$phNN for Classic Syllogistic Reasoning $(\mathtt{S}_3)$}
\label{algo:neuro_geo} 
\end{algorithm}

\subsection{Control process 3:  neighbourhood transition with restart}
\label{spatialising_negation}

With the two control processes above, $\mathcal{S}$phNN may mistakenly choose an unintended positive relation to spatialise a negative one. We illustrate this using the following example: $\neg\mathbf{P}(\mathcal{O}_X, \mathcal{O}_Y)$ can be spatialised by three possible positive spatial relations, namely, $\mathbf{D}(\mathcal{O}_X, \mathcal{O}_Y)$, $\mathbf{PO}(\mathcal{O}_X, \mathcal{O}_Y)$, $\overline{\mathbf{PP}}(\mathcal{O}_X, \mathcal{O}_Y)$, but only the relation $\mathbf{D}(\mathcal{O}_X, \mathcal{O}_Y)$ is correct, if $\mathcal{O}_X$ disconnects from $\mathcal{O}_Z$, $\mathbf{D}(\mathcal{O}_X, \mathcal{O}_Z)$, and $\mathcal{O}_Y$ is part of $\mathcal{O}_Z$, $\mathbf{P}(\mathcal{O}_Y, \mathcal{O}_Z)$. If  $\neg\mathbf{P}$ is mistakenly spatialised as $\mathbf{PO}$, $\mathcal{S}$phNN will not be successful in constructing a target configuration. The remedy is to restart the process by choosing a different sphere as the first fixed sphere  (Lemma~\ref{enn_lm1}, \ref{enn_syl}). Because an unintended relation is inconsistent with the other two relations, this unintended relation cannot appear, when the other two relations are realised first. 
For example, $\mathbf{PO}(\mathcal{O}_{X},\mathcal{O}_{Y})$ is inconsistent with $\mathbf{D}(\mathcal{O}_{X},\mathcal{O}_{Z})$ and $\mathbf{P}(\mathcal{O}_{Y},\mathcal{O}_{Z})$. If we fix $\mathcal{O}_Z$, and realise $\mathcal{O}_X$ to disconnect from $\mathcal{O}_Z$, and realise $\mathcal{O}_Y$ to be part of  $\mathcal{O}_Z$, $\mathcal{O}_X$ will disconnect from $\mathcal{O}_Y$, as illustrated in Figure~\ref{fig:loop0}.  
We outline the control process with this remedy in Algorithm~\ref{algo:neuro_geo}.  
We prove that for any three satisfiable syllogistic statements, this algorithm can construct a sphere configuration with one epoch and a maximum of one restart (Theorem~\ref{thmcyc}).  
\begin{Thm}
Let $p_1$, $p_2$, $p_3$ be three syllogistic statements, where $p_1$ can be either $r_1(X_1,X_{2})$ or $r_1(X_{2}, X_{1})$, $p_2$ can be either $r_2(X_2,X_{3})$ or $r_2(X_{3}, X_{2})$, and $p_3$ can be either $r_3(X_1,X_{3})$ or $r_3(X_{3}, X_{1})$, $r_1, r_2, r_3\in\{all, some, no, some\_not\}$. $\mathcal{S}$phNN can determine the satisfiability of $p_1$, $p_2$, $p_3$  in the first epoch, with at most one restart. 
\end{Thm}

\subsection{$\mathcal{S}$phNN determines the validity of a long-chained syllogistic reasoning}
\label{chain_sym}

Let $p_1$, \dots, $p_{N-1} \therefore q$ be a long-chained syllogistic reasoning with $N-1$ premises, where $p_i$ can be either $r_i(X_i,X_{i+1})$ or $r_i(X_{i+1}, X_{i})$, $q$ is fixed to $r_n(X_1,X_{N})$, $r_i\in\{all, some, no, some\_not\}$. Without loss of generality, $p_1$, \dots, $p_{N-1} \therefore q$ can be spatialised into $N$ spatial statements $\psi_1(\mathcal{O}_1,\mathcal{O}_2)$, \dots, $\psi_{N-1}(\mathcal{O}_{N-1}, \mathcal{O}_{N})$, $\psi_{N}(\mathcal{O}_{N}, \mathcal{O}_{1})$, where $\psi_i = \psi(r_i)$ if $r_i(X_i,X_{j})$ or $\psi_i = \overline{\psi(r_i)}$ if $r_i(X_{j}, X_i)$,  and $\psi_i\in\{\mathbf{D}, \mathbf{P}, \overline{\mathbf{P}}, \neg\mathbf{D},$ $ \neg\mathbf{P}, \neg\overline{\mathbf{P}}\}$. 
It will be easy to construct $N-1$ spheres satisfying $\psi_1(\mathcal{O}_1,\mathcal{O}_2)$, \dots, $\psi_{N-1}(\mathcal{O}_{N-1}, \mathcal{O}_{N})$ in one epoch as follows: We initialise $N-1$ coincided spheres; then, we fix the first one and move the second one till the relation $\psi_1(\mathcal{O}_{1}, \mathcal{O}_{2})$ is satisfied; then we fix the second sphere and move the third sphere till the relation $\psi_2(\mathcal{O}_{2}, \mathcal{O}_{3})$ is satisfied, \dots. Each step can be designed as a gradual descent process (Corollary~\ref{Cly1}). 
This process repeats till the target is reached.
In this way, $\mathcal{S}$phNN can correctly construct an Euler diagram (in the form of a sphere configuration) for non-cyclic syllogistic statements. We formally describe this in Theorem~\ref{lmanoncyc}.


$\mathcal{S}$phNN determines the {\em validity} of the reasoning by trying to construct a counter-example, namely, to construct a sphere configuration satisfying $N$ relations 
$\psi_1(\mathcal{O}_1, \mathcal{O}_2)$ \dots $\psi_{N-1}(\mathcal{O}_{N-1}, \mathcal{O}_{N})$, and $\neg\psi_{N}(\mathcal{O}_{N},\mathcal{O}_{1})$, where $\psi_i\in\{\mathbf{D}, \mathbf{P}, {\overline{\mathbf{P}}},\neg\mathbf{D}, \neg\mathbf{P}, \neg{\overline{\mathbf{P}}}\}$. We show that $\mathcal{S}$phNN can construct a sphere configuration if the $N$ relations are satisfiable, with the worst computational complexity ${O}(N)$. 

Firstly, $\mathcal{S}$phNN initialises $N$ coincided spheres (all relations are $\mathbf{EQ}$). If this trivial configuration is a counter-example, the validity of the original reasoning is refuted. We prove that for satisfiable relations, if one relation must be the $\mathbf{EQ}$ relation, all relations will be the $\mathbf{EQ}$ relation. Thus, after refuting the trivial configuration (all relations are the $\mathbf{EQ}$ relation), $\mathcal{S}$phNN restricts the target spatial relations to be $\{\mathbf{D}, \mathbf{PP}, {\overline{\mathbf{PP}}},\neg\mathbf{D}, \neg\mathbf{PP}, \neg{\overline{\mathbf{PP}}}\}$. Because suppose that several $\mathbf{EQ}$ relations are in a satisfiable configuration, each $\mathbf{EQ}$ shall be replaced by a non-$\mathbf{EQ}$ relation. Otherwise, all relations must be $\mathbf{EQ}$ relations. 

\begin{sidewaysfigure} 
\centering
\includegraphics[width=1\textwidth]{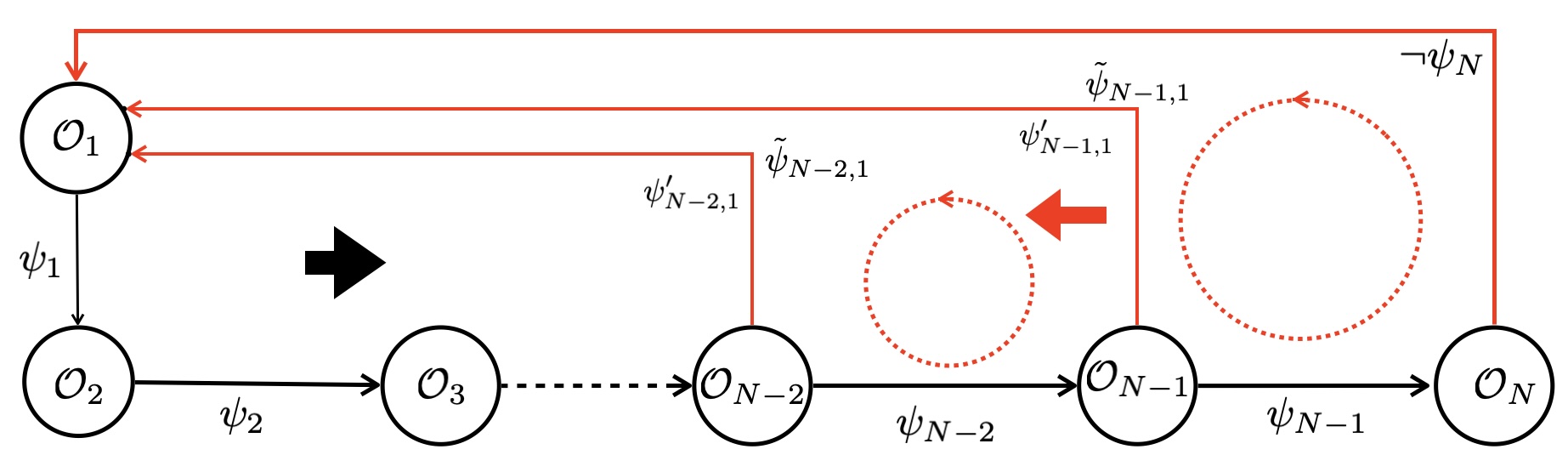}
\caption{Deterministic neural syllogistic reasoning with $N$ terms. {\bf The thick black arrow} is the forward construction process, starting from $\mathcal{O}_1$, constructing $\mathcal{O}_2$ that satisfies the relation $\psi_1$ to $\mathcal{O}_1$ \dots, till $\mathcal{O}_N$ that satisfies the relation $\psi_{N-1}$ to $\mathcal{O}_{N-1}$. If fixing $\mathcal{O}_1$ and $\mathcal{O}_{N-1}$, $\mathcal{O}_N$ satisfies the relations $\neg\psi_N$ and $\psi_{N-1}$, the original reasoning will be determined as 
being invalid; otherwise,  $\neg\psi_N$ and $\psi_{N-1}$ will determine the relation between $\mathcal{O}_1$ and $\mathcal{O}_{N-1}$, namely, $\psi'_{N-1,1}$ and trigger the backward updating process, as {\color{red}{\bf the thick red arrow}} shows.}
\label{nsnn}
\end{sidewaysfigure}

\begin{algorithm}[!t]
\SetKwComment{Comment}{/* }{ */}
\newcommand{\graycomments}[1]{\footnotesize\ttfamily\textcolor{gray}{#1}}
\SetCommentSty{graycomments}
\DontPrintSemicolon\KwIn{$N$ target relations: $\mathbf{T}_{1,2}(\mathcal{O}_{1},\mathcal{O}_{2})$\dots $\mathbf{T}_{N-1,N}(\mathcal{O}_{N-1},\mathcal{O}_{N})$, $\mathbf{T}_{N,1}(\mathcal{O}_{N},\mathcal{O}_{1})$, $\mathbf{T}_i\in\{\mathbf{P},\overline{\mathbf{P}},\neg\mathbf{P},\neg\overline{\mathbf{P}},\mathbf{D}, \neg\mathbf{D}\}$}  
\KwOut{$\mathtt{SAT}$ or $\mathtt{UNSAT}$ 
\algorithmiccomment{return $\mathtt{SAT}$ if $N$ relations are satisfiable.}} 
Initialise all $\mathcal{O}_{i}$ as being coincided $(1\le i\le N)$;\\ 
\lIf{$N$ coincided spheres satisfy the $N$ relations}{\Return{$\mathtt{SAT}$};}
\lIf{$N == 3$} {\Return {$\mathtt{S}_{3}$ for $\mathbf{T}_{1,2}(\mathcal{O}_{1},\mathcal{O}_{2})$, $\mathbf{T}_{2,3}(\mathcal{O}_{2},\mathcal{O}_{3})$, $\mathbf{T}_{3,1}(\mathcal{O}_{3},\mathcal{O}_{1})$};}
\For{$i=1\dots N-1$}{
fix $\mathcal{O}_{i}$, update $\mathcal{O}_{i+1}$ to satisfy $\mathbf{T}_{i,i+1}(\mathcal{O}_{i},\mathcal{O}_{i+1})$;\algorithmiccomment{Algorithm~\ref{algo:control1}}}  
do $COP^{\mathbf{T}_{N,1}}_{\mathbf{T}_{N,N-1}}(\mathcal{O}_{N}| \mathcal{O}_{1}, \mathcal{O}_{N-1})$; \algorithmiccomment{Algorithm~\ref{algo:cop}}\\
\lIf{$\mathbf{T}_{N,1}(\mathcal{O}_{N}, \mathcal{O}_{1})$}{\Return{$\mathtt{SAT}$};}
$\mathbf{R}\gets$ $\mathcal{I}(\mathcal{O}_{1},\mathcal{O}_{N-1})$, $\mathbf{R}\in\mathcal{T}_5$;\\
fix $\mathcal{O}_{1}$, update $\mathcal{O}_{N}$ to satisfy $\mathbf{T}_{N,1}(\mathcal{O}_{N},\mathcal{O}_{1})$; \algorithmiccomment{Algorithm~\ref{algo:control1}}\\
fix $\mathcal{O}_{N}$, update $\mathcal{O}_{N-1}$ to satisfy $\mathbf{T}_{N-1,N}(\mathcal{O}_{N-1},\mathcal{O}_{N})$; \algorithmiccomment{Algorithm~\ref{algo:control1}}\\
$i\gets N-1$;\\
\While{$i>3$}{
$\mathbf{R}'\gets$ $\mathcal{I}(\mathcal{O}_{1},\mathcal{O}_{i})$, $\mathbf{R}'\in\mathcal{T}_5$;\\
\lIf{$\mathbf{R}'\in\{\mathbf{D}, \mathbf{PP}, \overline{\mathbf{PP}}\}$} 
{$\mathbf{T}_{1,i}\gets\mathbf{R}'$}
\lElseIf{$\mathbf{R}==\mathbf{D}$}{$\mathbf{T}_{1,i}\gets\neg\mathbf{D}$}
\lElseIf{$\mathbf{R}==\mathbf{PP}$}{$\mathbf{T}_{1,i}\gets\neg\mathbf{P}$}
\lElseIf{$\mathbf{R}==\overline{\mathbf{PP}}$}{$\mathbf{T}_{1,i}\gets\neg\overline{\mathbf{P}}$}
fix $\mathcal{O}_{1}$, update $\mathcal{O}_{i}$ to satisfy $\mathbf{T}_{1,i}(\mathcal{O}_{1},\mathcal{O}_{i})$; \algorithmiccomment{Algorithm~\ref{algo:control1}}\\
\lIf{$\mathbf{T}_{i-1,i}(\mathcal{O}_{i-1},\mathcal{O}_{i})$ holds}{\Return{$\mathtt{SAT}$};}
$\mathbf{R}\gets$ $\mathcal{I}(\mathcal{O}_{1},\mathcal{O}_{i-1})$, $\mathbf{R}\in\mathcal{T}_5$;\\
fix $\mathcal{O}_{i}$, update $\mathcal{O}_{i-1}$ to satisfy $\mathbf{T}_{i-1,i}(\mathcal{O}_{i-1},\mathcal{O}_{i})$; \algorithmiccomment{Algorithm~\ref{algo:control1}}\\
$i\gets i -1$
}
\Return{$\mathtt{S}_3$ of $\mathbf{T}_{1,2}(\mathcal{O}_{1},\mathcal{O}_{2})$, $\mathbf{T}_{2,3}(\mathcal{O}_{2},\mathcal{O}_{3})$, and $\mathbf{R}(\mathcal{O}_{1},\mathcal{O}_{3})$}  
\caption{$\mathcal{S}$phNN for long Syllogistic Reasoning $(\mathtt{S}_N)$}
\label{algo:neuro_geoN} 
\end{algorithm}

Suppose that $\mathcal{S}$phNN has successfully constructed $N$ spheres $\mathcal{O}_1, \dots, \mathcal{O}_N$ satisfying $\psi_{1}(\mathcal{O}_{1},\mathcal{O}_{2}),\dots, \psi_{N-1}(\mathcal{O}_{N-1},\mathcal{O}_{N})$ (Theorem~\ref{lmanoncyc}). Let $\tilde{\psi}_{N-1,1}$ be the {\em inspected} spatial relation between $\mathcal{O}_{N-1}$ and $\mathcal{O}_{1}$, that is, $\tilde{\psi}_{N-1,1}\in\{\mathbf{D}, \mathbf{PO}, \mathbf{PP}, {\overline{\mathbf{PP}}}\}$. 
$\mathcal{S}$phNN determines the satisfiability of the three spheres $\mathcal{O}_{N-1}$, $\mathcal{O}_{1}$, and $\mathcal{O}_{N}$ with the relations
$\overline{\tilde{\psi}_{N-1,1}}(\mathcal{O}_{1},\mathcal{O}_{N-1})$, $\psi_{N-1}(\mathcal{O}_{N-1},\mathcal{O}_{N})$, and $\neg\psi_{N}(\mathcal{O}_{N},\mathcal{O}_{1})$ as follows: It fixes $\mathcal{O}_{1}$ and $\mathcal{O}_{N-1}$, and rotates $\mathcal{O}_{N}$ around $\mathcal{O}_{N-1}$ to reach the relation $\neg\psi_{N}(\mathcal{O}_{N},\mathcal{O}_{1})$. 
If $\neg\psi_{N}(\mathcal{O}_{N},\mathcal{O}_{1})$ is satisfied, $\mathcal{S}$phNN will find a counter-example. Otherwise, the relations $\overline{\tilde{\psi}_{N-1,1}}(\mathcal{O}_{1},\mathcal{O}_{N-1})$, $\psi_{N-1}(\mathcal{O}_{N-1},\mathcal{O}_{N})$, and $\neg\psi_{N}(\mathcal{O}_{N},\mathcal{O}_{1})$ are unsatisfiable, which follows that  $\psi_{N-1}(\mathcal{O}_{N-1},\mathcal{O}_{N})$ and $\neg\psi_{N}(\mathcal{O}_{N},\mathcal{O}_{1})$ will deduce the relation
$\neg\overline{\tilde{\psi}_{N-1,1}}(\mathcal{O}_{1},\mathcal{O}_{N-1})$. Accordingly,  
$\mathcal{S}$phNN starts the backward process by updating the relation between $\mathcal{O}_{1}$ and $\mathcal{O}_{N-1}$: It fixes $\mathcal{O}_1$ and updates $\mathcal{O}_N$ to satisfy  $\neg\psi_{N} (\mathcal{O}_N, \mathcal{O}_1)$, then fixes $\mathcal{O}_N$ and updates $\mathcal{O}_{N-1}$ to satisfy $\psi_{N-1}(\mathcal{O}_{N-1},\mathcal{O}_{N})$. After the two operations, the relation between $\mathcal{O}_{N-1}$ and $\mathcal{O}_{1}$ will be inspected and updated. Let  $\tilde{\psi}'_{N-1,1}$  be the inspected relation, $\tilde{\psi}'_{N-1,1}(\mathcal{O}_{N-1},\mathcal{O}_{1})$, where $\tilde{\psi}'_{N-1,1}\in\{\mathbf{D}, \mathbf{PP}, \overline{\mathbf{PP}}, \mathbf{PO}\}$. The relation between $\mathcal{O}_{N-1}$ and $\mathcal{O}_{1}$, $\psi'_{N-1,1}$, will be updated as follows.
$$
\psi'_{N-1,1}=
\begin{cases}
\tilde{\psi}'_{N-1,1}, \mbox{if }\tilde{\psi}'_{N-1,1}\in\{\mathbf{D}, \mathbf{PP}, {\overline{\mathbf{PP}}}\}\\
\neg\mathbf{D}, \mbox{if }\tilde{\psi}'_{N-1,1}=\mathbf{PO}\land\tilde{\psi}_{N-1,1}=\mathbf{D}\\
\neg\mathbf{PP}, \mbox{if }\tilde{\psi}'_{N-1,1}=\mathbf{PO}\land\tilde{\psi}_{N-1,1}=\mathbf{PP}\\
\neg{\overline{\mathbf{PP}}}, \mbox{if }\tilde{\psi}'_{N-1,1}=\mathbf{PO}\land\tilde{\psi}_{N-1,1}={\overline{\mathbf{PP}}}
\end{cases}
$$ 
In all cases, $\psi'_{N-1,1}$ is a syllogistic relation. This way, $\mathcal{S}$phNN 
reduces the task into the case of $N-1$. \dots If $\overline{\psi'_{2,1}}$ contradicts with $\psi_{1}$, $\mathcal{S}$phNN will conclude there is no counter-example and the original syllogistic reasoning is valid.  The construction process is illustrated in Figure~\ref{nsnn} and listed in Algorithm~\ref{algo:neuro_geoN} that allows us to prove {\em the principle of deterministic neural reasoning} Theorem~\ref{thmcyc1}.  
\setcounter{Thm}{4}
\begin{Thm}
(The principle of deterministic neural reasoning) Let $p_1$, \dots, $p_{N-1} \therefore q$ be a long-chained syllogistic reasoning with $N-1$ premises, where $p_i$ can be either $r_i(X_i,X_{i+1})$ or $r_i(X_{i+1}, X_{i})$ ($1\le i\le N-1$), $q$ is fixed to $r_n(X_1,X_{N})$, $r_j\in\{all, some, no, some\_not\}$ ($1\le j\le N$). 
$\mathcal{S}$phNN can determine its validity (or satisfiability) with the computational complexity of ${O}(N)$.
\end{Thm}

%% file: 9arxiv/proofs.tex
\begin{sidewaysfigure} 
\centering
\includegraphics[width=1
\textwidth]{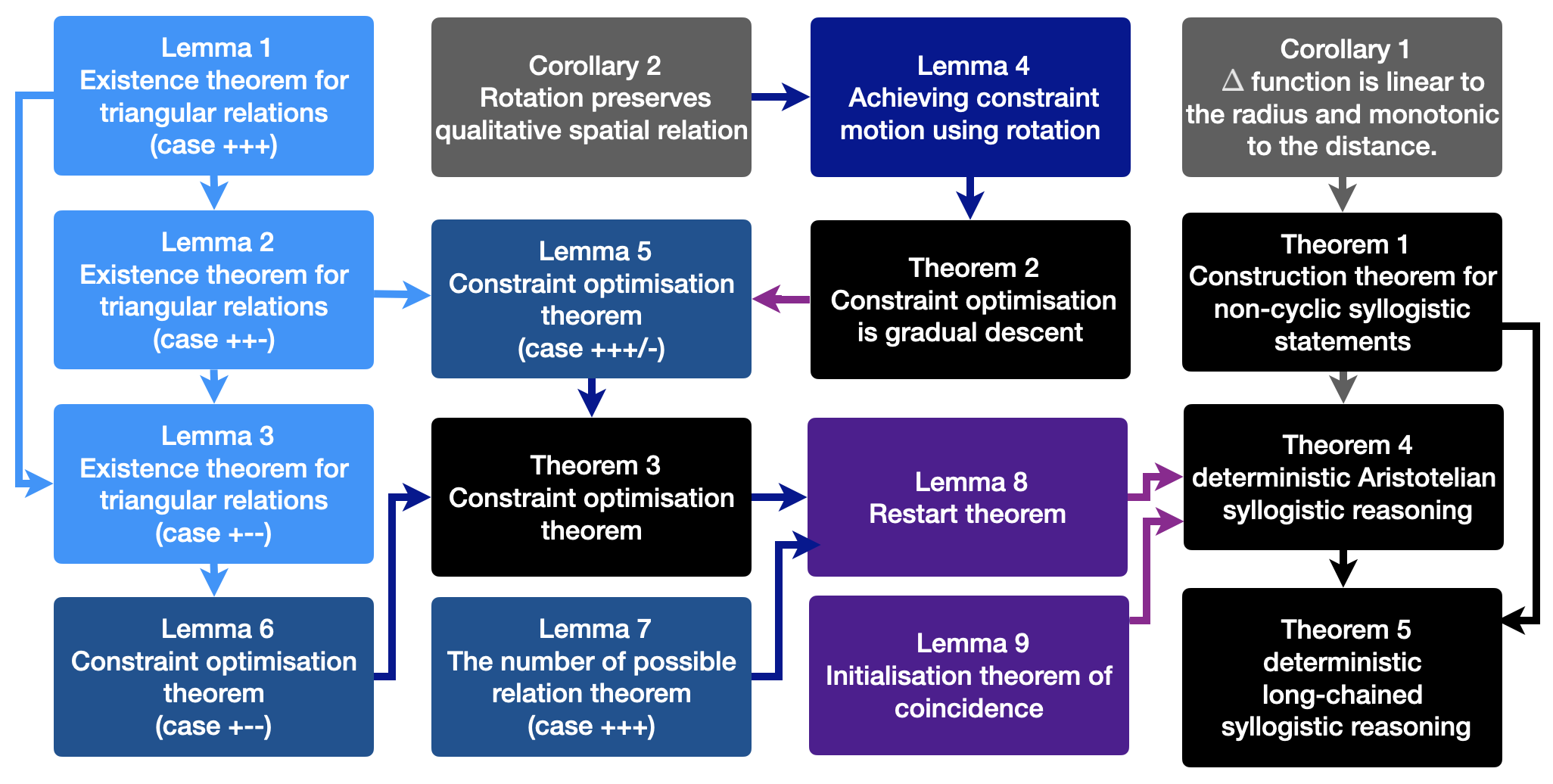}
\caption{The dependency relation among 16 theorems.  Theorem~\ref{thmcyc1} is the main theorem for {\bf the principle of deterministic neural reasoning} that $\mathcal{S}$phNN determines the validity (the satisfiability) of long-chained ($N\ge 3$) syllogistic reasoning in one epoch. Theorem~\ref{thmcyc} proves the case of Aristotelian syllogistic reasoning ($N=3$). The construction process initialises all spheres to be coinciding (The Initialisation  Lemma~\ref{thmeq}) and transforms to the target configuration by gradual descent operations (Theorem~\ref{mono_cop}), with a maximum of one restart (The Restart Lemma~\ref{enn_syl}). 
The constraint optimisation theorem guarantees the transformation process will gradually reduce the global loss without violating the constraints (Corollary~\ref{Cly2}
 and Lemma~\ref{rotate}); the restart theorem guarantees that if there is a satisfiable configuration of three spheres for any Aristotelian syllogistic reasoning, $\mathcal{S}$phNN can successfully construct by first constructing and fixing two spheres, with a maximum of one restart with another sphere 
 (Lemma~\ref{basic}-~\ref{lma_2neg_sat}, ~\ref{enn_thm0}-~\ref{enn_lm1}, where `$+$' and `$-$' represent a positive relation, e.g. $\mathbf{D}$, and a negative relation), e.g., $\neg\mathbf{P}$. Each $\Delta$ function (gradual descent function) of two spheres is linear to the radius and monotonic to the distance between the centres of spheres (Corollary~\ref{Cly1}), so gradual descent operation can construct sphere configuration for non-cyclic syllogistic statements in one epoch (Theorem~\ref{lmanoncyc}).}  
\label{theorems}
\end{sidewaysfigure}
  
\section{The proofs of the theorems}
\label{proofs}

\begin{quote} 
{\em 
It seems impossible to reach any definitive conclusions concerning human rationality in the absence of a detailed analysis of the sensitivity of the criterion and the cost involved in evaluating the alternatives.  
    
    \hfill --- Amos Tversky  \cite{Tversky69}
     } 
\end{quote}

We outline theorems and proofs whose dependency relations are diagrammed in Figure~\ref{theorems}. Proofs are independent of the dimension of spheres, so theorems hold for $n$-dimensional space ($n\ge 2$) (Corollary~\ref{thmdim}). With these theorems and proofs, $\mathcal{S}$phNN becomes the first neural model for deterministic logical reasoning.

\subsection{Basic theorems}
\label{basic_thms}
\setcounter{Cly}{0}
\begin{Cly}\label{Cly1}
Each $\Delta$ function is {\rm linear} concerning the radius and {\rm monotonic} concerning the distance between the centres. 
\end{Cly}
\setcounter{Prf}{0}
\begin{PrfCly}
Each $\Delta$ function is a function of the radius $r_X$ and distance between the centres $dis_{X,V}$. So, it is linear concerning the radius $r_X$ and monotonic concerning the distance between the centres $dis_{X,V}$, except $\Delta_{\mathbf{EQ}:\mathbf{PO}}^\mathbf{T}(\mathcal{O}_X, \mathcal{O}_V)$, where $\mathbf{T}\in\mathcal{T}$. When $\mathcal{O}_X$ coincides with $\mathcal{O}_V$ ($\mathbf{EQ}$), one step of update the length of $\|\vec{O}_X\|$ ($\|\vec{O}_X\|\neq 0$) will push $\mathcal{O}_X$ to partially overlap with $\mathcal{O}_V$, so $\Delta_{\mathbf{EQ}:\mathbf{PO}}^\mathbf{T}(\mathcal{O}_X, \mathcal{O}_V)$ can also be understood as monotonic.  \qed
\end{PrfCly}

\subsection{The satisfiability theorem for non-cyclic statements}
\label{control1}
\setcounter{Thm}{0}
\begin{Thm}
\label{lmanoncyc}
Let $p_1$, \dots, $p_{N-1}$ be $N-1$ premises of a long-chained syllogistic reasoning system, where $p_i$ can be either $r_i(X_i,X_{i+1})$ or $r_i(X_{i+1}, X_{i})$, $(1\le i\le N-1)$, $r_i\in\{all, some, no, some\_not\}$. $\mathcal{S}$phNN can construct a configuration of $N$ spheres, $\mathcal{O}_{i}$ ($i=1\dots N$), as an Euler diagram of the $N-1$ syllogistic statements, such that $X_i$ maps to $\mathcal{O}_{i}$, and $p_i$ maps to $\psi_i(\mathcal{O}_i, \mathcal{O}_{i+1})$, where $\psi_i = \psi(r_i)$ if $r_i(X_i,X_{i+1})$ or $\psi_i = \overline{\psi(r_i)}$ if $r_i(X_{i+1}, X_i)$,  where $\psi_i\in\{\mathbf{D}, \mathbf{P}, \overline{\mathbf{P}}, \neg\mathbf{D},$ $ \neg\mathbf{P}, \neg\overline{\mathbf{P}}\}$.
\end{Thm} 
\begin{Prf}
We show $\psi_1(\mathcal{O}_{1},\mathcal{O}_{2}),\dots, \psi_{N-1}(\mathcal{O}_{N-1},\mathcal{O}_N)$ are {\em satisfiable}. We prove this by inducting on the length of the sequence. 
\begin{enumerate}
    \item $N=1$. For any initial relation between $\mathcal{O}_1$ and $\mathcal{O}_2$, $\mathcal{S}$phNN can realise the target relation by using $\Delta$ functions in the neural transition map of qualitative spatial relations. 
    \item Suppose that it holds for $N\le K-1$.
    \item  $N=K$. Assume that $\mathcal{S}$phNN has constructed $K-1$ spheres $\mathcal{O}_{1}$,\dots, $\mathcal{O}_{{K-1}}$ satisfying first $K-2$ constraints. To optimise $\mathcal{O}_K$, $\mathcal{S}$phNN repeats the method used for $N=1$, as optimising $\psi_{K-1}(\mathcal{O}_{K-1},\mathcal{O}_K)$ will not hurt other relations.  \qed
\end{enumerate} 
\end{Prf}

\subsection{Existence theorems} 
\setcounter{Lma}{0}
\begin{Lma}
\label{basic}
Given $\mathbf{R}_1, \mathbf{R}_2, \mathbf{R}_3\in\{\mathbf{D}, \mathbf{EQ}, \mathbf{PO}, \mathbf{PP}, \overline{\mathbf{PP}}\}$. If the three relations are satisfiable, that is, $\exists\mathcal{O}_1, \mathcal{O}_2, \mathcal{O}_3[\mathbf{R}_1(\mathcal{O}_1, \mathcal{O}_2)\land \mathbf{R}_2(\mathcal{O}_2, \mathcal{O}_3)\land\mathbf{R}_3(\mathcal{O}_3, \mathcal{O}_1)]$, then for any fixed $\mathcal{O}_X$ and $\mathcal{O}_Y$ satisfying  $\mathbf{R}_1(\mathcal{O}_X, \mathcal{O}_Y)$, there will be $\mathcal{O}_Z$ such that $\mathbf{R}_2(\mathcal{O}_Y, \mathcal{O}_Z)$ and $\mathbf{R}_3(\mathcal{O}_Z, \mathcal{O}_X)$. 
\end{Lma}
\begin{PrfLma}
\label{existence}
We enumerate the combination of relations of $\mathbf{R}_2$ and $\mathbf{R}_3$.
\begin{enumerate}
\item $\mathbf{R}_3(\mathcal{O}_Z, \mathcal{O}_X)=\mathbf{EQ}(\mathcal{O}_Z, \mathcal{O}_X)$. A trivial case of Theorem~\ref{lmanoncyc}.
\item $\mathbf{R}_2(\mathcal{O}_Y, \mathcal{O}_Z)=\mathbf{EQ}(\mathcal{O}_Y, \mathcal{O}_Z)$. A trivial case of Theorem~\ref{lmanoncyc}.
\item $\mathbf{R}_2(\mathcal{O}_Y, \mathcal{O}_Z)=\mathbf{D}(\mathcal{O}_Y, \mathcal{O}_Z)$ and $\mathbf{R}_3(\mathcal{O}_Z, \mathcal{O}_X)=\mathbf{D}(\mathcal{O}_Z, \mathcal{O}_X)$.  For any fixed $\mathcal{O}_X$ and $\mathcal{O}_Y$, if
\begin{figure}
\centering
\includegraphics[width=0.8\textwidth]{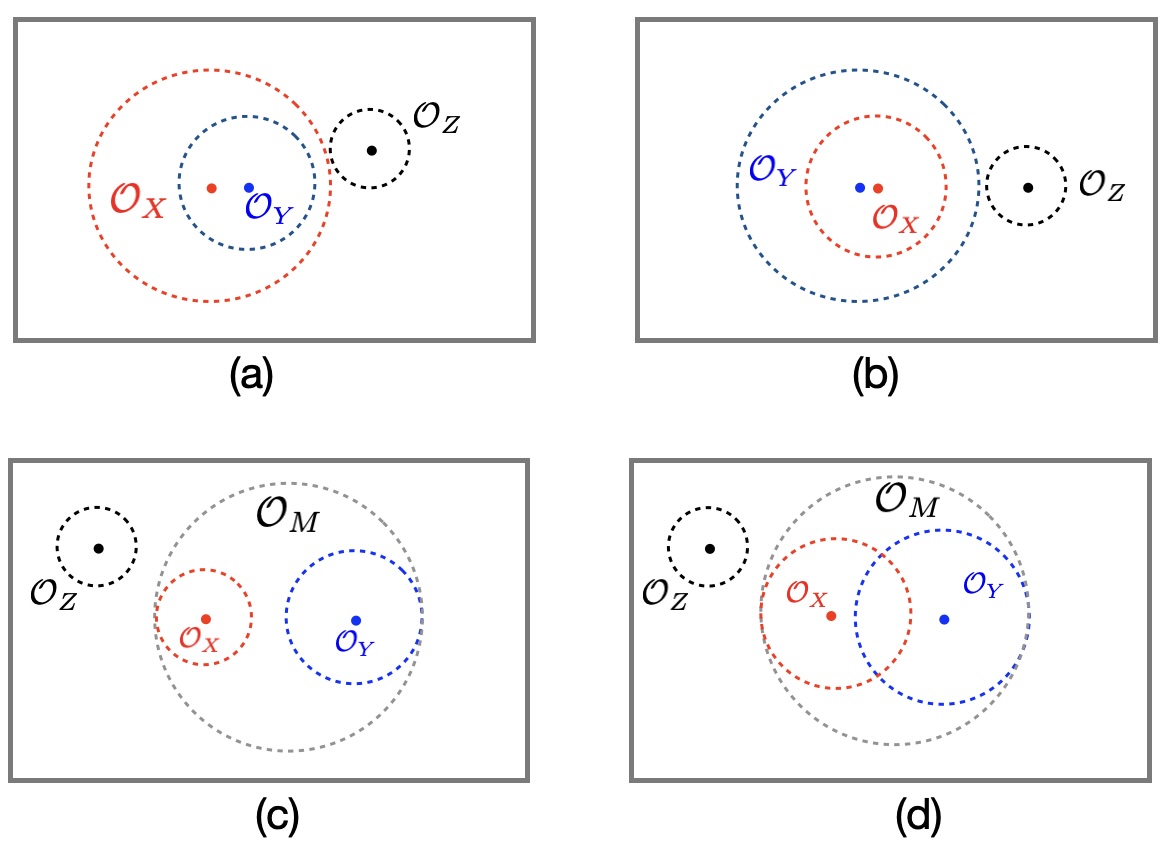}
\caption{(a) $\mathcal{O}_Y$ is proper proper part of $\mathcal{O}_X$; (b) $\mathcal{O}_X$ is proper part of $\mathcal{O}_Y$; (c) $\mathcal{O}_X$ disconnects from $\mathcal{O}_Y$; (d) $\mathcal{O}_X$ partially overlaps with $\mathcal{O}_Y$. In any situation, there is $\mathcal{O}_Z$ disconnecting from $\mathcal{O}_X$ and $\mathcal{O}_Y$.}
\label{cp_dd}
\end{figure}
\begin{enumerate}
    \item $\overline{\mathbf{PP}}(\mathcal{O}_X, \mathcal{O}_Y)$. Any $\mathcal{O}_Z$ disconnecting from $\mathcal{O}_X$ disconnects from $\mathcal{O}_Y$, shown in Figure~\ref{cp_dd}(a).
    \item $\mathbf{PP}(\mathcal{O}_X, \mathcal{O}_Y)$. Any $\mathcal{O}_Z$ disconnecting from $\mathcal{O}_Y$ disconnects from $\mathcal{O}_X$, shown in Figure~\ref{cp_dd}(b).
\begin{figure} 
\centering
\includegraphics[width=0.8\textwidth]{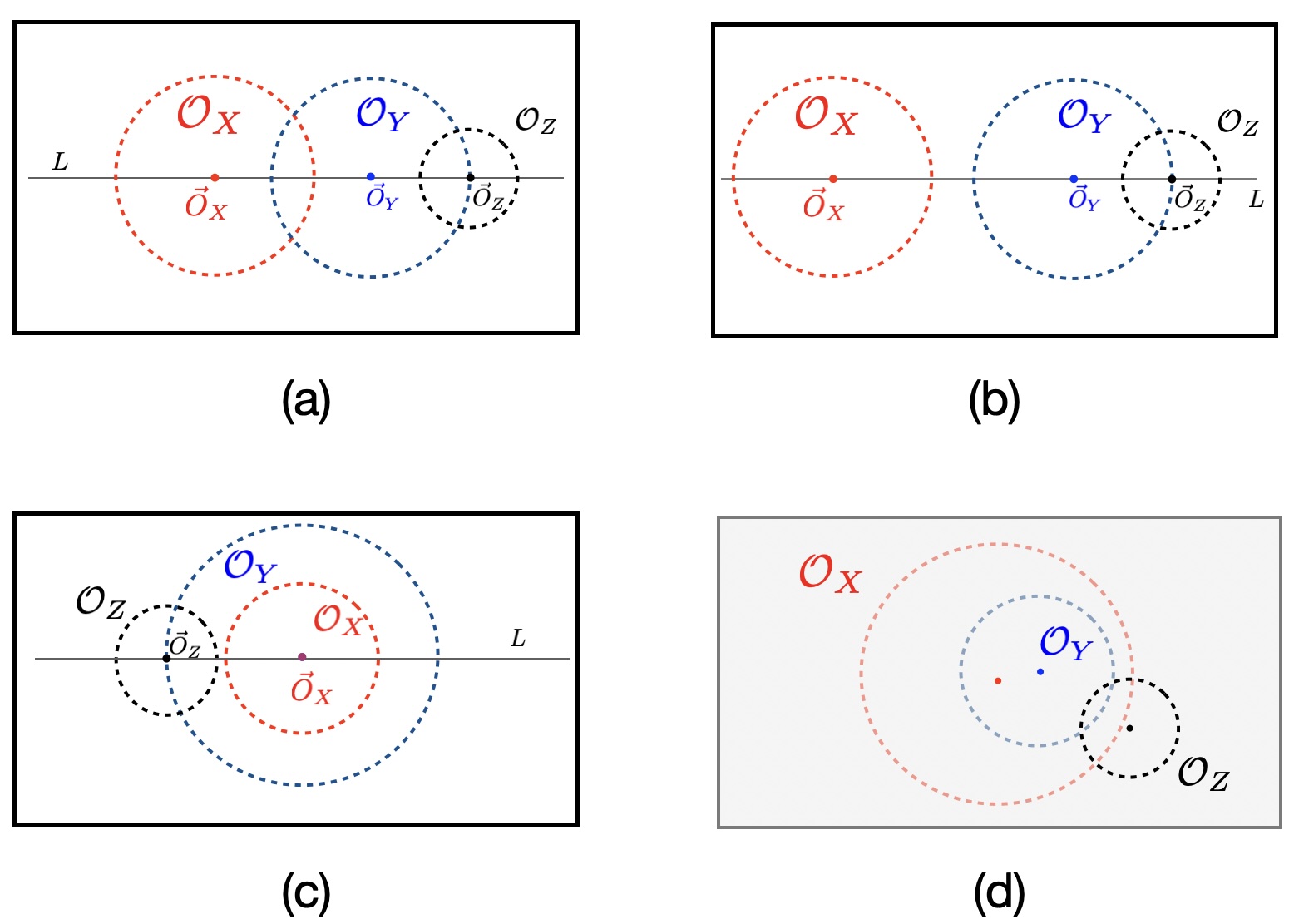}
\caption{(a-c) $\mathcal{O}_Y$ partially overlaps with $\mathcal{O}_Z$ and $\mathcal{O}_Z$ disconnects from $\mathcal{O}_X$; (d) $\mathcal{O}_X$ contains $\mathcal{O}_Y$, therefore, if $\mathcal{O}_Z$ connects with $\mathcal{O}_Y$, it will connect with $\mathcal{O}_X$. The grey background means an unsatisfiable case.}
\label{cp_pod}
\end{figure}
    \item $\mathbf{D}(\mathcal{O}_X, \mathcal{O}_Y)$. Let both $\mathcal{O}_X$ and  $\mathcal{O}_Y$ be inside $\mathcal{O}_M$, any $\mathcal{O}_Z$ disconnecting from $\mathcal{O}_M$ disconnects from $\mathcal{O}_X$ and $\mathcal{O}_Y$, shown in Figure~\ref{cp_dd}(c).
    \item $\mathbf{PO}(\mathcal{O}_X, \mathcal{O}_Y)$. The same as (c), shown in Figure~\ref{cp_dd}(d).
\end{enumerate}
\item $\mathbf{R}_2(\mathcal{O}_Y, \mathcal{O}_Z)=\mathbf{PO}(\mathcal{O}_Y, \mathcal{O}_Z)$ and $\mathbf{R}_3(\mathcal{O}_Z, \mathcal{O}_X)=\mathbf{D}(\mathcal{O}_Z, \mathcal{O}_X)$.  For any fixed $\mathcal{O}_X$ and $\mathcal{O}_Y$, if

\begin{enumerate}
    \item $\mathbf{PO}(\mathcal{O}_X, \mathcal{O}_Y)$. Let $\vec{O}_Z$ (the centre of $\mathcal{O}_Z$) be located at the boundary of  $\mathcal{O}_Y$ and be the apogee to $\mathcal{O}_X$. As $\mathcal{O}_X$ partially overlaps with $\mathcal{O}_Y$, $\mathbf{PO}(\mathcal{O}_X, \mathcal{O}_Y)$, they cannot be concentric, so there is $r_Z=\epsilon$ such that $\mathcal{O}_Z$ disconnects from $\mathcal{O}_X$ and partially overlaps with $\mathcal{O}_Z$, shown in Figure~\ref{cp_pod}(a).
    \item $\mathbf{D}(\mathcal{O}_X, \mathcal{O}_Y)$. The same as (a), shown in Figure~\ref{cp_pod}(b).
    \item  $\mathbf{PP}(\mathcal{O}_X, \mathcal{O}_Y)$. If $\mathcal{O}_X$ and $\mathcal{O}_Y$ are not concentric, the case is the same as (a); otherwise, $\mathcal{O}_X$ is a proper part of $\mathcal{O}_Y$ ($r_Y >r_X$), let $r_Z < r_Y -r_X$. shown in Figure~\ref{cp_pod}(c).
    \item $\overline{\mathbf{PP}}(\mathcal{O}_X, \mathcal{O}_Y)$. Any $\mathcal{O}_Z$ connecting with $\mathcal{O}_Y$ connects with $\mathcal{O}_X$. This contradicts with $\mathbf{D}(\mathcal{O}_Z,  \mathcal{O}_X)$, shown in Figure~\ref{cp_pod}(d).
\end{enumerate}  

\begin{figure} 
\centering
    \includegraphics[width=0.8\textwidth]{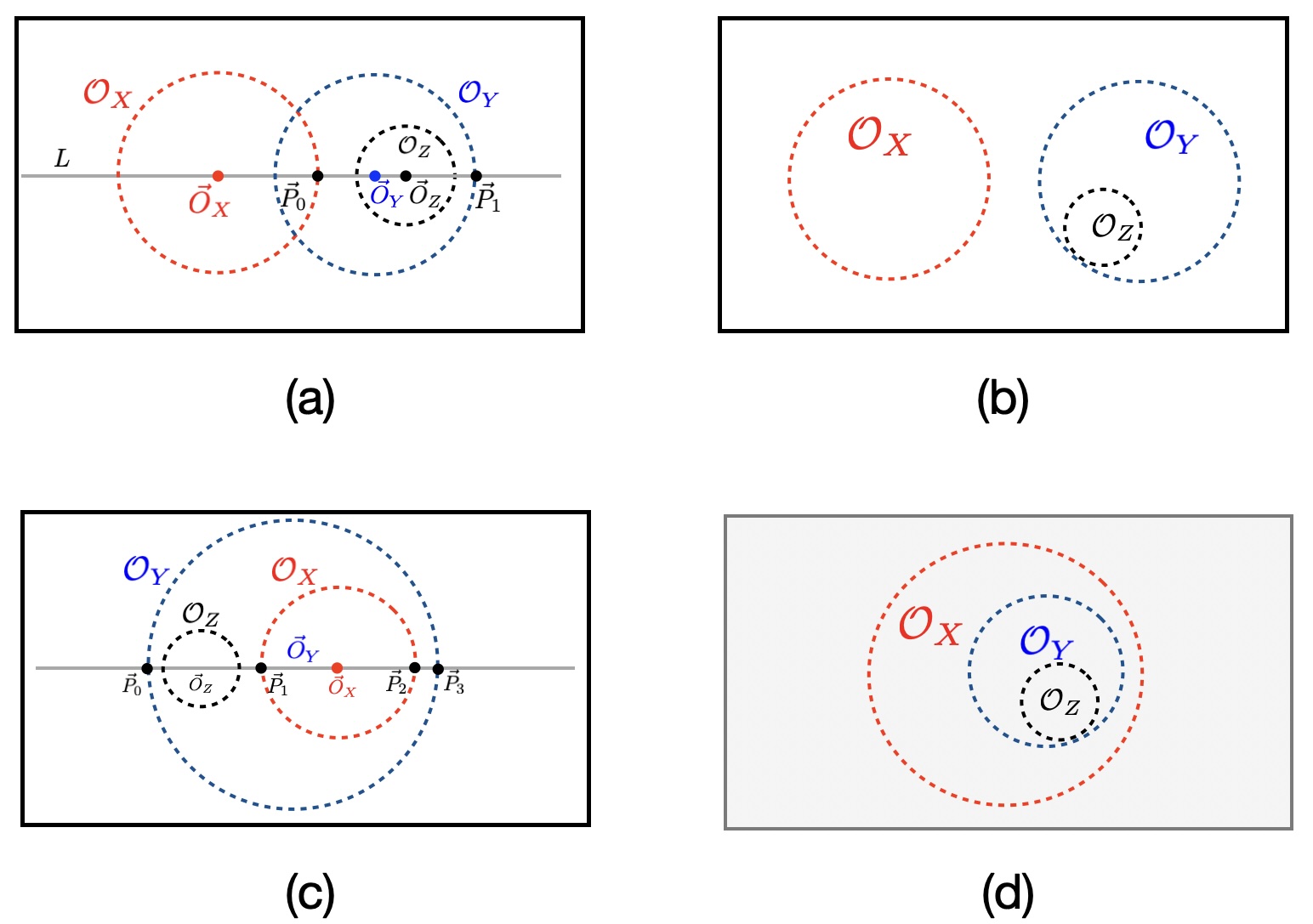}
\caption{(a-c) As long as $\mathcal{O}_X$ does not totally cover $\mathcal{O}_Y$, there will be $\mathcal{O}_Z$ that is proper part of $\mathcal{O}_Y$, and disconnects from $\mathcal{O}_X$; (d) $\mathcal{O}_X$ totally covers $\mathcal{O}_Y$, if $\mathcal{O}_Z$ is inside $\mathcal{O}_Y$, it will be inside $\mathcal{O}_X$.}
\label{cp_ppbard}
\end{figure}
\item  $\mathbf{R}_2(\mathcal{O}_Y, \mathcal{O}_Z)=\overline{\mathbf{PP}}(\mathcal{O}_Y, \mathcal{O}_Z)$ and $\mathbf{R}_3(\mathcal{O}_Z, \mathcal{O}_X)=\mathbf{D}(\mathcal{O}_Z, \mathcal{O}_X)$. For any fixed $\mathcal{O}_X$ and $\mathcal{O}_Y$, if 
\begin{enumerate}
    \item $\mathbf{PO}(\mathcal{O}_X, \mathcal{O}_Y)$. As $\mathcal{O}_X$ partially overlaps with $ \mathcal{O}_Y$, let the line $L$ pass the centres of $\mathcal{O}_X$ and $\mathcal{O}_Y$, intersect with the boundary of $\mathcal{O}_X$ at $\vec{P}_0$ ($\vec{P}_0$ is inside $\mathcal{O}_Y$), intersect with the boundary of $\mathcal{O}_Y$ at $\vec{P}_1$ ($\vec{P}_1$ is outside $\mathcal{O}_X$). Let $\mathcal{O}_Z$ be a sphere whose diameter is a segment between ${\vec{P}_0}$ and ${\vec{P}_1}$, shown in Figure~\ref{cp_ppbard}(a).
    \item $\mathbf{D}(\mathcal{O}_X, \mathcal{O}_Y)$. Let $\mathcal{O}_Z$ be any sphere inside $\mathcal{O}_Y$, shown in Figure~\ref{cp_ppbard}(b).
    \item  $\mathbf{PP}(\mathcal{O}_X, \mathcal{O}_Y)$. As $\mathcal{O}_X$ is proper part of $ \mathcal{O}_Y$, let the line $L$ pass the centres of $\mathcal{O}_X$ and $\mathcal{O}_Y$, intersect with the boundary of $\mathcal{O}_X$ at $\vec{P}_1$ and $\vec{P}_2$, intersect with the boundary of  $\mathcal{O}_Y$ at $\vec{P}_0$ and $\vec{P}_3$. 
    Without the loss of generality, let $|{\vec{P}_0}{\vec{P}_1}|\ge |{\vec{P}_2}{\vec{P}_3}|\ge 0$. As $\mathbf{PP}(\mathcal{O}_X, \mathcal{O}_Y)$, it is not possible that $|{\vec{P}_0}{\vec{P}_1}|= |{\vec{P}_2}{\vec{P}_3}|= 0$.  Let  $\mathcal{O}_Z$ be a sphere whose diameter is a segment between ${\vec{P}_0}$ and ${\vec{P}_1}$, shown in  Figure~\ref{cp_ppbard}(c).
    \item
    $\overline{\mathbf{PP}}(\mathcal{O}_X, \mathcal{O}_Y)$. As $\mathcal{O}_Z$ is part of $\mathcal{O}_Y$, $\mathcal{O}_Z$ will be inside $\mathcal{O}_X$, which contradicts with $\mathbf{D}(\mathcal{O}_X, \mathcal{O}_Z)$, shown in  Figure~\ref{cp_ppbard}(d).
\end{enumerate} 

\begin{figure} 
\centering
\includegraphics[width=0.8\textwidth]{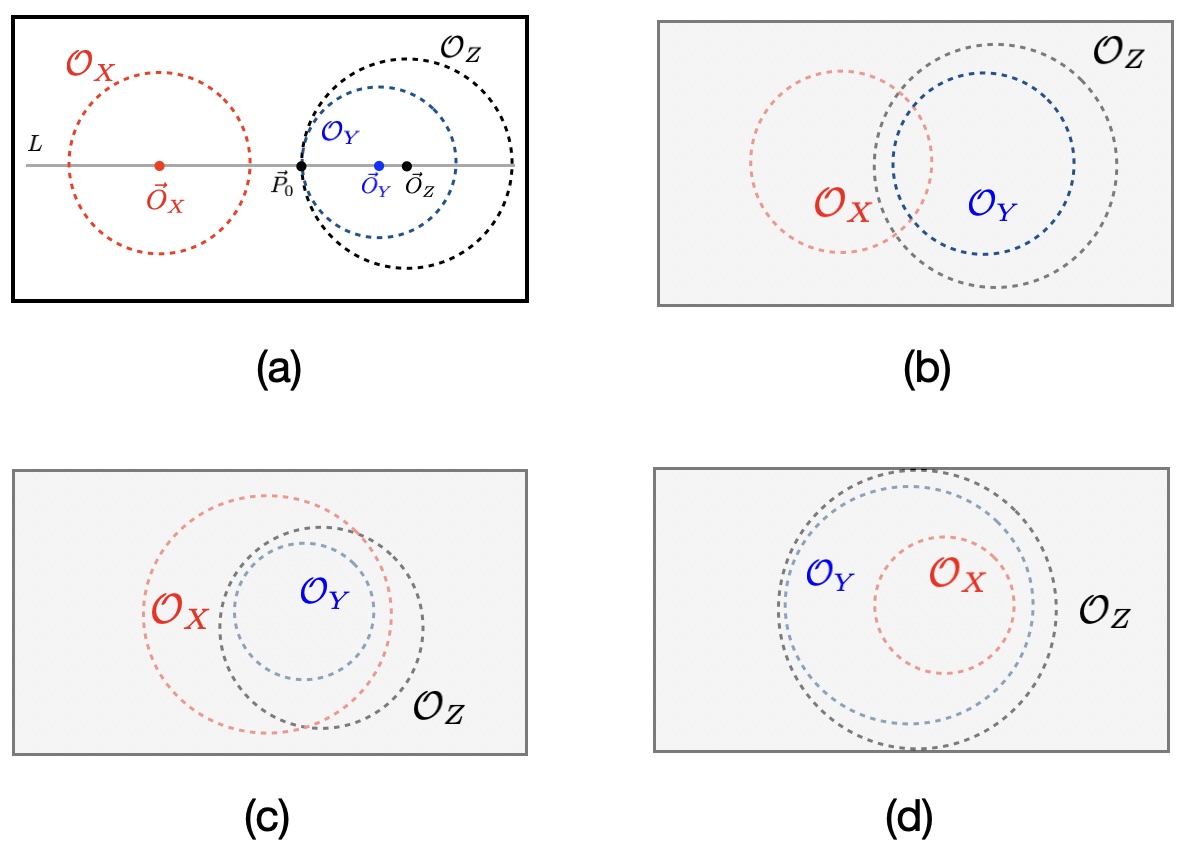}
\caption{$\mathcal{O}_Y$ is proper part of $\mathcal{O}_Z$ and $\mathcal{O}_Z$ disconnects from $\mathcal{O}_X$. This case is only possible when $\mathcal{O}_X$ disconnects from $\mathcal{O}_Y$ (a).}
\label{cp_ppd}
\end{figure}

\item   $\mathbf{R}_2(\mathcal{O}_Y, \mathcal{O}_Z)=\mathbf{PP}(\mathcal{O}_Y, \mathcal{O}_Z)$ and $\mathbf{R}_3(\mathcal{O}_Z, \mathcal{O}_X)=\mathbf{D}(\mathcal{O}_Z, \mathcal{O}_X)$. For any fixed $\mathcal{O}_X$ and $\mathcal{O}_Y$, if 
\begin{enumerate}
    \item $\mathbf{D}(\mathcal{O}_X, \mathcal{O}_Y)$. Let the line $L$ pass the centres of $\mathcal{O}_X$ and $\mathcal{O}_Y$, intersect with the boundary of $\mathcal{O}_Y$ at ${\vec{P}_0}$, the perigee of $\mathcal{O}_X$. Let  $\mathcal{O}_Z$ be the sphere that tangentially contains $\mathcal{O}_Y$ and ${\vec{P}_0}$ be the tangential point, shown in Figure~\ref{cp_ppd}(a).
    \item  $\mathbf{PO}(\mathcal{O}_X, \mathcal{O}_Y)\lor\overline{\mathbf{PP}}(\mathcal{O}_X, \mathcal{O}_Y)\lor\mathbf{PP}(\mathcal{O}_X, \mathcal{O}_Y)$. As $\mathcal{O}_Y$ is part of $\mathcal{O}_Z$, any sphere $\mathcal{O}_X$,  if $\mathcal{O}_X$ connects with $\mathcal{O}_Y$, $\mathcal{O}_X$ connects with $\mathcal{O}_Z$, which contradicts with $\mathbf{D}(\mathcal{O}_Z, \mathcal{O}_X)$, shown in Figure~\ref{cp_ppd}(b-d).
\end{enumerate} 

\item $\mathbf{R}_2(\mathcal{O}_Y, \mathcal{O}_Z)=\mathbf{D}(\mathcal{O}_Y, \mathcal{O}_Z)$ and $\mathbf{R}_3(\mathcal{O}_Z, \mathcal{O}_X)=\mathbf{PO}(\mathcal{O}_Z, \mathcal{O}_X)$. Case 4.
\begin{figure} 
\centering
\includegraphics[width=0.8\textwidth]{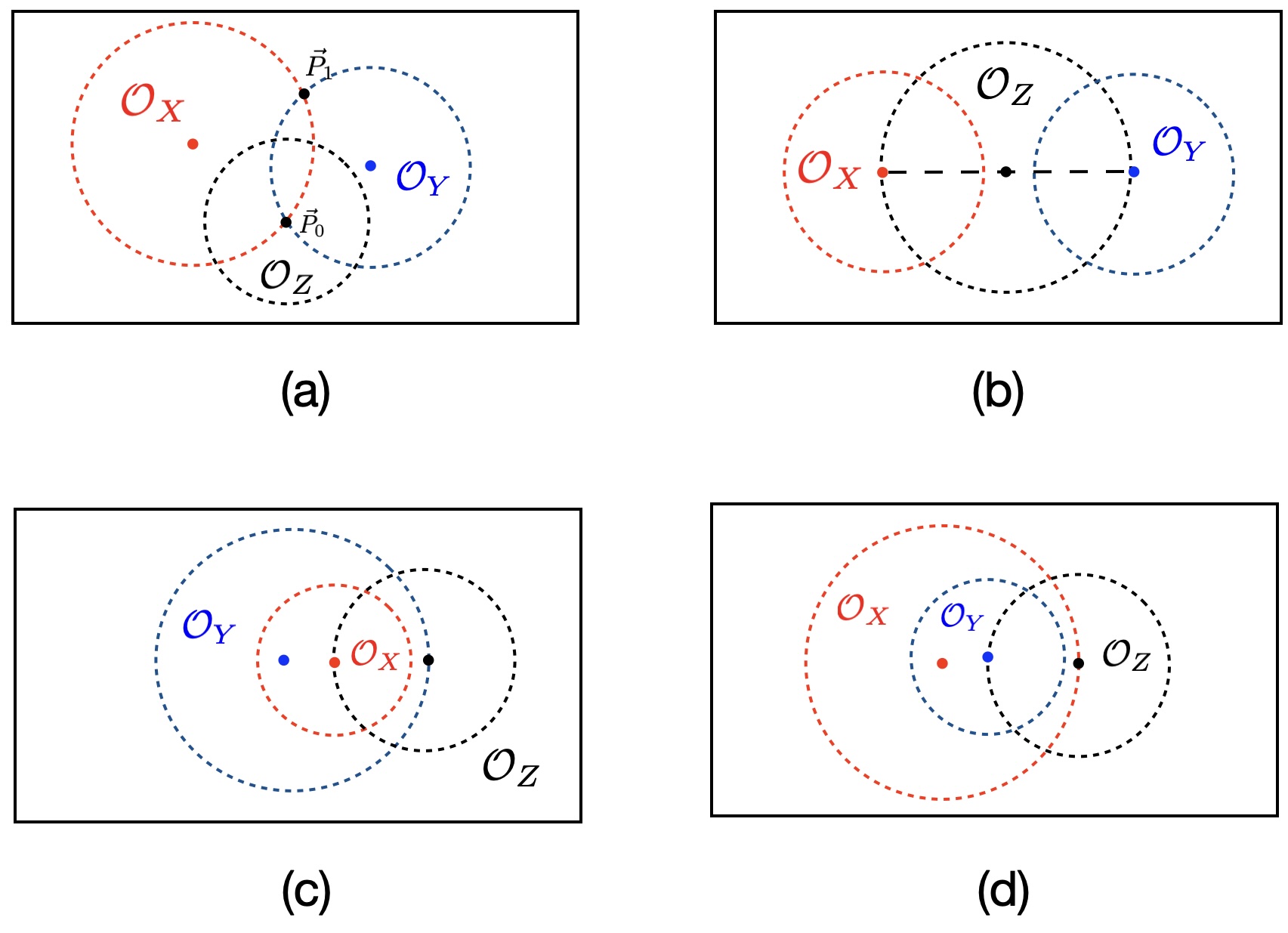}
\caption{(a) $\mathcal{O}_X$ partially overlaps with $\mathcal{O}_Y$, and their boundaries intersect at ${\vec{P}_0}$ and ${\vec{P}_1}$; (b) $\mathcal{O}_X$ disconnects from $\mathcal{O}_Y$; (c) $\mathcal{O}_X$ is proper part of $\mathcal{O}_Y$; (d) $\mathcal{O}_Y$ is proper part of $\mathcal{O}_X$.}
\label{cp_popo}
\end{figure}
\item $\mathbf{R}_2(\mathcal{O}_Y, \mathcal{O}_Z)=\mathbf{PO}(\mathcal{O}_Y, \mathcal{O}_Z)$ and $\mathbf{R}_3(\mathcal{O}_Z, \mathcal{O}_X)=\mathbf{PO}(\mathcal{O}_Z, \mathcal{O}_X)$. For any fixed $\mathcal{O}_X$ and $\mathcal{O}_Y$, if
\begin{enumerate}
    \item $\mathbf{PO}(\mathcal{O}_X, \mathcal{O}_Y)$. Let the boundaries of $\mathcal{O}_X$ and $\mathcal{O}_Y$ intersect at ${\vec{P}_0}$ and ${\vec{P}_1}$. Any sphere $\mathcal{O}_Z$ with ${\vec{P}_0}$ as the centre and with $r_Z$ less than $\min\{r_X, r_Y\}$ will partially overlap with $\mathcal{O}_X$ and $\mathcal{O}_Y$, shown in Figure~\ref{cp_popo}(a).
    \item $\mathbf{D}(\mathcal{O}_X, \mathcal{O}_Y)$. Let $\mathcal{O}_Z$ be the sphere with the segment $|\vec{O}_X\vec{O}_Y|$ as the diameter, shown in Figure~\ref{cp_popo}(b).
    \item  $\mathbf{PP}(\mathcal{O}_X, \mathcal{O}_Y)$. Let $\mathcal{O}_Z$ be a sphere whose centre is at the boundary of $\mathcal{O}_Y$ and whose boundary passes the centre of $\mathcal{O}_X$, shown in  Figure~\ref{cp_popo}(c).
    \item
    $\overline{\mathbf{PP}}(\mathcal{O}_X, \mathcal{O}_Y)$.  Let $\mathcal{O}_Z$ be a sphere whose centre is at the boundary of $\mathcal{O}_X$ and whose boundary passes the centre of $\mathcal{O}_Y$, shown in  Figure~\ref{cp_popo}(d).
\end{enumerate} 

\begin{figure} 
\centering
\includegraphics[width=0.8\textwidth]{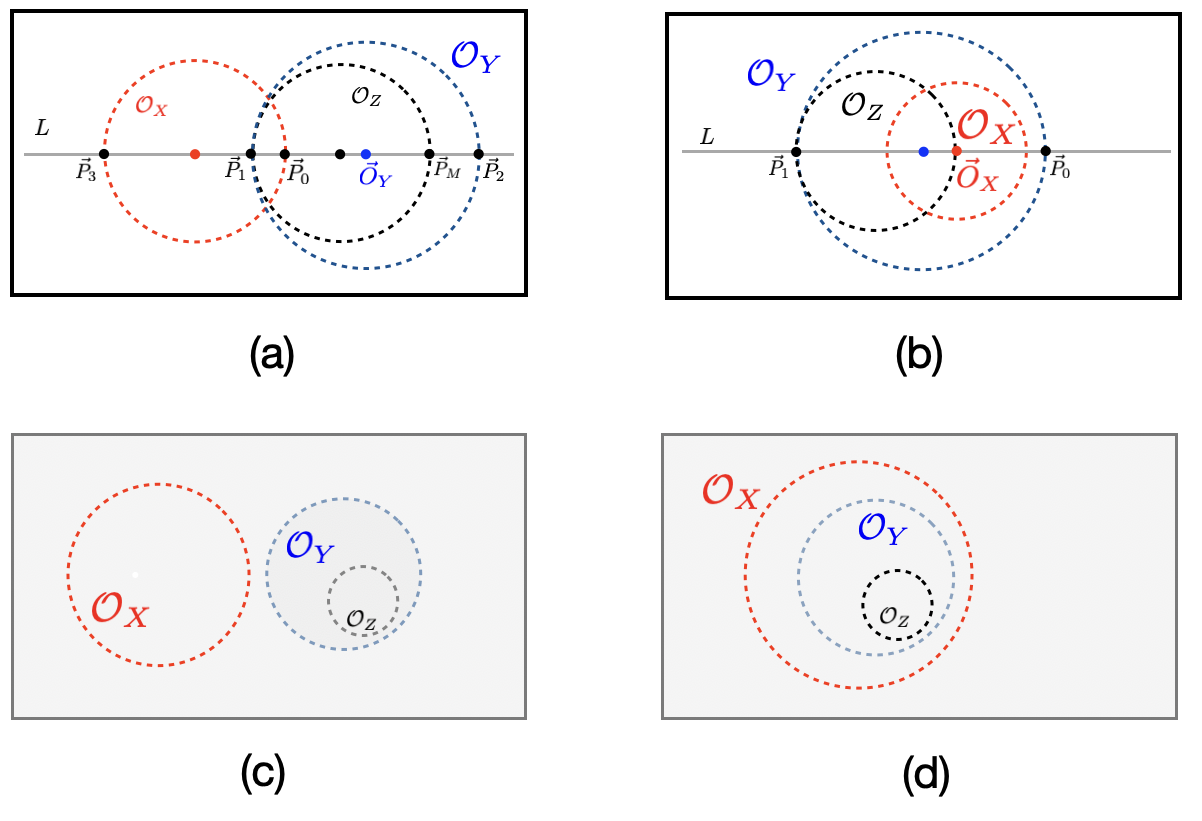}
\caption{(a) $\mathcal{O}_X$ partially overlaps with $\mathcal{O}_Y$; (b) $\mathcal{O}_X$ is proper part of $\mathcal{O}_Y$; (c) $\mathcal{O}_X$ disconnects from $\mathcal{O}_Y$; (d) $\mathcal{O}_Y$ is proper part of  $\mathcal{O}_X$.}
\label{cp_ppbarpo}
\end{figure}
\item $\mathbf{R}_2(\mathcal{O}_Y, \mathcal{O}_Z)=\overline{\mathbf{PP}}(\mathcal{O}_Y, \mathcal{O}_Z)$ and $\mathbf{R}_3(\mathcal{O}_Z, \mathcal{O}_X)=\mathbf{PO}(\mathcal{O}_Z, \mathcal{O}_X)$. For any fixed $\mathcal{O}_X$ and $\mathcal{O}_Y$, if 
\begin{enumerate}
    \item $\mathbf{PO}(\mathcal{O}_X, \mathcal{O}_Y)$. Let the line $L$ pass the centres of  $\mathcal{O}_X$ and $\mathcal{O}_Y$, intersect with the boundary of $\mathcal{O}_X$ at points ${\vec{P}_0}$ and ${\vec{P}_3}$, and intersect with the boundary of $\mathcal{O}_Y$ at points ${\vec{P}_1}$ and ${\vec{P}_2}$, $\vec{P}_M$ is located between ${\vec{P}_0}$ and ${\vec{P}_2}$. $\mathcal{O}_Z$ is the sphere with $|{\vec{P}_1}\vec{P}_M|$ as the diameter. It is easy to prove that $\mathcal{O}_Z$ partially overlaps with $\mathcal{O}_X$ and is a proper part of $\mathcal{O}_Y$, as shown in Figure~\ref{cp_ppbarpo}(a).
    \item $\mathbf{PP}(\mathcal{O}_X, \mathcal{O}_Y)$. Let the line $L$ pass the centres of $\mathcal{O}_X$ and $\mathcal{O}_Y$ and intersect with the boundary of $\mathcal{O}_Y$ at points ${\vec{P}_0}$ and ${\vec{P}_1}$.  Let ${\vec{O}_X}$ be closer to ${\vec{P}_0}$ than to ${\vec{P}_1}$. Let $\mathcal{O}_Z$ be the sphere whose diameter is $|{\vec{P}_1}\vec{O}_X|$, shown in Figure~\ref{cp_ppbarpo}(b).
    \item  $\mathbf{D}(\mathcal{O}_X, \mathcal{O}_Y)$. As $\mathcal{O}_X$ disconnects from $\mathcal{O}_Y$, $\mathcal{O}_X$ will disconnect from any sphere inside $\mathcal{O}_Y$. This contradicts with $\mathbf{PO}(\mathcal{O}_Z, \mathcal{O}_X)$, shown in  Figure~\ref{cp_ppbarpo}(c).
    \item
    $\overline{\mathbf{PP}}(\mathcal{O}_X, \mathcal{O}_Y)$.  As $\mathcal{O}_X$ contains $\mathcal{O}_Y$, $\mathcal{O}_X$ will contains any sphere inside $\mathcal{O}_Y$. This contradicts with $\mathbf{PO}(\mathcal{O}_Z, \mathcal{O}_X)$, shown in  Figure~\ref{cp_ppbarpo}(d).
\end{enumerate} 
\begin{figure} 
\centering
\includegraphics[width=0.8\textwidth]{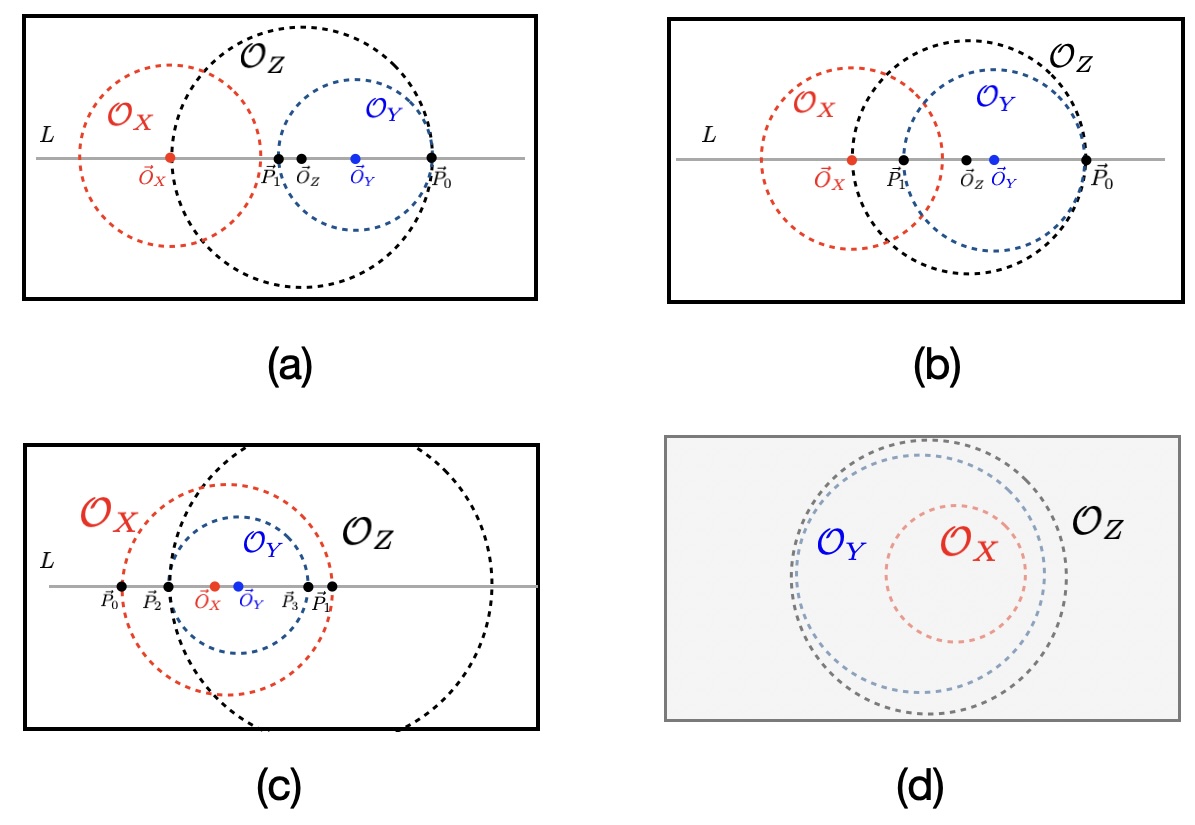}
\caption{(a) $\mathcal{O}_X$ disconnects from  $\mathcal{O}_Y$; (b) $\mathcal{O}_X$ partially overlaps with $\mathcal{O}_Y$; (c) $\mathcal{O}_Y$ is proper part of $\mathcal{O}_X$; (d) $\mathcal{O}_X$ is  part of $\mathcal{O}_Y$. }
\label{cp_pppo}
\end{figure} 
\begin{figure} 
\centering
\includegraphics[width=0.8\textwidth]{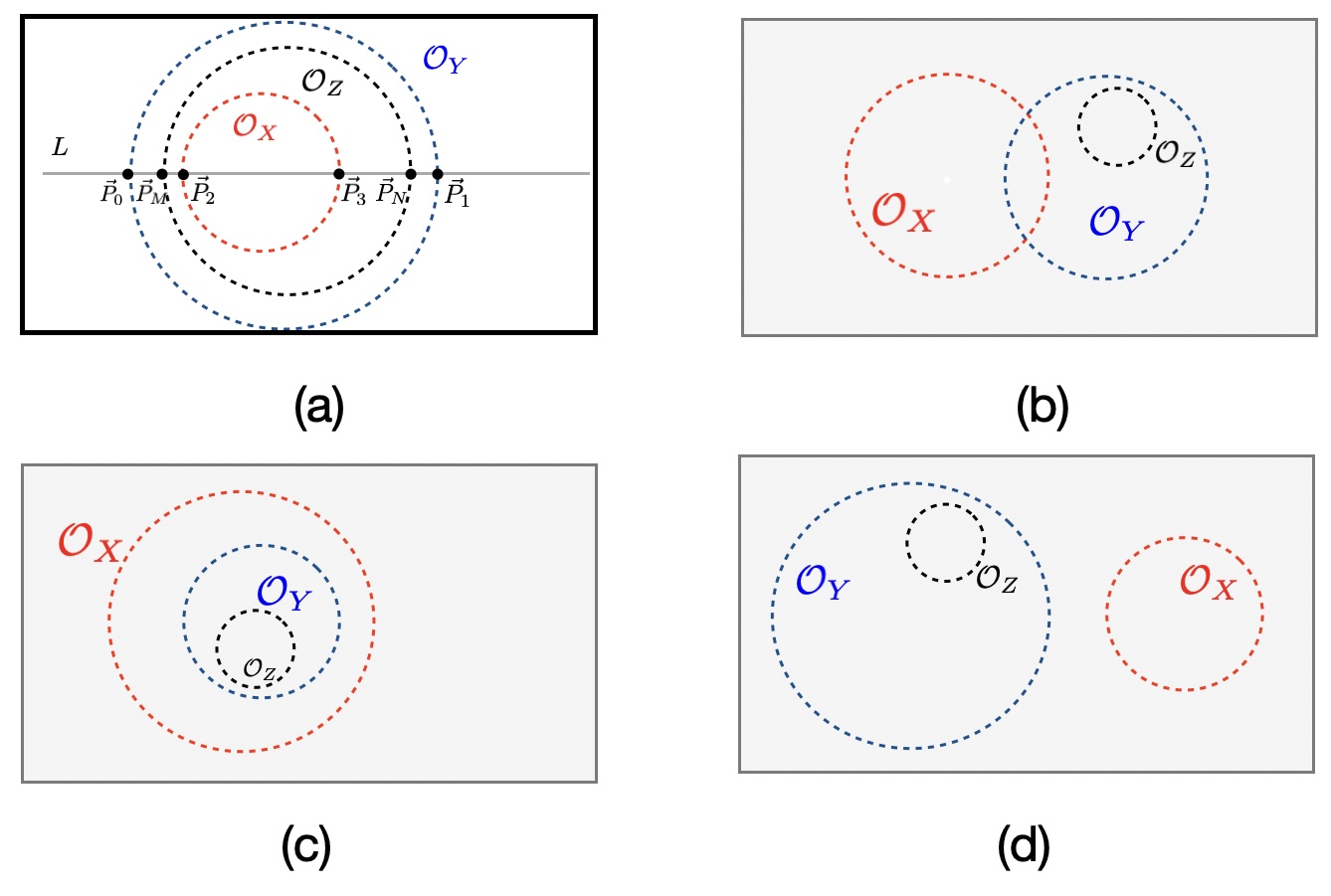}
\caption{(a) $\mathcal{O}_X$ is proper part of $\mathcal{O}_Y$; (b) $\mathcal{O}_X$ partially overlaps with $\mathcal{O}_Y$; (c) $\mathcal{O}_Y$ is proper part of $\mathcal{O}_X$; (d) $\mathcal{O}_X$ disconnects from $\mathcal{O}_Y$.}
\label{cp_ppbarppbar}
\end{figure} 
\item $\mathbf{R}_2(\mathcal{O}_Y, \mathcal{O}_Z)=\mathbf{PP}(\mathcal{O}_Y, \mathcal{O}_Z)$ and $\mathbf{R}_3(\mathcal{O}_Z, \mathcal{O}_X)=\mathbf{PO}(\mathcal{O}_Z, \mathcal{O}_X)$. For any fixed $\mathcal{O}_X$ and $\mathcal{O}_Y$, if 
\begin{enumerate}
    \item
 $\mathbf{D}(\mathcal{O}_X, \mathcal{O}_Y)$. Let ${\vec{P}_0}$ be located at the boundary of  $\mathcal{O}_Y$ and be the apogee to $\mathcal{O}_X$. Let $\mathcal{O}_Z$ be the sphere takes the segment $|\vec{O}_X{\vec{P}_0}|$ as the diameter. Then, $\mathcal{O}_Z$ contains $\mathcal{O}_Y$ and partially overlaps with $\mathcal{O}_X$, shown in Figure~\ref{cp_pppo}(a).
\item    
     $\mathbf{PO}(\mathcal{O}_X, \mathcal{O}_Y)$. The same as (a), shown in Figure~\ref{cp_pppo}(b).
\item$\overline{\mathbf{PP}}(\mathcal{O}_X, \mathcal{O}_Y)$. Let the line $L$ pass the centres of $\mathcal{O}_X$ and $\mathcal{O}_Y$, intersect with the boundary of $\mathcal{O}_X$ at ${\vec{P}_0}$ and ${\vec{P}_1}$, and intersect with the boundary of $\mathcal{O}_Y$ at ${\vec{P}_2}$ and ${\vec{P}_3}$, as shown in  Figure~\ref{cp_pppo}(c). Let $\mathcal{O}_Z$ take  ${\vec{P}_1}$ as the centre, and $|{\vec{P}_1}{\vec{P}_2}|$ as the radius, where $|{\vec{P}_1}{\vec{P}_2}|>|{\vec{P}_1}{\vec{P}_3}|$ . 
    \item  $\mathbf{PP}(\mathcal{O}_X, \mathcal{O}_Y)$. For any $\mathcal{O}_Z$ containing $\mathcal{O}_Y$, $\mathcal{O}_Z$ will contain $\mathcal{O}_X$. This contradicts with $\mathbf{PO}(\mathcal{O}_Z, \mathcal{O}_X)$, shown in Figure~\ref{cp_pppo}(d).
\end{enumerate} 
\item $\mathbf{R}_2(\mathcal{O}_Y, \mathcal{O}_Z)=\mathbf{D}(\mathcal{O}_Y, \mathcal{O}_Z)$ and $\mathbf{R}_3(\mathcal{O}_Z, \mathcal{O}_X)=\overline{\mathbf{PP}}(\mathcal{O}_Z, \mathcal{O}_X)$.  Case 5.
\item $\mathbf{R}_2(\mathcal{O}_Y, \mathcal{O}_Z)=\mathbf{PO}(\mathcal{O}_Y, \mathcal{O}_Z)$ and $\mathbf{R}_3(\mathcal{O}_Z, \mathcal{O}_X)=\overline{\mathbf{PP}}(\mathcal{O}_Z, \mathcal{O}_X)$.  Case 9.
\item $\mathbf{R}_2(\mathcal{O}_Y, \mathcal{O}_Z)=\overline{\mathbf{PP}}(\mathcal{O}_Y, \mathcal{O}_Z)$ and $\mathbf{R}_3(\mathcal{O}_Z, \mathcal{O}_X)=\overline{\mathbf{PP}}(\mathcal{O}_Z, \mathcal{O}_X)$.  For any fixed $\mathcal{O}_X$ and $\mathcal{O}_Y$, if
\begin{enumerate}
    \item $\mathbf{PP}(\mathcal{O}_X, \mathcal{O}_Y)$. Let the line $L$ pass the centres of $\mathcal{O}_X$ and $\mathcal{O}_Y$, intersect with the boundary of $\mathcal{O}_X$ at ${\vec{P}_2}$ and ${\vec{P}_3}$, and intersect with the boundary of $\mathcal{O}_Y$ at ${\vec{P}_0}$ and ${\vec{P}_1}$, where ${\vec{P}_0}$ is nearer to ${\vec{P}_2}$ than to ${\vec{P}_3}$.   Let $\vec{P}_M$ be a point between ${\vec{P}_0}$ and ${\vec{P}_2}$; $\vec{P}_N$ be a point between ${\vec{P}_1}$ and ${\vec{P}_3}$. Let $\mathcal{O}_Z$ be the sphere with $|\vec{P}_M\vec{P}_N|$ as the diameter, as shown in  Figure~\ref{cp_ppbarppbar}(a).  
    \item From $\overline{\mathbf{PP}}(\mathcal{O}_Y, \mathcal{O}_Z)$ and $ \overline{\mathbf{PP}}(\mathcal{O}_Z, \mathcal{O}_X)$ we have $\overline{\mathbf{PP}}(\mathcal{O}_Y, \mathcal{O}_X)$, which is equivalent with $\mathbf{PP}(\mathcal{O}_X, \mathcal{O}_Y)$. Therefore, other relations between $\mathcal{O}_X$ and $\mathcal{O}_Y$ are not possible, shown in Figure~\ref{cp_ppbarppbar}(b-d).
\end{enumerate}
\begin{figure} 
\centering
\includegraphics[width=0.8\textwidth]{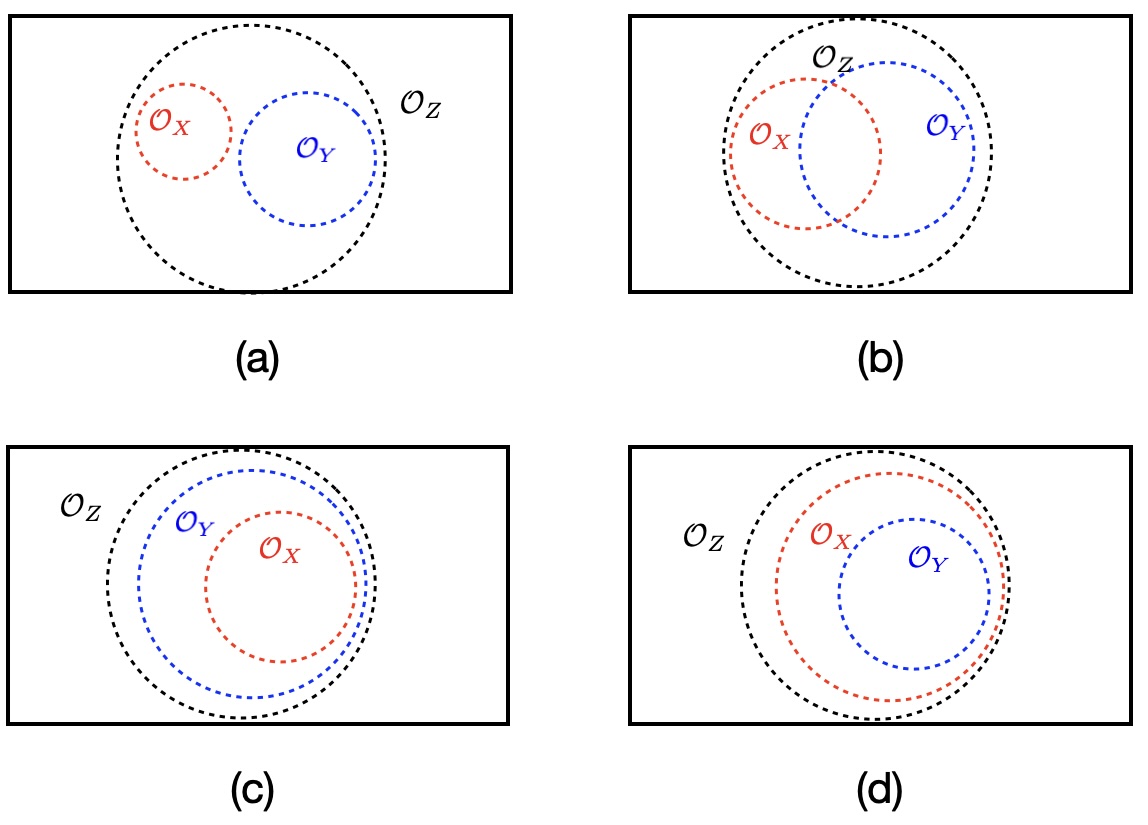}\
\caption{When $\mathcal{O}_Y$ is proper part of $\mathcal{O}_Z$ and $\mathcal{O}_Z$ contains $\mathcal{O}_X$, $\mathcal{O}_Y$ and  $\mathcal{O}_X$ can be of any relation.}
\label{cp_ppbarpp}
\end{figure}
\item $\mathbf{R}_2(\mathcal{O}_Y, \mathcal{O}_Z)=\mathbf{PP}(\mathcal{O}_Y, \mathcal{O}_Z)$ and $\mathbf{R}_3(\mathcal{O}_Z, \mathcal{O}_X)=\overline{\mathbf{PP}}(\mathcal{O}_Z, \mathcal{O}_X)$. For any $\mathcal{O}_X$, $\mathcal{O}_Y$, let $\mathcal{O}_Z$ be large enough to contain both $\mathcal{O}_X$ and $\mathcal{O}_Y$, as shown in Figure~\ref{cp_ppbarpp}. 
\item $\mathbf{R}_2(\mathcal{O}_Y, \mathcal{O}_Z)=\mathbf{D}(\mathcal{O}_Y, \mathcal{O}_Z)$ and $\mathbf{R}_3(\mathcal{O}_Z, \mathcal{O}_X)=\mathbf{PP}(\mathcal{O}_Z, \mathcal{O}_X)$. Case 6. 
\item $\mathbf{R}_2(\mathcal{O}_Y, \mathcal{O}_Z)=\mathbf{PO}(\mathcal{O}_Y, \mathcal{O}_Z)$ and $\mathbf{R}_3(\mathcal{O}_Z, \mathcal{O}_X)=\mathbf{PP}(\mathcal{O}_Z, \mathcal{O}_X)$. Case 10. 
\item $\mathbf{R}_2(\mathcal{O}_Y, \mathcal{O}_Z)=\overline{\mathbf{PP}}(\mathcal{O}_Y, \mathcal{O}_Z)$ and $\mathbf{R}_3(\mathcal{O}_Z, \mathcal{O}_X)=\mathbf{PP}(\mathcal{O}_Z, \mathcal{O}_X)$. Case 14.
\item $\mathbf{R}_2(\mathcal{O}_Y, \mathcal{O}_Z)=\mathbf{PP}(\mathcal{O}_Y, \mathcal{O}_Z)$ and $\mathbf{R}_3(\mathcal{O}_Z, \mathcal{O}_X)=\mathbf{PP}(\mathcal{O}_Z, \mathcal{O}_X)$. Case 13. \qed
\end{enumerate}
\end{PrfLma}

\begin{Lma}
\label{lma_1neg_sat}
Given $\mathbf{R}_1, \mathbf{R}_2\in\{\mathbf{D}, \mathbf{EQ}, \mathbf{PO}, \mathbf{PP}, \overline{\mathbf{PP}}\}$ and $\mathbf{R}_3\in\{\neg\mathbf{D},  \neg\mathbf{P}, \neg\overline{\mathbf{P}}\}$. If the three relations are satisfiable, that is, $\exists\mathcal{O}_1, \mathcal{O}_2, \mathcal{O}_3[\mathbf{R}_1(\mathcal{O}_1, \mathcal{O}_2)\land \mathbf{R}_2(\mathcal{O}_2, \mathcal{O}_3)\land\mathbf{R}_3(\mathcal{O}_3, \mathcal{O}_1)]$, for any fixed $\mathcal{O}_X$ and $\mathcal{O}_Y$ satisfying  $\mathbf{R}_1(\mathcal{O}_X, \mathcal{O}_Y)$, there will be $\mathcal{O}_Z$ such that $\mathbf{R}_2(\mathcal{O}_Y, \mathcal{O}_Z)$ and $\mathbf{R}_3(\mathcal{O}_Z, \mathcal{O}_X)$. 
\end{Lma}
\begin{PrfLma}
We enumerate the relations of $\mathbf{R}_1$ and $\mathbf{R}_2$. The negative value of $\mathbf{R}_3$ can be understood as the grouping of several positive relations, as follows.
\begin{eqnarray*}
&&\neg\mathbf{D}= \mathbf{EQ}\lor \mathbf{PO}\lor \mathbf{PP}\lor \overline{\mathbf{PP}}\\
&&\neg\mathbf{P}= \mathbf{D}\lor \mathbf{PO}\lor  \overline{\mathbf{PP}}\\
&&\neg\overline{\mathbf{P}}= \mathbf{D}\lor \mathbf{PO}\lor \mathbf{PP}
\end{eqnarray*}
The rest of the part is similar to the proof of Lemma~\ref{existence}. \qed
\end{PrfLma}

\begin{Lma}
\label{lma_2neg_sat}
Relations $\mathbf{T}_1$, $\mathbf{T}_2$, and $\mathbf{T}_3$ are satisfiable, where $\mathbf{T}_1, \mathbf{T}_2\in\{\neg\mathbf{D}, \neg\mathbf{P}, \neg\overline{\mathbf{P}}\}$ and $\mathbf{T}_3\in\{\mathbf{D}, \mathbf{P}, \overline{\mathbf{P}}, \mathbf{PO}\}$. For any fixed $\mathcal{O}_Z$ and $\mathcal{O}_X$ satisfying $\mathbf{T}_3(\mathcal{O}_Z, \mathcal{O}_X)$,  there is $\mathcal{O}_Y$ satisfying  $\mathbf{T}_1(\mathcal{O}_X, \mathcal{O}_Y)$, $\mathbf{T}_2(\mathcal{O}_Y, \mathcal{O}_Z)$. 
\end{Lma}
\begin{figure}
\centering
\includegraphics[width=0.8
\textwidth]{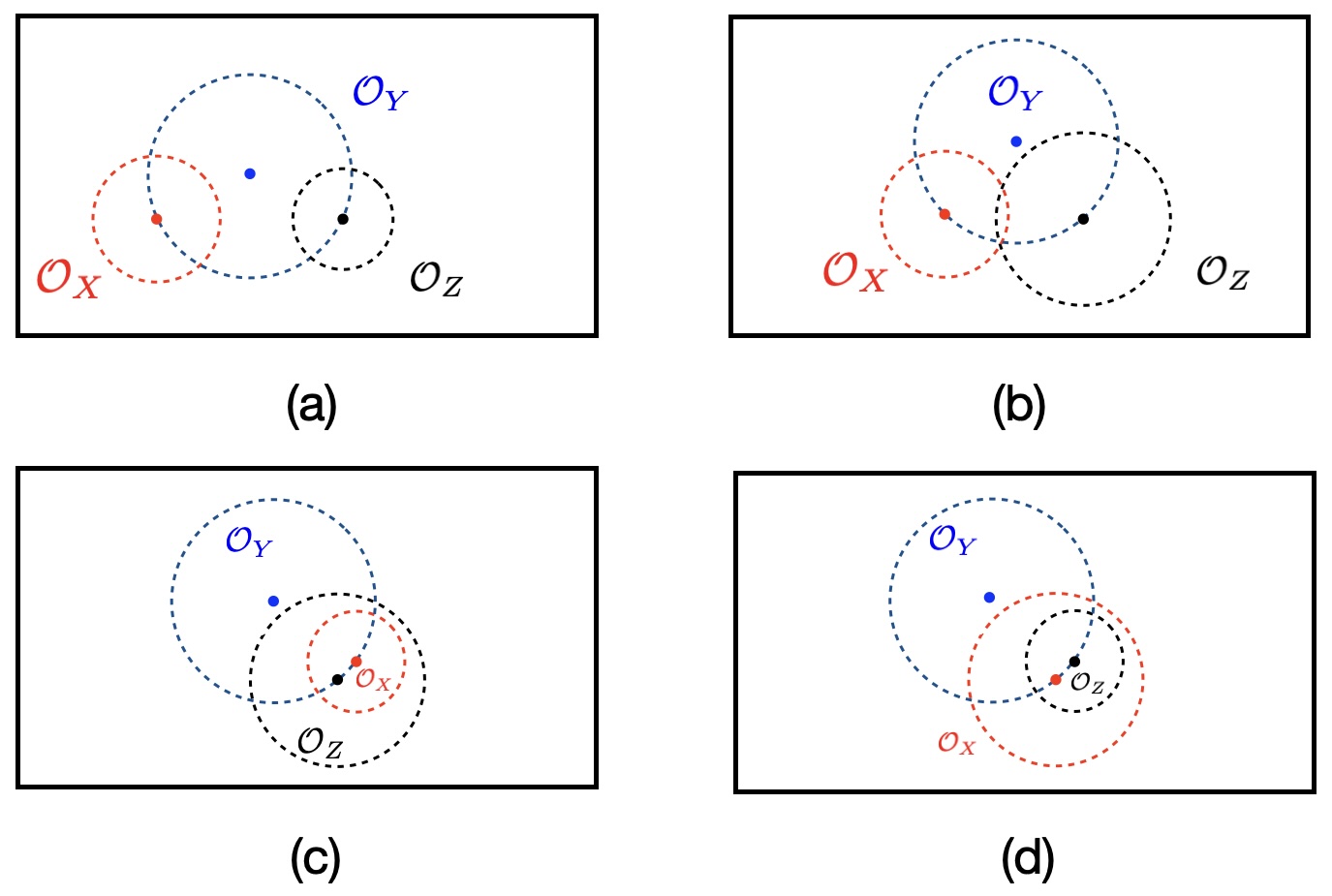} 
\caption{$\mathcal{O}_Y$ can partially overlap with $\mathcal{O}_X$ and $\mathcal{O}_Z$ in each case: (a) $\mathcal{O}_X$ disconnects from $\mathcal{O}_Z$, (b) $\mathcal{O}_X$ partially overlaps with $\mathcal{O}_Z$, (c) $\mathcal{O}_X$ is proper part of $\mathcal{O}_Z$, (d) $\mathcal{O}_Z$ is proper part of $\mathcal{O}_X$.}  
\label{tt_nn}
\end{figure}
\begin{PrfLma}
\begin{enumerate}
    \item $\mathbf{T}_3(\mathcal{O}_Z, \mathcal{O}_X) =\mathbf{D}(\mathcal{O}_Z, \mathcal{O}_X) $.
    
    Let $\mathcal{O}_Y$ be such a sphere whose centre is outside $\mathcal{O}_X$ and $\mathcal{O}_Z$ and whose boundary passes the centres of $\mathcal{O}_X$ and $\mathcal{O}_Z$. In this way,  $\mathcal{O}_Y$ partially overlaps with $\mathcal{O}_X$ and $\mathcal{O}_Z$. Therefore, for all $\mathbf{T}_1, \mathbf{T}_2\in\{\neg\mathbf{D}, \neg\mathbf{P}, \neg\overline{\mathbf{P}}\}$, we have $\mathbf{T}_1(\mathcal{O}_X, \mathcal{O}_Y)$ and $\mathbf{T}_2(\mathcal{O}_Y, \mathcal{O}_Z)$, as illustrated in Figure~\ref{tt_nn}(a).
\item  $\mathbf{T}_3(\mathcal{O}_Z, \mathcal{O}_X) = \mathbf{PO}(\mathcal{O}_Z, \mathcal{O}_X)$ or $ \mathbf{P}(\mathcal{O}_Z, \mathcal{O}_X)$ or $ \overline{\mathbf{P}}(\mathcal{O}_Z, \mathcal{O}_X) $.
The same as case 1,  illustrated in Figure~\ref{tt_nn}(b-d).  \qed
\end{enumerate}
\end{PrfLma}

\subsection{The relative qualitative space}

\begin{figure}[h]
\centering
\includegraphics[width=0.8
\textwidth]{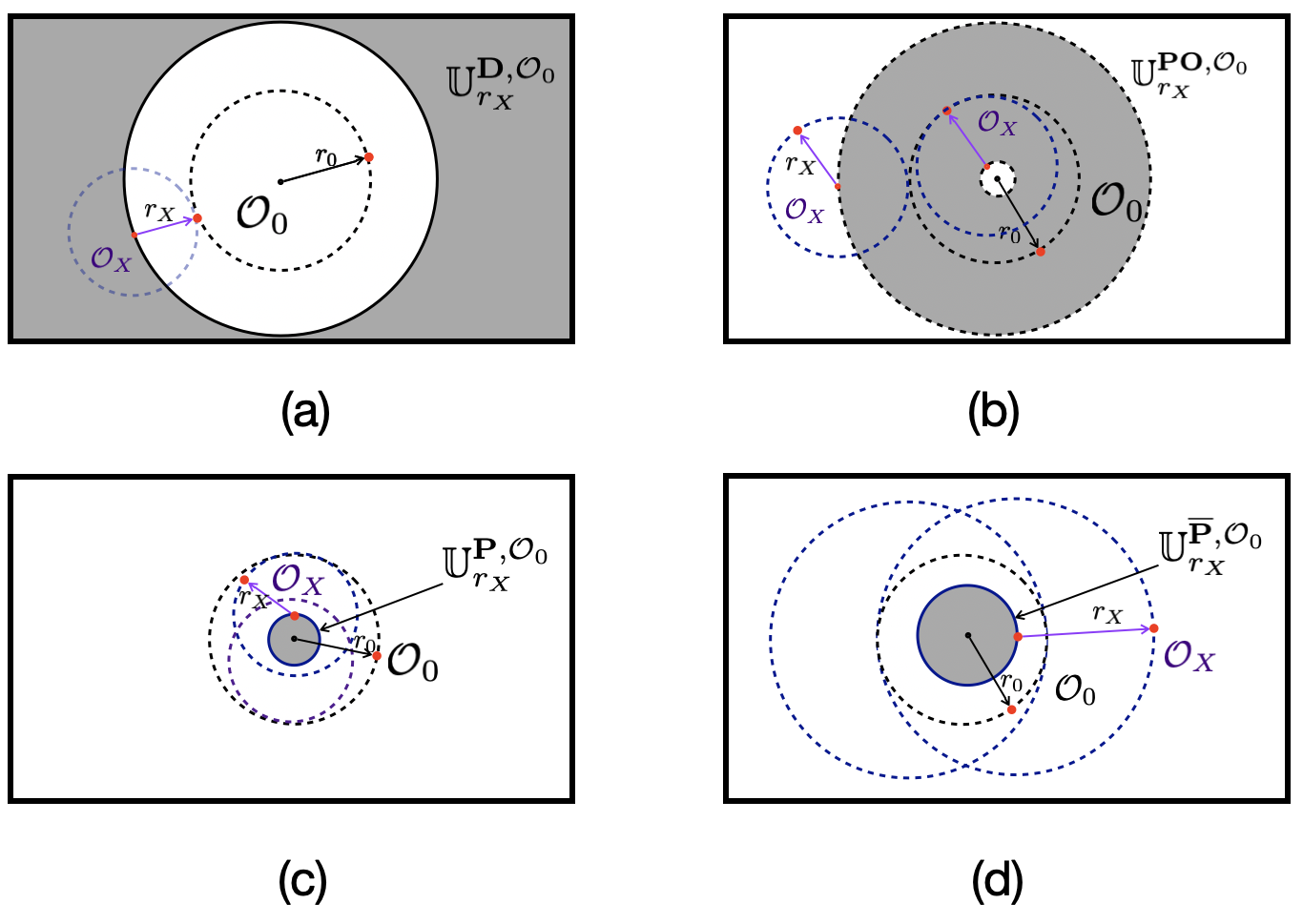}
\caption{(a) $\mathbb{U}^{\mathbf{D},\mathcal{O}_0}_{r_X}$ occupies the whole space except an open sphere with the centre $\vec{O}_0$ and the radius of $r_0 + r_X$. $\mathbb{U}^{\mathbf{D},\mathcal{O}_0}_{r_X}$ is concave; (b) $\mathbb{U}^{\mathbf{PO},\mathcal{O}_0}_{r_X}$ is an open ring concentric with $\mathcal{O}_0$ whose radius within the range of $|r_0 - r_X|$ and $r_0 + r_X$. $\mathbb{U}^{\mathbf{PO},\mathcal{O}_0}_{r_X}$ is concave, if $r_X<r_0$; (c) $\mathbb{U}^{\mathbf{P},\mathcal{O}_0}_{r_X}$ is a closed sphere with the centre $\vec{O}_0$ and the radius of $r_0 - r_X$. It is convex; (d)  $\mathbb{U}^{{\overline{\mathbf{P}}},\mathcal{O}_0}_{r_X}$ is a closed sphere that is concentric with $\mathcal{O}_0$ and with the radius of $r_X - r_0$. It is convex.}  
\label{cp_trace}
\end{figure}

\begin{Def}
\label{cp_tra}
Let $\mathcal{O}_0$ be a fixed sphere with radius $r_0$ and the centre $\vec{O}_0$, and let $\mathcal{O}_X$ be a moving sphere with fixed radius $r_X$, satisfying $\mathbf{R}(\mathcal{O}_X, \mathcal{O}_0)$, where $\mathbf{R}\in\{\mathbf{D}, \mathbf{PO}, \mathbf{P}, \overline{\mathbf{P}}\}$. All possible locations of the centre of $\mathcal{O}_X$ form a relative qualitative space $\mathbb{U}^{\mathbf{R},\mathcal{O}_0}_{r_X}$ as follows.
\begin{enumerate}
    \item $\mathbf{R}(\mathcal{O}_X, \mathcal{O}_0) = \mathbf{D}(\mathcal{O}_X, \mathcal{O}_0)$. $\mathbb{U}^{\mathbf{D},\mathcal{O}_0}_{r_X}$ is the space of all points $\vec{O}_X$ whose distance to $\vec{O}_0$ is greater than or equal to $r_X+r_0$. $\mathbb{U}^{\mathbf{D},\mathcal{O}_0}_{r_X}=\{\vec{O}_X : \|\vec{O}_X\vec{O}_0\|\ge r_X+r_0\}$, as illustrated in Figure~\ref{cp_trace}(a).
    \item $\mathbf{R}(\mathcal{O}_X, \mathcal{O}_0) = \mathbf{PO}(\mathcal{O}_X, \mathcal{O}_0)$. $\mathbb{U}^{\mathbf{PO},\mathcal{O}_0}_{r_X}$ is the space of all points $\vec{O}_X$ whose distance to $\vec{O}_0$ is less than $r_X+r_0$ and greater than the absolute difference between $r_X$ and $r_0$. $\mathbb{U}^{\mathbf{PO},\mathcal{O}_0}_{r_X}=\{\vec{O}_X : |r_X-r_0|<\|\vec{O}_X\vec{O}_0\|< r_X+r_0\}$. If $r_X\neq r_0$, $\mathbb{U}^{\mathbf{PO},\mathcal{O}_0}_{r_X}$ will be a ring, and thus is concave, as illustrated in Figure~\ref{cp_trace}(b).  If $r_X= r_0$, $\mathbb{U}^{\mathbf{PO},\mathcal{O}_0}_{r_X}$ will be a sphere expect its centre.
    \item $\mathbf{R}(\mathcal{O}_X, \mathcal{O}_0) = \mathbf{P}(\mathcal{O}_X, \mathcal{O}_0)$. If $r_0 \ge r_X$, $\mathbb{U}^{\mathbf{P},\mathcal{O}_0}_{r_X}$ is the space of all points $\vec{O}_X$ whose distance to $\vec{O}_0$ is less than or equal to $r_0 - r_X$. $\mathbb{U}^{\mathbf{P},\mathcal{O}_0}_{r_X}=\{\vec{O}_X : \|\vec{O}_X\vec{O}_0\|\le r_0 -r_X\}$, as illustrated in Figure~\ref{cp_trace}(c). If $r_0 < r_X$, $\mathbb{U}^{\mathbf{P},\mathcal{O}_0}_{r_X}$ is empty $\emptyset$.
    \item $\mathbf{R}(\mathcal{O}_X, \mathcal{O}_0) = \overline{\mathbf{P}}(\mathcal{O}_X, \mathcal{O}_0)$. If $r_X \ge r_0$, $\mathbb{U}^{{\overline{\mathbf{P}}},\mathcal{O}_0}_{r_X}$ is the space of all points $\vec{O}_X$ whose distance to $\vec{O}_0$ is less than or equal to $r_X - r_0$. $\mathbb{U}^{{\overline{\mathbf{P}}},\mathcal{O}_0}_{r_X}=\{\vec{O}_X : \|\vec{O}_X\vec{O}_0\|\le r_X -r_0\}$, as illustrated in Figure~\ref{cp_trace}(d). If $r_X < r_0$, $\mathbb{U}^{{\overline{\mathbf{P}}},\mathcal{O}_0}_{r_X}$ is empty $\emptyset$.
\end{enumerate}
\end{Def}

\begin{figure}[!t]
\centering
\includegraphics[width=1
\textwidth]{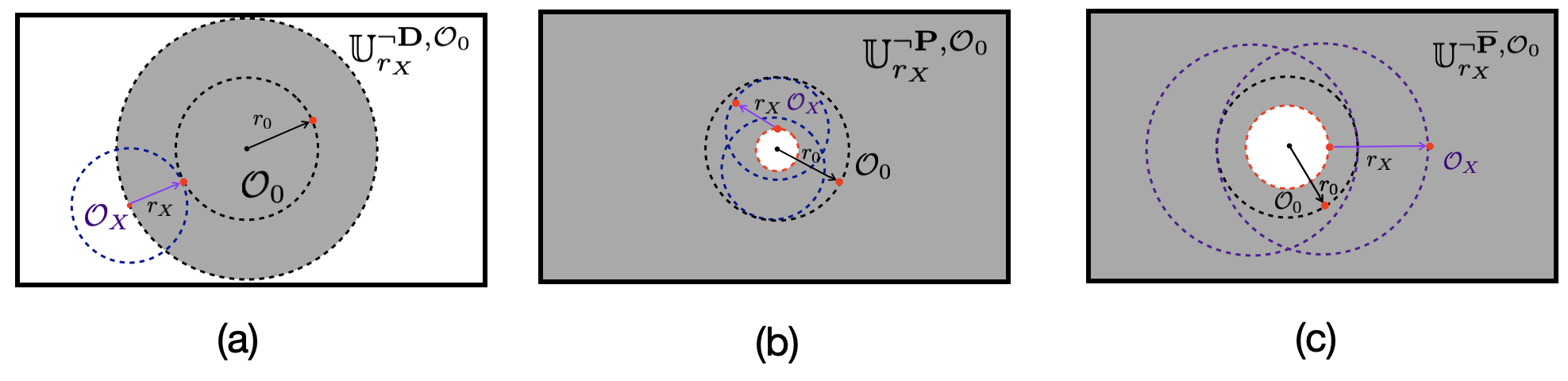}
\caption{(a) $\mathbb{U}^{\neg\mathbf{D},\mathcal{O}_0}_{r_X}$ is the complement space of $\mathbb{U}^{\mathbf{D},\mathcal{O}_0}_{r_X}$; (b) $\mathbb{U}^{\neg\mathbf{P},\mathcal{O}_0}_{r_X}$ is the complement space of $\mathbb{U}^{\mathbf{P}, \mathcal{O}_0}_{r_X}$; (c) $\mathbb{U}^{{\neg\overline{\mathbf{P}}},\mathcal{O}_0}_{r_X}$ is the complement space of $\mathbb{U}^{{\overline{\mathbf{P}}},\mathcal{O}_0}_{r_X}$.}  
\label{cp_trace_neg}
\end{figure}

\begin{Def}
\label{cp_tra_neg}
$\mathbb{U}^{\neg\mathbf{D},\mathcal{O}_0}_{r_X}$, $\mathbb{U}^{\neg\mathbf{P},\mathcal{O}_0}_{r_X}$, and $\mathbb{U}^{{\neg\overline{\mathbf{P}}},\mathcal{O}_0}_{r_X}$ are complement regions of  $\mathbb{U}^{\mathbf{D},\mathcal{O}_0}_{r_X}$, $\mathbb{U}^{\mathbf{P},\mathcal{O}_0}_{r_X}$, and $\mathbb{U}^{{\overline{\mathbf{P}}},\mathcal{O}_0}_{r_X}$, respectively. 
\begin{enumerate}
    \item  $\mathbb{U}^{\neg\mathbf{D},\mathcal{O}_0}_{r_X}$ is the space of all points $\vec{O}_X$ whose distance to $\vec{O}_0$ is less than $r_X+r_0$. $\mathbb{U}^{\neg\mathbf{D},\mathcal{O}_0}_{r_X}=\{\vec{O}_X : \|\vec{O}_X\vec{O}_0\|< r_X+r_0\}$, as illustrated in Figure~\ref{cp_trace_neg}(a). 
    \item If $r_0 \ge r_X$, $\mathbb{U}^{\neg\mathbf{P},\mathcal{O}_0}_{r_X}$ is the space of all points $\vec{O}_X$ whose distance to $\vec{O}_0$ is greater than  $r_0 - r_X$. $\mathbb{U}^{\neg\mathbf{P},\mathcal{O}_0}_{r_X}=\{\vec{O}_X : \|\vec{O}_X\vec{O}_0\|> r_0 -r_X\}$, as illustrated in Figure~\ref{cp_trace_neg}(b). If $r_0 < r_X$, $\mathbb{U}^{\neg\mathbf{P},\mathcal{O}_0}_{r_X}$ is the whole space $\mathbb{U}$.
    \item If $r_X \ge r_0$, $\mathbb{U}^{\neg\overline{\mathbf{P}},\mathcal{O}_0}_{r_X}$ is the space of all points $\vec{O}_X$ whose distance to $\vec{O}_0$ is greater than  $r_X - r_0$. $\mathbb{U}^{\neg\overline{\mathbf{P}},\mathcal{O}_0}_{r_X}=\{\vec{O}_X : \|\vec{O}_X\vec{O}_0\|> r_X -r_0\}$, as illustrated in Figure~\ref{cp_trace_neg}(c). If $r_X < r_0$, $\mathbb{U}^{\neg\overline{\mathbf{P}},\mathcal{O}_0}_{r_X}$ is the whole space $\mathbb{U}$.
\end{enumerate}
\end{Def}

\subsection{The rotation theorem in a relative qualitative space}

\begin{Cly}
\label{Cly2}
For any spheres $\mathcal{O}_X$ and $\mathcal{O}_V$, rotating $\mathcal{O}_X$ around the centre of $\mathcal{O}_V$ preserves the qualitative spatial relation between them. 
\end{Cly} 
\begin{PrfCly}
Each qualitative spatial relation is a function of the distance between their centres, $dis_{X,V}=\|\vec{O}_X-\vec{O}_V\|$, and the radii.
Rotating $\mathcal{O}_X$ around the centre of $\mathcal{O}_V$ preserves $dis_{X,V}$, and their radii $r_X$ and $r_V$. Therefore, it preserves the qualitative spatial relation. \qed
\end{PrfCly}

\begin{figure}
\centering
\includegraphics[width=1
\textwidth]{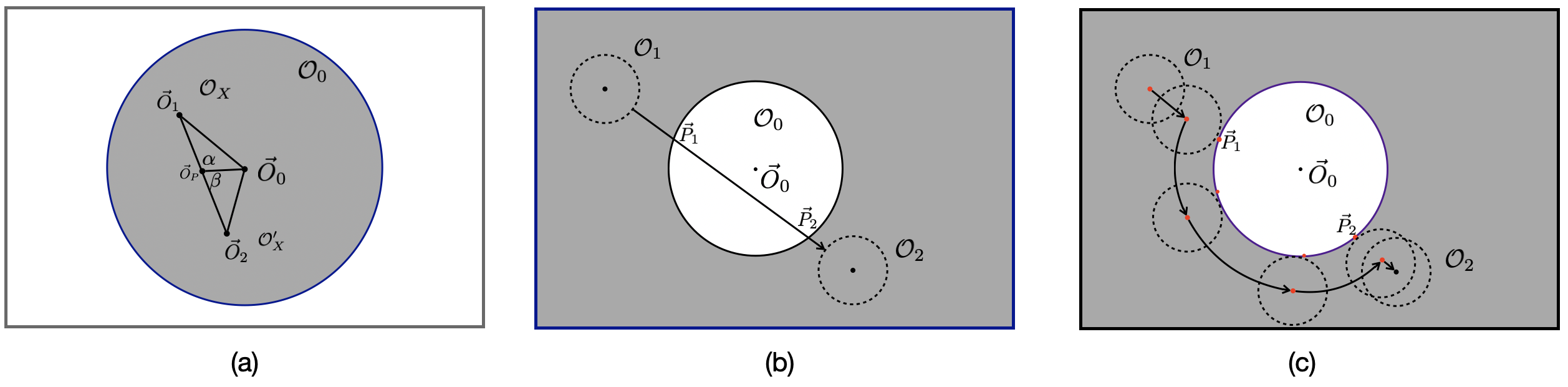}
\caption{(a) $\mathbb{U}^{\mathbf{P},\mathcal{O}}_{r_X}$ and $\mathbb{U}^{\overline{\mathbf{P}},\mathcal{O}}_{r_X}$ are convex spheres; (b) $\mathbb{U}^{\mathbf{D},\mathcal{O}}_{r_X}$ is concave ($\mathbb{U}^{\mathbf{PO},\mathcal{O}}_{r_X}$ can be concave), with a spherical hole; (c) this hole can be circumvented by rotating around it.}  
\label{cp_rotate}
\end{figure} 

\begin{Lma}
\label{rotate}
For any two different spheres $\mathcal{O}_1$ and $\mathcal{O}_2$ with the same radius $r$ and satisfying $\mathbf{R}(\mathcal{O}_1, \mathcal{O}_0)$ and $\mathbf{R}(\mathcal{O}_2, \mathcal{O}_0)$, if $\mathbf{R} \in\{\mathbf{P}, \overline{\mathbf{P}}\}$, directly move $\mathcal{O}_1$ to $\mathcal{O}_2$, the relation $\mathbf{R}$ will always hold during the movement process;  if $\mathbf{R} \in\{\mathbf{D}, \mathbf{PO}\}$, directly shifting $\mathcal{O}_1$ to $\mathcal{O}_2$ may violate the relation $\mathbf{R}$ during the movement process. To preserve $\mathbf{R}(\mathcal{O}_1, \mathcal{O}_0)$ during the process of shifting, $\mathcal{O}_1$ may need to rotate around the centre of $\mathcal{O}_0$.
\end{Lma}

\begin{PrfLma}
\begin{enumerate}
\item $\mathbf{R} \in\{\mathbf{P}, \overline{\mathbf{P}}\}$. $\mathbb{U}^{\mathbf{R},\mathcal{O}_0}_{r}$ is a close sphere $\mathcal{O}$ concentric with $\mathcal{O}_0$. Let the centre of $\mathcal{O}_1$ move from $\vec{O}_1$ to $\vec{O}_2$. Then, both $\vec{O}_1$ and $\vec{O}_2$ are inside $\mathcal{O}$, as shown in Figure~\ref{cp_rotate}(a). Let $\vec{O}_P$ be any point along the segment $|\vec{O}_1\vec{O}_2|$. Let $\alpha=\angle\vec{O}_0\vec{O}_P\vec{O}_1$ and $\beta=\angle\vec{O}_0\vec{O}_P\vec{O}_2$, $\alpha+\beta=180^\circ$, one of them is greater than or equal to $90^\circ$, therefore, $\|\vec{O}_0 - \vec{O}_P\|$ is less than $\max\{|\vec{O}_0\vec{O}_1|, |\vec{O}_0\vec{O}_2|\}$, so, $\vec{O}_P$ is inside $\mathcal{O}$. 
\item $\mathbf{R} \in\{\mathbf{D}, \mathbf{PO}\}$.  $\mathbb{U}^{\mathbf{D},\mathcal{O}_0}_{r}$ and $\mathbb{U}^{\mathbf{PO},\mathcal{O}_0}_{r}$ are concave. They may encompass a concentric sphere that does not belong to itself. Suppose that direct shifting $\mathcal{O}_1$ to $\mathcal{O}_2$ intersects with this concentric sphere at points $\vec{P}_1$ and $\vec{P}_2$, as illustrated in Figure~\ref{cp_rotate}(b). At point $\vec{P}_1$ the relation $\mathbf{R}$ exactly holds, with Corollary~\ref{topo_preserve}, rotating $\mathcal{O}_1$ at point $\vec{P}_1$ around the centre of $\mathcal{O}_0$ to point $\vec{P}_2$ will preserve the relation $\mathbf{R}$, as illustrated in Figure~\ref{cp_rotate}(c). \qed
\end{enumerate}
\end{PrfLma}

\subsection{The constraint optimisation is gradual descent}
\begin{figure} 
\centering
\includegraphics[width=0.9\textwidth]{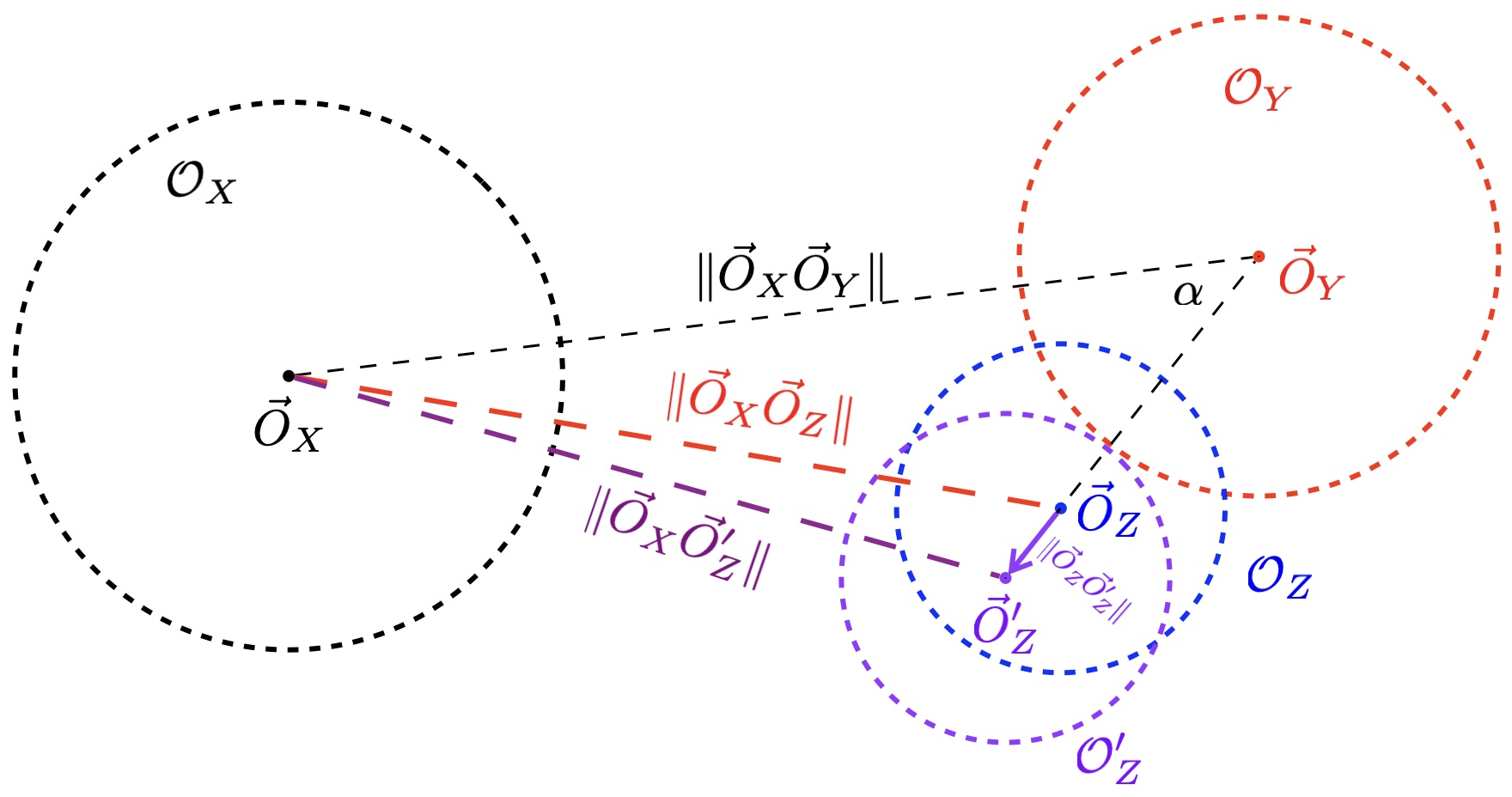}
\caption{The gradual descent along the line segment $\|\vec{O}_Z\vec{O}'_Z\|$ may cause the gradual ascent of  $| \|\vec{O}_X\vec{O}'_Z\|- \|\vec{O}_X\vec{O}_Z\| |$, however, $| \|\vec{O}_X\vec{O}'_Z\|- \|\vec{O}_X\vec{O}_Z\| | \le \|\vec{O}_Z\vec{O}'_Z\|$.}
\label{cop_mono}
\end{figure}

\begin{Thm}
\label{mono_cop}
Let $\mathcal{O}_X$ and $\mathcal{O}_Y$ be two fixed non-concentric spheres; $\mathcal{O}_Z$ be a movable sphere; $\mathbf{T}_{ZY}$ and $\mathbf{T}_{ZX}$ be the target relations of $\mathcal{O}_Z$ to $\mathcal{O}_Y$ and $\mathcal{O}_X$, respectively, $\mathbf{T}_{ZY}, \mathbf{T}_{ZX}\in\mathcal{T}=\{\mathbf{D}, \mathbf{P}, \overline{\mathbf{P}}, \neg\mathbf{D}, \neg\mathbf{P}, \neg\overline{\mathbf{P}}\}$.  
$COP^{\mathbf{T}_{ZX}}_{\mathbf{T}_{ZY}}(\mathcal{O}_{Z}| \mathcal{O}_{X}; \mathcal{O}_{Y})$ 
is gradual descent. 
\end{Thm} 

\begin{Prf} 
$\mathbf{S}_{ZY}$ and $\mathbf{S}_{ZX}$ be the actual relations of $\mathcal{O}_Z$ to $\mathcal{O}_Y$ and $\mathcal{O}_X$, respectively. $\mathbf{S}_{ZX}\in f_{tsp}(\mathbf{T}_{ZX})$ and $\mathbf{S}_{ZY}\in f_{tsp}(\mathbf{T}_{ZY})$; the relation  $\tilde{\mathbf{R}}_2(\mathcal{O}_Z, \mathcal{O}_Y)$ and $\mathbf{R}_3(\mathcal{O}_Z, \mathcal{O}_X)$ be the next target relations of $\mathcal{O}_Z$ to $\mathcal{O}_Y$ and to $\mathcal{O}_X$, respectively, where $\tilde{\mathbf{R}}_2$, $\mathbf{R}_3\in\mathcal{S}$. $COP^{\mathbf{T}_{ZX}}_{\mathbf{T}_{ZY}}(\mathcal{O}_{Z}| \mathcal{O}_{X}; \mathcal{O}_{Y})$ repeatedly performs two steps as follows: (1) it gradually decreases the value of the function $\Delta_{\mathbf{S}_{ZX}}^{\mathbf{R}_3}(\mathcal{O}_Z, \mathcal{O}_X)  + \Delta_{\mathbf{S}_{ZY}}^{\tilde{\mathbf{R}}_2}(\mathcal{O}_Z, \mathcal{O}_Y)$;  (2) while $\Delta_{\mathbf{S}_{ZY}}^{\tilde{\mathbf{R}}_2}(\mathcal{O}_Z, \mathcal{O}_Y)>0$, gradual descent operations will be applied for $\Delta_{\mathbf{S}_{ZY}}^{\tilde{\mathbf{R}}_2}(\mathcal{O}_Z, \mathcal{O}_Y)$. 

Consider the case that the radius of $\mathcal{O}_Z$ is fixed, and repeatedly perform gradual descent operation on $\Delta_{\mathbf{S}_{ZY}}^{\tilde{\mathbf{R}}_2}(\mathcal{O}_Z, \mathcal{O}_Y)$ until it equals zero, at this time $\Delta_{\mathbf{S}_{ZX}}^{\mathbf{R}_3}(\mathcal{O}_Z, \mathcal{O}_X)$ may increase a value $\delta^{*}_0$. We need to prove that the decreased value related to $\mathcal{O}_Y$, is no less than $\delta^{*}_0$.

When $\mathcal{O}_Y$ is fixed, with Theorem~\ref{mono_delta}, the gradual descent operation on $\Delta_{\mathbf{S}_{ZY}}^{\tilde{\mathbf{R}}_2}(\mathcal{O}_Z, \mathcal{O}_Y)$ will move $\mathcal{O}_Z$ along the straight line $\vec{O}_Y\vec{O}_Z$, with the decrease of $\delta^{'}_1=\|\vec{O}_Z\vec{O}'_Z\|$. This may cause a maximum increase of $\Delta_{\mathbf{S}_{ZX}}^{\mathbf{R}_3}(\mathcal{O}_Z, \mathcal{O}_X)$ with the value of $\delta^{*}_0=| \|\vec{O}_X\vec{O}'_Z\|- \|\vec{O}_X\vec{O}_Z\| |$, shown in Figure~\ref{cop_mono}. We have $\delta^{'}_1 = \|\vec{O}_Z\vec{O}'_Z\|\ge |\|\vec{O}_X\vec{O}'_Z\|- \|\vec{O}_X\vec{O}_Z\| |= \delta^{*}_0$. The equal relation holds if $\vec{O}_X$, $\vec{O}_Z$, and $\vec{O}'_Z$ are collinear.

Consider the case that 
the gradual descent operation on $\Delta_{\mathbf{S}_{ZY}}^{\tilde{\mathbf{R}}_2}(\mathcal{O}_Z, \mathcal{O}_Y)$ also update the radius of $\mathcal{O}_Z$, with the change of $\delta_{r_Z}$. This value helps to reduce the value of $\Delta_{\mathbf{S}_{ZY}}^{\tilde{\mathbf{R}}_2}(\mathcal{O}_Z, \mathcal{O}_Y)$. With Corollary~\ref{mono_delta}, the same amount of the value may increase the value of $\Delta_{\mathbf{S}_{ZX}}^{\mathbf{R}_3}(\mathcal{O}_Z, \mathcal{O}_X)$, which exactly counteracts the decreased value from the relation to $\mathcal{O}_Y$. Therefore, $COP^{\mathbf{T}_{ZX}}_{\mathbf{T}_{ZY}}(\mathcal{O}_Z| \mathcal{O}_X; \mathcal{O}_Y)$ is gradual descent.     \qed
\end{Prf} 
  
\subsection{Theorems about constraint optimisation}

\begin{Lma}
\label{enn_thm0}
Let $\mathbf{R}_1$, $\mathbf{R}_2$, and $\mathbf{R}_3$ be {\em satisfiable}, where $\mathbf{R}_1, \mathbf{R}_2\in\{\mathbf{D}, \mathbf{P},\mathbf{PO}, \overline{\mathbf{P}}\}$, $\mathbf{R}_3\in\{\mathbf{D}, \mathbf{P}, \overline{\mathbf{P}}, \neg\mathbf{D}, \neg\mathbf{P}, \neg\overline{\mathbf{P}}\}$. Let spheres $\mathcal{O}_X$ and $\mathcal{O}_Y$ be fixed and satisfy the relation $\mathbf{R}_1(\mathcal{O}_X, \mathcal{O}_Y)$.
$\mathcal{S}$phNN can construct $\mathcal{O}_Z$ such that  $\mathbf{R}_2(\mathcal{O}_Y, \mathcal{O}_Z)$ and $\mathbf{R}_3(\mathcal{O}_Z, \mathcal{O}_X)$.
\end{Lma}
\begin{PrfLma}  
With Theorem~\ref{lmanoncyc}, $\mathcal{S}$phNN can construct $\mathcal{O}_Z$ such that  $\mathbf{R}_2(\mathcal{O}_Y, \mathcal{O}_Z)$.
With Theorem~\ref{mono_cop}, $COP(\mathcal{O}_Z|
\mathcal{O}_X; \mathcal{O}_Y)$ is gradual descent. 
\begin{enumerate}
    \item $\mathbf{R}_3\in\{\mathbf{D}, \mathbf{P}, \overline{\mathbf{P}}\}$.
    
    With Lemma~\ref{basic}, there is $\mathcal{O}^*_Z$ satisfying  $\mathbf{R}_2(\mathcal{O}_Y, \mathcal{O}^*_Z)$ and $\mathbf{R}_3(\mathcal{O}^*_Z, \mathcal{O}_X)$. That is, $COP(\mathcal{O}^*_{Z}| \mathcal{O}_X; \mathcal{O}_Y)=0$.
    \item $\mathbf{R}_3\in\{ \neg\mathbf{D}, \neg\mathbf{P}, \neg\overline{\mathbf{P}}\}$.
    
    With Lemma~\ref{lma_1neg_sat}, there is $\mathcal{O}'_Z$ satisfying  $\mathbf{R}_2(\mathcal{O}_Y, \mathcal{O}'_z)$ and $\mathbf{R}_3(\mathcal{O}'_Z, \mathcal{O}_X)$. That is, $COP(\mathcal{O}'_{Z}| \mathcal{O}_X; \mathcal{O}_Y)=0$.
\end{enumerate}

In both cases, $COP(\mathcal{O}_Z| \mathcal{O}_X; \mathcal{O}_Y)$ will reach 0. Therefore, $\mathcal{S}$phNN can construct $\mathcal{O}_Z$ such that  $\mathbf{R}_2(\mathcal{O}_Y, \mathcal{O}_Z)$ and $\mathbf{R}_3(\mathcal{O}_Z, \mathcal{O}_X)$ by gradual descending the function $COP(\mathcal{O}_Z| \mathcal{O}_X; \mathcal{O}_Y)$. \qed
\end{PrfLma}

\begin{Lma}
\label{lma_2neg}
Let $\mathcal{O}_X$, $\mathcal{O}_Y$, and $\mathcal{O}_Z$ be spheres that satisfy three relations $\mathbf{T}_1(\mathcal{O}_X, \mathcal{O}_Y)$, $\mathbf{T}_2(\mathcal{O}_Y, \mathcal{O}_Z)$ and $\mathbf{T}_3(\mathcal{O}_Z, \mathcal{O}_X)$, where $\mathbf{T}_1, \mathbf{T}_2\in\{ \neg\mathbf{D}, \neg\mathbf{P}, \neg\overline{\mathbf{P}}\}$, $\mathbf{T}_3\in\{\mathbf{D}, \mathbf{P}, \overline{\mathbf{P}}, \mathbf{PO}\}$. $\mathcal{S}$phNN can construct an Euler Diagram by first realising $\mathbf{T}_3(\mathcal{O}_Z, \mathcal{O}_X)$, then fix $\mathcal{O}_Z$ and $\mathcal{O}_X$, and constructs $\mathcal{O}_Y$ to satisfy both $\mathbf{T}_1(\mathcal{O}_X, \mathcal{O}_Y)$ and $\mathbf{T}_2(\mathcal{O}_Y, \mathcal{O}_Z)$.
\end{Lma}

\begin{PrfLma} Let $\mathbf{S}_1$ be the actual relation between
$\mathcal{O}_X$ and $\mathcal{O}_Y$, and $\mathbf{S}_2$ be the actual relation between
$\mathcal{O}_Y$ and $\mathcal{O}_Z$, where   $\mathbf{S}_1\in\ f_{tsp}(\mathbf{T}_1) =\{\mathbf{T}_1, \neg\mathbf{T}_1\}$, $\mathbf{S}_2\in\ f_{tsp}(\mathbf{T}_2) =\{\mathbf{T}_2, \neg\mathbf{T}_2\}$, 
$\mathbf{T}_1, \mathbf{T}_2\in\{ \neg\mathbf{D}, \neg\mathbf{P}, \neg\overline{\mathbf{P}}\}$. With Lemma~\ref{lma_2neg_sat},  for any fixed $\mathcal{O}_Z$ and $\mathcal{O}_X$ satisfying $\mathbf{T}_3(\mathcal{O}_Z, \mathcal{O}_X)$, there exists $\mathcal{O}^*_Y$ satisfying $\mathbf{T}_1(\mathcal{O}_X, \mathcal{O}^*_Y)$ and $\mathbf{T}_2(\mathcal{O}^*_Y, \mathcal{O}_Z)$, in which $\mathbf{T}_1, \mathbf{T}_2\in\{\neg\mathbf{D}, \neg\mathbf{P}, \neg\overline{\mathbf{P}}\}$.
So, $COP(\mathcal{O}^*_{Y}| \mathcal{O}_{X}; \mathcal{O}_{Z})=0$. 
Therefore, $\mathcal{S}$phNN can construct an Euler Diagram by first realising $\mathbf{T}_3(\mathcal{O}_Z, \mathcal{O}_X)$ (Theorem~\ref{lmanoncyc}), then fix $\mathcal{O}_Z$ and $\mathcal{O}_X$, and constructs $\mathcal{O}_Y$ to satisfy both $\mathbf{T}_1(\mathcal{O}_X, \mathcal{O}_Y)$ and $\mathbf{T}_2(\mathcal{O}_Y, \mathcal{O}_Z)$ by gradual descending the function $COP(\mathcal{O}_{Y}| \mathcal{O}_{X}; \mathcal{O}_{Z})$.   \qed
\end{PrfLma}

\begin{Thm}
\label{thm_control2}
Let $\mathbf{R}_1$, $\mathbf{T}_2$, and $\mathbf{T}_3$ be {\em satisfiable}, where $\mathbf{R}_1\in\{\mathbf{D}, \mathbf{P},\mathbf{PO}, \overline{\mathbf{P}}\}$, $\mathbf{T}_2, \mathbf{T}_3\in\mathcal{T}=\{\mathbf{D}, \mathbf{P}, \overline{\mathbf{P}}, \neg\mathbf{D}, \neg\mathbf{P}, \neg\overline{\mathbf{P}}\}$. Let $\mathcal{O}_X$ and $\mathcal{O}_Y$ be fixed and satisfy $\mathbf{R}_1(\mathcal{O}_X, \mathcal{O}_Y)$.
$\mathcal{S}$phNN can construct $\mathcal{O}_Z$ such that  $\mathbf{T}_2(\mathcal{O}_Y, \mathcal{O}_Z)$, and $\mathbf{T}_3(\mathcal{O}_Z, \mathcal{O}_X)$.
\end{Thm}
\begin{Prf} 
Lemma~\ref{enn_thm0} and Lemma~\ref{lma_2neg}. \qed
\end{Prf}

\subsection{The restart theorem}

\begin{Lma}
\label{enn_lm1}
Let $\mathcal{O}_X$ and $\mathcal{O}_Y$ be fixed, satisfying $\mathbf{T}_1(\mathcal{O}_X, \mathcal{O}_Y)$, and $\mathcal{O}_Z$ be movable, satisfying $\mathbf{T}_2(\mathcal{O}_Y, \mathcal{O}_Z)$, where $\mathbf{T}_1, \mathbf{T}_2\in\{\mathbf{D}, \mathbf{P}, \overline{\mathbf{P}}\}$. Let the relation between $\mathcal{O}_Z$ and $\mathcal{O}_X$ be $\mathbf{R}_3(\mathcal{O}_Z, \mathcal{O}_X)$ and the three relations $\mathbf{T}_1, \mathbf{T}_2, \mathbf{R}_3$ are satisfiable, where $\mathbf{R}_3\in\{\mathbf{D}, \mathbf{PO}, \mathbf{P}, \overline{\mathbf{P}}\}$. The number of possible relations of $\mathbf{R}_3$ can not be 2. \end{Lma}
\begin{PrfLma}
We enumerate relations of $\mathbf{T}_1$ and $\mathbf{T}_2$.
\begin{enumerate}
\begin{figure}
\centering
\includegraphics[width=0.8
\textwidth]{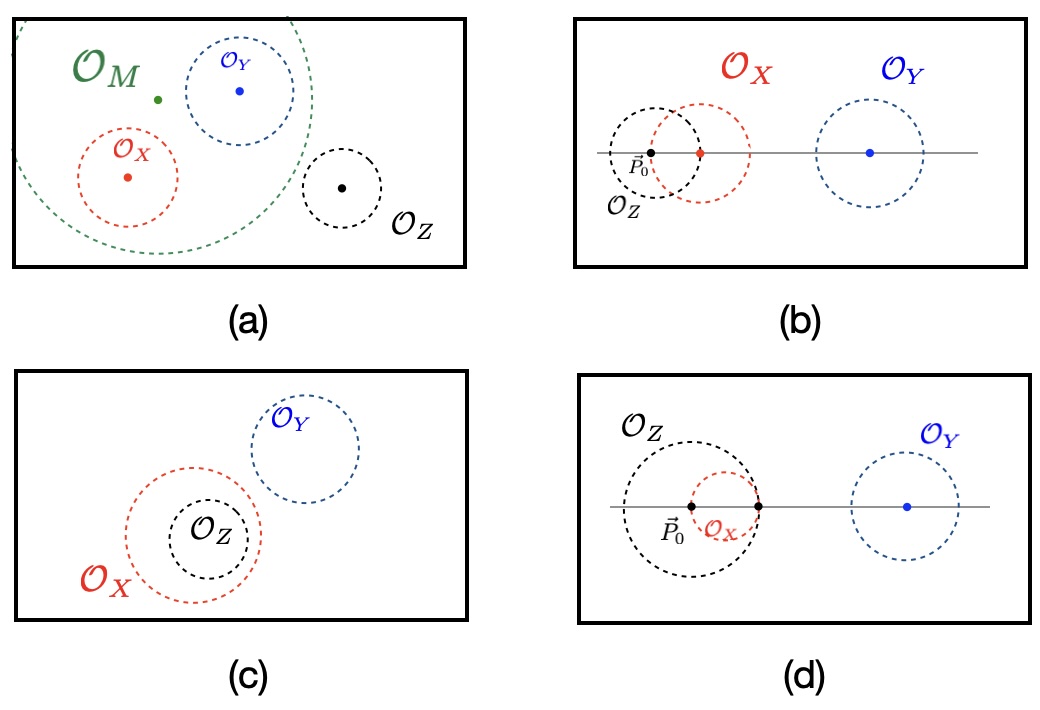} 
\caption{When $\mathcal{O}_X$ disconnects from $\mathcal{O}_Y$, and $\mathcal{O}_Y$ disconnects from $\mathcal{O}_Z$, all 4 qualitative relations between $\mathcal{O}_X$ and $\mathcal{O}_Z$ are possible.}  
\label{tt_dd}
\end{figure}
\item $\mathbf{T}_1(\mathcal{O}_X, \mathcal{O}_Y)=\mathbf{D}(\mathcal{O}_X, \mathcal{O}_Y)$ and $\mathbf{T}_2(\mathcal{O}_Y, \mathcal{O}_Z)=\mathbf{D}(\mathcal{O}_Y, \mathcal{O}_Z)$. 
 
\begin{enumerate}
    \item Let $\mathcal{O}_M$ be a sphere that contains $\mathcal{O}_X$ and $\mathcal{O}_Y$. Any $\mathcal{O}_Z$ that disconnects from $\mathcal{O}_M$ disconnects from $\mathcal{O}_X$ and $\mathcal{O}_Y$, $\mathbf{D}(\mathcal{O}_Z, \mathcal{O}_X)$ shown in Figure~\ref{tt_dd}(a).
    \item  ${\vec{P}_0}$ be the apogee to $\mathcal{O}_Y$ at the boundary of $\mathcal{O}_X$. Let $\mathcal{O}_Z$ take ${\vec{P}_0}$ as the centre and have the same radius as $\mathcal{O}_X$, then $\mathcal{O}_Z$ partially overlaps with $\mathcal{O}_X$ and disconnects from $\mathcal{O}_Y$, $\mathbf{PO}(\mathcal{O}_Z, \mathcal{O}_X)$ shown in Figure~\ref{tt_dd}(b).
    \item Let  $\mathcal{O}_Z$ be part of $\mathcal{O}_X$, $\mathcal{O}_Z$ will disconnect from $\mathcal{O}_Y$, $\mathbf{P}(\mathcal{O}_Z, \mathcal{O}_X)$, shown in Figure~\ref{tt_dd}(c).
    \item ${\vec{P}_0}$ be the apogee to $\mathcal{O}_Y$ at the boundary of $\mathcal{O}_X$. Let $\mathcal{O}_Z$ take ${\vec{P}_0}$ as the centre and the diameter of $\mathcal{O}_X$ as the radius, then $\mathcal{O}_X$ is part of $\mathcal{O}_Z$ and disconnects from $\mathcal{O}_Y$, $\overline{\mathbf{P}}(\mathcal{O}_Z, \mathcal{O}_X)$, shown in Figure~\ref{tt_dd}(d). 
\end{enumerate}
So, the number of possible relations of $\mathbf{R}_3$ is 4.
\begin{figure}
\centering
\includegraphics[width=0.8
\textwidth]{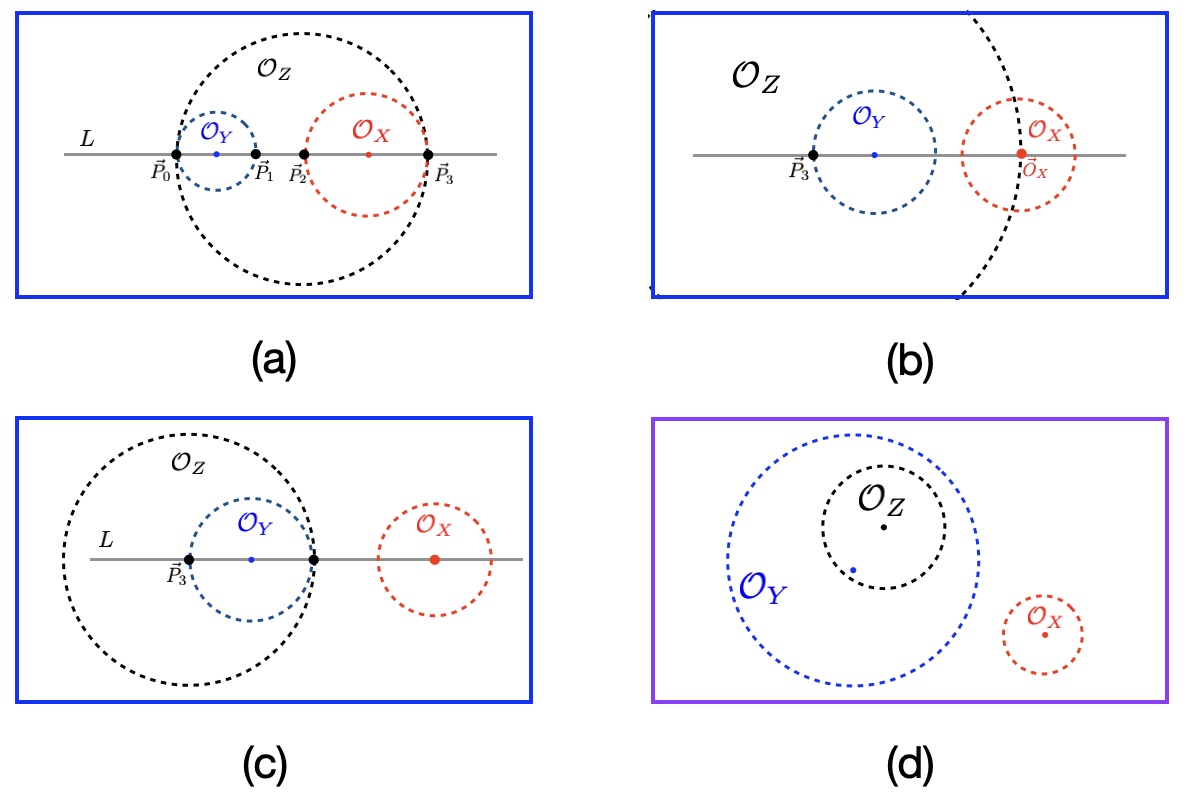} 
\caption{(a-c) $\mathcal{O}_X$ disconnects from  $\mathcal{O}_Y$ and  $\mathcal{O}_Y$ is proper part of  $\mathcal{O}_Z$, then, $\mathcal{O}_Z$ cannot be part of $\mathcal{O}_X$, and other qualitative relations between $\mathcal{O}_X$ and $\mathcal{O}_Z$ are possible; (d) if $\mathcal{O}_X$ disconnects from  $\mathcal{O}_Y$ and  $\mathcal{O}_Z$ is proper part of  $\mathcal{O}_Y$, $\mathcal{O}_X$ will disconnect from $\mathcal{O}_Z$.}  
\label{tt_ppbard}
\end{figure}
\begin{figure}
\centering
\includegraphics[width=0.8
\textwidth]{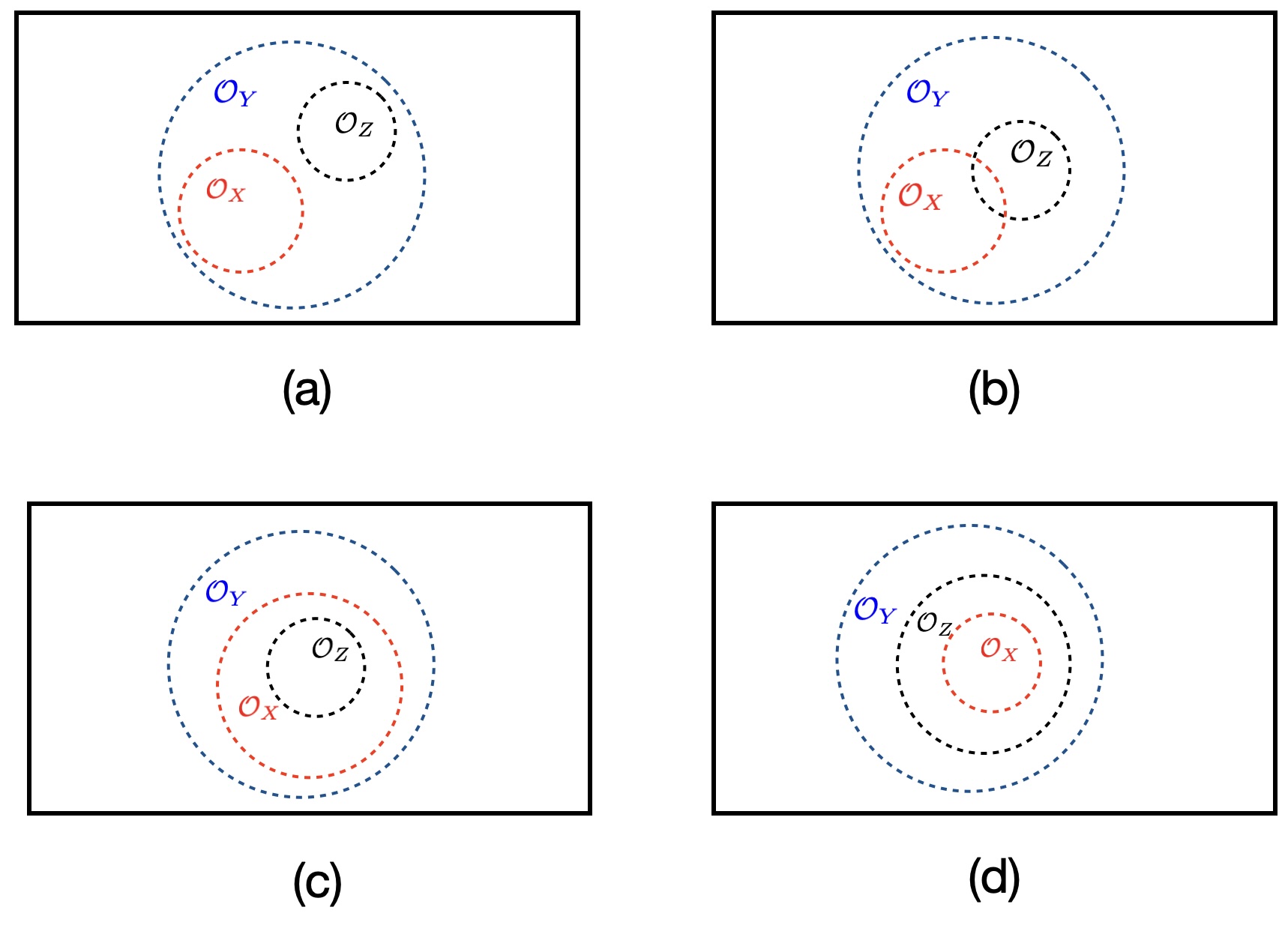} 
\caption{If $\mathcal{O}_Y$ contains both $\mathcal{O}_X$ and  $\mathcal{O}_Z$, $\mathcal{O}_X$ and  $\mathcal{O}_Z$ can be of any relations.}
\label{tt_ppbar}
\end{figure}
\item $\mathbf{T}_1(\mathcal{O}_X, \mathcal{O}_Y)=\mathbf{D}(\mathcal{O}_X, \mathcal{O}_Y)$ and $\mathbf{T}_2(\mathcal{O}_Y, \mathcal{O}_Z)=\mathbf{P}(\mathcal{O}_Y, \mathcal{O}_Z)$.
\begin{enumerate}
    \item Let the line $L$ pass the centres of $\mathcal{O}_X$ and $\mathcal{O}_Y$, intersect with  boundaries of $\mathcal{O}_X$ and $\mathcal{O}_Y$ at ${\vec{P}_0}$, ${\vec{P}_1}$, ${\vec{P}_2}$, and ${\vec{P}_3}$, respectively, shown in Figure~\ref{tt_ppbard}(a). Let $\mathcal{O}_Z$ be the sphere with $\vec{P}_0\vec{P}_3$ as diameter, $\overline{\mathbf{P}}(\mathcal{O}_Z, \mathcal{O}_X)$;
    \item  Let the line $L$ pass the centres of $\mathcal{O}_X$ and $\mathcal{O}_Y$, intersect with the  boundary of  $\mathcal{O}_Y$ at  ${\vec{P}_3}$, the apogee to $\mathcal{O}_X$.  Let $\mathcal{O}_Z$ take ${\vec{P}_3}$ as the centre and  $\mathcal{O}_Z$'s boundary pass the centre of $\mathcal{O}_X$, then $\mathcal{O}_Z$ contains $\mathcal{O}_Y$ and partially overlaps with $\mathcal{O}_X$, $\mathbf{PO}(\mathcal{O}_Z, \mathcal{O}_X)$, shown in Figure~\ref{tt_ppbard}(b);
    \item  Let $\mathcal{O}_Z$ take ${\vec{P}_3}$ (created in case (b)) as the centre and take the diameter of $\mathcal{O}_Y$ as the radius, then $\mathcal{O}_Z$ contains $\mathcal{O}_Y$ and disconnects from  $\mathcal{O}_X$, $\mathbf{D}(\mathcal{O}_Z, \mathcal{O}_X)$, shown in Figure~\ref{tt_ppbard}(c).
    \item  If $\mathcal{O}_Z$ is proper part of $\mathcal{O}_X$, $\mathcal{O}_X$ disconnects from  $\mathcal{O}_Y$, $\mathcal{O}_Z$ will disconnect from $\mathcal{O}_Y$. This contradicts with the relation that $\mathcal{O}_Y$ is proper part of $\mathcal{O}_Z$,  $\mathbf{P}(\mathcal{O}_Y, \mathcal{O}_Z)$.
\end{enumerate}
 So, the number of possible relations of $\mathbf{R}_3$ is 3.  
\item $\mathbf{T}_1(\mathcal{O}_X, \mathcal{O}_Y)=\mathbf{D}(\mathcal{O}_X, \mathcal{O}_Y)$ and $\mathbf{T}_2(\mathcal{O}_Y, \mathcal{O}_Z)=\overline{\mathbf{P}}(\mathcal{O}_Y, \mathcal{O}_Z)$. As $\mathcal{O}_X$ disconnects from $\mathcal{O}_Y$ and   $\mathcal{O}_Z$ is inside $\mathcal{O}_Y$, so $\mathcal{O}_X$ disconnects from $\mathcal{O}_Z$, as shown in  Figure~\ref{tt_ppbard}(d). So, the number of possible relations of $\mathbf{R}_3$ is 1.
\item $\mathbf{T}_1(\mathcal{O}_X, \mathcal{O}_Y)=\mathbf{P}(\mathcal{O}_X, \mathcal{O}_Y)$ and $\mathbf{T}_2(\mathcal{O}_Y, \mathcal{O}_Z)=\mathbf{D}(\mathcal{O}_Y, \mathcal{O}_Z)$. Case 2.

\item $\mathbf{T}_1(\mathcal{O}_X, \mathcal{O}_Y)= \mathbf{P}(\mathcal{O}_X, \mathcal{O}_Y)$ and $\mathbf{T}_2(\mathcal{O}_Y, \mathcal{O}_Z)=\overline{\mathbf{P}}(\mathcal{O}_Y, \mathcal{O}_Z)$. $\mathcal{O}_X$ and  $\mathcal{O}_Z$ are part of  $\mathcal{O}_Y$, $\mathcal{O}_X$ and  $\mathcal{O}_Z$ can be of any relations, as shown in Figure~\ref{tt_ppbar}. So, the number of possible relations of $\mathbf{R}_3$ is 4. 
\item $\mathbf{T}_1(\mathcal{O}_X, \mathcal{O}_Y)=\mathbf{P}(\mathcal{O}_X, \mathcal{O}_Y)$ and $\mathbf{T}_2(\mathcal{O}_Y, \mathcal{O}_Z)=\mathbf{P}(\mathcal{O}_Y, \mathcal{O}_Z)$. When $\mathcal{O}_X$ is proper part of $\mathcal{O}_Y$ and $\mathcal{O}_Y$ is proper part of $\mathcal{O}_Z$, $\mathcal{O}_X$ will be part of $\mathcal{O}_Z$, as shown in  Figure~\ref{tt_pppbarp}(a). So, the number of possible relations of $\mathbf{R}_3$ is 1.
\item $\mathbf{T}_1(\mathcal{O}_X, \mathcal{O}_Y)=\overline{\mathbf{P}}(\mathcal{O}_X, \mathcal{O}_Y)$ and $\mathbf{T}_2(\mathcal{O}_Y, \mathcal{O}_Z)=\mathbf{D}(\mathcal{O}_Y, \mathcal{O}_Z)$. Case 3.
\begin{figure}
\centering
\includegraphics[width=0.8
\textwidth]{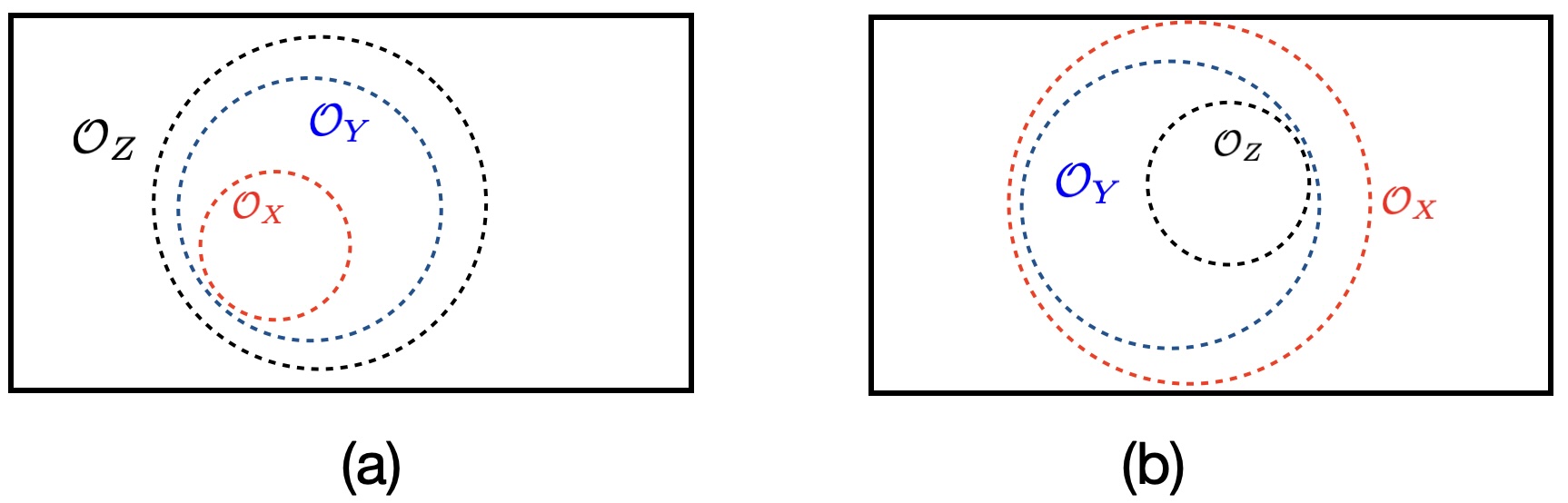} 
\caption{(a) if $\mathcal{O}_X$ is proper part of $\mathcal{O}_Y$ and $\mathcal{O}_Y$ is proper part of $\mathcal{O}_Z$, $\mathcal{O}_X$ will be part of $\mathcal{O}_Z$; switching $\mathcal{O}_X$ and $\mathcal{O}_Z$ will result in the case (b).}  
\label{tt_pppbarp}
\end{figure}
\item $\mathbf{T}_1(\mathcal{O}_X, \mathcal{O}_Y)=\overline{\mathbf{P}}(\mathcal{O}_X, \mathcal{O}_Y)$ and $\mathbf{T}_2(\mathcal{O}_Y, \mathcal{O}_Z)=\mathbf{P}(\mathcal{O}_Y, \mathcal{O}_Z)$. Case 5.
\item $\mathbf{T}_1(\mathcal{O}_X, \mathcal{O}_Y)=\overline{\mathbf{P}}(\mathcal{O}_X, \mathcal{O}_Y)$ and $\mathbf{T}_2(\mathcal{O}_Y, \mathcal{O}_Z)=\overline{\mathbf{P}}(\mathcal{O}_Y, \mathcal{O}_Z)$. This is equivalent to Case 6, as shown in Figure~\ref{tt_pppbarp}(b). So, the number of possible relations of $\mathbf{R}_3$ is 1.
\end{enumerate}
Therefore, the number of possible relations of $\mathbf{R}_3$ can not be 2. \qed
\end{PrfLma}

\begin{Lma}
\label{enn_syl}
Let three relations $\mathbf{T}_0$, $\mathbf{T}_1$, and $\mathbf{T}_2$ be {\em satisfiable}, which means that there are three spheres $\mathcal{O}_0$, $\mathcal{O}_1$, and $\mathcal{O}_2$ satisfying the relations $\mathbf{T}_0(\mathcal{O}_0, \mathcal{O}_1)$, $\mathbf{T}_1(\mathcal{O}_1, \mathcal{O}_2)$, and $\mathbf{T}_2(\mathcal{O}_2, \mathcal{O}_0)$,  where $\mathbf{T}_0,\mathbf{T}_1,\mathbf{T}_2\in\{\mathbf{D}, \mathbf{P}, \overline{\mathbf{P}}, \neg\mathbf{D}, \neg\mathbf{P}, \neg\overline{\mathbf{P}}\}$. 
Let $i\in\{0,1,2\}$, $j=(i+1) \mod\ 3$, $k=(j+1) \mod\ 3$. $\mathcal{S}$phNN can successfully construct a sphere configuration with a maximum of one restart: first construct $\mathcal{O}_i$ and $\mathcal{O}_j$ satisfying $\mathbf{T}_i(\mathcal{O}_i, \mathcal{O}_j)$; fix $\mathcal{O}_j$ and construct $\mathcal{O}_k$ satisfying $\mathbf{T}_j(\mathcal{O}_j, \mathcal{O}_k)$; fix $\mathcal{O}_i$ and $\mathcal{O}_j$ and update $\mathcal{O}_k$ to satisfy $\mathbf{T}_k(\mathcal{O}_k, \mathcal{O}_i)$ while preserving $\mathbf{T}_j(\mathcal{O}_j, \mathcal{O}_k)$. If the last step fails, $\mathcal{S}$phNN restarts the process by firstly fixing $\mathcal{O}_k$ and realising $\mathbf{T}_k(\mathcal{O}_k, \mathcal{O}_i)$ and $\mathbf{T}_j(\mathcal{O}_j, \mathcal{O}_k)$.  
\end{Lma}

\begin{PrfLma}
\begin{enumerate}
    \item  $\mathbf{T}_i\in\{\mathbf{D}, \mathbf{P},   \overline{\mathbf{P}}\}$.
\begin{enumerate}
    \item   
    at most one of  $\mathbf{T}_j$ and $\mathbf{T}_k$ is a member of $\{\neg\mathbf{D}, \neg\mathbf{P}, \neg\overline{\mathbf{P}}\}$. $\mathbf{T}_0$, $\mathbf{T}_1$, and $\mathbf{T}_2$ are {\em satisfiable}, with Lemma~\ref{enn_thm0}, $\mathcal{S}$phNN will construct an Euler diagram.
    \item $\mathbf{T}_j,\mathbf{T}_k\in\{\neg\mathbf{D}, \neg\mathbf{P}, \neg\overline{\mathbf{P}}\}$.  With Lemma~\ref{lma_2neg}, $\mathcal{S}$phNN will construct an Euler diagram. 
\end{enumerate}

    \item $\mathbf{T}_i\in\{\neg\mathbf{D}, \neg\mathbf{P}, \neg\overline{\mathbf{P}}\}$.
    
$\mathbf{T}_i$ is consistent with three relations in the set $\{\mathbf{D}, \mathbf{PO}, \mathbf{P}, \overline{\mathbf{P}}\}$. Let $\mathsf{consis}(\mathbf{T}_i)$ denote the three consistent relations: $\mathsf{consis}(\neg\mathbf{D})=\{\mathbf{PO}, \mathbf{P}, \overline{\mathbf{P}}\}$, $\mathsf{consis}(\neg\mathbf{P})=\{\mathbf{PO}, \mathbf{D}, \overline{\mathbf{P}}\}$, and $\mathsf{consis}(\neg\overline{\mathbf{P}})=\{\mathbf{PO}, \mathbf{P}, \mathbf{D}\}$.

Let the relation between $\mathcal{O}_i$ and $\mathcal{O}_j$ be $\mathbf{R}_i(\mathcal{O}_i, \mathcal{O}_j)$, $\mathbf{R}_i\in\{\mathbf{D}, \mathbf{PO}, \mathbf{P}, \overline{\mathbf{P}}\}$ and $\mathbf{R}_i$ is consistent with $\mathbf{T}_i$. 
\begin{enumerate}
    \item If $\mathbf{R}_i$, $\mathbf{T}_j$, and $\mathbf{T}_k$ are {\em satisfiable}, the same proof structure as case 1, as Lemma~\ref{enn_thm0} and Lemma~\ref{lma_2neg} hold for $\{\mathbf{D}, \mathbf{P}, \mathbf{PO},   \overline{\mathbf{P}}\}$. \item If $\mathbf{R}_i$, $\mathbf{T}_j$, and $\mathbf{T}_k$ are {\em unsatisfiable},  then the following syllogistic reasoning is valid.
    \syllogism{$\mathbf{T}_j(\mathcal{O}_j, \mathcal{O}_k)$\\  $\mathbf{T}_k(\mathcal{O}_k, \mathcal{O}_i)$}{$\neg\mathbf{R}_i(\mathcal{O}_i, \mathcal{O}_j)$} 
\begin{enumerate}
    \item $\mathbf{T}_j, \mathbf{T}_k\in\{\mathbf{D}, \mathbf{P}, \overline{\mathbf{P}}\}$.
    
Only relations in $\mathsf{consis}(\mathbf{T}_i)$ except $\mathbf{R}_i$ can be consistent with $\mathbf{T}_j$ and $\mathbf{T}_k$. The size of $\mathsf{consis}(\mathbf{T}_i)/\mathbf{R}_i$ is less than or equals to 2. With Lemma~\ref{enn_lm1}, the value cannot be 2, so, exactly one relation $\mathbf{R}^*_i$ is consistent with $\mathbf{T}_i$ and  $\mathbf{T}_j$ and $\mathbf{T}_k$. With Theorem~\ref{lmanoncyc}, let $\mathcal{S}$phNN fix $\mathcal{O}_k$, then optimise $\mathcal{O}_j$ to the relation  $\mathbf{T}_j(\mathcal{O}_j, \mathcal{O}_k)$, and optimise $\mathcal{O}_i$ to the relation $\mathbf{T}_k(\mathcal{O}_k, \mathcal{O}_i)$, then the relation between $\mathcal{O}_i$ and $\mathcal{O}_j$ can only be $\mathbf{R}^*_i$.  

\item  $\mathbf{T}_j, \mathbf{T}_k\in\{\neg\mathbf{D}, \neg\mathbf{P}, \neg\overline{\mathbf{P}}\}$.
There is no valid syllogism with three negative forms (All valid syllogisms are listed in \ref{syl24}).

So, $\mathbf{T}_j, \mathbf{T}_k\in\{\neg\mathbf{D}, \neg\mathbf{P}, \neg\overline{\mathbf{P}}\}$ is not possible. 
  
\item exactly one of $\mathbf{T}_j$ and $ \mathbf{T}_k$ is the member of $\{\neg\mathbf{D}, \neg\mathbf{P}, \neg\overline{\mathbf{P}}\}$.

Without the loss of generality, let $\mathbf{T}_j\in\{\neg\mathbf{D}, \neg\mathbf{P}, \neg\overline{\mathbf{P}}\}$. Therefore, $\mathbf{T}_k\in\{\mathbf{D}, \mathbf{P}, \overline{\mathbf{P}}\}$, and  $\mathbf{T}_i, \mathbf{T}_j\in\{\neg\mathbf{D}, \neg\mathbf{P}, \neg\overline{\mathbf{P}}\}$.
     With Lemma~\ref{lma_2neg}, $\mathcal{S}$phNN will construct an Euler diagram without error. 
     \qed 
\end{enumerate}
\end{enumerate}
\end{enumerate}
\end{PrfLma}

\begin{Lma}
\label{thmeq}
Let $\psi_1(\mathcal{O}_1,\mathcal{O}_2)$\dots $\psi_{N-1}(\mathcal{O}_{N-1},\mathcal{O}_N)$,  $\psi_{N}(\mathcal{O}_{N},\mathcal{O}_{1})$ be $N$ syllogistic statements, where $\psi_i\in\{\mathbf{D}, \mathbf{P}, {\overline{\mathbf{P}}},\neg\mathbf{D}, \neg\mathbf{P}, \neg{\overline{\mathbf{P}}}\}$. If the $N$ statements are satisfiable, and one of $\psi_k$ must be $\mathbf{EQ}$, then, every $\psi_i$ must be $\mathbf{EQ}$.
\end{Lma}
\begin{figure}[!t]
\centering
\includegraphics[width=1\textwidth]{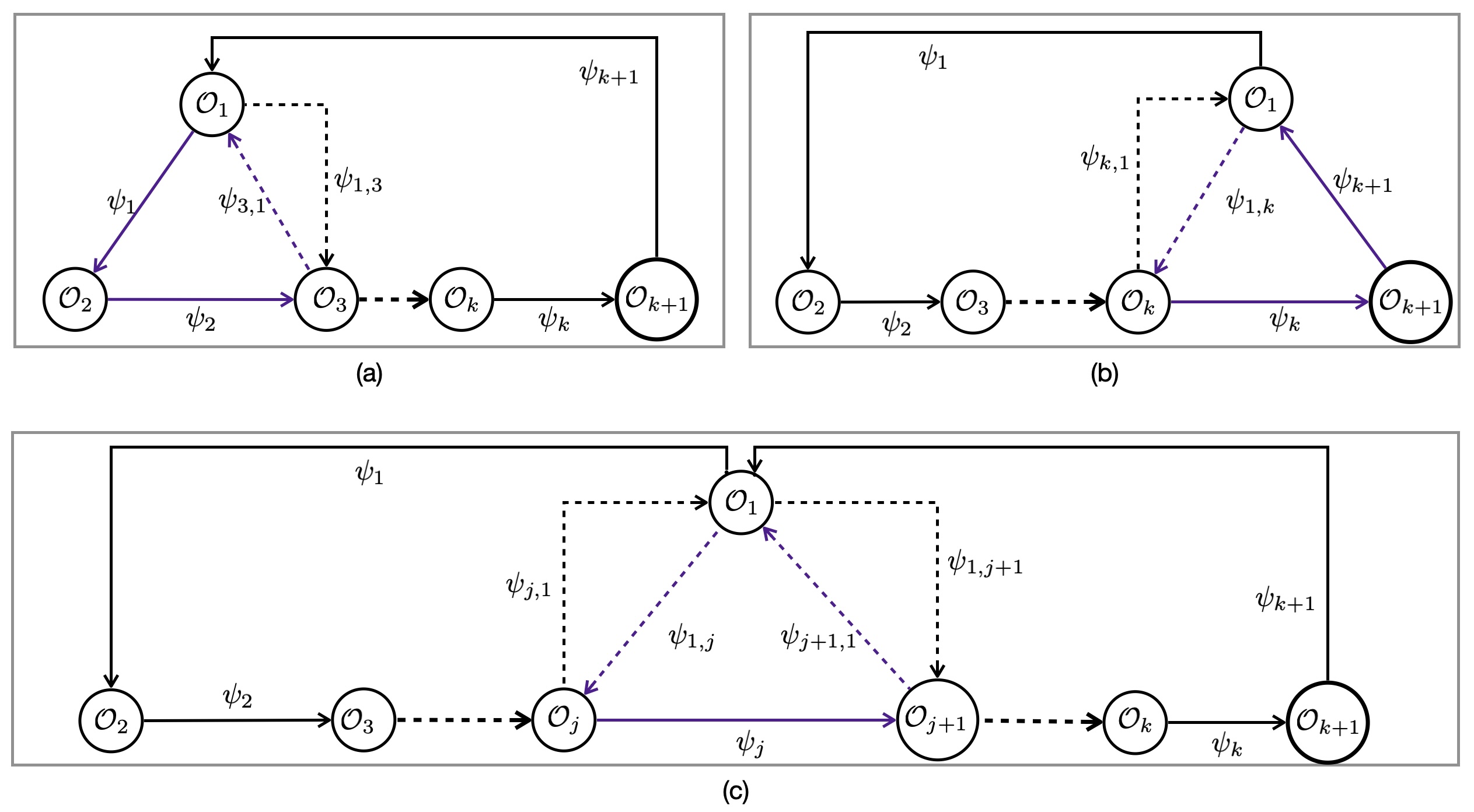}
\caption{Case $N=k+1$: (a) $\psi_1$ or $\psi_2$ is $\mathbf{EQ}$. The $k+1$ spheres are partitioned into two circular chains (1) $\mathcal{O}_1$,  $\mathcal{O}_2$ and  $\mathcal{O}_3$, and (2) $\mathcal{O}_1$, $\mathcal{O}_3$ \dots; (b) $\psi_k$ or $\psi_{k+1}$ is $\mathbf{EQ}$. The $k+1$ spheres are partitioned into two circular chains (1) $\mathcal{O}_1$,  $\mathcal{O}_k$ and  $\mathcal{O}_{k+1}$, and (2) $\mathcal{O}_k$, $\mathcal{O}_1$, $\mathcal{O}_2$ \dots$\mathcal{O}_{k-1}$; (c) $\psi_j  $ is $\mathbf{EQ}$ ($2<j<k$). The $k+1$ spheres are partitioned into three circular chains (1) $\mathcal{O}_1$,  $\mathcal{O}_j$ and  $\mathcal{O}_{j+1}$, (2) $\mathcal{O}_j$, $\mathcal{O}_1$, $\mathcal{O}_2$ \dots$\mathcal{O}_{j-1}$, (3) $\mathcal{O}_1$, $\mathcal{O}_{j+1}$, \dots.}  
\label{eqn_diagram}
\end{figure}

\begin{PrfLma}
We induct on $N$.
\begin{enumerate}
    \item $N=3.$ Without loss of generality, let $\psi_3(\mathcal{O}_3, \mathcal{O}_1)$ must be $\mathbf{EQ}(\mathcal{O}_3, \mathcal{O}_1)$. That is, $\mathbf{P}(\mathcal{O}_3, \mathcal{O}_1)$ and $\mathbf{P}(\mathcal{O}_1, \mathcal{O}_3)$ are valid. From $\mathbf{P}(\mathcal{O}_3, \mathcal{O}_1)$, we infer $\mathbf{P}(\mathcal{O}_3, \mathcal{O}_2)$ and $\mathbf{P}(\mathcal{O}_2, \mathcal{O}_1)$ and from $\mathbf{P}(\mathcal{O}_1, \mathcal{O}_3)$, we infer $\mathbf{P}(\mathcal{O}_1, \mathcal{O}_2)$ and $\mathbf{P}(\mathcal{O}_2, \mathcal{O}_3)$, as for valid conclusion $\mathbf{P}(\mathcal{O}_S, \mathcal{O}_P)$ there is only one kind of premises $\mathbf{P}(\mathcal{O}_S, \mathcal{O}_M)$ and $\mathbf{P}(\mathcal{O}_M, \mathcal{O}_P)$ (All valid syllogisms are listed in \ref{syl24}). 
    \item If the theorem holds, when $N\le k$.
    \item   $N=k+1$. Let $\psi_{j}$ must be $\mathbf{EQ}$. 
    \begin{enumerate}
        \item Case $j=1$ or $j=2$. As $N$ statements are satisfiable, there will be an Euler diagram in the form of a sphere configuration with $N$ spheres. Consider three spheres $\mathcal{O}_{1}$, $\mathcal{O}_{2}$ and  $\mathcal{O}_{3}$. They form an Euler diagram for three syllogistic statements ($\psi_{3,1}\in\mathcal{T}$ is an observed syllogistic relation), and $\psi_{j}$ must be $\mathbf{EQ}$. So, three relations among $\mathcal{O}_{1}$, $\mathcal{O}_{2}$ and  $\mathcal{O}_{3}$ must be $\mathbf{EQ}$. Consider $k$  relations among $k$ spheres: $\mathcal{O}_{1}$, $\mathcal{O}_{3}$, \dots $\mathcal{O}_{k+1}$, where $\psi_{1,3}\in\mathcal{T}$ is an observed syllogistic relation. The relation between $\mathcal{O}_{1}$ and $\mathcal{O}_{3}$ must be $\mathbf{EQ}$. With the inductive assumption, all relations must be $\mathbf{EQ}$. Figure~\ref{eqn_diagram}(a).  
        \item Case $j=k$ or $j=k+1$. Consider three spheres $\mathcal{O}_{k}$, $\mathcal{O}_{k+1}$ and  $\mathcal{O}_{1}$. Three relations among them must be  $\mathbf{EQ}$. The rest proof is similar to the Case of $j=1$ or $j=2$. Figure~\ref{eqn_diagram}(b).
        \item Case $2<j<k$.   Consider three spheres $\mathcal{O}_{1}$, $\mathcal{O}_{j}$ and  $\mathcal{O}_{j+1}$, where $\psi_{1,j}$, $\psi_{j+1,1}\in\mathcal{T}$ are observed syllogistic relations. They form an Euler diagram for three syllogistic statements, and $\psi_{j}$ must be $\mathbf{EQ}$. So, three relations among $\mathcal{O}_{1}$, $\mathcal{O}_{j}$ and  $\mathcal{O}_{j+1}$ must be $\mathbf{EQ}$. Consider $j$ ($< k$) relations among $j$ spheres: $\mathcal{O}_{1}$, \dots $\mathcal{O}_{j}$. The relation between $\mathcal{O}_{1}$ and $\mathcal{O}_{j}$ must be $\mathbf{EQ}$. With the inductive assumption, all $j$ relations must be $\mathbf{EQ}$. Consider $k-j+2$ ($<k$) relations among $\mathcal{O}_{1}$, $\mathcal{O}_{j+1}$\dots $\mathcal{O}_{k+1}$. The relation between $\mathcal{O}_{1}$ and $\mathcal{O}_{j+1}$ must be $\mathbf{EQ}$. With the inductive assumption, all $k-j+2$ relations must be $\mathbf{EQ}$. Figure~\ref{eqn_diagram}(c).  \qed
    \end{enumerate}
\end{enumerate}
\end{PrfLma}
  
\subsection{The theorem of deterministic neural syllogistic reasoning}
\begin{Thm}
\label{thmcyc}
Let $p_1$, $p_2$, $p_3$ be three syllogistic statements, where $p_1$ can be either $r_1(X_1,X_{2})$ or $r_1(X_{2}, X_{1})$, $p_2$ can be either $r_2(X_2,X_{3})$ or $r_2(X_{3}, X_{2})$, and $p_3$ can be either $r_3(X_1,X_{3})$ or $r_3(X_{3}, X_{1})$, $r_1, r_2, r_3\in\{all, some, no, some\_not\}$. $\mathcal{S}$phNN can determine the satisfiability of $p_1$, $p_2$, $p_3$  in the first epoch, with at most one restart. 
\end{Thm}  
\begin{Prf}
We map $X_i$ to $\mathcal{O}_i$ ($i=1,2,3$) and map $p_i$ to $\mathbf{T}_{ij}(\mathcal{O}_i, \mathcal{O}_j)$, where $i,j=1,2,3$, $i\neq j$, $\mathbf{T}_{ij} = \psi(r_i)$ if $r_i(X_i,X_{j})$ or $\mathbf{T}_{ij} = \overline{\psi(r_i)}$ if $r_i(X_{j}, X_i)$,  and $\mathbf{T}_{ij}\in\{\mathbf{D}, \mathbf{P}, \overline{\mathbf{P}}, \neg\mathbf{D},$ $ \neg\mathbf{P}, \neg\overline{\mathbf{P}}\}$. $\mathcal{S}$phNN first initialises three coincided spheres (line 1 in Algorithm~\ref{algo:neuro_geo}); if this configuration is a model that satisfies the three target relations, done (line 2 in Algorithm~\ref{algo:neuro_geo}). If not, $\mathcal{S}$phNN fixes $\mathcal{O}_1$, then updates $\mathcal{O}_2$ and $\mathcal{O}_3$, to satisfy $\mathbf{T}_{12}(\mathcal{O}_1, \mathcal{O}_2)$ and $\mathbf{T}_{31}(\mathcal{O}_3, \mathcal{O}_1)$, respectively (line 3, 4 in the algorithm), then $\mathcal{S}$phNN performs $COP^{\mathbf{T}_{32}}_{\mathbf{T}_{31}}(\mathcal{O}_{3}| \mathcal{O}_{2}, \mathcal{O}_{1})$ (line 5 in the algorithm). If the global loss reaches zero, done; otherwise, $\mathcal{S}$phNN repeats the process by fixing $\mathcal{O}_{3}$ (line 6 - 9, in the algorithm).   $COP^{\mathbf{T}_{13}}_{\mathbf{T}_{12}}(\mathcal{O}_{1}| \mathcal{O}_{3}, \mathcal{O}_{2})$ will reach zero, if the input is satisfiable (Theorem~\ref{mono_cop} and Lemma~\ref{enn_syl}). \qed  
\end{Prf}

\begin{Thm}
\label{thmcyc1} {\em (The principle of deterministic neural reasoning)} Let $p_1$, \dots, $p_{N-1} \therefore q$ be a long-chained syllogistic reasoning with $N-1$ premises, where $p_i$ can be either $r_i(X_i,X_{i+1})$ or $r_i(X_{i+1}, X_{i})$ ($1\le i\le N-1$), $q$ is fixed to $r_n(X_1,X_{N})$, $r_j\in\{all, some, no, some\_not\}$ ($1\le j\le N$). 
$\mathcal{S}$phNN can determine its validity (or satisfiability) in the first epoch, with maximum once restart, with the computational complexity of ${O}(N)$. 
\end{Thm}

\begin{Prf} 

Without loss of generality, $p_1$, \dots, $p_{N-1} \therefore q$ can be spatialised into $N$ spatial statements $\psi_1(\mathcal{O}_1,\mathcal{O}_2)$, \dots, $\psi_{N-1}(\mathcal{O}_{N-1}, \mathcal{O}_{N})$, $\psi_{N}(\mathcal{O}_{N}, \mathcal{O}_{1})$, where $\psi_i = \psi(r_i)$ if $r_i(X_i,X_{j})$ or $\psi_i = \overline{\psi(r_i)}$ if $r_i(X_{j}, X_i)$,  and $\psi_i\in\{\mathbf{D}, \mathbf{P}, \overline{\mathbf{P}}, \neg\mathbf{D},$ $ \neg\mathbf{P}, \neg\overline{\mathbf{P}}\}$. 

Any non-cyclic chain
$r_1(X_1,X_2)$\dots $r_{N-1}(X_{N-1},X_N)$ is satisfiable (Theorem~\ref{lmanoncyc}). Consequently, $\mathcal{S}$phNN can construct a satisfiable model with $N-1$ steps (line 4-5 in Algorithm~\ref{algo:neuro_geoN}).
$\mathcal{S}$phNN determines the satisfiability of relations among  $\mathcal{O}_{N-1}$, $\mathcal{O}_N$, and $\mathcal{O}_1$, if not, $\mathcal{S}$phNN will determine the relation between $\mathcal{O}_1$ and $\mathcal{O}_{N-1}$ (line 11 -17). With Theorem~\ref{thmcyc}, the complexity of this part is $\mathcal{O}(1)$. This way, $\mathcal{S}$phNN reduces the satisfiability with $N$ terms to the case of $N-1$ terms, and starts the backward procedure (line 22 in Algorithm~\ref{algo:neuro_geoN}) whose worst complexity (backwards to the first three spheres $\mathcal{O}_1$, $\mathcal{O}_2$, $\mathcal{O}_3$) is the same as the forward procedure. Therefore, the worst complexity is proportional to $[(N-1) *\mathcal{O}(1)] + [(N-1)*\mathcal{O}(1)]$, which is proportional to ${O}(N)$.      
\qed
\end{Prf}

\begin{Cly}
\label{thmdim}
If $\mathcal{S}$phNN can construct an Euler diagram in 2-dimensional space, it can construct an Euler diagram in $n$-dimensional space ($n>2$), and vice versa.
\end{Cly}
\begin{PrfCly} 
Trivial, as all proofs hold for dimension $n\ge 2$.\qed
\end{PrfCly} 


For the question of how the standard theories of global rationality can be simplified to render them more tractable, Herbert A. Simon proposed to use a simpler decision criterion that he called {\em satisficing} and models with better predictive power. Compared with traditional neural models, $\mathcal{S}$phNN adopts a simple and strict criterion, namely, $\mathcal{S}$phNN shall reach the global loss of zero within one epoch for satisfiable long-chained syllogistic statements. This endows $\mathcal{S}$phNN with the ability to achieve the rigour of syllogistic reasoning. 


%% file: 9arxiv/power_of_rep.tex
\begin{sidewaysfigure}
\centering 
\includegraphics[width=1.\textwidth]{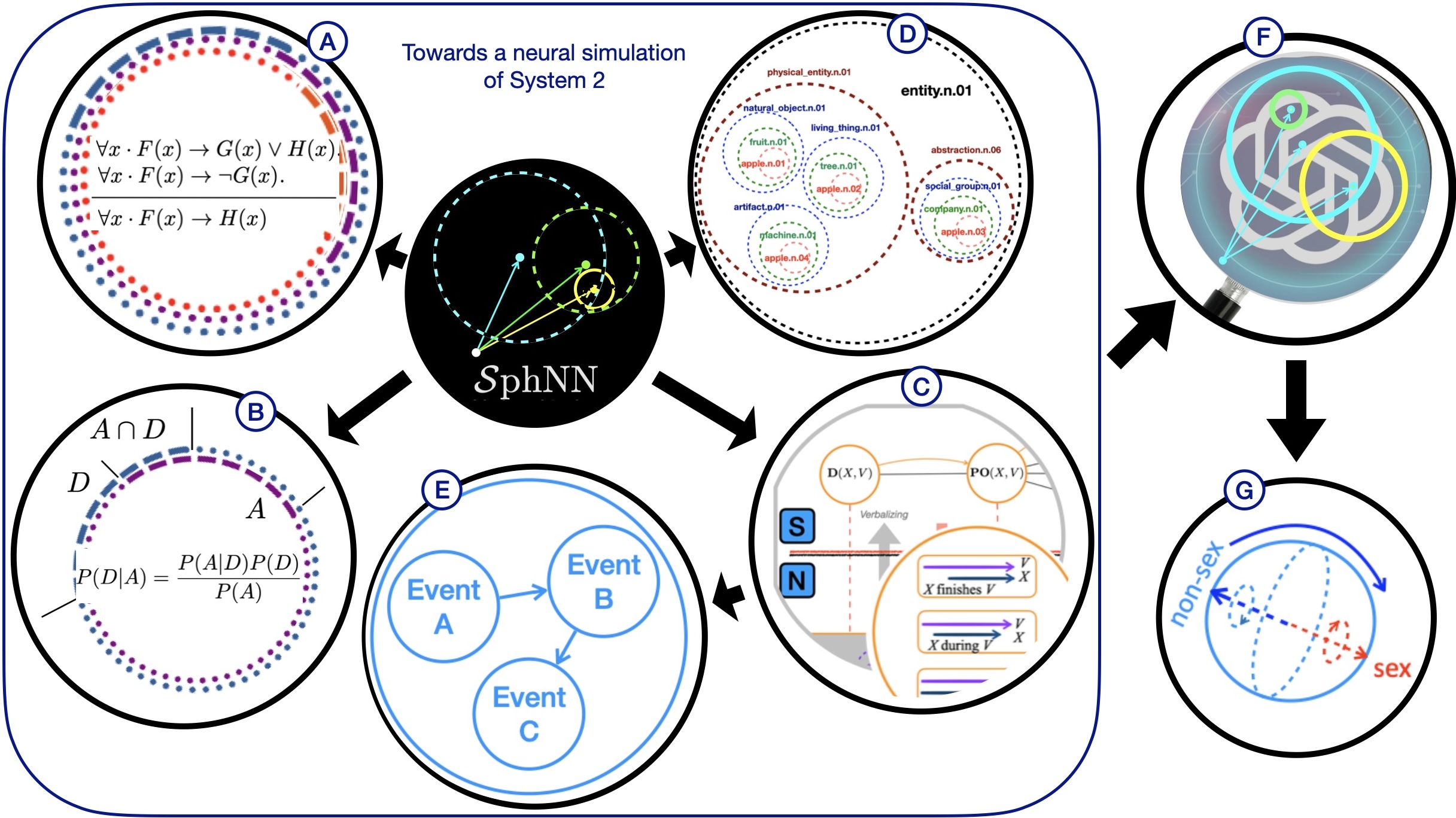}
\caption{Evolution of Sphere Neural-Networks: (A) deterministic neural logic deduction; (B) Bayesian reasoning; (C) neuro-symbolic reasoning for spatio-temporal relations; (D)  neuro-symbolic unification; (E) event reasoning; (F) a neural simulation of System 2, to communicate with, and examine and govern LLMs; (G) Rotating spheres for humour interpretation.}
\label{sphapp}
\end{sidewaysfigure} 

\section{$\mathcal{S}$phNN and human rational reasoning}
\label{sphere_power}

\begin{quote} 
    {\em  
        
The central task of a natural science is to make the wonderful commonplace: to show that complexity, correctly viewed, is only a mask for
simplicity; to find pattern hidden in apparent chaos.

    \hfill --- Herbert A. Simon \cite{Simon19} 
      
}
\end{quote}

Geometrically, a sphere can be understood as a set of points in {\em a universe}, whose distances to {\em a fixed point} (the centre of the sphere) is within {\em a constant} (the radius). Sphere configurations may represent a variety of conceptual structures. If the universe is a line, spheres turn out to be line segments and can represent temporal relations. If the universe is a circle, spheres turn out to be arcs and can represent complement concepts and evolve to spatial semantics for logical reasoning with negation, Bayesian reasoning. Descartes's product of spheres represents heterogeneous knowledge and serves as spatial semantics for neuro-symbolic spatio-temporal reasoning, pure neuro-symbolic unification, and event reasoning. All these lead to a neural model of System 2 that can examine, instruct, and govern hallucinative LLMs. Synergistic collaboration of neural models of System 1 and System 2 can be represented by rotating spheres, which shapes a way for humour understanding (the highest-level cognition), as illustrated in Figure~\ref{sphapp}.

\subsection{Logical reasoning with negation and disjunctions} 

\begin{figure}[t]
\centering
\includegraphics[width=0.9\textwidth]{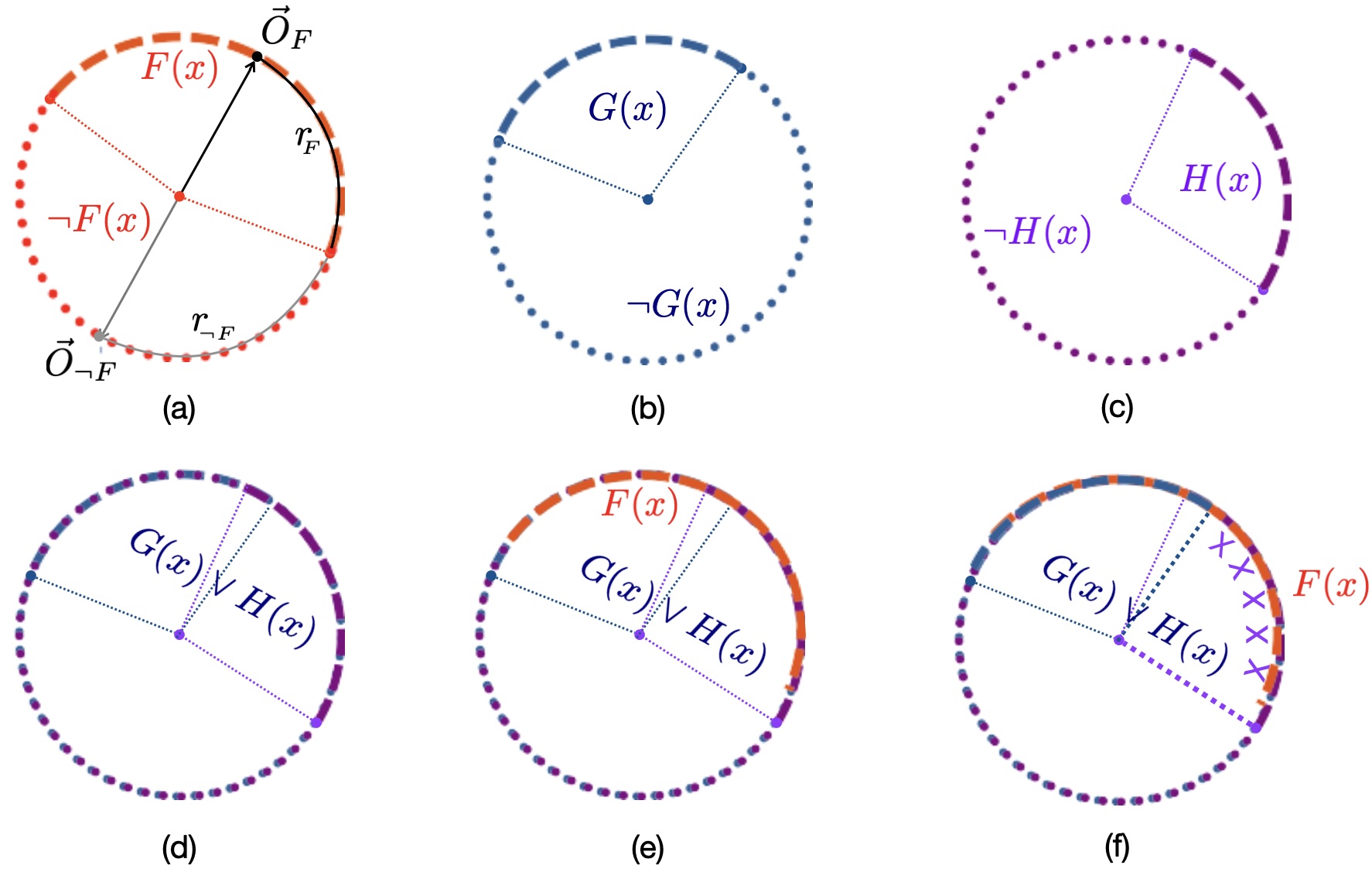}
\caption{Arc embedding can be used to decide the validity of logical reasoning with negation and disjunctions. (a)-(c) illustrates the arc embedding of $F(x)$, $G(x)$, and $H(x)$; (d) illustrates the arc embedding of $G(x)\lor H(x)$; (e) $\forall x\cdot F(x)\rightarrow G(x)\lor H(x)$: the $F(x)$ arc is part of the arc of $G(x)\lor H(x)$; (f)  $\forall x\cdot F(x)\rightarrow \neg G(x)$: the arc of $F(x)$ disconnects from the arc of $G(x)$; As the $F(x)$ arc  is part of the arc of $G(x)\lor H(x)$, it will be part of the $H(x)$ arc. Thus, $\forall x\cdot F(x)\rightarrow H(x)$.}
\label{fig:arc_sphere}
\end{figure}

If the universe is a circle, a sphere becomes an arc. Geometrically, we can represent a set $F$ as 
an arc, with the centre $\vec{O}_F$ and the radius $r\!_{\!_F}$, the complement set $\neg F$ as the arc with the centre $\vec{O}_{\neg F}$ and the radius $r\!_{\!_{\neg F}}$ ($\vec{O}_{F}$ and $\vec{O}_{\neg F}$ have the same length and point at the opposite direction, the sum of $r\!_{\!_{F}}$  and $r\!_{\!_{\neg F}}$ is half of the perimeter), as shown in Figure~\ref{fig:arc_sphere}(a-c). 

The arc embedding (Descartes's product of several 2-dimensional arcs) can represent logical reasoning with negations and disjunctions. We illustrate this by using a first-order logic deduction with disjunction and negation, as follows. 
\syllogism{$\forall x\cdot F(x)\rightarrow G(x)\lor H(x)$.\\ $\forall x\cdot F(x)\rightarrow \neg G(x)$.}{$\forall x\cdot F(x)\rightarrow H(x)$.}
Geometrically, the first premise describes the relation that each component of Descartes's product of $F$ arcs is part of the union of corresponding components of $G$ arcs and $H$ arcs, read as {\em for any $x$, if $x$ is a member of $F$, $x$ is either a member of $G$ or a member of $H$}, as shown in Figure~\ref{fig:arc_sphere}(d-e); the second premise describes the relation that each component of Descartes's product of $F$ arcs is part of the corresponding component of $\neg G$ arcs, read as {\em for any $x$, if $x$ is a member of $F$, $x$ is not a member of $G$}, as shown in Figure~\ref{fig:arc_sphere}(f). The conclusion describes the relation that each component of Descartes's product of $F$ arcs is part of the corresponding component of $H$ arcs, read as {\em for any $x$, if $x$ is a member of $F$, $x$ is a member of $H$}. In this way, $\mathcal{S}$phNN realises logical reasoning with negation and disjunction as motions (rotation and resizing) of arcs targeting an arc configuration. Arc embedding and the methodology of reasoning through model construction may simulate ``reasoning as the motion of mind'' \cite{tversky19} and pave a new way of cognitive modelling. For example, it can be used to simulate how clever monkeys perform disjunctive syllogistic reasoning \cite{Ferrigno21}.

\begin{figure}
\centering  
\includegraphics[width=0.6\textwidth]{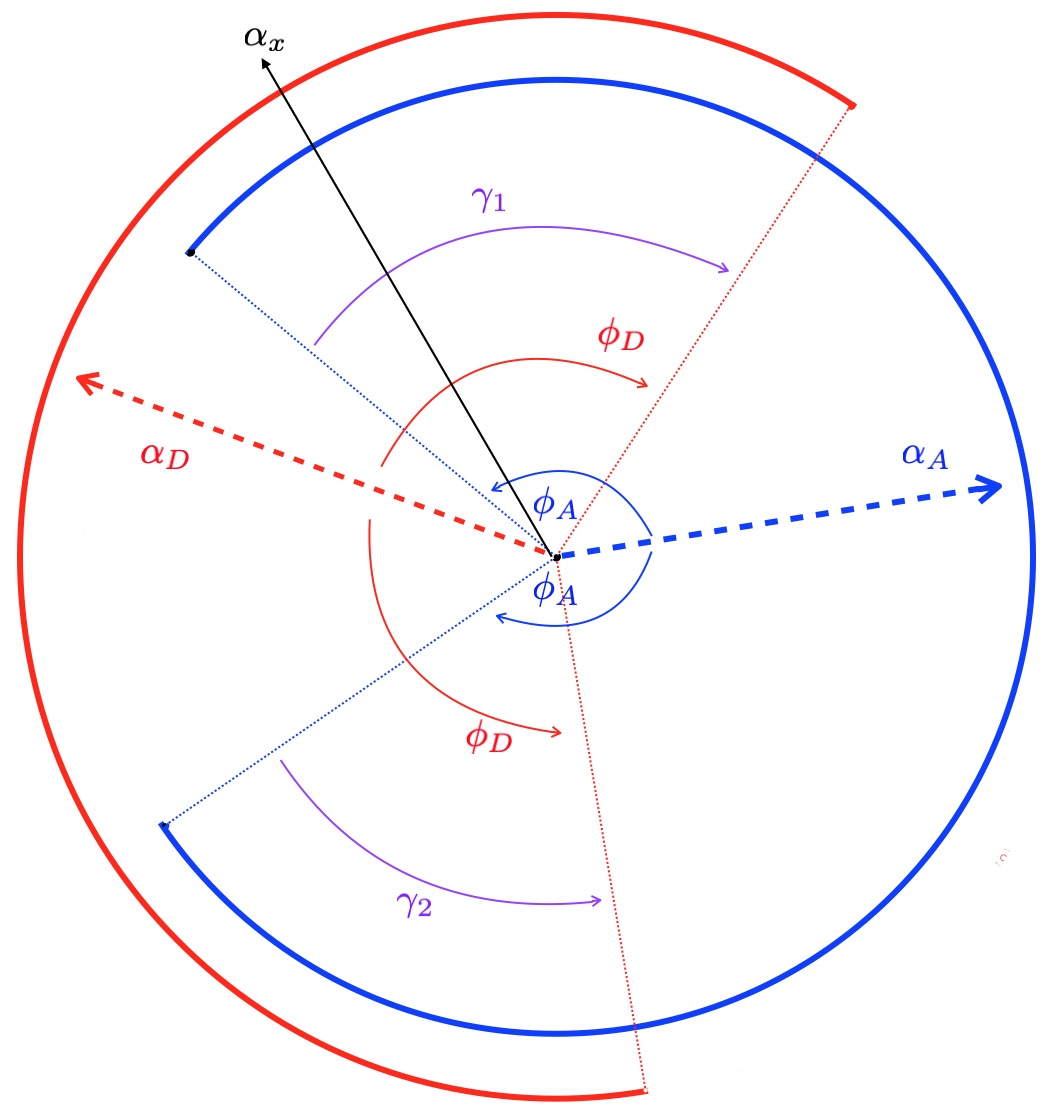}  
\caption{The blue arc represents Event $A$; the red arc represents Event $D$; the probability that ``Event $D$ occurs if Event $A$ occurs'' is the length of the intersection of both arcs divided by the length of arc $A$. Both red and blue arcs shall be part of the same circle.}
\label{monkey}
\end{figure}

\begin{figure}
 \centering 
\begin{subfigure}{0.5\columnwidth}\centering 
\includegraphics[width=0.9\textwidth]{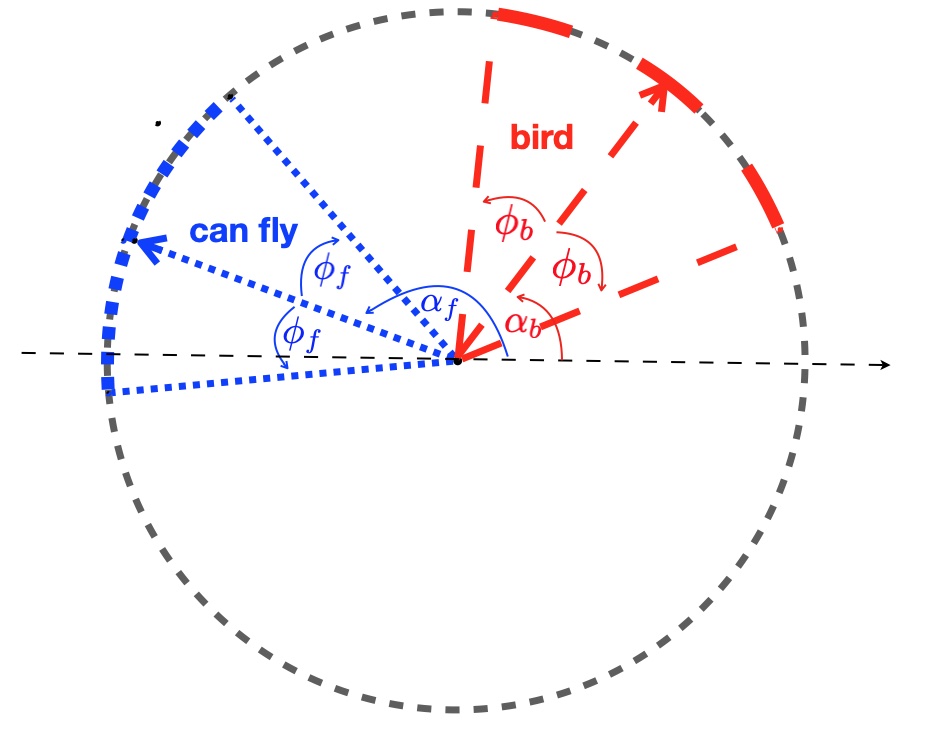}
  \subcaption{}
\end{subfigure} 
\begin{subfigure}{0.45\columnwidth}
\centering
\includegraphics[width=1\textwidth]{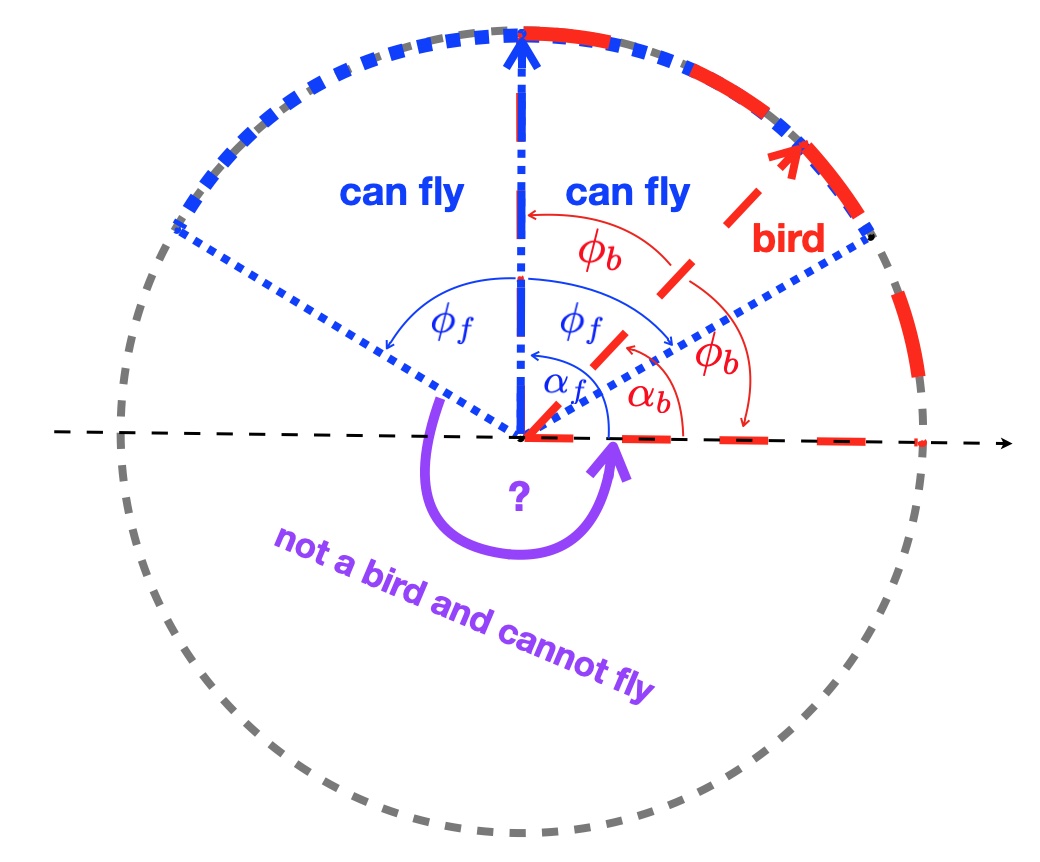}
\subcaption{}  
\end{subfigure} 
\caption{(a) Initializing two arcs, one representing the set of {\em birds} and the other representing the set of {\em animals that can fly}; (b) A final configuration of the two arcs, after $\mathcal{S}$phNN updates the locations and the sizes of the arcs. 
}
\label{fig:birds}
\end{figure}

\subsection{Bayesian reasoning and probability judgment}

Learning of new concepts may fall into the paradigm of Bayesian induction \cite{Tenenbaum15}.  Like syllogistic reasoning having the Euler diagram as its spatial semantics, Bayesian rules can have a configuration of arcs as its spatial semantics as follows. Let arc $A$ with the centre $\alpha_A$ and the offset angle $\phi_A$ represent event $A$. The intersection of arc $A$ and arc $D$  represents the part that event $A$ co-occurs with event $D$, which consists of two parts: (1) $\gamma_1=\max\{0, \phi_{A} + \phi_{D} - \arccos{\cos(\alpha_1, \alpha_2)}\}$, and (2) $\gamma_2=\max\{0, \phi_{A} + \phi_{D} - (2\pi -\arccos{\cos(\alpha_1, \alpha_2))}\}$ (see Figure~\ref{monkey}). Let $\alpha_x$ be a ray starting from the centre $\vec{O}$. The chance of $\alpha_x$ to hit both arcs is $\frac{\gamma_1+\gamma_2}{2\pi}$, written as $P(A\cap D)$. We can decompose the hit into two steps: $\alpha_x$ hits arc $A$ with the chance $\frac{2\phi_A}{2\pi}$, written as $P(A)$, then under this condition, hit arc $D$ with the chance $\frac{\gamma_1 + \gamma_2}{2\phi_A}$, written as $P(D|A)$. We have $P(A\cap D)=\frac{\gamma_1+\gamma_2}{2\pi}=\frac{2\phi_A}{2\pi}\frac{\gamma_1 + \gamma_2}{2\phi_A}=P(A)P(D|A)$. In the same way, we have $P(D)P(A|D)=\frac{\gamma_1+\gamma_2}{2\pi}$. Put together, we have the Bayesian rule: $$P(D|A)=\frac{P(A|D)P(D)}{P(A)}.$$ This introduces a novel neural approach to solving statistical problems. We illustrate this by using the flying bird problem in \cite{Jamnik21}: {\em One-quarter of all animals are birds. Two-thirds of all birds can fly. Half of all flying animals are birds. Birds have feathers. If X is an animal, what is the probability that it's not a bird, and it cannot fly?}
We represent the set of {\em birds} and the set of {\em animals that can fly} as two arcs: arc {\em bird} with the centre $\alpha_b$ and the offset angle $ \phi_b$ and arc {\em flying} with the centre $\alpha_f$ and the offset angle $\phi_f$, as shown in Figure~\ref{fig:birds}(a). Animals except birds will be the arc with the centre $\pi+\alpha_b$ and the offset angle $ \pi- \phi_b$. Animals that cannot fly will be the arc with the centre $\pi+\alpha_f$ and the offset angle $ \pi- \phi_f$.  We have three relations as follows: (1)~one-quarter of all animals are birds, that is,  $2\phi_b = \frac{\pi}{2}$; (2)~two-thirds of all birds can fly, that is, $\frac{2}{3}2\phi_b = \gamma$, where $\gamma$ is the sum of the angle(s) of the intersected sectors, and $\gamma=\max\{0, \phi_b + \phi_f - \arccos\cos(\alpha_b - \alpha_f)\}$ $+ \max\{0, \phi_b + \phi_f - (2\pi - \arccos\cos(\alpha_b - \alpha_f))\} $; and (3)~half of all flying animals are birds, that is, $\frac{1}{2}2\phi_f=\gamma$. The arc for the set of {\em an animal that is not a bird and cannot fly} will be $2\pi-2\phi_b -\phi_f=2\pi - \frac{\pi}{2}-\frac{4}{3}\frac{\pi}{4}=\frac{7\pi}{6}$, shown in Figure~\ref{fig:birds}(b).

\begin{sidewaysfigure} 
\centering
\includegraphics[width=1\textwidth]{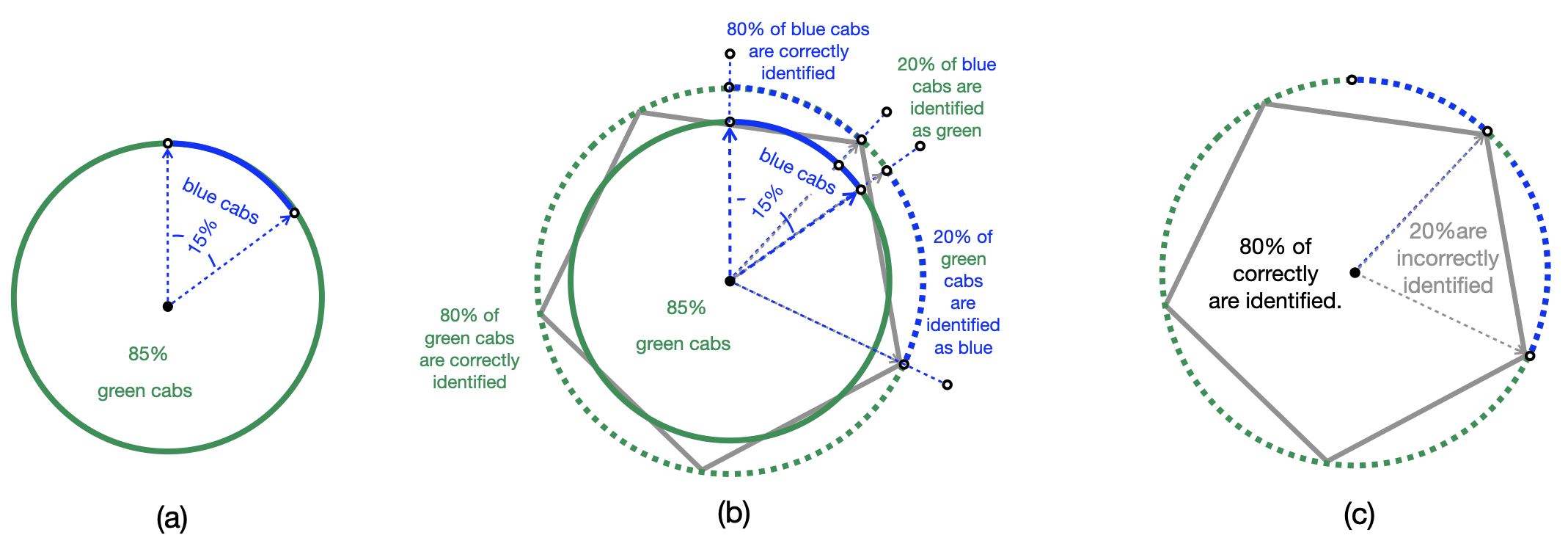}
\caption{(a) 15\% of cabs are Blue; (b) the witness made correct identifications in 80\% of the cases and erred in 20\% of the cases. A witness identified the cab as blue, and the probability that the cab was Blue is $\frac{80\%\times 15\%}{80\%\times 15\%+20\%\times 85\%}=\frac{12}{29}$; (c) Tversky and Kahneman's experiments show that most people ignored the base-rate, and gave the probability of $80\%$.}
\label{fig:cab}
\centering
\includegraphics[width=1\textwidth]{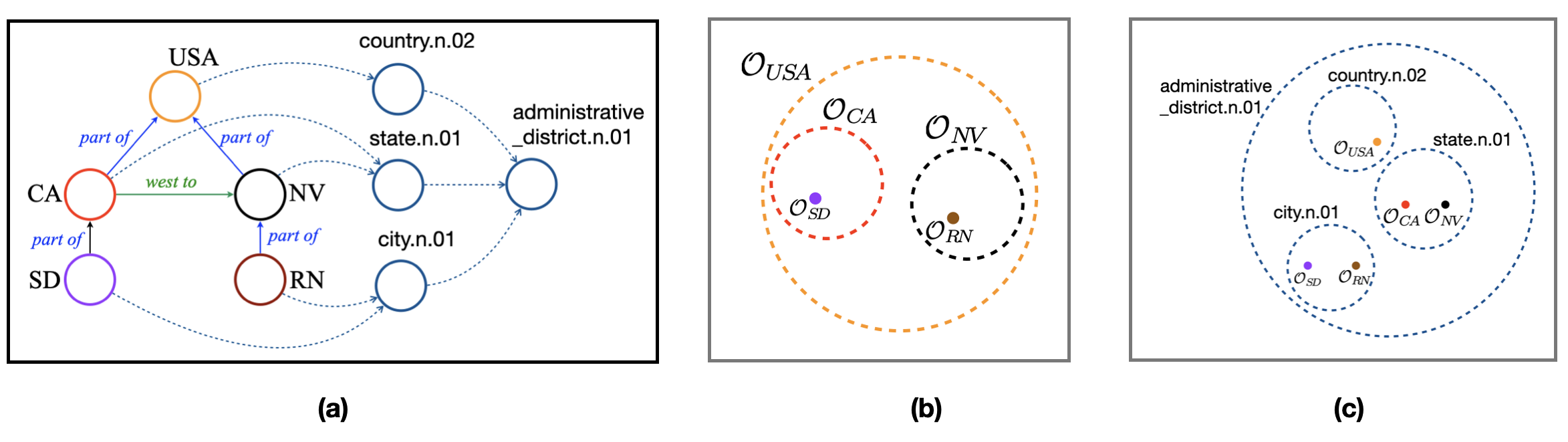}
\caption{(a) A knowledge graph with heterogeneous information, the solid arrow represents a geospatial relation, the dotted arrow represents a category relation; (b) The sphere configuration for the hierarchical spatial locations; (c) The sphere configuration for the category structure.}
\label{fig:sf}
\end{sidewaysfigure} 
 
We revisit Tversky and Kahneman’s Taxi-cab problem \cite{amos_kahnmann77} to show how $\mathcal{S}$phNN may explain and simulate heuristic reasoning.
\begin{quote}{\em
A cab was involved in a hit-and-run accident at night. Two cab companies, the Green and the Blue, operate in the city. You are given the following data:
\begin{enumerate}
    \item 85\% of the cabs in the city are Green and 15\% are Blue.
    \item A witness identified the cab as a Blue cab. The court tested his ability to identify cabs under the appropriate visibility conditions. When presented with a sample of cabs (half of which were Blue and half of which were Green) the witness made correct identifications in 80\% of the cases and erred in 20\% of the cases.
\end{enumerate}
Question: What is the probability that the cab involved in the accident was Blue rather than Green?
}
\end{quote}

We represent blue cabs and green cabs as two arcs in a circle, as shown in Figure~\ref{fig:cab}(a). Probabilities of human judgements are represented by arcs of a concentric circle, as shown in Figure~\ref{fig:cab}(b). That the witness made correct identifications in 80\% of the cases and erred in 20\% of the cases covers two cases - the cab is Blue, and the cab is Green. So the probability that the cab involved in the accident was Blue shall count in the case when the involved cab is Green and mistakenly identified as Blue. Tversky and Kahneman's experiments reported that people ignored this and gave the probability $80\%$, as shown in Figure~\ref{fig:cab}(c). In this way, a judgment process can be simulated as a process of creating and rotating arcs to reach a configuration and computing relations among them.


\subsection{Descartes's product of spheres to embed heterogeneous knowledge} 
Descartes's product of spheres may represent heterogeneous knowledge. In the knowledge graph in Figure~\ref{fig:sf}(a), there are two different types: (1) geospatial relations, e.g., San Diego is in California; (2) category relations, e.g., Nevada is a state. Geographically, Reno is inside Nevada, and San Diego is inside California. Categorically, so, Reno and San Diego are inside the city sphere; Nevada and  California are inside the state sphere. Both the city sphere and the state sphere are inside the administrative sphere. Let each entity $e$ can be represented by a Descartes product of two spheres $(\mathcal{O}_e^{(1)}, \mathcal{O}_e^{(2)})$.  Spheres in the first position represent geospatial relations; for example, $\mathcal{O}_{C\!A}^{(1)}$ is inside $\mathcal{O}_{U\!S\!A}^{(1)}$, as California is part of the USA; spheres in the second position represent category relations, for example, $\mathcal{O}_{C\!A}^{(2)}$ is inside $\mathcal{O}_{\mathrm{state.n.01}}^{(2)}$, as illustrated in 
 Figure~\ref{fig:sf}(b, c).

\begin{figure} 
 \centering 
\includegraphics[width=1\textwidth]{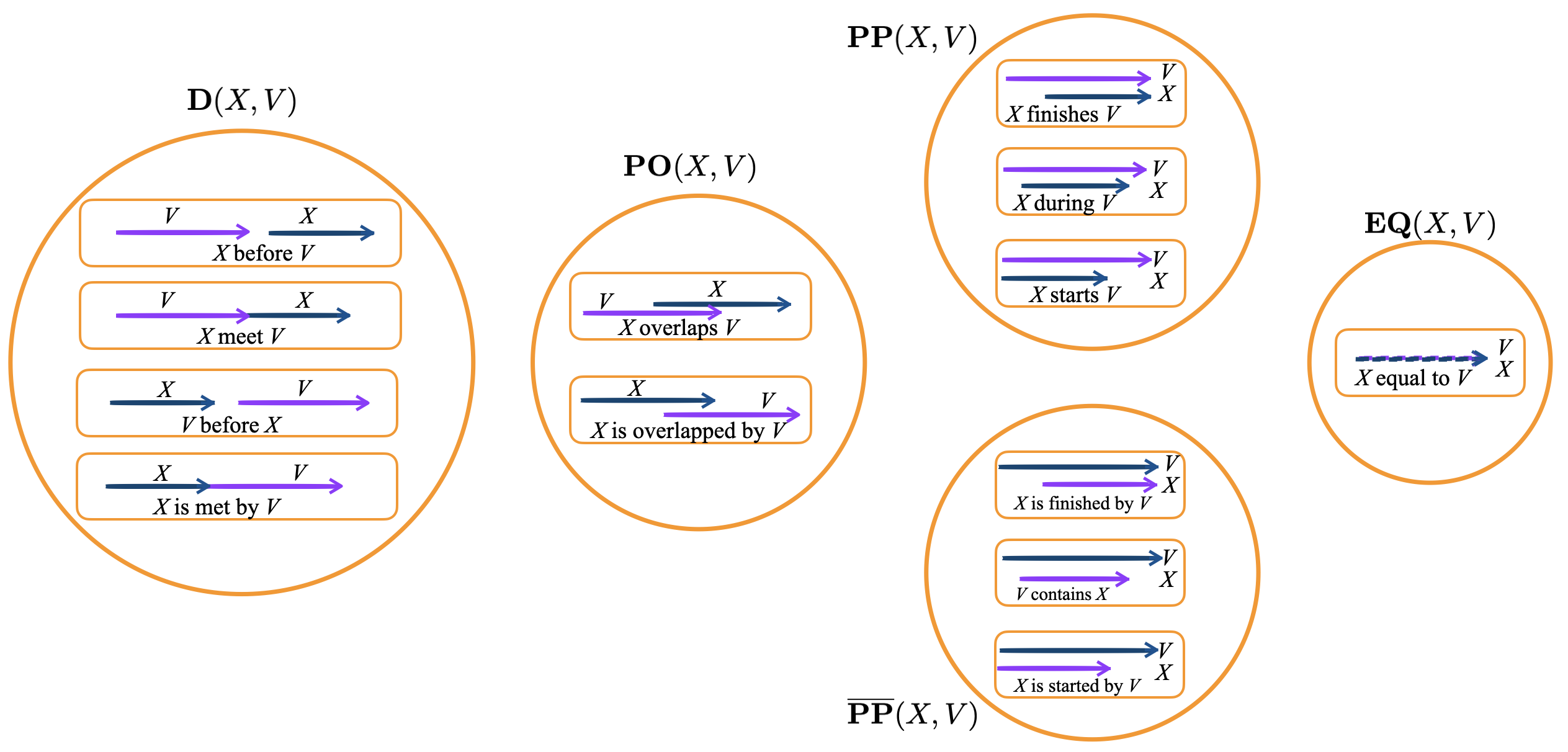}
\caption{The five qualitative spatial relations turn into 13 temporal interval relations in \cite{Allen83}.
}
\label{fig:tmap0}
\end{figure}

\subsection{Neuro-symbolic temporal reasoning}

\begin{quote} 
{\em  Automatic driving will come, but not in the way we have been led to believe.
    
    \hfill --- Gerd Gigerenzer\cite{Gigerenzer22}
     } 
\end{quote}

If we project two closed spheres into the temporal line, they will become temporal intervals. The neighbourhood transition map in Section~\ref{transition_map} will become a transition map for temporal relations with two additional properties: (1) the temporal arrow will introduce the order between two intervals; (2) temporal intervals introduce new relations by the coincide relation of their endpoints. With the two properties, the five qualitative spatial relations between spheres in Figure~\ref{fig:hsgnn}(a) turn out to be 13 relations between temporal intervals \cite{Allen83}, as illustrated in 
Figure~\ref{fig:tmap0}
. This way, $\mathcal{S}$phNN can reason with spatio-temporal relations by using a transition map structured as a Descartes product of a spatial transition map and a temporal transition map, as illustrated in Figure~\ref{sphapp}(B). This spatio-temporal reasoning is neuro-symbolic and will bring advantages of neural computing into the symbolic qualitative spatio-temporal reasoning\footnote{The literature of symbolic qualitative spatio-temporal has some limitations: {\em there has been active work in this area [qualitative spatial reasoning] for more than 20 years, and more than 1,000 research papers have been published, but very little connects to any common-sense reasoning problem that might ever arise} \cite{GaryACM15}, in part because symbolic rules have not completely governed the connection relation -- The two axioms in the main-stream literature of qualitative spatial reasoning also allows distance comparison relation \cite{DongJPL}. $\mathcal{S}$phNN suggests that it would be easier to solve the problems in the vector space.}
and solve tasks in real applications that must be addressed from both perspectives \cite{sun2023}, for example, complex traffic scenarios of self-driving cars \footnote{\href{https://www.economist.com/science-and-technology/is-it-smarter-than-a-seven-month-old/21804141 }{Is self-driving car smarter than a seven-month-old?} The Economist, September 4, 2021.} \cite{Mehul2021}.

\subsection{Event reasoning} 

Events are four-dimensional entities in three-dimensional physical space and one-dimensional temporal space and are closely tied with causalities and goals \cite{ZacksTversky01}.
Spatio-temporal reasoning is the foundation of the reasoning of events. Temporally, an event can be partitioned into three parts: a start, a middle, and an end. 
Spatially, it can be partitioned into objects and spatial relations among objects. Driven by goals or forced by certain causality, agents within an event may perform actions that transform the event's start to the end through the middle. As spatiotemporal relations among events are the backbone of an event structure and closely related to causal relations \cite{tversky19}, $\mathcal{S}$phNN can be developed to reason relations among components of  events, 
as illustrated in Figure~\ref{sphapp}(E), and further develop causal reasoning, e.g.,  \cite{Scholkopfetal21}.  

\subsection{Towards a neural model of System 2}

\begin{quote} 
    {\em The simplest scheme of evolution depends on two processes: a generator and a test. The generator produces variety, new forms that have not existed previously, whereas the test culls out the generated forms so that only those
that are well fitted to the environment will survive. 
    
    \hfill --- Herbert A. Simon \cite{Simon19} 
        } 
\end{quote}

LLMs hallucinate and can neither stop nor notice it by itself. Their rationality can be improved by dividing a reasoning task into several subtasks, such as Chain-of-Thought \cite{creswell2022selectioninference,wei2023COT} (Multiple CoT \cite{LLM2023Cot2}), Tree-of-Thought, and Graph-of-Thought \cite{LLM2023graph}. Though this divide-and-conquer strategy improves performances, each sub-thought and atomic reasoning are still carried out within the black box paradigm of traditional deep learning \cite{creswell2022selectioninference,uesato2023solving,lightman2023lets}, which lack determinacy and responsibility. For example, each sub-thought in the chain of reasoning utilises supervised reward model processes \cite{uesato2023solving,lightman2023lets}. 
Challenging research in AI is to move from simulating associative thinking (System 1) to simulating higher-level cognition (rational reasoning of System 2) \cite{Bengio22}. Theorem~\ref{thmcyc1} guarantees $\mathcal{S}$phNN to be a neural simulation of System 2 for deterministic syllogistic reasoning and can examine the reasoning results of LLMs in two different ways: (1)~if LLMs can prompt the steps of its reasoning process, $\mathcal{S}$phNN checks the existence of the final sphere configuration by following the reasoning steps; (2)~LLMs provide internal vector embeddings of the reasoning results. $\mathcal{S}$phNN checks the existence of the final sphere configuration by using these vectors as the orientations of sphere centres. The non-existence of the final configuration refutes the reasoning result of LLMs. The interaction between $\mathcal{S}$phNN and ChatGPT mirrors a micro-world of the neural dual-process model of the mind.  
When $\mathcal{S}$phNN evolves to various kinds of rational reasoning, it will serve as the deterministic neural reasoners that examine and instruct the outputs of LLMs, as illustrated in Figure~\ref{sphapp}(F). 

\subsection{Neuro-symbolic unification, supporting both heuristic and deliberative reasoning}

\begin{quote}
    {\em 
Our ambition, anyhow, is to offer something clearly better. 
More relevant to us than the varieties of dual-process theories is the way the whole approach has shaken and in some sense shattered the psychology of reasoning. 

\hfill--- Mercier and Sperber \cite{enigma2018}

        }
\end{quote}

In the main literature of neural-symbolic AI \cite{NeurosymbolicAI3}, the neural module and the symbolic module are two separate modules with incompatible semantics. They are loosely bridged in a probabilitistic way that the symbolic module provides semantic loss functions to optimise the neural module \cite{asp15,Xu2018ASL,neurASP20,Cunnington23}. In contrast, $\mathcal{S}$phNN 
demonstrates the possibility of creating continuous set-theoretic semantics that explicitly cohabit with the latent vector semantics of the neural module. Thus, $\mathcal{S}$phNN can genrate to pure neuro-symbolic unification \cite{NeurosymbolicAI3}, a new {\em artifact} \cite{Simon19}, described as follows: Firstly, it initialises the orientations of sphere centres by using latent feature vectors provided by LLMs. Then, it tries to optimise the lengths of the centre vectors and radii of spheres to reach the target sphere configuration with minimal rotations of sphere centres. If no satisfiable configuration is found, it refutes the output of LLMs; if there is a satisfiable configuration, it confirms the outputs of LLMs, meanwhile directly updating the vector outputs of LLMs and, consequently, forcing LLMs to gear their parameters. This way, sphere semantics can be viewed as being promoted and specified from the latent vector semantics. The existence of such pure neuro-symbolic spheres has been created and explored by utilising geometric construction \cite{dong19iclr,DongSpringer21}. They loyally inherit the preciseness of symbolic structure and heuristics of vector embedding, so that symbolic structures can govern out-of-distribution data. This frees neural computing from the stable world assumption and can solve hard AI problems with performances beyond the glass ceiling of traditional deep learning neural networks~\cite{dong19,dong2023word}, 
as illustrated in Figure~\ref{sphapp}(D). 


\begin{sidewaysfigure}
\centering 
\includegraphics[width=1.\textwidth]{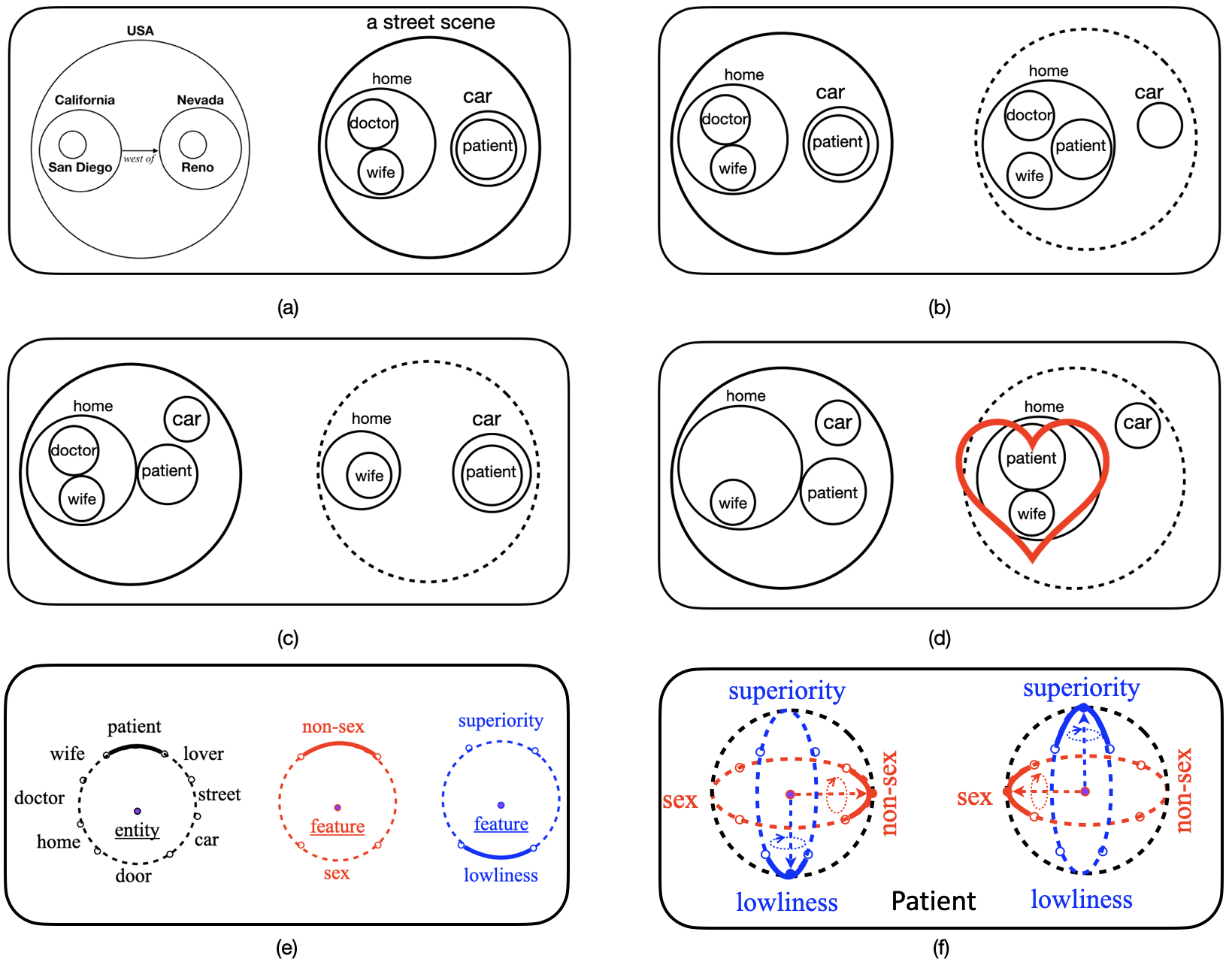}
\caption{(a) The Euler diagram for the geographic relation between San Diego and Reno can be transformed for street scenes; (b) the {\em start} and the {\em end} parts of a patient-visiting-doctor event; (c) the {\em start} and the {\em end} parts of a patient-doctor's wife dialogue event: {\em ``Is the doctor at home?''  ``No,''}; (d) doctor's wife replies, ``come right in''. This signals the lover-meet event between doctor's wife and the patient; (e) features and roles of the patient can be embedded as a Descartes product of arcs in a 2-dimensional circle; (f) two 2-dimensional circles (one blue, one red) can be integrated into a $3$-dimensional sphere; a rotating axis of this sphere will represent the centre vector of an arc. The switching of a rotating axis will simulate feature opposition.}
\label{sphere4humor}
\end{sidewaysfigure}

\subsection{Towards humour understanding, the highest level of cognition}

\begin{quote}
    {\em 
Automatic inference in perception and deliberate inference in reasoning are at the two ends of a continuum. Between them, there is a great variety of inferential processes doing all kinds of jobs. 

\hfill--- Mercier and Sperber  \cite{enigma2018}

        }
\end{quote}

Humour is usually regarded as the highest level of cognition that interweaves both heuristic and rational reasoning -- if an AI system can simulate humour, it can simulate any cognitive activity \cite{joke2020}. 
The understanding of humour can be described by the Script-Based Semantic Theory Of Humour (SSTH) \cite{raskin85} and its improved versions \cite{attado97,attardo2002script,Attardo17,Julia20}, as follows: A humour encodes two scripts; the first schematizes a normal event appearing with high frequency in everyday life, and then a punch triggers the second event. The new event schematizes an abnormal story (out-of-distribution) in which an object in the first event has an opposite feature that surprises the reader. A computation model for humour understanding shall be capable of acquiring both normal and abnormal events from the same text. Though this may frustrate both classic AI and traditional neural networks, we propose that the two inconsistent scripts can be synergistically unified by rotating spheres \cite{Dong_dag21Humor,dong_et_al22}.
 We show that humour understanding also has the root in spatial reasoning by using the classic joke of the SSTH theory \cite{raskin85} as follows. 
\begin{itemize}
 \item[-] {\em ``Is the doctor at home?''} the patient asked in his bronchial whisper. 
\item[-] {\em ``No''}, the doctor’s young and pretty wife whispered in reply, ``Come right in.''
\end{itemize}
    
We transform the Euler diagram of San Diego and Reno scenario into a diagram of a street scene: a doctor and his wife are at home, and a patient is in a car, as shown in Figure~\ref{sphere4humor}(a). This is the {\em start} of a patient-visit-doctor event. The expected {\em end} of the event is the patient at home with the doctor, as shown in Figure~\ref{sphere4humor}(b). However, the wife tells the patient, ``The doctor is not at home.''  The expected {\em end} part of this event is that the patient gets in his car and leaves, as shown in Figure~\ref{sphere4humor}(c). The pretty wife continues, ``Come right in.'' This triggers the {\em start} of an affair event between the wife and the patient, as shown in Figure~\ref{sphere4humor}(d), in which the target of the patient switches from ``no sex'' to ``sex'', the doctor loses the superiority and becomes the loser to the patient. 
We represent the features of an object as Descartes's product of 2-dimensional arcs, as shown in Figure~\ref{sphere4humor}(e), and assemble them into an $n$-dimensional sphere. The centre vector of an arc can be represented by a rotating axis of the $n$-dimensional sphere, and the switching of features will be physically simulated by the rotation of an axis, as shown in Figure~\ref{sphere4humor}(f). 
Let $\mathbb{H}$ be the signature of a humour script, then $\mathbb{H}$ will be structured as a pair of event scripts $(\mathbb{S},\mathbb{S})$. An event script has temporal parts (the start, mid, and end of an event); each can be represented as a configuration sphere $\mathbb{E}$. Thus, $\mathbb{H}$ has a form of  
$((\mathbb{E}, \mathbb{E}, \mathbb{E}),(\mathbb{E}, \mathbb{E}, \mathbb{E}))$.

%% file: 9arxiv/conclusion.tex
\section{Conclusions}

\begin{quote} 
    {\em The intuitive mind is a sacred gift and the rational mind is a faithful servant. 
    
    \hfill ---Albert Einstein  
        } 
\end{quote}
 

\begin{sidewaysfigure}
\centering
\includegraphics[width=1.\textwidth]{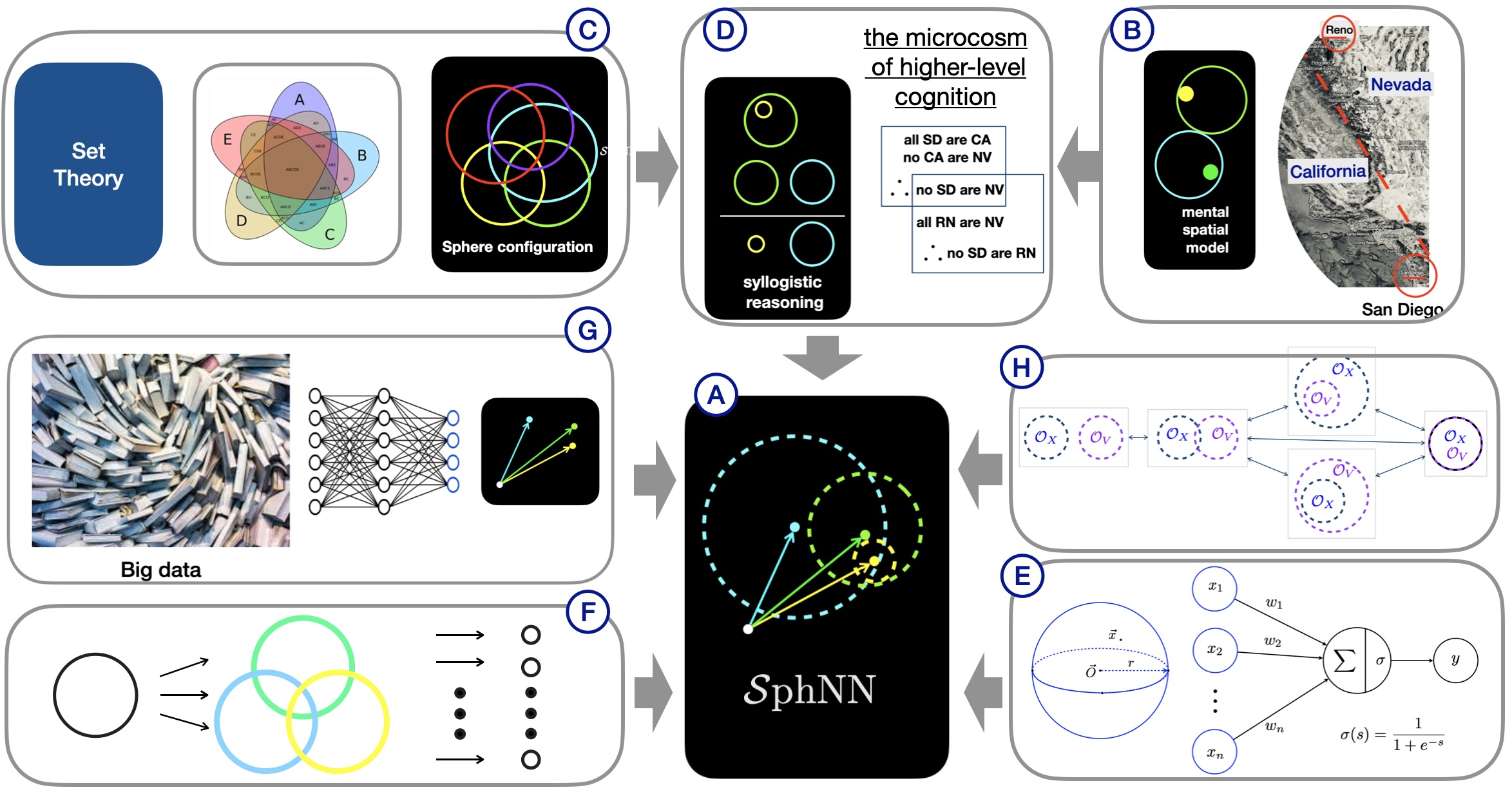}
\caption{(A) $\mathcal{S}$phNN is a deterministic neural network for long-chain syllogistic deduction by constructing Euler diagram in the vector space; (B)  syllogistic reasoning can have its root in spatial reasoning \cite{tversky19}; (C) Euler diagram has its root in set theory \cite{hammer98}; (D) Syllogistic reasoning is the microcosm of human rationality;  
(G) Orientations of sphere centres can be learned from traditional neural networks; (E) one ancestor of $\mathcal{S}$phNN is the {\em diameter-limited perceptron} \cite{minsky69}; (F) the other ancestor is the set-diagram network architecture \cite{rosenblatt62}; (H) The ancestor of the neuro-symbolic transition map of $\mathcal{S}$phNN is the neighbourhood graph in qualitative spatial reasoning \cite{Randell92a,Freksa91,CohnRenz07,DongJPL}.}
\label{visual_conclusion}
\end{sidewaysfigure}


We adopt the minimalist approach to qualitatively extending traditional neural networks by generalising the computational building block from vectors to spheres, and develop  
$\mathcal{S}$phNN for deterministic syllogistic reasoning. $\mathcal{S}$phNN has the genealogy of the {\em diameter-limited perceptron}, in the sense that the input of $\mathcal{S}$phNN is the input domain of a {\em diameter-limited perceptron}, as illustrated in Figure~\ref{visual_conclusion}(E). $\mathcal{S}$phNN can also be understood as a deviation of the set-diagram network architecture, in the sense the $\mathcal{S}$phNN utilises Euler diagram configuration in the vector space, Figure~\ref{visual_conclusion}(C), while Rosenblatt's set-diagram network used Venn diagram, as illustrated in Figure~\ref{visual_conclusion}(F). Traditional deep neural networks learn latent feature vectors from corpora, and $\mathcal{S}$phNN can host these vectors in sphere centres. In this way, traditional neural networks can be a part of Sphere Neural Networks, which optimise orientations of sphere centres from data, Figure~\ref{visual_conclusion}(G).  
Domain-general reasoning is rooted in spatial reasoning, which can be realised by constructing iconic mental models, Figure~\ref{visual_conclusion}(B). The construction process is carried out by repeatedly transforming the current sphere configuration to its neighbour till the target is achieved, Figure~\ref{visual_conclusion}(H).  \begin{sidewaysfigure}
\centering 
\includegraphics[width=1.\textwidth]{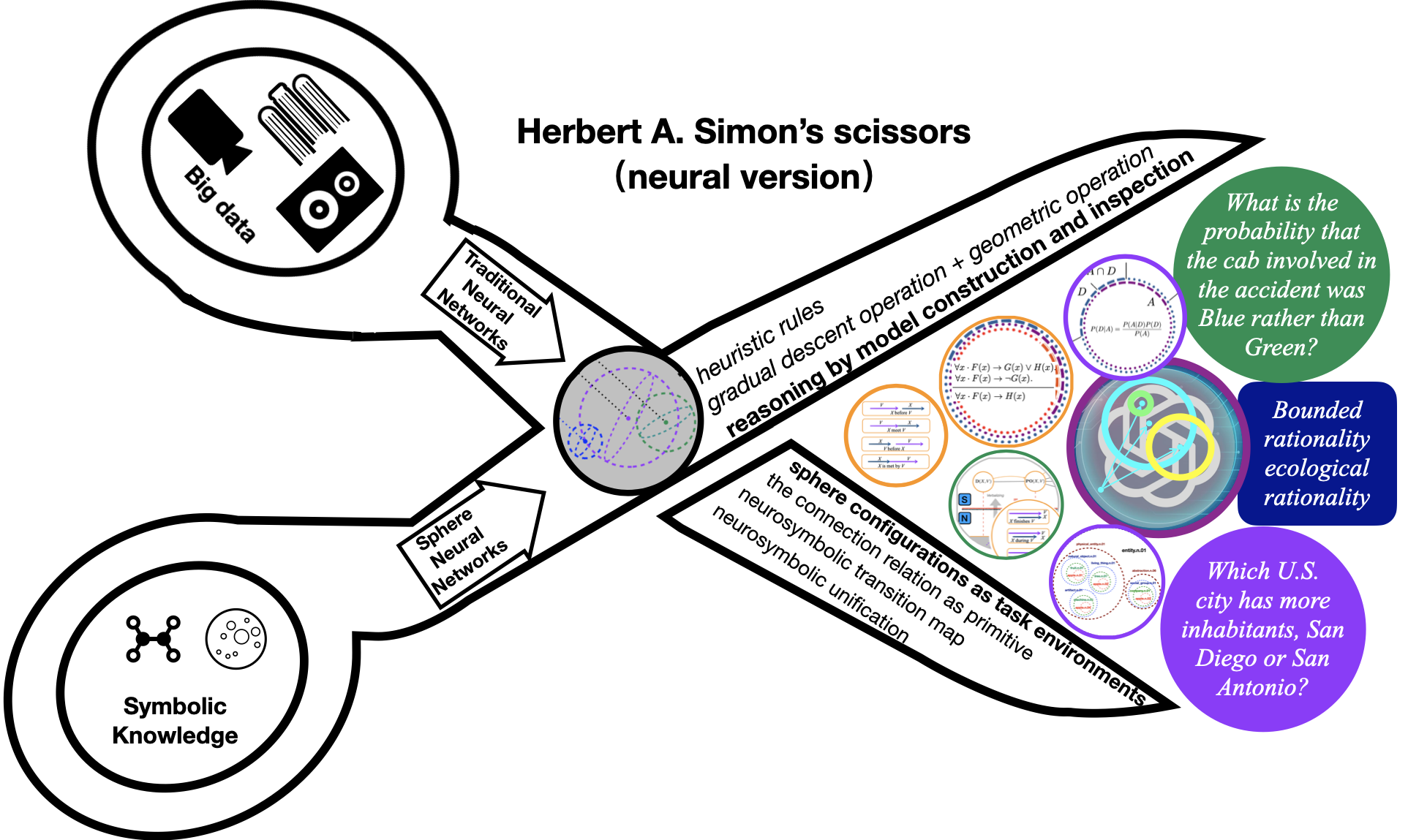}
\caption{Our neural version of Herbert A. Simon's scissors: the environment blade is structured by configurations of spheres that represent task environments in mind; the blade of rational reasoning consists of various processes that select parts of sphere configuration and solve problems by transforming towards the target configuration under certain conditions.}
\label{simon_scissor}
\end{sidewaysfigure}We develop $\mathcal{S}$phNN, Figure~\ref{visual_conclusion}(A), the first neural model that achieves the determinacy of long-chained syllogistic reasoning, the microcosm of human rationality and can evolve into simulate humour reasoning, the highest level of cognition. Syllogism and humour serve as the ends of the continuum of high-level cognition \cite{Dong_dag21syl,Dong_dag21Wheel}. Sphere Neural Networks promise to simulate a variety of rational reasoning in between, pave neural ways to implement Herbert A. Simon's scissors \cite{simon90scissor,Gigerenzer99,Gigerenzer02,Simon19} for heuristic reasoning under uncertainty and bounded resources (shown in Figure~\ref{simon_scissor}) and create a neural path for psychological AI that {\em process rules faster and
without errors} \cite[p.26]{Gigerenzer22}. 

The world of AI is filled with deep-learning skyscrapers, among which Foundation Models and Large Language Models (FM/LLMs) are the highest. They have demonstrated remarkable success in simulating various human intelligence, and 
when their parameters reach a large scale, e.g., 100 billion, the reasoning phenomenon can be observed \cite{wei2023COT,suzgun2022}. This seems to suggest that the larger the number of parameters is, the more powerful reasoning LLMs will have. But, this may be a mirage \cite{schaeffer2023}. 
Being able to communicate does not follow being able to reason \cite{language4commu}. Enhancing the decision-making capability of foundation models faces significant challenges, and certain components may be missing in current foundation models and decision-making paradigms \cite{foundation2023_yang}. Deterministic syllogistic reasoning, the microcosm of human rationality that dominated logical research for over 2000 years, will probably be an unreachable horizon for deep learning skyscrapers for another thousand years, let alone other rational reasoning. Alternatively, if we use spheres as the computational building block and adopt the methodology of reasoning by {\em model construction and inspection}, a very small-scaled neural network ($\mathcal{S}$phNN) already achieves human-like {\em determinate} logical deduction without training data, in which spheres play the role of the hub to connect with (1) traditional neural computing, (2) set-theoretic knowledge representation, (3) mental model theories, and (4) qualitative spatial reasoning. Traditional neural networks can be understood as special Sphere Neural-Networks where all radii are fixed to zero. The non-zero radii turn out to be the missing stakes that cause deep-learning skyscrapers to be trapped in the swamp of hallucination and prevent them from marching to the realm of rationality.   



%% file: 9arxiv/experiments_new.tex
\section{Experiments}
 
\subsection{Experiment 1}
\label{exp2}
This experiment examines whether $\mathcal{S}$phNN can determine every {\em valid} Aristotelian syllogistic deduction among 256 possibilities using the method of reasoning by model construction. 

\subsubsection{Method}

To determine the validity of a Aristotelian syllogistic deduction 
\syllogism{$r_1(S, M)$.\\ $r_2(M, P)$.}{$r_3(S, P)$.} $\mathcal{S}$phNN will try to refute it by constructing three spheres $\mathcal{O}_S$, $\mathcal{O}_M$, and $\mathcal{O}_P$, satisfying $\psi(r_1)(\mathcal{O}_S, \mathcal{O}_M)$, $\psi(r_2)(\mathcal{O}_M, \mathcal{O}_P)$, and $\neg\psi(r_3)(\mathcal{O}_S, \mathcal{O}_P)$. If it fails, $\mathcal{S}$phNN will conclude the original deduction is valid. {\em The principle of deterministic neural reasoning} guarantees $\mathcal{S}$phNN to correctly construct a sphere model for each satisfiable Aristotelian syllogistic deduction in one epoch.  

\subsubsection{Setting of experiments} We set the learning rate to 0.0001, the maximum number of epochs to 1, and set different dimensions of spheres (dim) $=2, 3,15, 30, 100, 200, 2000, 10000$. All spheres are initialised as being coincided, the radius being one and the length of the centre point being 10. All experiments were conducted on MacBook Pro Apple M1 Max (10C CPU/24C GPU), 32 GB memory. 

\subsubsection{Experiment results}  

Experiment results show that $\mathcal{S}$phNN accurately constructs Euler diagrams for each {\em satisfiable} Aristotelian syllogistic deduction in the first epoch. For any syllogistic structure whose global loss is greater than zero, $\mathcal{S}$phNN concludes the premises and the negation of the conclusion is valid. This totals exactly $256-232=24$ {\em valid} syllogistic reasoning structures. This shows that the transition map of neighbourhood relations successfully guides $\mathcal{S}$phNN to construct target sphere configurations and that three control processes are effective and efficient.

\begin{figure}[h]  
\begin{subfigure}[c]{0.5\textwidth}
\includegraphics[width=1\textwidth]{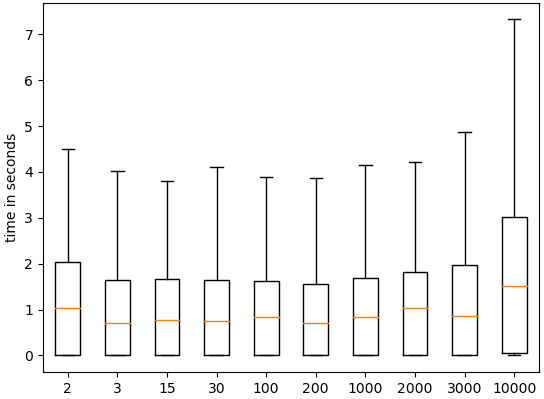} 
\subcaption{Time costs in seconds to determine a syllogistic reasoning being invalid.}
\end{subfigure}\hspace{24pt}
\begin{subfigure}[r]{0.5\textwidth}  
\centering
\includegraphics[width=1\textwidth]{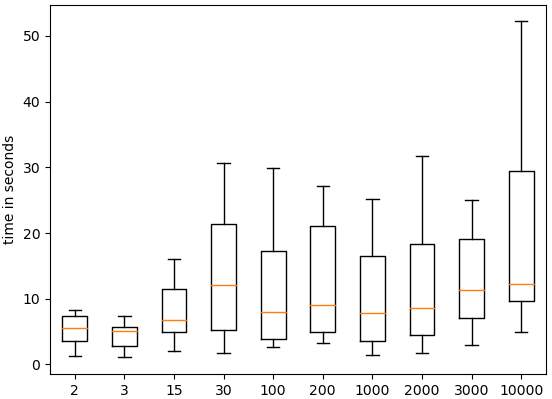}
\subcaption{Time costs in seconds to determine a syllogistic reasoning being valid.}
\end{subfigure}
\caption{Time costs for constructing spheres with different dimensions, from 2-dimensional circles to 10000-dimensional spheres. (a) The time costs do not change much as the dimensions of spheres increase; (b) There are more fluctuations in time costs as the sphere dimensions increase.}
\label{tc_sphnn}
\end{figure}

$\mathcal{S}$phNN took much more time to determine a reasoning being valid (the meantime is $43.68$ seconds) than to determine a reasoning being invalid (the meantime is $7.51$ seconds), as illustrated in Table~\ref{timecost_sphnn} and Figure~\ref{tc_sphnn}.

\begin{table}[t]
\caption{The statistics of time costs for $\mathcal{S}$phNN to determine the validity of an Aristotelian syllogistic reasoning. Sphere dimensions range from 2 to 10000.}
\label{timecost_sphnn}
\centering
\hspace*{-1.2em}
\scalebox{0.8}{
\begin{tabular}{c|c|c|c|c}
\hline  
time costs in seconds &{\bf max}&{\bf min} &{\bf mean}&{\bf median}\\\hline 
valid reasoning&467.19&1.44&43.68&24.48\\
\hline 
invalid reasoning&335.50&0.00018& 7.51 &3.78\\
\hline 
\end{tabular}
}
\end{table}

\subsubsection{Discussions}
\label{discussion}

If spheres were randomly initialised, $\mathcal{S}$phNN can successfully construct Euler diagrams for all {\em satisfiable} syllogistic statements, except one as follows.
\syllogism{$all$ M are S.\\ $all$ P are M.}{$all$ S are P.}
The only model of this syllogism is all three spheres being coincided. Our strict criterion (the global loss of zero) requires $\mathcal{S}$phNN to optimise the three spheres to have the same size and to be located at the same place, which gradual descent operations are hard to construct. Consequently, $\mathcal{S}$phNN mistakenly concluded the negation of the conclusion, namely, ``some $S$ are not $P$'', is valid. 


\subsection{Experiment 2}
\label{exp3}

This experiment evaluates the performance of $\mathcal{S}$phNN for long-chained syllogistic reasoning, compared with ChatGPTs (``gpt-3.5-turbo'' and ``gpt-4o'').

\subsubsection{Testing datasets}

For syllogistic reasoning with $N$ terms, we create 120 tasks in 24 groups; each group has 5 tasks, among which there is one valid and four invalid syllogistic reasoning. We range $N$ from 3 to 12, totalling 1200 tasks. 

\begin{figure}[h]  
\begin{subfigure}[c]{0.2\textwidth}
\includegraphics[width=0.8\textwidth]{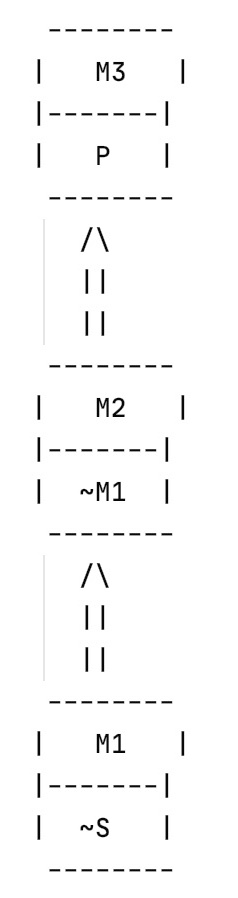} 
\subcaption{ChatGPT argues for the validity by constructing a satisfiable model. This is wrong.}
\end{subfigure}\hspace{24pt}
\begin{subfigure}[r]{0.6\textwidth}  
\centering
\includegraphics[width=1\textwidth]{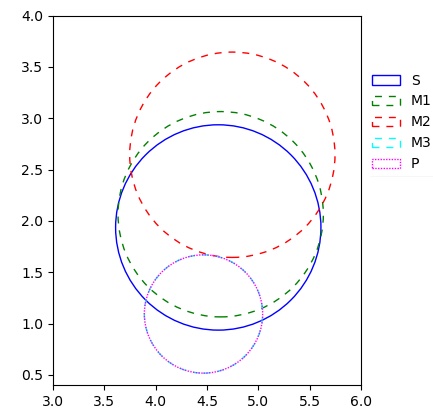}
\subcaption{$\mathcal{S}$phNN refutes the validity by finding a counter Euler diagram. The circle of P, centred at $( 4.4659342765808105,1.0935691595077515 )$ with the radius $0.5761237553232794$, is part the circle of M3 that is centred at $( 4.465934753417969,1.09356689453125)$ with the radius $0.5761421616706572$.}
\end{subfigure}
\caption{Both ChatGPT and $\mathcal{S}$phNN use diagrams to check the validity of the syllogistic reasoning: {\em some M1 are not S, some M2 are not M1, no M3 are M2, all P are M3}, $\therefore$ {\em no S are P}, which is invalid.}
\label{fig:neural_diagram}
\end{figure}
\subsubsection{Testing with ChatGPT and $\mathcal{S}$phNN}

We ask ChatGPT to decide the validity of syllogistic reasoning using prompts.

If we simply ask ChatGPT (gpt-3.5-turbo), ``is this reasoning always true? {\em some M1 are not S, some M2 are not M1, no M3 are M2, all P are M3, therefore, no S are P.}'' 
A typical answer of ChatGPT starts with its decision, followed by an explanation. 
\begin{quote}{\em
The reasoning is valid, and the conclusion is logically correct. To see why, we can draw a diagram (shown in Figure~\ref{fig:neural_diagram}(a)).
The diagram shows that there is no overlap between the S circle and the P circle and that all S are outside the P circle. This means that the statement "no S are P" is true based on the given premises.
}         
\end{quote}
Without detailed instruction, ChatGPT may provide partial explanations and diagrams as described above, which are hard for us to evaluate. So, we used the Level 6 TELeR prompt structure \citep{prompt23} to send syllogistic reasoning tasks with detailed instructions to ChatGPT, which consists of 6 parts: (1) assigning ChatGPT the role of a {\em professional logician}, (2) a detailed instruction, including output format, (3) a question, (4) a context, (5) output explanation, (6) an example. Our TELeR prompts instruct ChatGPT to decide the validity of syllogistic reasoning. 
If ChatGPT determines that the given reasoning is invalid, it will be asked to output the configuration of spheres as a counter-example.  We tell ChatGPT to give replies as short as possible and that most statements in the tasks are invalid. For the task above,  it replied, {\em ``valid. (circle S, inside, circle M1), (circle M2, inside, circle M1), (circle M3, inside, circle M2), (circle P, inside, circle M3), (circle S, outside, circle P).''} We can clearly see that ChatGPT made an incorrect decision and with an incorrect explanation for two reasons: (1) {\em some M2 are not M1} shall not be diagrammed as {\em circle M2 inside circle M1}; {\em no M3 are M2} shall not be diagrammed as {\em circle M3 are inside circle M2}; (2) methodologically, even ChatGPT correctly diagrammed all the statements, it is not sufficient to conclude the reasoning being valid. To show the reasoning is valid, ChatGPT shall explain that the premises and the negation of the conclusion are unsatisfiable. 

We feed the same syllogistic reasoning to $\mathcal{S}$phNN, and let it decide the validity. $\mathcal{S}$phNN first negates the conclusion into {\em some S are P}, then tries to construct an Euler diagram. Finally, $\mathcal{S}$phNN successfully constructs a counter-model to refute the validity of the input syllogistic reasoning, shown in Figure~\ref{fig:neural_diagram}(b).

\subsubsection{Testing results}
The best performance of ChatGPT (gpt-3.5-turbo) is $80\%$ (96 among 120 tasks when $N=4$), and the performances drop to $51.67\%$ (62 among 120 tasks) as the length of chains increases; the best performance of ChatGPT (gpt-4o) is $95.83\%$ ( among 120 tasks when $N=6$), and the performances drop to $89.17\%$ (107 among 120 tasks) as the length of chains increases, as listed in Table~\ref{chatgpt0}. 

\begin{table}[b]
\caption{Performances of {\bf ChatGPT}.}
\label{chatgpt0}
\centering
\hspace*{-1.2em}
\scalebox{0.8}{
\begin{tabular}{c|c|c|c|c|c|c|c|c|c|c|c}
\hline  
Syllogism with {\bf N} terms &{\bf 3}&{\bf 4} &{\bf 5}&{\bf 6}&{\bf 7}&{\bf 8}&{\bf 9}&{\bf 10}&{\bf 11}&{\bf 12}&total\\\hline 
Num. of correct decision of gpt-3.5-turbo&95&96&74&72&87&74&62&64&67&64& 120\\
Num. of correct decision of gpt-4o&110&108&109&115&114&115&112&110&107& 113&120\\
\hline 
\end{tabular}
}
\end{table}
\begin{table}
\caption{1200 tasks are grouped by the number of terms $N$ in a reasoning task. Each group has 120 reasoning tasks. The time limit affects the performance of {\bf $\mathcal{S}$phNN}.}
\label{sphnn1}
\centering
\hspace*{-1.2em}
\scalebox{0.8}{
\begin{tabular}{l|c|c|c|c|c|c|c|c|c|c}
\hline  
time limit &{\bf N} = {\bf 3}&{\bf 4} &{\bf 5}&{\bf 6}&{\bf 7}&{\bf 8}&{\bf 9}&{\bf 10}&{\bf 11}&{\bf 12}\\\hline
15&115&114&108&106&100&96&93&94&94&94\\
30&116&116&115&114&114&113&109&102&98&97\\
45&116&117&116&115&115&116&115&115&116&111\\
60&120&120&120&120&117&119&118&117&117&113\\
75&-&-&-&-&120&120&120&120&118&116\\
90&-&-&-&-&-&-&-&-&120&119\\
105&-&-&-&-&-&-&-&-&-&120\\
\hline 
\end{tabular}
}
\end{table}
In contrast, without a time limit, the performance of $\mathcal{S}$phNN is perfect. 
$\mathcal{S}$phNN does reasoning by constructing models, so generally, it needs more time for longer chained reasoning. This is also well revealed in the experiments. Table~\ref{sphnn1} listed the performance with time limits in making yes-no decisions. The longer the chain is, the more time $\mathcal{S}$phNN will need. When the chain length is 3, $\mathcal{S}$phNN can reach $95.83\%$ within 15 seconds; when the chain length is less than 7, $\mathcal{S}$phNN can finish a reasoning task with 100\% accuracy within 60 seconds. When the chain length increases to 12, $\mathcal{S}$phNN can reach $76.67\%$ within 15 seconds, and reach 100\% accuracy within 2 minutes (120 seconds). 

\subsubsection{Analysis}

The number of possible syllogistic structures increases exponentially with the number of terms -- there are $2^{3N-1}$ different syllogistic reasoning structures for $N$ terms. It is not possible for supervised deep learning to reach the rigour of syllogistic reasoning by increasing the amount of training data. A promising alternative approach is to construct models, as advocated by main-stream cognitive psychologists, e.g., \cite{RagniMarkus13,Khemlani2021, LairdByrne91,laird87}, which ChatGPT has often used in its answers and makes its answer look professional. The limitation seems that ChatGPT does not examine whether a model is correctly constructed.

\subsection{Experiment 3}
\label{exp4}

The human mind is a synergistic collaboration of associative fast thinking (System 1) and slow logic reasoning that follows rules (System 2) \citep{Kahneman11}. System 1 thinking generates coherent stories at the cost of logic, and System 2 thinking may check and refute the outcomes of System 1. LLMs' hallucination somehow mirrors features of System 1 and serves as creative foundations to develop modern AI products. This experiment uses syllogistic reasoning as a micro-world of the human mind, in which $\mathcal{S}$phNN serves as System 2 to check and improve the outputs of ChatGPT (System 1 simulators) through prompt engineering.


\subsubsection{The design of the experiment}

We test the performances of two ChatGPT versions (gpt-3.5-turbo and gpt-4o) with and without the support from $\mathcal{S}$phNN, in the task of deciding the satisfiability of 256 Aristotelian syllogistic statements, for example, given ``{\tt all S are M0. all M0 are P. all S are P.}'', ChatGPT shall decide whether the three statements are satisfiable.

We continued using the Level 6 TELeR prompt structure \citep{prompt23} to send ChatGPT syllogistic reasoning tasks. Our TELeR prompts give ChatGPT detailed instructions to decide the satisfiability of each Aristotelian syllogistic reasoning. 
If ChatGPT decides that a given syllogistic reasoning is satisfiable, it will output the configuration of spheres. $\mathcal{S}$phNN will check whether the decision is correct, and if correct, whether the decision is supported by the explanation. If one of the two is incorrect, $\mathcal{S}$phNN will send feedback to ChatGPT and let ChatGPT re-do the task. We tell ChatGPT to give replies as short as possible and that most tasks are satisfiable. For example, to check the satisfiability of {\em 'no M0 are S', 'all P are M0', 'some S are not P'}, we feed ChatGPT the original prompt, as follows. 

\begin{verbatim}
f"""We represent '''all X are Y''' as circle X being inside
circle Y, '''no X are Y''' as circle X disconnecting from 
circle Y. '''some X are Y''' as one of the three possible 
configurations:  (1) circle X is inside circle Y; (2) circle 
X partially overlaps with circle Y; (3) circle Y is inside 
circle X. '''some X are not Y''' as one of the three possible
configurations: (1) circle X disconnects from circle Y; 
(2) circle X properly contains circle Y; (3) circle X partially
overlaps with circle Y. If '''all X are Y''', then '''some X are Y'''. 

Can the statements '''no M0 S', 'all P M0', 'some S are not P''' be 
represented by relations among three circles? 

If they cannot be represented by relations among three circles, 
only reply '''cannot''', otherwise, reply '''yes''', and give 
the relations in the list of triple forms:  
(circle X, inside, circle Y). 

Do not reply both '''cannot''' and '''yes'''.
\end{verbatim}
ChatGPT replied as follows.

\begin{verbatim}
 yes (circle s, disconnects from, circle m0), (circle p, inside,
 circle m0), (circle s, partially overlaps with, circle p)
\end{verbatim}

$\mathcal{S}$phNN checks the correctness by constructing a model\footnote{For this experiment, we extended the neuro-symbolic transition map to have the partial overlapping relation as a target relation.} for the output, and failed. This information is sent back to ChatGPT by adding the following line to the end of the above prompt. 
\begin{verbatim}
It is not correct that '''(circle s, disconnects from, 
circle m0), (circle p, inside, circle m0), (circle s, partially
overlaps with, circle p)'''
\end{verbatim}
With this feedback from $\mathcal{S}$phNN, ChatGPT repeats the task and replied, \verb+cannot+.

$\mathcal{S}$phNN then checks the satisfiability of the three original syllogistic statements {\em `no M0 are S', `all P are M0', `some S are not P'}, and concludes that they are satisfiable, so, append the following line to the end of the last prompt. 
\begin{verbatim}
It is not correct that '''cannot'''.
\end{verbatim}
With continued feedback from $\mathcal{S}$phNN, ChatGPT replied as follows.
\begin{verbatim}
Yes, the statements can be represented by relations among
three circles. The relations in the triple form are:
(circle m0, inside, circle p)
(circle s, disconnects from, circle p)
(circle s, disconnects from, circle m0)
\end{verbatim}
This time, $\mathcal{S}$phNN successfully constructed a model for the three statements from ChatGPT. After two rounds of communication with $\mathcal{S}$phNN, ChatGPT correctly answered the question. 

\begin{table}
\caption{$\mathcal{S}$phNN checks decisions and explanations of ChatGPT (gpt-3.5-turbo) and sends feedback. `$\checkmark$EXPL' for correct explanation, `{\color{gray}\ding{55} H-}' for hallucination. The gray colour represents irrational cases.}
  \label{gpt35sphnn}
  \centering
  \begin{tabular}{lllllllr}
    \hline
 gpt-3.5-turbo  &\multicolumn{4}{c}{\#correct}&\multicolumn{2}{c}{\#incorrect }  & {\color{gray}Irrationality}                 \\    \cline{2-5}\cline{6-7}Explanation     & $\checkmark$EXPL &  {\color{gray}\ding{55} H0} & {\color{gray}\ding{55} H1} & {\color{gray}\ding{55} H2} & {\color{gray}$\checkmark$EXPL} & \ding{55}& {\color{gray}ratio}\\
    \hline 
no feedback&  121 ( 47.27\%)  & {\color{gray}{1}}   
  & {\color{gray}{66}}   & {\color{gray}{43}} &{\color{gray}{13}}  & 12  & {\color{gray}48.05\%}\\
$\le$10 feedback& 145 (56.64\%)  & {\color{gray}{3}}  & {\color{gray}{60}}  & {\color{gray}{40}} & {\color{gray}{0}} & 8&{\color{gray}40.23\%}\\
    \hline 
  \end{tabular}
\end{table}

\begin{table}
\caption{With how many times ($<10$) of $\mathcal{S}$phNN's feedback, ChatGPT (gpt-3.5-turbo) made a correct decision with a correct explanation. ChatGPT is not sensitive to $\mathcal{S}$phNN's feedback.}
  \centering
  \begin{tabular}{clllllllllll}
    \hline
  \#Number of feedback& 0  & 1 
  & 2  & 3&4 & 5  & 6& 7 
  & 8  & 9&10 \\
    \hline 
\#Tasks&  121 & 14 & 2 & 4 & 1 & 0 & 0 & 2 & 0 &1 &111\\
    \hline 
  \end{tabular}
\end{table}

\begin{table}
\caption{$\mathcal{S}$phNN checks decisions and explanations of ChatGPT (gpt-4o) and sends feedback. `$\checkmark$EXPL' for correct explanation, `{\color{gray}\ding{55} H-}' for hallucination. The gray colour represents irrational cases.}
  \label{gpt4osphnn}
  \centering
  \begin{tabular}{lllllllr}   
\hline
 gpt-4o  &\multicolumn{4}{c}{\#correct}&\multicolumn{2}{c}{\#incorrect}  & {\color{gray}Irrationality}                 \\  \cline{2-5}\cline{6-7} Explanation     & $\checkmark$EXPL &  {\color{gray}\ding{55} H0} & {\color{gray}\ding{55} H1} & {\color{gray}\ding{55} H2} & {\color{gray}$\checkmark$EXPL} & \ding{55}& {\color{gray}ratio}\\
    \hline 
no feedback&  198 (77.39\%)  & {\color{gray}{13}} 
  & {\color{gray}{3}}   & {\color{gray}{7}} &{\color{gray}{22}}  & 3 & {\color{gray}21.48\%}\\
$\le$10 feedback&  205 (80.01\%)  & {\color{gray}{26}}  & {\color{gray}{0}}  & {\color{gray}{4}} & {\color{gray}{14}} & 7&{\color{gray}17.19\%}\\
\hline
  \end{tabular}
\end{table}

\begin{table}
\caption{With how many times ($<10$) of $\mathcal{S}$phNN's feedback, ChatGPT (gpt-4o) made a correct decision with a correct explanation. ChatGPT is not sensitive to $\mathcal{S}$phNN's feedback.}
  \centering
  \begin{tabular}{clllllllllll}
    \hline
  \#Number of feedback& 0  & 1 
  & 2  & 3&4 & 5  & 6& 7 
  & 8  & 9&10 \\
    \hline 
\#Tasks&  198 & 4 & 2 & 0 & 0 & 1 & 0 & 0 & 0 &0 & 51  \\
    \hline 
  \end{tabular}
\end{table}

\subsubsection{Evaluation} 
$\mathcal{S}$phNN is used as a neural model to automatically classify the decision and the explanation of ChatGPT into six classes: (1-4) correct decision with correct or hallucinating explanations; (5) incorrect decision with correct explanation; (6) incorrect decision with incorrect explanation 
We distinguish three types of hallucinations: hallucinating with partial information (HALLU 0), hallucinating with incorrect relations (HALLU 1), and hallucinative reasoning (HALLU 2). The irrationality ratio is defined as the percentage of the sum of hallucinated correct decisions plus incorrect decisions (grey columns in Table~\ref{gpt35sphnn} and Table~\ref{gpt4osphnn}) with correct explanations in 256 cases.

\paragraph{HALLU 0: Correct decision with partial explanation}

For example, given the task of deciding the satisfiability of the three statements: {\tt some S are not M0. all M0 are P. all S are P.} 

ChatGPT decided {\em satisfiable}, but gave an explanation  {\tt (circle S, inside, circle P)
(circle P, inside, circle M0)
(circle M0, inside, circle P)}. This explanation is partial, as it neglected the relation between {\tt circle S} and {\tt circle M0}. Note: the relation {\tt (circle P, inside, circle M0)} and {\tt (circle M0, inside, circle P)} is not contradictory. The two relations will follow that {\tt M0} and {\tt P} are coincided.

\paragraph{HALLU 1: Correct decision, with at least one misexplanation of a single statement} 

For example, given the task of deciding the satisfiability of the three statements: {\tt no M0 are S. no M0 are P. all S are P.} 

ChatGPT decided {\em satisfiable}, but gave an explanation {\tt (circle P, inside, circle M0) 
(circle M0, outside, circle S)
(circle S, inside, circle P)}. Here, ChatGPT mistakenly interpreted no M0 are P as {\tt (circle P, inside, circle M0)}. 

\paragraph{HALLU 2: Correct decision with a logically incorrect explanation}

For example, given the task of deciding the satisfiability of the three statements: {\tt some M0 are not S. all M0 are P. some S are not P.} 

ChatGPT decided {\em satisfiable}, and gave an explanation  {\tt (circle M0, overlaps, circle S)
(circle M0, inside, circle P)
(circle S, outside, circle P)}. Here, it is correct that {\tt some M0 are not S} can be explained as {\tt (circle M0, overlaps, circle S)} and {\tt all M0 are P} can be explained as {\tt (circle M0, inside, circle P)} and {\tt some S are not P} can be explained as {\tt (circle S, outside, circle P)}. However, the relations among the three circles are not satisfiable.

\subsubsection{Testing results and lessons learned}

$\mathcal{S}$phNN not only successfully improves the performance of ChatGPT in deciding the satisfiability of syllogistic reasoning with three terms but also confirms the outputs of ChatGPT -- those approved by $\mathcal{S}$phNN are no more hallucinations. Using a maximum of 10 times feedback, ChatGPT (gpt-3.5-turbo) improves its performance (correct decision with correct explanation) from 47.27\% to 56.64\%, and reduced its irrationality ratio from 48.05\% to 40.23\%, as shown in Table~\ref{gpt35sphnn}. Although both ratios are greatly improved, ChatGPT often ignored $\mathcal{S}$phNN's feedback and could not correct all the mistakes. In testing the most advanced ChatGPT (version gpt-4o), without feedback, it achieved 77.39\% accuracy and 21.48\% irrationality. After a maximum of 10 times feedback, it improved the accuracy to 80.01\% and reduced the irrationality to 17.19\%, as shown in Table~\ref{gpt4osphnn}. ChatGPT still ignored $\mathcal{S}$phNN's feedback and repeat earlier incorrect answers. This suggests that prompt engineering may not be the most effective communication for System 2 simulators (either symbolic or neural systems) to correct errors of System 1 simulators. This also suggests that in addition to prompt engineering, there should be effective communication in the vector space through which LLMs cannot ignore $\mathcal{S}$phNN's feedback.

\subsection{Experiment 4}
\label{exp5}

Effective communication will be realised, when $\mathcal{S}$phNN's conclusion explicitly changes ChatGPT's vector embeddings. This requires $\mathcal{S}$phNN to use ChatGPT's vector embeddings as components of spheres. The aim of this experiment is to examine the possibility of whether pre-trained vector embedding can be understood as pointers that point to the orientations of spheres. The positive result may suggest a continuum between traditional neural networks and our novel sphere neural networks, from observing the locations of stars at night and drawing a star chart (pre-trained vector embedding) to observing the orbits of stars and planets, understanding the structures of galaxies (configurations of sphere embedding), and developing astronomy. 

This experiment shows that pre-trained vector embeddings are good approximations for the orientations of the sphere centres and the symbolic syllogistic relations can be precisely encoded by boundary relations among spheres. 

\subsubsection{Testing dataset}
We group 24 {\em valid} Aristotelian syllogism types into 14 groups, as ``{\em no X are Y}'' has the same meaning as ``{\em no Y are X}'' and ``{\em some X are Y}'' has the same meaning as ``{\em some Y are X}''. We collected 500 interpretations for each group using the hypernym relations in WordNet-3.0 \cite{MillerWordNet95},  totalling 7000 syllogism reasoning tasks. For example,   \syllogism{$\mbox{{\em all clarinetist.n.01 are musician.n.01}}$.\\$\mbox{{\em all  musician.n.01 are performer.n.01}}.$}{$\mbox{{\em  all clarinetist.n.01 are performer.n.01}}.$}
is an interpretation of the valid syllogistic reasoning 
\syllogism{$\mbox{{\em all S are M}}$.\\$\mbox{{\em all  M are P}}.$}{$\mbox{{\em  all S are P}}.$}
From the 7000 syllogistic reasoning tasks, we select those whose word stems, e.g., {\em clarinetist, musician, performer}, have different vector embeddings in 50-D GLOVE, in 1024-D BERT, in OpenAI text-embedding-ada-002 1536-D, OpenAI text-embedding-3-small, and OpenAI text-embedding-3-large, totalling $2537$ tasks remained. Next, for each task, we enumerate the other three possible conclusions. The other three conclusions in the above example are as follows. 
\begin{enumerate}[i]
    \item \mbox{{\em some clarinetist.n.01 are not performer.n.01}}
    \item \mbox{{\em no clarinetist.n.01 are performer.n.01}}
    \item \mbox{{\em some clarinetist.n.01 are performer.n.01}}
\end{enumerate}
This totals $2537\times 4=10148$ syllogistic reasoning tasks, among which, 6479 tasks are satisfiable, and 145 tasks are interpretations of the following syllogistic structure. 
\syllogism{$\mbox{{\em all S are M}}$.\\$\mbox{{\em all  M are P}}.$}{$\mbox{{\em  all P are S}}.$}
or
\syllogism{$\mbox{{\em all M are P}}$.\\$\mbox{{\em all  S are M}}.$}{$\mbox{{\em  all P are S}}.$}
The two types of syllogistic reasoning are satisfiable, only when the three terms refer to the same concept. Equivalently, when the three corresponding spheres are coinciding. 

  
\subsubsection{The design of the experiment}

\begin{figure} 
\centering 
\centering
\includegraphics[width=0.5\linewidth]{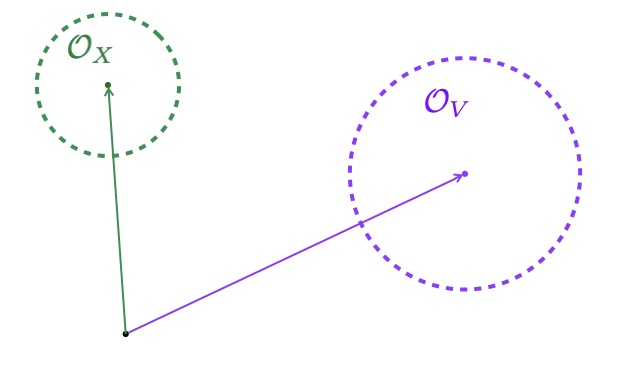} 
\caption{If $\mathcal{O}_V$  and the orientation of the centre of $\mathcal{O}_X$ are fixed, $\mathcal{O}_X$ cannot be inside $\mathcal{O}_V$.}
\label{fig:exp4}
\end{figure}

We create two experiment settings: (1) fix the orientation of a sphere centre to the pre-trained vector; (2) no orientation restriction and all spheres are initialised as being coincided in 2-dimensional space. When the orientations of two spheres are fixed, we cannot use the constraint optimisation algorithm, because this will change its orientation. We also notice that not each relation can be achieved by fixing one sphere. For example, if we fix $\mathcal{O}_V$  and the orientation of the centre of $\mathcal{O}_X$, $\mathcal{O}_X$ cannot be optimised to be inside $\mathcal{O}_V$, as illustrated in Figure~\ref{fig:exp4}. For this reason, the single relation optimisation may not be realised (Theorem 1). To solve this problem, we allow iterations and set the maximum iteration number to 2. 
This way, $\mathcal{O}_X$ being inside $\mathcal{O}_V$ will be realised in the next iteration, where $\mathcal{O}_X$ is fixed, and $\mathcal{O}_V$ enlarges its radius to contain $\mathcal{O}_X$.

\begin{table}
\caption{The orientation of pre-trained vectors very well approximates the orientation of spheres}
\label{ptv} 
\centering
\hspace*{-1.2em}
\scalebox{0.8}{
\begin{tabular}{c|c|c|c}
\hline 
\multicolumn{3}{c|}{using pre-trained vectors as the orientations of centres} &
\multicolumn{1}{c}{no restrictions}  \\
\hline
GLOVE 50-D &BERT 1024-D& OpenAI ada-002 & all spheres are initialised as being coincided\\
  & & OpenAI 3-small & \\
 & & OpenAI 3-large &  \\\hline
6479/6624 & 6479/6624 &6479/6624 &6624/6624\\ 
\hline 
$97.81\%$ & $97.81\%$ & $97.81\%$ & $100.00\%$ \\ 
\hline 
\end{tabular}
}
\end{table}

\subsubsection{Experiment results and analysis}

Without restriction, $\mathcal{S}$phNN successfully constructed models for every satisfiable syllogistic reasoning (6624 tasks). When orientations of sphere centres are fixed to pre-trained vectors, $\mathcal{S}$phNN successfully constructed models for all 6479 tasks, with 145 unsuccessful cases, as expected. Results are listed in shown in Table~\ref{ptv}. This positive result suggests the continuum between traditional neural networks and our novel sphere neural networks. This follows an efficient way for $\mathcal{S}$phNN to construct sphere configurations by fixing the orientation of a sphere centre to the pre-trained vector learned from the corpus. As listed in Table~\ref{tcptv}, $\mathcal{S}$phNN needs 1.48 seconds in average to construct a sphere configuration in 2-dimensional space or to determine there is no satisfiable sphere configuration. In contrast, by fixing centre orientation to pre-trained vectors, $\mathcal{S}$phNN only needs around 10\% of the time in average (from 0.07 seconds to 0.16 seconds).

\begin{table}
\caption{Average time costs by fixing centre orientation to pre-trained vectors or not. Time costs are measured in seconds (s).}
\label{tcptv} 
\centering
\hspace*{-1.2em}
\scalebox{0.8}{
\begin{tabular}{c|c|c|c|c|c} 
\hline
GLOVE 50-D &BERT 1024-D& OpenAI ada-002& OpenAI 3-small & OpenAI 3-large & no restriction \\\hline 
0.07s &0.12s& 0.13s& 0.11s & 0.16s & 1.48s\\ 
\hline 
\end{tabular}
}
\end{table}


%% file: 9arxiv/syllogism_list.tex
\newpage
\section{24 valid types of classic syllogistic reasoning}
\label{syl24}
Each valid syllogism is given a name whose vowels indicate types of moods, e.g., `C\underline{E}L\underline{A}R\underline{E}NT' indicates types of moods are `E', `A', `E', respectively. `A' for {\em universal affirmative}, all $X$ are $Y$, `I' for {\em particular affirmative}, some $X$ are $Y$; `E' for {\em universal negative}, no $X$ are $Y$, `O' for {\em particular negative}, some $X$ are not $Y$. 
\begin{table} [!b]
  \caption{List of all valid syllogisms, each is mapped to a qualitative spatial statement.}
\label{syllogismList}
  \centering 
\hspace*{-4em}
  \scalebox{0.85}{
  \setlength{\tabcolsep}{4.0pt} 
  \begin{tabular}{clllr}
    \hline 
   Num & Name     & Premise &Conclusion    & Qualitative spatial relations statement \\
    \hline 
   1& BARBARA & all $s$ are $m$, all $m$ are $p$ & all $s$ are $p$  & $\mathbf{P}(\mathcal{O}_s,\mathcal{O}_m)\wedge\mathbf{P}(\mathcal{O}_m,\mathcal{O}_p)\rightarrow\mathbf{P}(\mathcal{O}_s,\mathcal{O}_p)$     \\
   2& BARBARI & all $s$ are $m$, all $m$ are $p$ & some $s$ are $p$  & $\mathbf{P}(\mathcal{O}_s,\mathcal{O}_m)\wedge\mathbf{P}(\mathcal{O}_m,\mathcal{O}_p)\rightarrow\neg\mathbf{D}(\mathcal{O}_s,\mathcal{O}_p)$      \\
   3&CELARENT& no $m$ is $p$, all $s$ are $m$ & no $s$ is $p$& $\mathbf{D}(\mathcal{O}_m,\mathcal{O}_p)\land \mathbf{P}(\mathcal{O}_s,\mathcal{O}_m)\rightarrow\mathbf{D}(\mathcal{O}_s,\mathcal{O}_p)$     \\
    4& CESARE& no $p$ is $m$, all $s$ are $m$ & no $s$ is $p$& $\mathbf{D}(\mathcal{O}_p,\mathcal{O}_m)\land \mathbf{P}(\mathcal{O}_s,\mathcal{O}_m)\rightarrow\mathbf{D}(\mathcal{O}_s,\mathcal{O}_p)$     \\
   5& CALEMES &  all $p$ are $m$, no $m$ is $s$ & no $s$ is $p$& $\mathbf{P}(\mathcal{O}_p,\mathcal{O}_m)\land\mathbf{D}(\mathcal{O}_m,\mathcal{O}_s)\rightarrow\mathbf{D}(\mathcal{O}_s,\mathcal{O}_p)$ \\
   6& CAMESTRES & all $p$ are $m$, no $s$ is $m$ & no $s$ is $p$& $\mathbf{P}(\mathcal{O}_p,\mathcal{O}_m)\land \mathbf{D}(\mathcal{O}_s,\mathcal{O}_m)\rightarrow\mathbf{D}(\mathcal{O}_s,\mathcal{O}_p)$ \\
   7& DARII & all $m$ are $p$, some $s$ are $m$ & some $s$ are $p$  & $\mathbf{P}(\mathcal{O}_m,\mathcal{O}_p)\land \neg\mathbf{D}(\mathcal{O}_s,\mathcal{O}_m)\rightarrow\neg\mathbf{D}(\mathcal{O}_s,\mathcal{O}_p)$      \\
   8&DATISI& all $m$ are $p$, some $m$ are $s$ & some $s$ are $p$  & $\mathbf{P}(\mathcal{O}_m,\mathcal{O}_p)\land \neg\mathbf{D}(\mathcal{O}_m,\mathcal{O}_s)\rightarrow\neg\mathbf{D}(\mathcal{O}_s,\mathcal{O}_p)$      \\
   9&  DARAPTI & all $m$ are $s$, all $m$ are $p$ & some $s$ are $p$  & $\mathbf{P}(\mathcal{O}_m,\mathcal{O}_s)\wedge\mathbf{P}(\mathcal{O}_m,\mathcal{O}_p)\rightarrow\neg\mathbf{D}(\mathcal{O}_s,\mathcal{O}_p)$     \\
   10& DISAMIS & some $m$ are $p$, all $m$ are $s$ & some $s$ are $p$  & $\neg\mathbf{D}(\mathcal{O}_m,\mathcal{O}_p)\land \mathbf{P}(\mathcal{O}_m,\mathcal{O}_s)\rightarrow\neg\mathbf{D}(\mathcal{O}_s,\mathcal{O}_p)$      \\
  11& DIMATIS &  some $p$ are $m$, all $m$ are $s$ & some $s$ are $p$  & $\neg\mathbf{D}(\mathcal{O}_p,\mathcal{O}_m)\land \mathbf{P}(\mathcal{O}_m,\mathcal{O}_s)\rightarrow\neg\mathbf{D}(\mathcal{O}_s,\mathcal{O}_p)$      \\
   12& BAROCO& all $p$ is $m$, some $s$ are not $m$ & some $s$ are not $p$& $\mathbf{P}(\mathcal{O}_p,\mathcal{O}_m)\land \neg\mathbf{P}(\mathcal{O}_s,\mathcal{O}_m)\rightarrow\neg\mathbf{P}(\mathcal{O}_s,\mathcal{O}_p)$     \\
    13& CESARO& no $p$ is $m$, all $s$ are $m$ & some $s$ are not $p$  & $\mathbf{D}(\mathcal{O}_p,\mathcal{O}_m)\land \mathbf{P}(\mathcal{O}_s,\mathcal{O}_m)\rightarrow\neg\mathbf{P}(\mathcal{O}_s,\mathcal{O}_p)$     \\
   14&CAMESTROS & all $s$ are $m$, no $m$ is $p$ & some $s$ are not $p$  & $\mathbf{P}(\mathcal{O}_s,\mathcal{O}_m)\wedge\mathbf{D}(\mathcal{O}_m,\mathcal{O}_p)\rightarrow\neg\mathbf{P}(\mathcal{O}_s,\mathcal{O}_p)$     \\
   15&CELARONT& no $s$ is $m$, all $p$ are $m$ & some $s$ are not $p$& $\mathbf{D}(\mathcal{O}_s,\mathcal{O}_m)\wedge\mathbf{P}(\mathcal{O}_p,\mathcal{O}_m)\rightarrow\neg\mathbf{P}(\mathcal{O}_s,\mathcal{O}_p)$     \\
   16& CALEMOS& all $p$ are $m$, no $m$ is $s$ & some $s$ are not $p$& $\mathbf{P}(\mathcal{O}_p,\mathcal{O}_m)\land \mathbf{D}(\mathcal{O}_m,\mathcal{O}_s)\rightarrow\neg\mathbf{P}(\mathcal{O}_s,\mathcal{O}_p)$     \\
   17& BOCARDO & some $m$ are not $p$, all $m$ are $s$ & some $s$ are not $p$  & $\neg\mathbf{P}(\mathcal{O}_m,\mathcal{O}_p)\land \mathbf{P}(\mathcal{O}_m,\mathcal{O}_s)\rightarrow\neg\mathbf{P}(\mathcal{O}_s,\mathcal{O}_p)$     \\
   18& BAMALIP & all $m$ are $s$, all $p$ are $m$ & some $s$ are $p$  & $\mathbf{P}(\mathcal{O}_m,\mathcal{O}_s)\wedge\mathbf{P}(\mathcal{O}_p,\mathcal{O}_m)\rightarrow\neg\mathbf{D}(\mathcal{O}_s,\mathcal{O}_p)$      \\
   19&FERIO& some $s$ are $m$, no $m$ is $p$ & some $s$ are not $p$& $\neg\mathbf{D}(\mathcal{O}_s,\mathcal{O}_m)\wedge\mathbf{D}(\mathcal{O}_m,\mathcal{O}_p)\rightarrow\neg\mathbf{P}(\mathcal{O}_s,\mathcal{O}_p)$     \\
   20&FESTINO& some $s$ are $m$, no $p$ is $m$ & some $s$ are not $p$& $\neg\mathbf{D}(\mathcal{O}_s,\mathcal{O}_m)\wedge\mathbf{D}(\mathcal{O}_p,\mathcal{O}_m)\rightarrow\neg\mathbf{P}(\mathcal{O}_s,\mathcal{O}_p)$     \\
   21& FERISON& some $m$ are $s$, no $m$ is $p$ & some $s$ are not $p$& $\neg\mathbf{D}(\mathcal{O}_m,\mathcal{O}_s)\wedge\mathbf{D}(\mathcal{O}_m,\mathcal{O}_p)\rightarrow\neg\mathbf{P}(\mathcal{O}_s,\mathcal{O}_p)$     \\
  22&FRESISON& some $m$ are $s$, no $p$ is $m$ & some $s$ are not $p$& $\neg\mathbf{D}(\mathcal{O}_m,\mathcal{O}_s)\wedge\mathbf{D}(\mathcal{O}_p,\mathcal{O}_m)\rightarrow\neg\mathbf{P}(\mathcal{O}_s,\mathcal{O}_p)$     \\

   23&FELAPTON& all $m$ are $s$, no $m$ is $p$ & some $s$ are not $p$& $\mathbf{P}(\mathcal{O}_m,\mathcal{O}_s)\wedge\mathbf{D}(\mathcal{O}_m,\mathcal{O}_p)\rightarrow\neg\mathbf{P}(\mathcal{O}_s,\mathcal{O}_p)$     \\
   24&FESAPO& all $m$ are $s$, no $p$ is $m$ & some $s$ are not $p$& $\mathbf{P}(\mathcal{O}_m,\mathcal{O}_s)\wedge\mathbf{D}(\mathcal{O}_p,\mathcal{O}_m)\rightarrow\neg\mathbf{P}(\mathcal{O}_s,\mathcal{O}_p)$     \\
 \hline 
  \end{tabular}
  }
  \end{table}

%% file: 9arxiv/map_arch.tex
\section{The neuro-symbolic map for neighbourhood transition}
\label{tos}

Here, we list the complete structure of
$\mathcal{M}\triangleq (\mathcal{T}, f_{tsp}, \mathcal{I}, \mathcal{S},  f_{tnt}, \Delta)$.

$\mathcal{T}$  is the set of target relations.  $\mathcal{T}\triangleq\{\mathbf{D}, \neg\mathbf{D},\mathbf{P},\neg\mathbf{P},\overline{\mathbf{P}},\neg\overline{\mathbf{P}}\}$. 

$f_{tsp}$ is the function of {\em t}arget-oriented {\em s}patial {\em p}artitions. Given a target $\mathbf{T}\in\mathcal{T}$, $f_{tsp}(\mathbf{T})$ returns the set of qualitative spatial relations, as follows.
\begin{eqnarray*}
&&f_{tsp}(\neg\mathbf{D})\triangleq \{\neg\mathbf{D}, \mathbf{D}\}\\
&&f_{tsp}(\neg\mathbf{P})\triangleq \{\neg\mathbf{P},  \mathbf{P}\}\\
&&f_{tsp}(\neg\overline{\mathbf{P}})\triangleq \{\neg\overline{\mathbf{P}},  \overline{\mathbf{P}}\}\\
&&f_{tsp}(\mathbf{P})\triangleq \{\mathbf{P}, \mathbf{D}, \mathbf{PO}_1, \mathbf{PO}_2, \overline{\mathbf{PP}}\}\\
&&f_{tsp}(\mathbf{D})\triangleq \{\mathbf{D}, \mathbf{EQ}, \mathbf{PO}_1, \mathbf{PO}_2, \overline{\mathbf{PP}}, \mathbf{PP}\}\\
&&f_{tsp}(\overline{\mathbf{P}})\triangleq \{\overline{\mathbf{P}}, \mathbf{D}, \mathbf{PO}_3, \mathbf{PO}_4, \mathbf{PP}\}
\end{eqnarray*}

$\mathcal{I}$ is the set of inspection functions. Its element takes the form of $\mathcal{I}^\mathbf{R}(\mathcal{O}_1, \mathcal{O}_2)$. If $\mathbf{R}(\mathcal{O}_1, \mathcal{O}_2)$, $\mathcal{I}^\mathbf{R}(\mathcal{O}_X, \mathcal{O}_V) = 0$, otherwise $\mathcal{I}^\mathbf{R}(\mathcal{O}_X, \mathcal{O}_V) > 0$.  
\begin{eqnarray*}
&&\mathcal{I}^\mathbf{D}(\mathcal{O}_X, \mathcal{O}_V)=\max\{0, -dis_{X,V} + r_W + r_V\}\\
&&\mathcal{I}^{\neg\mathbf{D}}(\mathcal{O}_X, \mathcal{O}_V)=\max\{0, dis_{X,V} - r_W - r_V\}\\
&&\mathcal{I}^\mathbf{P}(\mathcal{O}_X,\mathcal{O}_V)\triangleq \max\{0, dis_{X,V} + r_X - r_V\}\\
&&\mathcal{I}^{\neg\mathbf{P}}(\mathcal{O}_X,\mathcal{O}_V)\triangleq \max\{0,  r_V - dis_{X,V} - r_X\}\\
&&\mathcal{I}^{\overline{\mathbf{P}}}(\mathcal{O}_X,\mathcal{O}_V)\triangleq\max\{0, dis_{X,V}+ r_V- r_X\}\\
&&\mathcal{I}^{\neg\overline{\mathbf{P}}}(\mathcal{O}_X,\mathcal{O}_V)\triangleq\max\{0,  r_X - dis_{X,V} - r_V\}\\
&&\mathcal{I}^{\mathbf{PO}}(\mathcal{O}_{X}, \mathcal{O}_{V})\triangleq\max\{0,  |r_X - r_V| - dis_{X,V} + \epsilon\}+\max\{0,  dis_{X,V}-r_V -r_X +\epsilon\}\\
&&\mathcal{I}^{\mathbf{PO}_1}(\mathcal{O}_{X}, \mathcal{O}_{V})\triangleq\mathcal{I}^{\mathbf{PO}}(\mathcal{O}_{X}, \mathcal{O}_{V})+\max\{0, r_V - dis_{X,V}+\epsilon\}\\
&&\mathcal{I}^{\mathbf{PO}_2}(\mathcal{O}_{X}, \mathcal{O}_{V})\triangleq\mathcal{I}^{\mathbf{PO}}(\mathcal{O}_{X}, \mathcal{O}_{V})+\max\{0,  dis_{X,V}-r_V \}\\
&&\mathcal{I}^{\mathbf{PO}_3}(\mathcal{O}_{X}, \mathcal{O}_{V})\triangleq\mathcal{I}^{\mathbf{PO}}(\mathcal{O}_{X}, \mathcal{O}_{V})+\max\{0,  r_V -r_X +\epsilon\}\\
&&\mathcal{I}^{\mathbf{PO}_4}(\mathcal{O}_{X}, \mathcal{O}_{V})\triangleq\mathcal{I}^{\mathbf{PO}}(\mathcal{O}_{X}, \mathcal{O}_{V})+\max\{0,  r_X -r_V\}
\end{eqnarray*}
\begin{eqnarray*}
&&\mathcal{I}^{\mathbf{EQ}}(\mathcal{O}_{X}, \mathcal{O}_{V})\triangleq  \|r_X -r_V\| + dis_{X,V}\\
&&\mathcal{I}^\mathbf{PP}(\mathcal{O}_X,\mathcal{O}_V)\triangleq \max\{0, dis_{X,V} + r_X - r_V + \epsilon\}\\
&&\mathcal{I}^{\overline{\mathbf{PP}}}(\mathcal{O}_X,\mathcal{O}_V)\triangleq\max\{0, dis_{X,V}+ r_V- r_X + \epsilon\}\\
\end{eqnarray*}
$\mathcal{S}$ is the set of all qualitative spatial relations in $\mathcal{M}$. $\mathcal{S}=\bigcup f_{tsp}(\mathbf{T})$, $\mathbf{T}\in\mathcal{T}$. 

Given a target relation $\mathbf{T}\in\mathcal{T}$, $\mathcal{S}$phNN inspects the current relation $\mathbf{R}\in f_{tsp}(\mathbf{T})$, and looks up the table of neighbourhood transitions to get the transformation function $f_{tnt}(\mathbf{T}, \mathbf{R})$, as listed in Table~\ref{trt_tb}. Possible operation sets of each transformation function are listed in Table~\ref{op_table}. 

$\Delta$ is the set of neighbourhood transition functions. Each transforms the current relation to the neighbourhood on the route to the target relation. 
\begin{table}
\caption{Possible operations of the $\Delta$ functions in Table~\ref{trt_tb}. $\mathcal{O}_V$ is fixed.}
\label{op_table}
\centering
\hspace*{-1.2em}
\scalebox{1}{
\begin{tabular}{l|c||l|c||l|c} 
\hline  
$\Delta^{-}_{-}(\mathcal{O}_X, \mathcal{O}_V)$& $operations$   & $\Delta_{\overline{\mathbf{PP}}:\mathbf{PO}}^{\mathbf{P}}$   &$dis^\downarrow$,  $r_X^\downarrow$  &$\Delta_{\mathbf{PO}_3:\mathbf{PO}_4}^{\overline{\mathbf{P}}}$   &  $r_X^\uparrow$ \\\hline   
$\Delta_{\mathbf{PO}_1}^\mathbf{D}$  & $dis^\uparrow$, $r_X^\downarrow$ & {\color{red}{$\Delta$}}$_{\mathbf{PO}_1:\mathbf{PO}_2}^{\mathbf{P}}$   & $dis^\downarrow$ &$\Delta_{\mathbf{PP}:\mathbf{PO}}^{\overline{\mathbf{P}}}$   & $r_X^\uparrow$ \\ \hline  
$\Delta_{\mathbf{PO}_2:\mathbf{PO}_1}^{\mathbf{D}}$  &  $dis^\uparrow$ & $\Delta^{\mathbf{P}}_{\mathbf{D}:\mathbf{PO}}$ & $dis^\downarrow$   & $\Delta_{\mathbf{PO}_4}^{\overline{\mathbf{P}}}$ & $dis^\uparrow$,  $r_X^\downarrow$  \\\hline  
$\Delta_{\mathbf{PP}:\mathbf{PO}}^{\mathbf{D}}$ &  $dis^\uparrow$   & $\Delta_{\mathbf{PO}_2}^{\mathbf{P}}$  & $dis^\downarrow$,  $r_X^\downarrow$   & $\Delta^{\overline{\mathbf{P}}}_{\mathbf{D}:\mathbf{PO}}$   &  $dis^\downarrow$, $r_X^\uparrow$ \\\hline   $\Delta_{\mathbf{EQ}:\mathbf{PO}}^{\mathbf{D}}$ & $dis^\uparrow$&$\Delta_{\mathbf{EQ}:\mathbf{PO}}^{\neg\mathbf{P}}$ & $dis^\uparrow$&$\Delta_{\mathbf{EQ}:\mathbf{PO}}^{\neg\overline{\mathbf{P}}}$ & $dis^\uparrow$
\\  \hline   $\Delta_{\overline{\mathbf{PP}}:\mathbf{PO}}^{\mathbf{D}}$ & $dis^\uparrow$,  $r_X^\downarrow$&$\Delta^{\neg\mathbf{P}}_\mathbf{P}$ & $dis^\uparrow$, $r_X^\uparrow$  & $\Delta^{\neg\overline{\mathbf{P}}}_{\overline{\mathbf{P}}}$  & $dis^\uparrow$, $r_X^\downarrow$ \\ 
\hline
$\Delta^{\neg\mathbf{D}}_\mathbf{D}$ &$dis^\downarrow$, $r_X^\uparrow$ &$\Delta^{\neg\mathbf{P}}_{{\overline{\mathbf{PP}}:\mathbf{PO}}}$ & $dis^\uparrow$, $r_X^\uparrow$  & $\Delta^{\neg\overline{\mathbf{P}}}_{\mathbf{PP}:\mathbf{PO}}$  & $dis^\uparrow$, $r_X^\downarrow$ \\ 
\hline
\end{tabular}
}
\end{table}
\begin{eqnarray*}
&&\Delta_{\mathbf{PO}_1}^\mathbf{D}(\mathcal{O}_X, \mathcal{O}_V)\triangleq  \max\{0, r_X + r_V  - dis_{X,V}\}\\
&&\Delta_{\mathbf{PO}_2:\mathbf{PO}_1}^{\mathbf{D}}(\mathcal{O}_{X}, \mathcal{O}_{V})\triangleq   \max\{0, r_V - dis_{X,V}\}\\
&&\Delta_{\mathbf{EQ}:\mathbf{PO}}^{\mathbf{D}}(\mathcal{O}_{X}, \mathcal{O}_{V})\triangleq   \Delta_{\mathbf{EQ}}^\mathbf{PO}(\mathcal{O}_{X}, \mathcal{O}_{V})\\
&&\Delta_{\mathbf{PP}:\mathbf{PO}}^{\mathbf{D}}(\mathcal{O}_{X}, \mathcal{O}_{V})\triangleq   \Delta_{\mathbf{PP}}^\mathbf{PO}(\mathcal{O}_{X}, \mathcal{O}_{V})\\
&&\Delta_{\overline{\mathbf{PP}}:\mathbf{PO}}^{\mathbf{D}}(\mathcal{O}_{X}, \mathcal{O}_{V})\triangleq   \Delta_{\overline{\mathbf{PP}}}^\mathbf{PO}(\mathcal{O}_{X}, \mathcal{O}_{V})\\
&&\Delta^{\mathbf{PO}}_\mathbf{EQ}(\mathcal{O}_X, \mathcal{O}_V)\triangleq  \vec{O}_X + \vec{\delta}
\end{eqnarray*}
\begin{eqnarray*}
&&\Delta^{\mathbf{PO}}_{\overline{\mathbf{PP}}}(\mathcal{O}_{X},\mathcal{O}_{V})\triangleq\max\{0,  r_X -r_V - dis_{X,V}\}\\
&&\Delta^{\mathbf{PO}}_\mathbf{PP}(\mathcal{O}_X, \mathcal{O}_V)\triangleq\max\{0, r_V -r_X - dis_{X,V}\}\\
&&\Delta_{\mathbf{D}:\mathbf{PO}}^{\overline{\mathbf{P}}}(\mathcal{O}_{X},\mathcal{O}_{V})\triangleq\Delta^{\mathbf{PO}}_\mathbf{D}(\mathcal{O}_X, \mathcal{O}_V)\\
&&\Delta_{\mathbf{PO}_4}^{\overline{\mathbf{P}}}(\mathcal{O}_X, \mathcal{O}_V)\triangleq\max\{0, dis_{X,V} + r_V - r_X\}\\
&&\Delta_{\mathbf{PO}_3:\mathbf{PO}_4}^{\overline{\mathbf{P}}}(\mathcal{O}_{X},\mathcal{O}_{V})\triangleq\Delta_{\mathbf{PO}_3}^{\mathbf{PO}_4}(\mathcal{O}_X, \mathcal{O}_V)\\
&&\Delta_{\mathbf{PP}:\mathbf{PO}}^{\overline{\mathbf{P}}}(\mathcal{O}_X, \mathcal{O}_V)\triangleq\Delta_{\mathbf{PP}}^\mathbf{PO}(\mathcal{O}_{X}, \mathcal{O}_{V})\\
&&\Delta^{\mathbf{PO}}_\mathbf{D}(\mathcal{O}_X, \mathcal{O}_V)\triangleq\max\{0, dis_{X,V} - r_V -r_X\}\\
&&\Delta_{\mathbf{PO}_3}^{\mathbf{PO}_4}(\mathcal{O}_X, \mathcal{O}_V)\triangleq \max\{0, r_V - r_X\}\\
&&\Delta_{\mathbf{D}:\mathbf{PO}}^\mathbf{P}(\mathcal{O}_X, \mathcal{O}_V)\triangleq\Delta^{\mathbf{PO}}_\mathbf{D}(\mathcal{O}_X, \mathcal{O}_V)\\
&&{\color{red}{\Delta}}_{\mathbf{PO}_1:\mathbf{PO}_2}^{\mathbf{P}}(\mathcal{O}_X, \mathcal{O}_V)\triangleq{\color{red}{\Delta}}_{\mathbf{PO}_1}^{\mathbf{PO}_2}(\mathcal{O}_X, \mathcal{O}_V)\\
&&{\color{red}{\Delta}}_{\mathbf{PO}_1}^{\mathbf{PO}_2}(\mathcal{O}_X, \mathcal{O}_V)\triangleq\max\{0,  \|\vec{O}_X-\vec{O}_V\| - r_V\}\\
&&\Delta_{\mathbf{PO}_2}^\mathbf{P}(\mathcal{O}_X, \mathcal{O}_V)\triangleq\max\{0, \|\vec{O}_X-\vec{O}_V\| + r_X -r_V\}
\\
&&\Delta_{\overline{\mathbf{PP}}:\mathbf{PO}}^\mathbf{P}(\mathcal{O}_X, \mathcal{O}_V)\triangleq\Delta^{\mathbf{PO}}_{\overline{\mathbf{PP}}}(\mathcal{O}_X, \mathcal{O}_V)\\
&&\Delta_\mathbf{D}^{\neg\mathbf{D}}(\mathcal{O}_X, \mathcal{O}_V)\triangleq\max\{0, \|\vec{O}_X-\vec{O}_V\| - r_V - r_X +\epsilon \}\\
&&\Delta_\mathbf{P}^{ \neg\mathbf{P}}(\mathcal{O}_X, \mathcal{O}_V)\triangleq\max\{0, r_V  - \|\vec{O}_X-\vec{O}_V\|- r_X +\epsilon \}\\
&&\Delta_{\overline{\mathbf{P}}}^{\neg\overline{\mathbf{P}}}(\mathcal{O}_X, \mathcal{O}_V)\triangleq\max\{0, r_X  - \|\vec{O}_X-\vec{O}_V\|- r_V +\epsilon \}
\end{eqnarray*}